\documentclass{article}

\PassOptionsToPackage{numbers, compress}{natbib}

\PassOptionsToPackage{numbers, compress}{natbib}

\usepackage[final]{neurips_data_2023}

\usepackage[utf8]{inputenc} 
\usepackage[T1]{fontenc}    
\usepackage{url}            
\usepackage{booktabs}       
\usepackage{amsfonts}       
\usepackage{nicefrac}       
\usepackage{microtype}      
\usepackage[dvipsnames,table]{xcolor}         

\usepackage{xspace}
\usepackage{paralist}
\usepackage{multirow}
\usepackage{paralist}
\usepackage{multicol}

\usepackage{float}
\usepackage{wrapfig}

\usepackage{amsmath}
\usepackage{longtable}

\usepackage{placeins}

\usepackage{algorithm}
\usepackage{algorithmic}

\usepackage{fontawesome5}
\newcommand{\faSearchPluss}{\reflectbox{\faSearchPlus}}
\newcommand{\faSearchMinuss}{\reflectbox{\faSearchMinus}}

\usepackage{subcaption}

\usepackage[export]{adjustbox}

\usepackage[thinlines]{easytable}
\usepackage{collcell}
\usepackage{tikz}

\newcommand{\layer}[1]{\ensuremath{\mathsf{#1}\xspace}}
\newcommand{\increase}[1]{(\textcolor{ForestGreen}{+#1})}
\newcommand{\increasenoparent}[1]{\textcolor{ForestGreen}{+#1}}
\newcommand{\decrease}[1]{(\textcolor{red}{-#1})}
\newcommand{\decreasenoparent}[1]{\textcolor{red}{-#1}}

\newcommand{\rrc}[0]{\texttt{RRC}\xspace}
\newcommand{\zoomin}[0]{\textit{zoom-in}\xspace}
\newcommand{\zoomout}[0]{\textit{zoom-out}\xspace}
\newcommand{\zoomless}[0]{\textit{zoom-224}\xspace}
\newcommand{\torchvision}[0]{\texttt{torchvision}\xspace}
\newcommand{\class}[1]{{\small\texttt{#1}\xspace}}

\newcommand{\papertitle}[0]{ImageNet-Hard: The Hardest Images Remaining from a Study of the Power of Zoom and Spatial Biases in Image Classification}

\makeatletter
\g@addto@macro{\endtabular}{\rowfont{}}
\makeatother
\newcommand{\rowfonttype}{}
\newcommand{\rowfont}[1]{
   \gdef\rowfonttype{#1}#1%
}
\newcolumntype{L}{>{\rowfonttype}l}
\usepackage{listings}

\definecolor{codegreen}{rgb}{0,0.6,0}
\definecolor{codegray}{rgb}{0.5,0.5,0.5}

\definecolor{backcolour}{RGB}{245,248,250}
\definecolor{emph}{RGB}{166,88,53}
\definecolor{nightblue}{RGB}{9,49,105}
\definecolor{keywords}{RGB}{207,33,46}
\definecolor{lightpurple}{RGB}{130,81,223}

\definecolor{MyLightGray}{rgb}{0.95, 0.95, 0.95}
\definecolor{CLIPBlue}{rgb}{0.192, 0.454, 0.643}
\newcommand{\clip}[0]{\textcolor{CLIPBlue}{CLIP\xspace}}

\lstdefinestyle{mystyle}{
    backgroundcolor=\color{backcolour},   
    commentstyle=\color{codegreen},
    keywordstyle=\color{keywords},
    stringstyle=\color{nightblue},
    basicstyle=\ttfamily\footnotesize,
    breakatwhitespace=false,         
    breaklines=true,                 
    captionpos=b,                    
    keepspaces=true,                 
    showspaces=false,                
    showstringspaces=false,
    showtabs=false,                  
    tabsize=2,
    frame=shadowbox,
    emph={AutoTokenizer,AutoModelForSequenceClassification,Explainer},
    emphstyle={\color{emph}},
    emph={[2]from_pretrained,compute_table},
    emphstyle={[2]\color{lightpurple}}
}

\usepackage{xcolor}

\newcommand\rebuttal[1]{\textcolor{black}{#1}}

\newcommand{\correct}[1]{\textcolor{ForestGreen}{#1}}
\newcommand{\wrong}[1]{\textcolor{red}{#1}}

\definecolor{mydarkblue}{rgb}{0,0.53,0.96}
\usepackage[pagebackref,breaklinks,colorlinks,citecolor=mydarkblue]{hyperref}

\usepackage[capitalize]{cleveref}
\crefname{section}{Sec.}{Secs.}
\Crefname{section}{Section}{Sections}
\Crefname{table}{Table}{Tables}
\crefname{table}{Tab.}{Tabs.}

\newcommand{\subsec}[1]{\noindent\textbf{#1}~~}
\newcommand{\ie}{\emph{i.e.}\xspace}
\newcommand{\eg}{\emph{e.g.}\xspace}

\definecolor{specialcolor}{rgb}{0.62, 0.32, 0.17}

\title{\papertitle}

\author{%
Mohammad Reza Taesiri \\
University of Alberta \\
\texttt{mtaesiri@gmail.com} \\
\And
Giang Nguyen \\
Auburn University\\
 \texttt{nguyengiangbkhn@gmail.com} \\
\And
Sarra Habchi \\
Ubisoft \\
\texttt{sarra.habchi@ubisoft.com} \\
\And
Cor-Paul Bezemer \\
University of Alberta \\
\texttt{bezemer@ualberta.ca} \\
\And
Anh Nguyen \\
Auburn University \\
\texttt{anh.ng8@gmail.com} \\
}

\begin{document}

\maketitle

\begin{abstract}
Image classifiers are information-discarding machines, by design.
Yet, how these models discard information remains mysterious.
We hypothesize that one way for image classifiers to reach high accuracy is to zoom to the most discriminative region in the image and then extract features from there to predict image labels, discarding the rest of the image.
Studying six popular networks ranging from AlexNet to CLIP, we find that proper framing of the input image can lead to the correct classification of 98.91\% of ImageNet images.
Furthermore, we 
uncover positional biases in various datasets, especially a strong center bias in two popular datasets: ImageNet-A and ObjectNet.
Finally, leveraging our insights into the potential of zooming, we propose a test-time augmentation (TTA) technique that improves classification accuracy by forcing models to explicitly perform zoom-in operations before making predictions.
Our method is more interpretable, accurate, and faster than MEMO, a state-of-the-art (SOTA) TTA method.
We introduce ImageNet-Hard, a new benchmark that challenges SOTA classifiers including large vision-language models even when optimal zooming is allowed.
\end{abstract}

\section{Introduction}
\label{sec:intro}

Since the release of AlexNet in 2012~\cite{krizhevsky2012imagenet}, deep neural networks have set many ImageNet (IN)~\cite{russakovsky2015imagenet} accuracy records \cite{krizhevsky2012imagenet,he2016deep}. 
While many papers reported improved learning algorithms or architectures, little is known about how the inner workings of image classifiers actually evolve.
The success is often attributed to a network's ability to detect more objects \cite{bau2017network} and a variety of facets of each object (\ie, invariance to style, pose, and form changes) \cite{nguyen2016multifaceted,goh2021multimodal}.
By aggregating the information from all the visual cues in a scene, a classifier somehow chooses a better label for the image.
\rebuttal{
For example,  Figs.~\href{https://link.springer.com/article/10.1007/s11263-019-01228-7}{13--14} in \cite{Ramprasaath2020GradCAM} show that a model detects both dogs and cats in the same image and only discards the dog features right before the classification layer to arrive at a \class{tiger cat} prediction.
}


\begin{figure}[ht]
    \centering
    \begin{subfigure}[b]{0.48\textwidth}
        \centering
        \begin{flushleft}
            \hskip -0.1in
            \scriptsize\rotatebox{90}{\kern -13.00pc IN-ReaL\kern 2.5pc IN-Sketch\kern 2.35pc IN-A}
        \end{flushleft}
        \includegraphics[width=\textwidth]{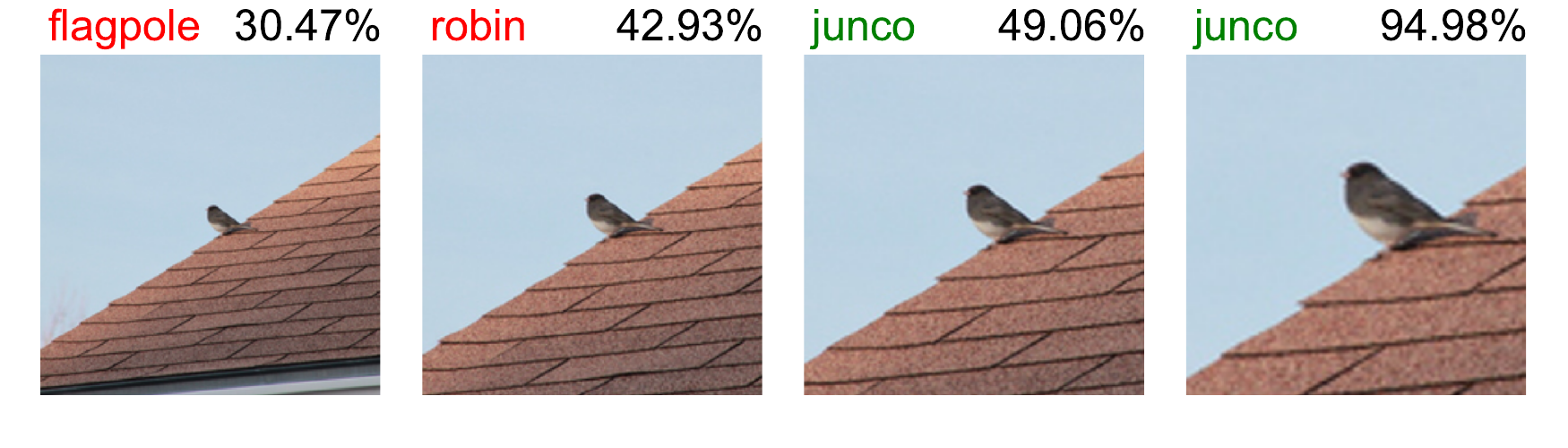}
        \includegraphics[width=\textwidth]{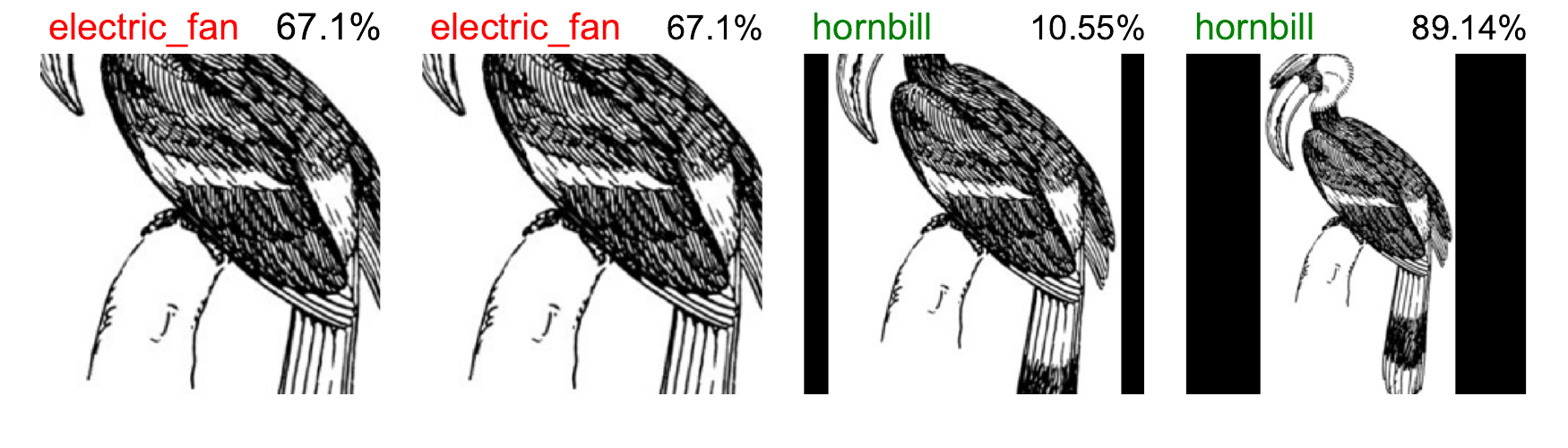}
        \includegraphics[width=\textwidth]{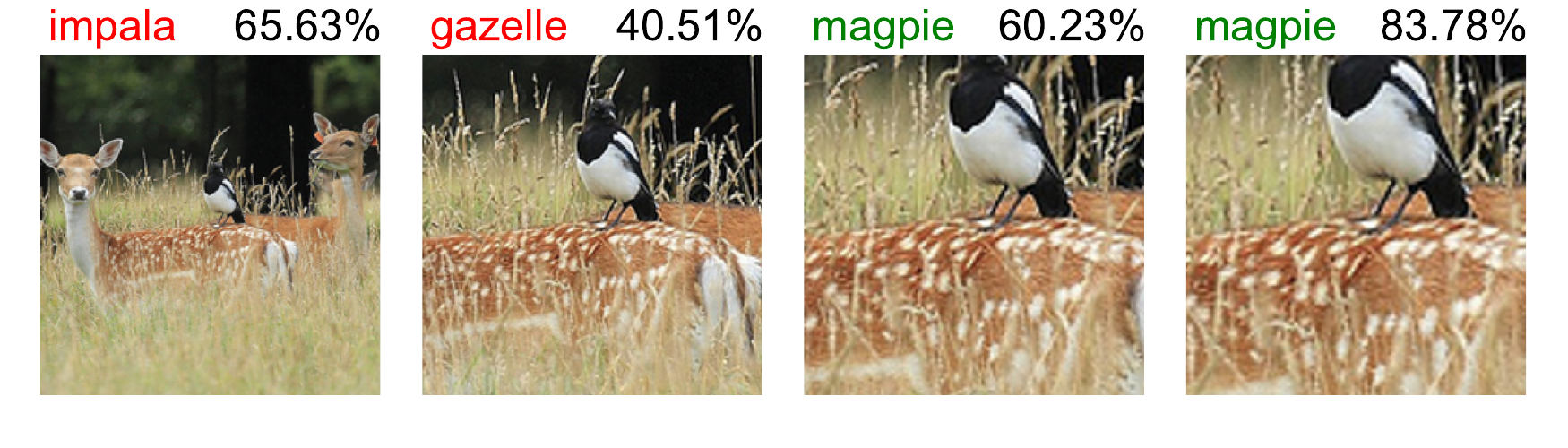}
        \caption{
        }
        \label{fig:teaser}
    \end{subfigure}
    \hfill
    \begin{subfigure}[b]{0.495\textwidth}
        \centering        \includegraphics[width=1.05\textwidth]{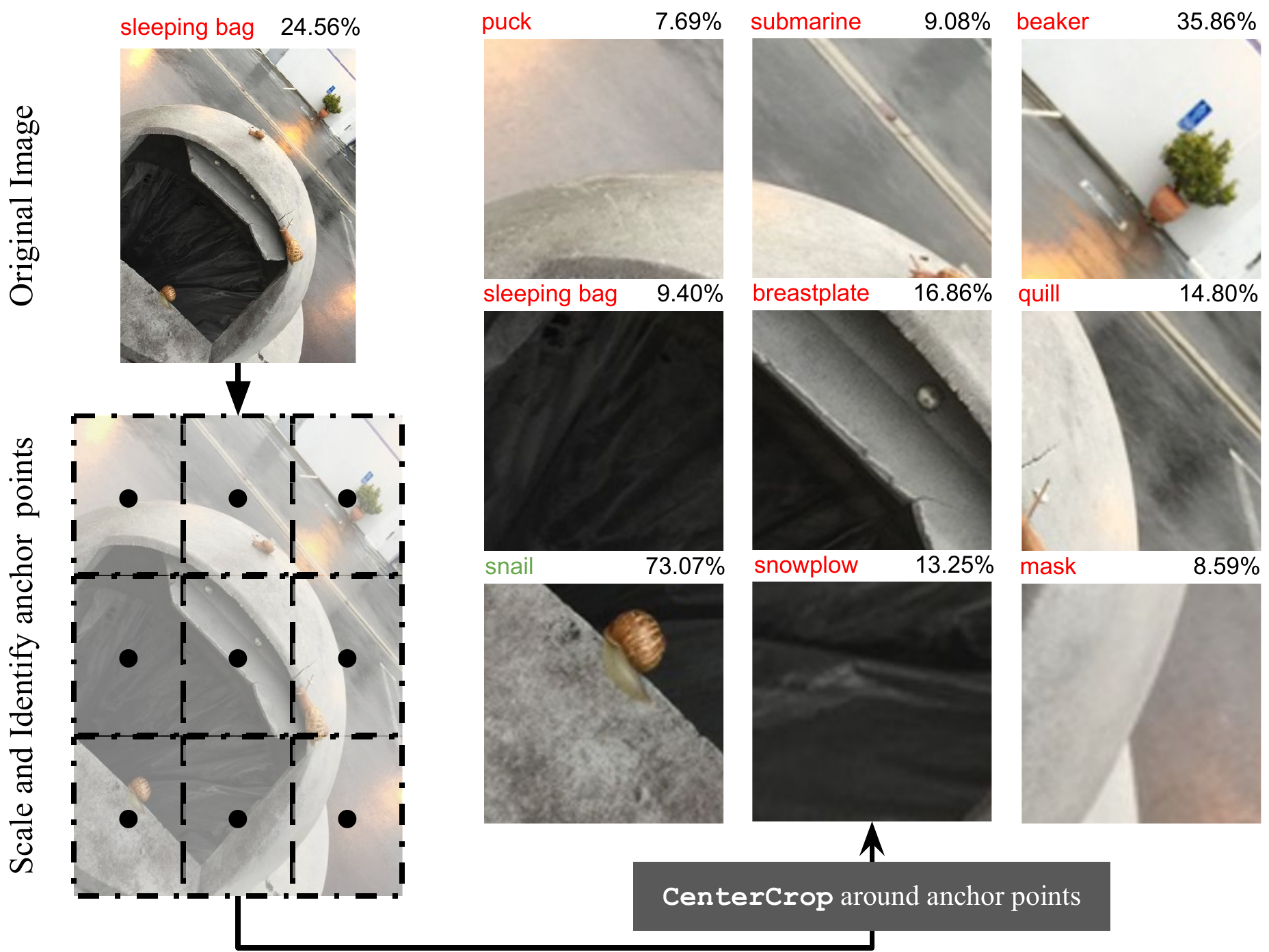}
        \caption{}
        \label{fig:better_overview}
    \end{subfigure}
    
    \caption{ \textbf{(a)} Each subfigure shows an input image, predicted label, and confidence score from an ImageNet classifier (top and middle: ResNet-50 \cite{he2016deep}; bottom: ViT-B/32 \cite{dosovitskiy2020image}). With the standard center-crop transform, all 3 samples were \wrong{misclassified} (left-most column). Adjusting framing via zooming yields \correct{correct} predictions. \textbf{(b)}  The zooming process correctly classifies a \class{snail} ImageNet-A image. We uniformly adjust the input query image's smaller dimension to match the target scale $S$. We then partition the image into a $3 \times 3$ grid, generating $9$ crops centered at grid-cell centers (\ie, $\bullet$ anchor points) and feed each crop to the original image classifier.
    }
\end{figure}

\rebuttal{
When processing an image, a network may implicitly \textbf{zoom} in or out (defined in \cref{sec:method}) to the most discriminative image region \emph{ignoring} the rest of the image (\cref{fig:teaser}), and then extract that localized region's features to predict image labels.
We hypothesize that the improved image classification may largely be due to the networks accurately zooming to the discriminative areas (\eg, \class{junco} and \class{magpie} birds in \cref{fig:teaser}) rather than more accurately describing them (\ie generating better features of these two birds).
}
In this work, we present supporting evidence for our zooming hypothesis.

We conduct a thorough study to test the effects of zooming in and out on the classification accuracy of six network architectures on six ImageNet-scale benchmarks. 
Our main findings also include:

\begin{compactenum}
    \item A major, surprising finding is that \textbf{state-of-the-art, IN-trained models can accurately predict up to $98.91\%$ of ImageNet samples when an optimally-zoomed image is provided}. 
    The remaining few hundred IN images ($0.39\%$) that are never correctly labeled by any model (despite optimal zooming) include mostly \emph{ill-posed} and \emph{rare} images (\cref{sec:maxpossibleaccuracy}).

    \item ImageNet-A~\cite{hendrycks2021natural} and ObjectNet~\cite{zhu2016object} both exhibit a substantial center bias. 
    For example, by only upsampling and center-cropping each ImageNet-A image, ResNet-50's accuracy increases dramatically from $0.09\%$ to $14.58\%$ (\cref{sec:center_bias}).

    \item Zooming can be leveraged as an inductive bias at test time to improve ImageNet classification accuracy. 
    That is, integrating zoom transformations into MEMO \cite{zhang2021memo}, a leading test-time augmentation method, yields consistently higher accuracy than the baseline ResNet-50 models and also MEMO with default transformations on multiple datasets (\cref{sec:memo_zoom}).

\end{compactenum}


Our findings show that the accuracy of image classifiers can be improved by finding an optimal zoom setting first and then classifying that crop alone (\cref{fig:teaser}).
Motivated by this insight, we build \textbf{ImageNet-Hard}\footnote{Code and data are available on \url{https://taesiri.github.io/ZoomIsAllYouNeed}.}, a new 1000-way classification benchmark that challenges state-of-the-art (SOTA) classifiers despite the application of optimal zooming (\cref{sec:imagenet_hard}).
In other words, we collect images from seven existing ImageNet-scale benchmarks where OpenAI's CLIP ViT-L/14~\cite{radford2021learning} misclassifies even when allowed to try 324 zooming settings.
Interestingly, SOTA classifiers that operate at 224$\times$224 resolution perform poorly on ImageNet-Hard (below 19\% accuracy).
Analyzing misclassifications on ImageNet-Hard reveals a major remaining challenge in the era of SOTA classifiers of Transformers~\cite{dosovitskiy2020image}, EfficientNets~\cite{tan2019efficientnet, kong2022efficient}, and large vision-language models (\cref{sec:analysis_errors}).

\section{Related Work}
\label{sec:related_works}


\subsec{Learning to Zoom in image classification} 
Leveraging zoom-in or crops of an image has a long history of improving fine-grained image classification with approaches varying from combining multiple crops at different resolutions \cite{fu2017look,uzkent2020learning}, using multiple crops of the object (i.e., part-based classification) \cite{donnelly2022deformable,krause2015fine,zheng2019looking,taesiri2022visual,zheng2019looking,kong2022efficient} to warping the input image \cite{recasens2018learning,jaderberg2015spatial,jin2021learning}.
We note that a common prior definition of ``zoom'' \cite{recasens2018learning,jaderberg2015spatial,jin2021learning,thavamani2023learning} is to first divide an input image into a grid and then warp the image, distorting the aspect ratio of the objects in the image.
In contrast, our zoom procedure utilizes only two functions: resize and crop, maintaining the original aspect ratio.

Furthermore, to our knowledge, we are the first to perform a zoom study on ImageNet-scale datasets (ImageNet \cite{russakovsky2015imagenet}, ImageNet-A \cite{hendrycks2021natural}, ObjectNet \cite{barbu2019objectnet}, etc) while prior zoom approaches \cite{donnelly2022deformable,krause2015fine,zheng2019looking,taesiri2022visual,zheng2019looking,kong2022efficient,jin2021learning} exclusively focus on non-ImageNet, fine-grained classification (e.g. classifying birds or dogs).
Due to such differences in the image distribution of interests, prior works mostly study \emph{zooming in} (which benefits fine-grained classification) while we study \emph{both} zooming in and out.

\subsec{Test-time data augmentation (TTA)} 
is a versatile technique that could help estimate uncertainty~\cite{smith2018understanding, bahat2020classification, ayhan2018test} and improve classification accuracy~\cite{krizhevsky2012imagenet, szegedy2015going, he2016deep, pang2019mixup, lyzhov2020greedy, shanmugam2021better, kim2020learning, chun2022cyclic}. 
When test inputs are sampled from unseen, non-training distributions, augmenting the data often improve a model's generalization to new domains~\cite{wang2020tent, zhang2021memo}.
A simple TTA method is 10-crop evaluation~\cite{krizhevsky2012imagenet} 
in which 5 patches of $224 \times 224$ px along with their horizontal reflections (resulting in 10 patches) are extracted from the original image.  
An alternative way to leverage the marginal output distributions over augmented data is to use them as gradient signals to update the classifier's parameters~\cite{wang2019learning, zhang2021memo}.
We employ this approach to update the model during test time, with patches being zoom-based augmentations.

\subsec{Biases in ImageNet and datasets}
Resize-then-center-crop has been a pre-processing standard for ImageNet classification since AlexNet \cite{krizhevsky2012imagenet}. 
This pre-processing exploits the known center bias of ImageNet.
While ImageNet has been shown to contain a variety of biases in image labels~\cite{beyer2020we, tsipras2020imagenet}, object poses~\cite{alcorn2019strike}, image quality~\cite{leung2021oowl500}, we are the first to examine the \emph{positional} biases of the out-of-distribution (OOD) benchmarks for ImageNet classifiers and find a strong center bias in ImageNet-A and ObjectNet that could affect how the community interprets progress on these OOD benchmarks.

Perhaps the closest to us is a preprint by Li et al.~\cite{li2021rethinking} that shows that cropping to the main object can improve model accuracy on ImageNet-A \cite{hendrycks2021natural}.
Yet, unlike \cite{li2021rethinking}, we study six ImageNet-scale datasets, both zooming in and zooming out, and we propose a new dataset of ImageNet-Hard.

\section{Method}
\label{sec:method}

\subsec{Zoom definition}
To zoom in or out of the image, we only use \textit{resize} and \textit{crop} operations. 
Initially, we uniformly resize the test image so that the smaller dimension matches the target scale of $S$.
Then, we define a $3 \times 3$ grid on the image to divide it into $9$ patches.
We perform a \href{https://pytorch.org/vision/main/generated/torchvision.transforms.CenterCrop.html}{\textit{CenterCrop}} operation at the center ($\bullet$ in \cref{fig:better_overview}) of each patch to extract a $224 \times 224$~px crop from each of the nine locations (see Python code in~\cref{sec:sample_python}).
In the \textit{CenterCrop} step, zero-padding is used when the content to be cropped is smaller than $224 \times 224$.
Overall, at each target scale $S$, we generate 9 crops (\cref{fig:better_overview}).

We test $36$ different values of $S$ ranging from $10$ to $1024$ px, resulting in a total of $36 \times 9=324$ different zoomed versions for each image.
Based on initial scale factor $S$, we define three groups: 
(1) \zoomout group contains all augmented crops where $S<224$; (2) \zoomin group contains all augmented crops where $S>224$; and (3) \zoomless group contains the $9$ patches where $S = 224$.

\subsec{Benchmark datasets}
We use the ImageNet~(IN)~\cite{russakovsky2015imagenet} dataset with both the original and ImageNet-ReaL~\cite{beyer2020we} (ReaL) labels. 
For each IN image, we use the union of the IN and ReaL labels (IN+ReaL) to complement each other to reduce noise in IN labels.
We further examine the effects of zoom-based transformations on four popular OOD benchmarks: (a) natural adversarials (ImageNet-A~\cite{hendrycks2021natural}), (b) image renditions (ImageNet-R~\cite{hendrycks2021many}), (c) black-and-white sketches (ImageNet-Sketch~\cite{wang2019learning}), and (d) viewpoint-and-background-controlled samples (ObjectNet~\cite{barbu2019objectnet}).
We refer to these as benchmarks as IN-A, IN-R, IN-S, and ON, respectively, in the rest of the paper.

\subsec{Classifiers}
We study the effects of zoom-based transformations on six popular image classifiers in the last decade: AlexNet \cite{krizhevsky2012imagenet}, VGG-16 \cite{simonyan2014very}, ResNet-18 \& ResNet-50 \cite{he2016deep}, ViT-B/32 \cite{dosovitskiy2020image}, and OpenAI's CLIP-ViT-L/14 \cite{radford2021learning}.  
The inclusion of the 11-year-old AlexNet provides a baseline for the power of deep features (when given the right region to look at). 
Predicted labels from CLIP-ViT-L/14 are acquired using its standard zero-shot classification setup (\cref{suppsec:zeroshot_clip_setup}).

\section{Experimental Results}
\label{sec:experimental_results_main}

\subsection{Zooming has the potential to substantially improve image classification accuracy}
\label{sec:maxpossibleaccuracy}

\begin{figure}[ht]
    \begin{minipage}{0.535\textwidth}
        \centering
        \captionof{table}{
        On in-distribution data (IN \& ReaL) there exists a substantial improvement when models are provided with an optimal zoom, either selected from $36$ (b) or $324$ pre-defined zoom crops (c).
        In contrast, \colorbox{MyLightGray}{OOD benchmarks} still pose a significant challenge to \textcolor{specialcolor}{IN-trained models} even with optimal zooming (i.e., all upper-bound accuracy scores < $80\%$).
        }
        \label{tab:1_main_results}
        \resizebox{\textwidth}{!}{%
\begin{tabular}{lrrcrrrr}
 & \multicolumn{1}{c}{IN} & \multicolumn{1}{c}{ReaL} & \multicolumn{1}{l}{IN+ReaL} & \multicolumn{1}{c}{\cellcolor{MyLightGray} IN-A} & \multicolumn{1}{c}{\cellcolor{MyLightGray} IN-R} & \multicolumn{1}{c}{\cellcolor{MyLightGray} IN-S} & \multicolumn{1}{c}{\cellcolor{MyLightGray} ON} \\ \bottomrule
\multicolumn{8}{l}{(a) \textit{Standard top-1 accuracy based on $N$ = \textbf{1} crop}} \\ 
{\textcolor{specialcolor}{AlexNet}} & 56.16 & 62.67 & 61.76 & 1.75 & 21.10 & 10.05 & 14.23 \\ 
{\textcolor{specialcolor}{VGG-16}} & 71.37 & 78.90 & 78.52 & 2.69 & 26.98 & 16.78 & 28.32 \\
{\textcolor{specialcolor}{ResNet-18}} & 69.45 & 76.94 & 76.47 & 1.37 & 32.14 & 19.41 & 27.59 \\
{\textcolor{specialcolor}{ResNet-50}} & 75.75 & 82.63 & 82.97 & 0.21 & 35.39 & 22.91 & 36.18 \\
{\textcolor{specialcolor}{ViT-B/32}} & 75.75 & 81.89 & 82.59 & 9.64 & 41.29 & 26.83 & 30.89 \\
\textbf{\footnotesize{\clip-ViT-L/14}} & 75.03 & 80.68 & 81.95 & 71.28 & 87.74 & 58.23 & 66.32 \\  \bottomrule 
\multicolumn{8}{l}{(b) \textit{Upper-bound accuracy using $N$ = \textbf{36} crops}} \\ \rowcolor{yellow}
{Random}     & 3.60  &       3.60 &            3.60 &  18.00 &  18.00   &   3.60    &   31.85    \\
{\textcolor{specialcolor}{AlexNet}}       & 85.19 & 90.30 & 89.74 & 31.37 & 47.04 & 24.40 & 49.17 \\
{\textcolor{specialcolor}{VGG-16}}        & 92.30 & 96.08 & 95.81 & 46.69 & 52.86 & 34.34 & 62.94 \\
{\textcolor{specialcolor}{ResNet-18}}     & 92.08 & 95.97 & 95.73 & 47.48 & 58.85 & 37.91 & 63.08 \\
{\textcolor{specialcolor}{ResNet-50}}     & 94.46 & 97.36 & 97.40 & 55.68 & 61.42 & 41.71 & 69.60 \\
{\textcolor{specialcolor}{ViT-B/32}}     & 95.05 & 97.61 & 97.88 & 68.43 & 68.77 & 49.10 & 70.30 \\
\textbf{\footnotesize{\clip-ViT-L/14}}  & 94.19 & 97.32 & 97.56 & 97.16 & 98.60 & 83.77 & 89.59 \\
\bottomrule
\multicolumn{8}{l}{(c) \textit{Upper-bound accuracy using $N$ = \textbf{324} crops}} \\
\rowcolor{yellow} {Random}     & 32.40  &       32.40 &            32.40 & 100.00  &  100.00 & 32.40      & 100.00       \\
{\textcolor{specialcolor}{AlexNet}} & 90.03 & 93.85 & 93.48  & 42.23 & 55.52 & 29.53 & 59.65 \\ 
{\textcolor{specialcolor}{VGG-16}} & 95.30 & 97.90 & 97.66 & 58.27 & 60.88 & 39.90 & 71.85 \\
{\textcolor{specialcolor}{ResNet-18}} & 95.15 & 97.76 & 97.55	 & 58.87 & 66.89 & 43.68 & 71.44 \\
{\textcolor{specialcolor}{ResNet-50}} & 96.78 & 98.62 & 98.57 & 66.68 & 68.84 & 47.64 & 76.83 \\
{\textcolor{specialcolor}{ViT-B/32}} & \textbf{97.19} & \textbf{98.75} & \textbf{98.91} & 78.03 & 75.58 & 55.99 & 79.28 \\
\textbf{\footnotesize{\clip-ViT-L/14}} & 96.78 & 98.69 & 98.80 & \textbf{98.45} & \textbf{99.20} & \textbf{89.00} & \textbf{93.13} \\ \bottomrule
\end{tabular}
        }
    \end{minipage}
    \hfill
 \begin{minipage}{0.435\textwidth}
        \centering
        \includegraphics[width=0.9\textwidth]{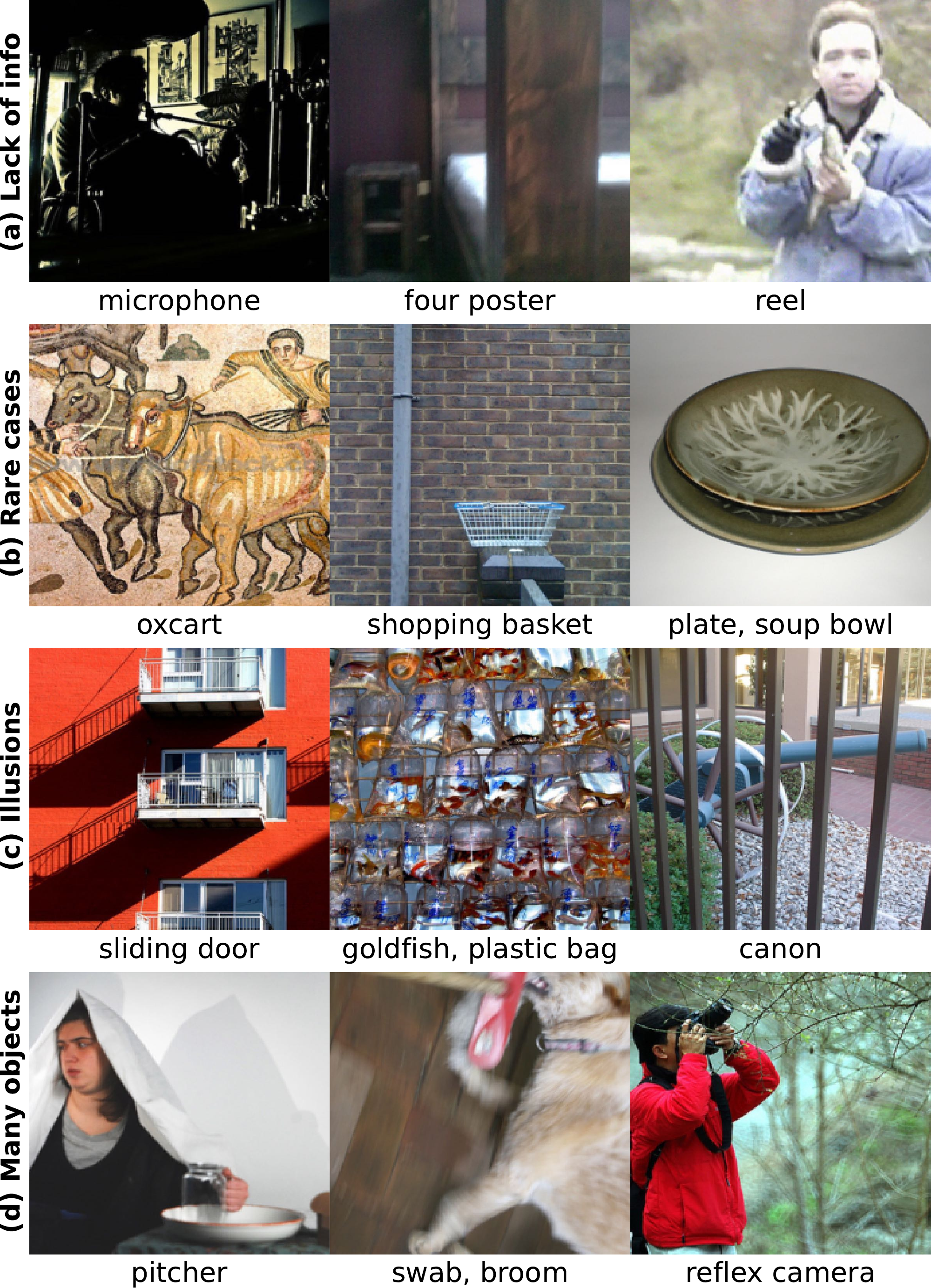}
        \captionof{figure}{IN+Real samples that are not correctly classifiable by \textcolor{specialcolor}{IN-trained models} using any of the 324 zoom transforms.}
        \label{fig:hard_images_reaL}
    \end{minipage}
\end{figure}

To understand the potential of zooming in improving image classification accuracy, first, we establish an \textbf{upper-bound accuracy} (\ie when an ``optimal'' zoom is given).
That is, we apply $36$ scales $\times$ $9$ anchors = $324$ zoom transformations (\cref{sec:method}) to each image to generate $324$ zoomed versions of the input.
We then feed all $N = 324$ versions to each network and label an image ``correctly classified given the optimal zoom'' if at least 1 of the $324$ is correctly labeled. 
We call such maximum possible top-1 accuracy ``upper-bound accuracy'' in \cref{tab:1_main_results}.
Our experiment also informs the community of the type of image that \emph{cannot} be correctly labeled even with an optimal zooming strategy.

\subsec{Results} 
Table~\ref{tab:1_main_results} shows upper-bound accuracy over different values of $N = \{1, 36, 324 \}$.
First, the \colorbox{yellow}{random} baselines (given $N = 324$ attempts per image) are at $32.4\%$ for 1,000 classes (IN, ReaL, IN+ReaL, and IN-Sketch), and $100\%$ for 200 and 313 classes (IN-A and ON, respectively). 
Yet, the accuracy of models with optimal zooming is far from random---\eg ResNet-50 largely outperforms not only the random baseline but also the 1-crop baseline on IN ($96.78\%$ vs. $75.75\%;$ \cref{tab:1_main_results}a vs. c).

On the IN, ReaL, and IN+ReaL datasets, there is a substantial gap for all models (around \increasenoparent{20} to \increasenoparent{35} points) between the 1-crop and the optimal zooming setting (\cref{tab:1_main_results}a vs. c).
Surprisingly, given optimal zooming, the 11-year-old AlexNet actually can correctly label over $90\%$ of IN images, which is roughly the 1-crop accuracy ($87.8\%$)  of the 2022 state-of-the-art ConvNexts \cite{liu2022convnet}.
This result is consistent with our hypothesis: One way for state-of-the-art classifiers to obtain their current accuracy is to simply learn \emph{how to zoom} on top of the same, old feature extractors (\eg that of AlexNet).

\subsec{Unclassifiable IN images} 
Interestingly, even with optimal zooming, no model reaches $100\%$ on IN images.
We find that $0.39\%$ of the IN+ReaL images were not classified correctly by any of the \textcolor{specialcolor}{IN-trained classifiers} and these images are similar to natural adversarial images (\cref{fig:hard_images_reaL}) and can be categorized into four groups:

\begin{enumerate}
    \item \textbf{Lack of information} (\cref{fig:hard_images_reaL}a): Images lack adequate signals for classification due to low light, occlusion, blurriness, or noise.

    \item \textbf{Rare cases} (\cref{fig:hard_images_reaL}b): Images depict the primary object but in an uncommon form, pose, or rendition.

    \item \textbf{Illusions} (\cref{fig:hard_images_reaL}c): Images have misleading elements, like a shadow appearing as a staircase, leading to misclassification.

    \item \textbf{Many objects} (\cref{fig:hard_images_reaL}d): Images displaying several classes of objects but not all classes are listed in the set of groundtruth labels.
\end{enumerate}





\subsec{OOD datasets pose a significant challenge to \textcolor{specialcolor}{IN-trained models} despite optimal zooming.}
Across IN-A, IN-R, and ON, all \textcolor{specialcolor}{IN-trained models} perform far below the 324-crop random baseline ($100\%$) with the highest score being $79.28\%$ (\cref{tab:1_main_results}).
In contrast, CLIP reaches far better scores than IN-trained models ($98.45\%$ on IN-A and $99.20\%$ on IN-R; \cref{tab:1_main_results}).
Our result suggests that OOD images (\eg objects in unusual poses or renditions) require a more robust feature extractor to recognize besides zooming.
And that CLIP was trained on an Internet-scale dataset \cite{radford2021learning} and thus is much more familiar with a variety of  poses, styles, and shapes of objects \cite{goh2021multimodal}.


\rebuttal{
Among the 324 zoom transformations, for each (classifier, dataset) pair, we initially construct a bipartite graph connecting transforms to images based on their correct classification.
With this graph, we employ the iterative, greedy minimum-set cover algorithm~\cite{petr1996setcover,algo-design} to compute the minimum set of transforms required to achieve the upper-bound accuracy detailed in~\cref{sec:maxpossibleaccuracy}.
Through this process, we discover that, on average, only 70\% of the transforms are essential.
Furthermore, we identify the \textbf{top-36 zoom transforms most important to classification} (see visualizations in~\cref{supp:viz_36crop}). More details on this process can be found in~\cref{sec:mincover}.
}



The upper-bound accuracy using 36 crops (\cref{tab:1_main_results}b) is only \emph{slightly lower} than that when using all $324$ crops but is substantially higher than (1) the standard 1-crop, \eg $85.19\%$ vs. $56.16\%$ for AlexNet on IN (\cref{tab:1_main_results}b); and (2) the random baseline (\ie $3.6\%$ for IN).
Our result confirms that these 36 zoom transforms are indeed important to classification (not because models are given 36 random trials per image) and that studying them might reveal interesting insights into the datasets.

As our 324 transforms include both zoom-in and zoom-out, we further analyze the contribution of each zoom type to each dataset.
We find that, across 7 datasets, zoom-in is more useful than zoom-out.
And that \textbf{zoom-out is the most important to abstract images} \ie, of IN-R and IN-S (\cref{sec:disentagling_zoom}).


\subsection{ImageNet-A and ObjectNet suffer from a severe center bias}
\label{sec:center_bias}



The standard image pre-processing for IN-trained models involves resizing the image so its smaller dimension is $256$, then taking the center $224 \times 224$ crop of the resized image \cite{center-crop,krizhevsky2012imagenet}.
While suitable for ImageNet, this pre-processing may not be optimal for every OOD dataset, not allowing a model to fully utilize off-center visual cues (which optimal zooming could).
Leveraging the minimum set of transforms obtained in \cref{sec:mincover}, we quantify which spatial locations (out of $9$ anchors; \cref{fig:better_overview}) contain the most discriminative features in each dataset.
That is, we compute the upper-bound accuracy for each of the $9$ anchor points per dataset and discover biases in some benchmarks.


\begin{figure}
\centering

\begin{subfigure}{0.1925\columnwidth}
    \resizebox{\columnwidth}{!}{
    \includegraphics[]{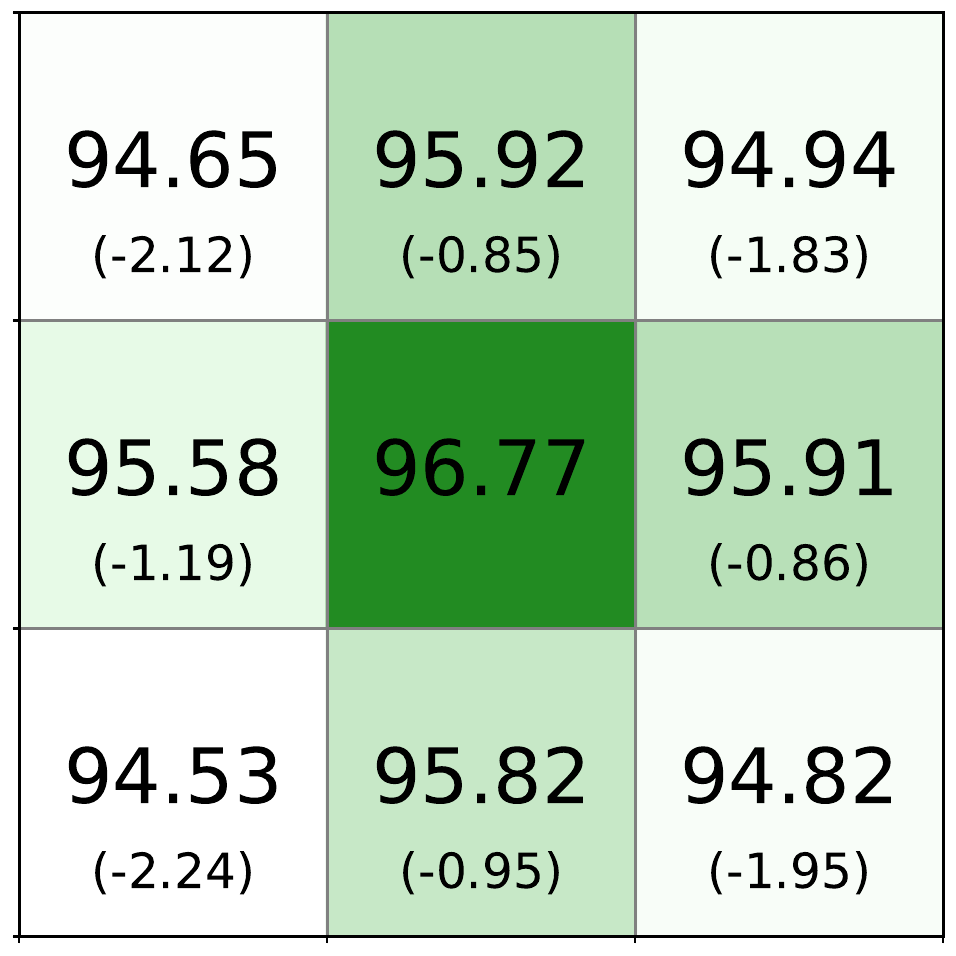}
    }
    \caption{\scriptsize{ImageNet-ReaL}}
    \label{fig:main_3x3_REAL_1}
\end{subfigure}
\hfill
\begin{subfigure}{0.1925\columnwidth}
    \resizebox{\columnwidth}{!}{
    \includegraphics[]{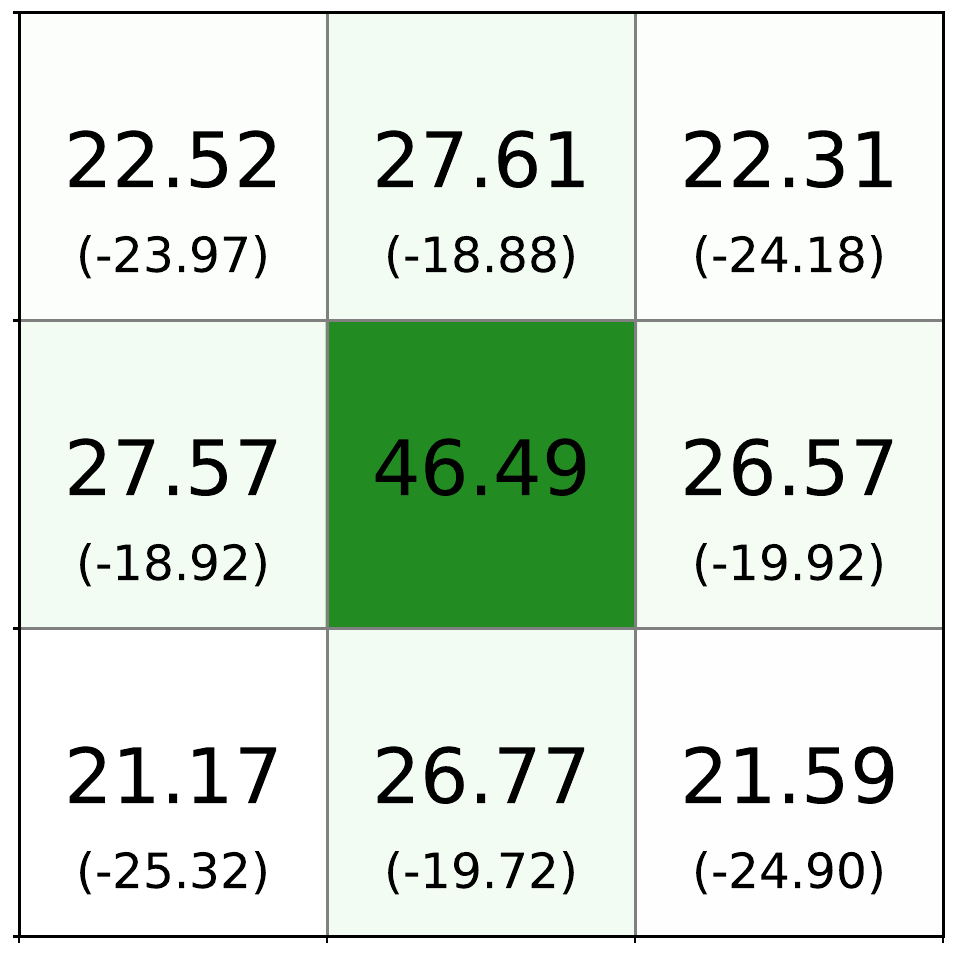}
    }
    \caption{\scriptsize{ImageNet-A}}
    \label{fig:main_3_3_a}
\end{subfigure}
\hfill
\begin{subfigure}{0.1925\columnwidth}
    \resizebox{\columnwidth}{!}{
       \includegraphics[]{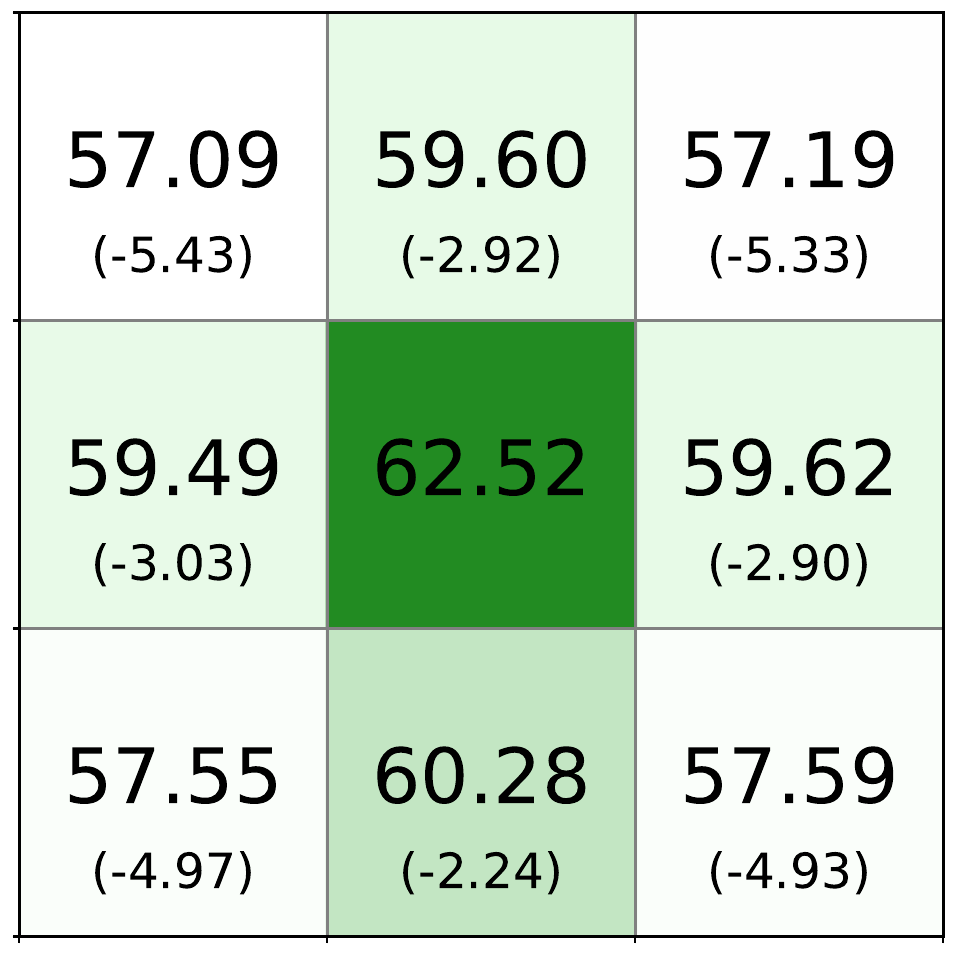}
}
    \caption{\scriptsize{ImageNet-R}}
    \label{fig:main_3_3_r}
\end{subfigure}
\hfill
\begin{subfigure}{0.1925\columnwidth}
    \resizebox{\columnwidth}{!}{
    \includegraphics[]{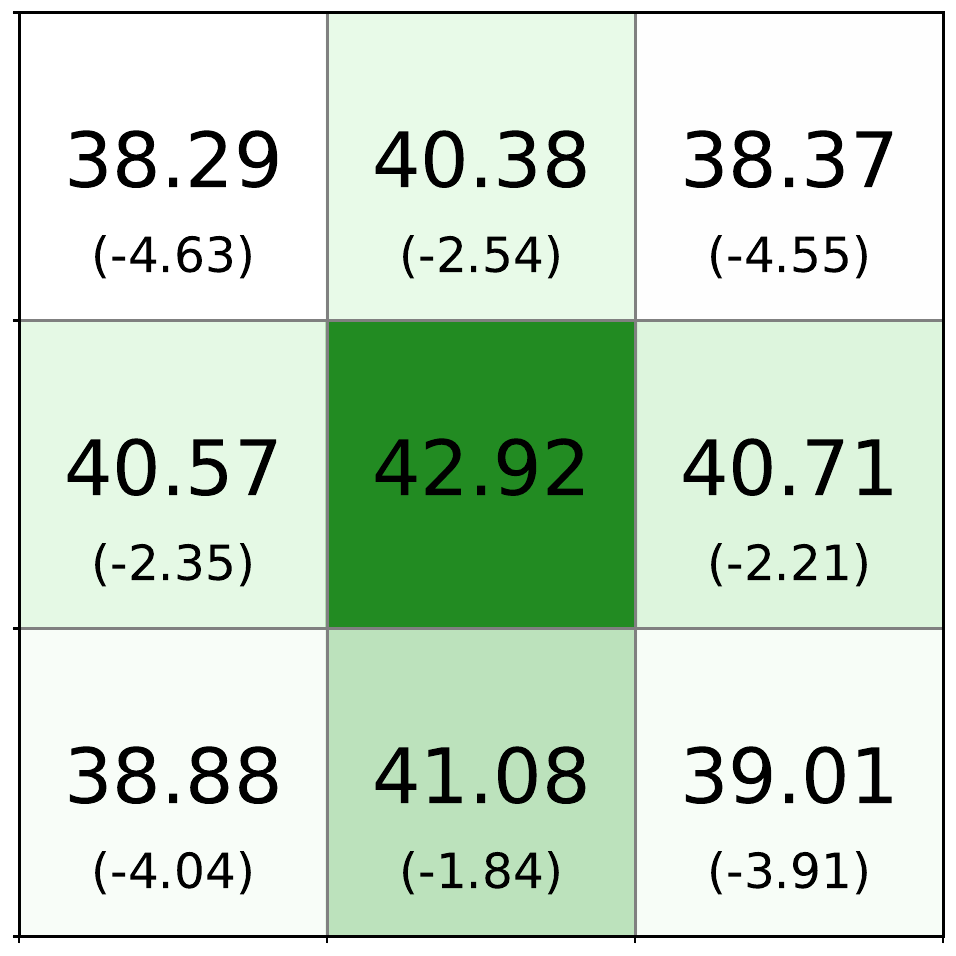}}
    \caption{\scriptsize{ImageNet-Sketch}}
    \label{fig:main_3_3_sketch}
\end{subfigure}
\hfill
\begin{subfigure}{0.1925\columnwidth}
\resizebox{\columnwidth}{!}{
     \includegraphics[]{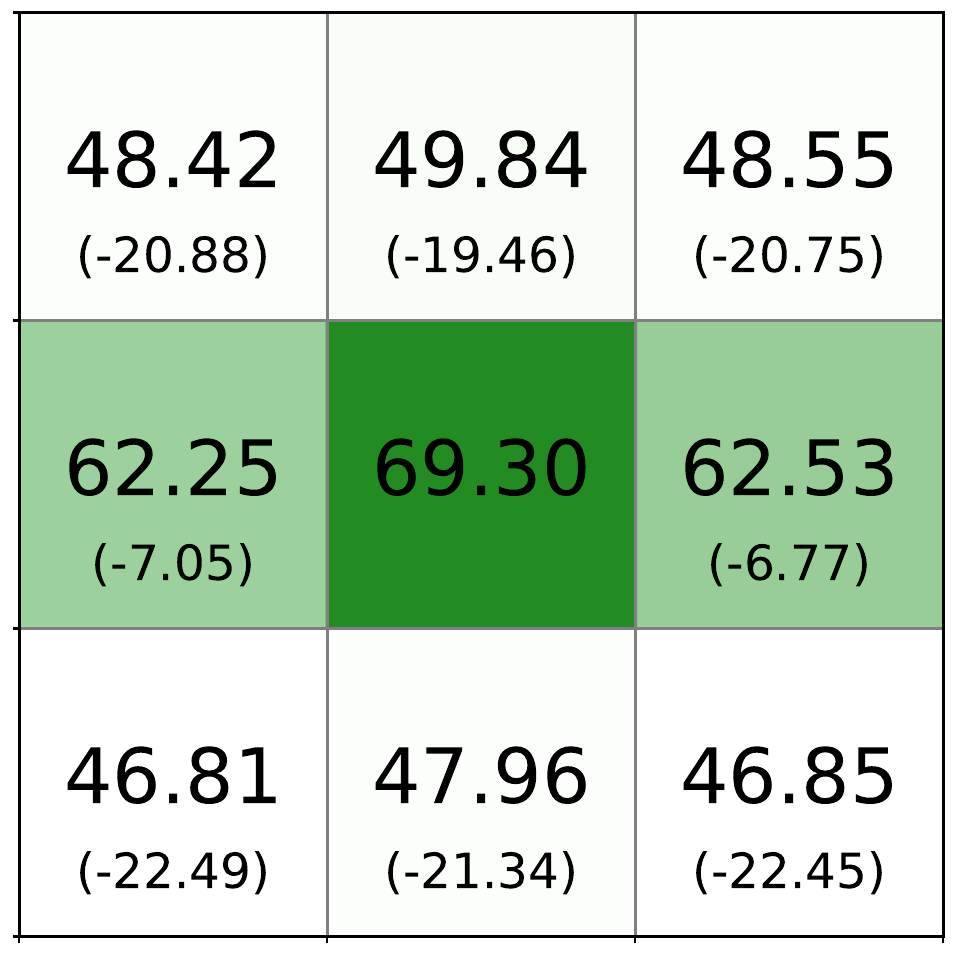}
    }
    \caption{\scriptsize{ObjectNet}}
    \label{fig:main_3_3_real}
\end{subfigure}
\caption{
    Upper-bound accuracy (\%) of ResNet-50 at each of the $9$ zoom locations. 
    The large gaps between the center and eight off-center locations on IN-A and ON demonstrate a center bias, which is 
    much smaller in IN (\cref{supp:anchor_based_analys}) and IN-R (b).
    \rebuttal{The values in parentheses indicate the delta with respect to the center crop.}
    }
\label{fig:anchor_based_analysis_all}
\vspace*{-0.5cm}
\end{figure}

\subsec{Experiment}
For each image, we have $9$ anchors (\cref{fig:better_overview}) and the originally $K=36$  zoomed versions per anchor as defined in \cref{sec:method}.
Yet, after reducing to the minimum set (\cref{sec:mincover}), $K$ averages at $25$, over all datasets, and $10 \leq K \leq 31$.
Here, we count the probability that the $K$ zoomed versions per anchor lead to a correct prediction.
In other words, we compute the upper-bound accuracy as in \cref{sec:maxpossibleaccuracy} but for each anchor separately.

\subsec{Results} First, as expected, the upper-bound accuracy for each anchor (\cref{fig:anchor_based_analysis_all}) is consistently lower than when all $9$ anchors are allowed (\cref{tab:1_main_results}c).
Second, across all $6$ datasets, the center anchor consistently achieves the highest upper-bound accuracy versus the other $8$ locations (\cref{fig:anchor_based_analysis_all} and \cref{supp:anchor_based_analys}), indicating a center bias in all datasets.
However, we find this bias is small in IN, IN-R, and IN-S but large in IN-A and ON (\ie the largest difference between center accuracy and the lowest off-center accuracy is around \decreasenoparent{25} and \decreasenoparent{23} points, respectively; whereas for other datasets, it is around \decrease{1} to \decrease{5} points, as shown in~\cref{fig:anchor_based_analysis_all}).

The center bias in ObjectNet can be explained by the fact that the images were captured using smartphones with aspect ratios of 2:3 or 9:16 (\cref{suppsec:imagenet_objectnet}).
Overall, such strong center bias in IN-A and ON may not be desirable since improvements on these two benchmarks may be attributed to learning to zoom to the center as opposed to the intended quest of recognizing objects in unusual forms (IN-A) or poses (ON).
By merely upscaling the image and center cropping, we can achieve higher accuracy using nearly all the same models on these two datasets~(\cref{suppfig:waterfall_imagenet_a,suppfig:waterfall_objectnet5k}).

We also find that, during test time, center-zooming (\cref{sec:clip_zoom_invariance}) increases the top-1 accuracy of all \textcolor{specialcolor}{IN-trained models} but not CLIP, even on IN-A and ON images. 
This observation is intriguing considering these OOD datasets contain more distracting objects than ImageNet images (\cref{sec:dataset_statistics}) and therefore, center-zooming \emph{should} de-clutter the scene for more accurate classification.
However, CLIP prefers a specific zoom scale that provides sufficient background for object recognition—it struggles to identify a single object in a tightly-cropped image \cite{zhong2022regionclip}. Future research should examine whether this ``zoom bias'' of CLIP is due to its image- or text-encoder, or both.


\subsection{Test-time augmentation of MEMO with \emph{only} zoom-in transforms improves accuracy}
\label{sec:memo_zoom}

Aggregating model predictions over zoom-in versions of the input image during test time leads to higher top-1 accuracy on IN, IN-ReaL, IN-A and ON, but lower accuracy on IN-R and IN-S (\cref{supsec:naive_aggregation}).
However, interestingly, always zooming out on IN-R and IN-S abstract images also hurts accuracy, suggesting that an adaptive zooming strategy might be a better approach.

\rebuttal{
Here, we test building such an adaptive test-time zooming strategy by modifying MEMO \cite{zhang2021memo}, a SOTA test-time augmentation method that finetunes a pre-trained classifier at \emph{test} time to achieve a more accurate prediction. Specifically, MEMO finds a network that produces a low-entropy predicted label over a set of $K=16$ augmented versions of the test image $I$  and then runs this finetuned model on $I$ again to produce the final prediction. It does this by applying different augmentations to the test point $I$ to get augmented points $I_1, \ldots, I_K$, passing these through the model to obtain predictive distributions, and updating the model parameters by minimizing the entropy of the averaged marginal distribution over predictions. While improving accuracy, MEMO requires a pre-defined set of diverse augmentation transforms (\eg sheer, rotate, and solarize in AugMix~\cite{hendrycks2021many}). Yet, the hyperparameters for each type of transform are hard-coded, and the contribution of each transform to improved classification accuracy is unknown.}




We improve MEMO's accuracy and interpretability by replacing AugMix transforms with only zoom-in functions.
Intuitively, a model first looks at all zoomed-in frames of the input image (at different zoom scales and locations) and then decides to achieve the most confident prediction.

\subsec{Experiment}
MEMO relies on AugMix~\cite{hendrycks2019augmix}, which applies a set of 13 image transforms, such as translation, rotation, and color distortion, to an original image at varying intensities, and then \emph{chains} them together to create $K = 16$ new augmented images (examples in \cref{suppviz:augmix-vs_rrc}).

We replace AugMix with \texttt{RandomResizedCrop} \cite{randomresizecrop} (\rrc), which takes a random crop of the input image (\ie at a random location, random rectangular area, and a random aspect ratio) and then resizes it to the fixed 224$\times$224 (\ie the network input size).
\rrc basically implements a random zoom-in function (examples in \cref{suppviz:augmix-vs_rrc}).

We compare the original MEMO~\cite{zhang2021memo} (which uses AugMix) and our version that uses \rrc on five benchmarks (IN, IN-A, IN-R, IN-S, and ON).
We follow the same experimental setup as in~\cite{zhang2021memo} (\eg $K = 16$). 
Specifically, we test three ResNet-50 variants that were pre-trained using distinct augmentation techniques.\footnote{The ResNet-50 model used as a baseline in this~\cref{sec:memo_zoom} is different from that in our other (non-MEMO) sections of the paper.}

\rebuttal{
We utilize Grad-Cam~\cite{Ramprasaath2020GradCAM} to understand the impact of MIMO on the network's attentions within the final layer, both before and after modification. 
Specifically, our investigation seeks test our hypothesis concerning the model's focus on the regions of interest within an image.
}


\subsec{Results} 
Both our MEMO + \rrc and the original MEMO + AugMix \cite{zhang2021memo} consistently outperform the baseline models, which do not use MEMO, on all five datasets (\cref{tab:memo_augmix_randomcropresize}).
That is, when combined with MEMO, zoom-in transforms implemented via \rrc are also helpful in classifying IN-S and IN-R images---where we previously find zoom-in to \emph{not} help in \layer{mean}/\layer{max} aggregation (\cref{supsec:naive_aggregation}).

On average, over all three models and five datasets, our \rrc outperforms AugMix by \increasenoparent{$0.28$} points, with a larger impact on IN-A, where it achieves a mean improvement of \increasenoparent{$1.10$} points (\cref{tab:memo_augmix_randomcropresize}).
Our results show that zoom-in alone can be a useful inductive bias, helping improve downstream image classification.
In contrast, some of the transformations among the 13 transform functions in AugMix may not be essential to the results of Zhang et al.~\cite{zhang2021memo} (no ablation studies of transformations were provided in \cite{zhang2021memo}) and are less effective than our zoom-in.

\begin{table}[ht]
\centering
\caption{
MEMO + \texttt{RRC} (\ie random zoom-in transforms) \textbf{outperforms} baselines and MEMO \cite{zhang2021memo}. 
}
\label{tab:memo_augmix_randomcropresize}
\resizebox{\textwidth}{!}{%
\begin{tabular}{@{}lcrrrr|crrrr|crrrr@{}}
\toprule
 & \multicolumn{5}{c|}{Baseline (1-crop)} & \multicolumn{5}{c|}{MEMO + AugMix \cite{zhang2021memo}} & \multicolumn{5}{c}{MEMO + \texttt{RRC} (\textbf{Ours})} \\ \cmidrule(l){2-16} 
 & IN & \multicolumn{1}{c}{IN-A} & \multicolumn{1}{c}{IN-R} & \multicolumn{1}{c}{IN-S} & \multicolumn{1}{c|}{ON} & IN & \multicolumn{1}{c}{IN-A} & \multicolumn{1}{c}{IN-R} & \multicolumn{1}{c}{IN-S} & \multicolumn{1}{c|}{ON} & IN & \multicolumn{1}{c}{IN-A} & \multicolumn{1}{c}{IN-R} & \multicolumn{1}{c}{IN-S} & \multicolumn{1}{c}{ON} \\ \cmidrule(l){2-16} 
ResNet-50~\cite{he2016deep} & \multicolumn{1}{r}{76.13} & 0.00 & 36.17 & 24.09 & \multicolumn{1}{r|}{35.92} & \multicolumn{1}{r}{77.27} & 0.83 & \textbf{41.28} & \textbf{27.63} & \multicolumn{1}{r|}{38.38} & \multicolumn{1}{r}{\textbf{77.50}} & \textbf{1.31} & 40.81 & 27.53 & \textbf{38.85} \\
DeepAug+AugMix~\cite{hendrycks2021many} & \multicolumn{1}{r}{75.82} & 3.87 & 46.77 & 32.62 & \multicolumn{1}{r|}{34.81} & \multicolumn{1}{r}{76.27} & 5.35 & 50.79 & \textbf{35.70} & \multicolumn{1}{r|}{36.42} & \multicolumn{1}{r}{\textbf{76.38}} & \textbf{5.76} & \textbf{50.88} & 35.65 & \textbf{36.64} \\
MoEx+CutMix~\cite{li2021feature}  & \multicolumn{1}{r}{79.04} & 7.97 & 35.52 & 23.96 & \multicolumn{1}{r|}{38.59} & \multicolumn{1}{r}{79.38} & 11.21 & \textbf{40.65} & \textbf{27.07} & \multicolumn{1}{r|}{40.62} & \multicolumn{1}{r}{\textbf{79.49}} & \textbf{13.61} & 40.41 & 26.80 & \textbf{41.43} \\ \cmidrule(l){2-16} 
\emph{mean} $\pm$ \emph{std} & \multicolumn{5}{c|}{36.75 $\pm$ 24.75} & \multicolumn{5}{c|}{39.26 \increase{2.51} $\pm$ 24.32} & \multicolumn{5}{c}{\textbf{39.54} \increase{2.79} $\pm$ \textbf{24.10}} \\ \bottomrule
\end{tabular}%
}
\end{table}

\begin{figure}[ht]
    \centering
    \begin{subfigure}[b]{1\textwidth}
        \includegraphics[width=0.329\textwidth]{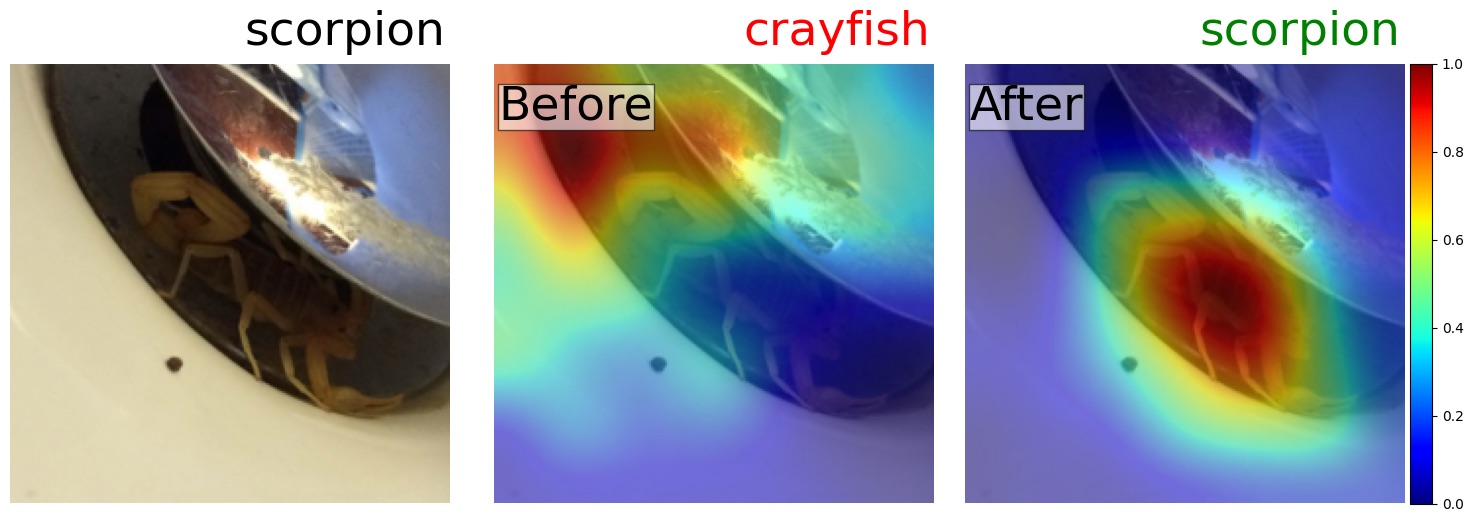}
        \includegraphics[width=0.329\textwidth]{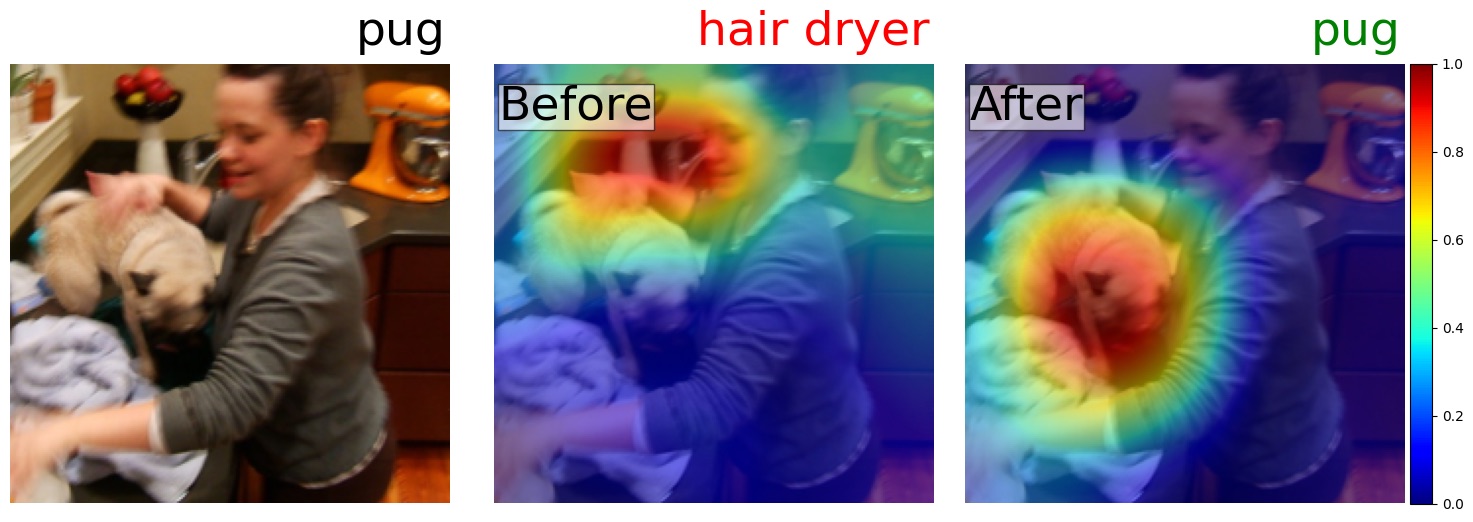}
        \includegraphics[width=0.329\textwidth]{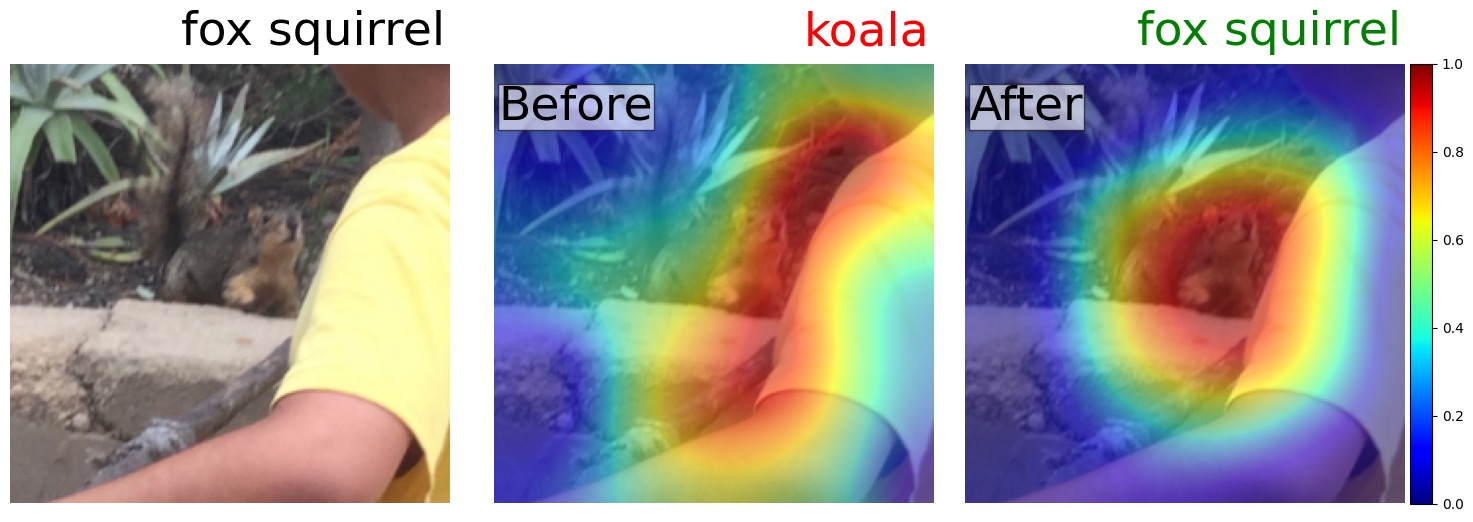}
    \end{subfigure}
    \caption{
        \rebuttal{Grad-CAM for the activation of the last convolutional layer of a ResNet-50 before and after the MEMO update suggests that the network attends to the object of interest after the update.}
    }
    \label{fig:gradcam_memo}
\end{figure}

\rebuttal{
Figure~\ref{fig:gradcam_memo} shows Grad-CAM visualizations for three samples, providing evidence of how the network's behavior changes before and after the MEMO update.
For an image of a \class{pug}, the network initially focused on a kitchen appliance, failing to detect the object correctly. After applying the MEMO modification, it refocused on the dog, classifying it accurately.
Similarly, in an image of a \class{fox squirrel}, the network initially had a diffuse focus but refocused on the fox squirrel after the update.
These results demonstrate the effectiveness of the MEMO modification in guiding the network's attention or encouraging the model to perform an implicit zoom on the regions of interest, thereby improving its classification performance.
}

\subsection{\rebuttal{ImageNet-Hard: A benchmark with images that remain unclassifiable, even after 324 zoom attempts}}
\label{sec:imagenet_hard}

Existing ImageNet-scale benchmarks followed one of the following three construction methods: 
(1) perturbing real images with the aim of making them harder for models to classify (\eg, ImageNet-C \cite{hendrycks2019benchmarking} and DAmageNet \cite{chen2019damagenet});
(2) collecting the real images that models misclassify (\eg, IN-A, ImageNet-O \cite{hendrycks2021natural}); or 
(3) setting up a highly-controlled data collection process (\eg, IN-S and ON).
Yet, none of such benchmarks explicitly challenge models on the ability to recognize a well-framed object in an image (\ie, no zooming required).
For example, ON is supposed to test the recognition of objects in unusual poses but the cluttered background in ON images is actually a major reason for misclassification (\cref{sec:center_bias}).
Furthermore, the results in \cref{tab:1_main_results} suggest that given optimal zooming, these existing benchmarks only challenge \textcolor{specialcolor}{IN-trained models} but not the Internet-scale vision-language models (\eg \clip) anymore. 
We propose ImageNet-Hard, a novel ImageNet-scale benchmark that challenges existing and future SOTA models.
ImageNet-Hard is a collection of images that the SOTA \clip-ViT-L/14~fails to correctly classify even when 324 zooming attempts are provided.

\subsubsection{ImageNet-Hard construction}

\subsec{Initial data collection}
We take \clip-ViT-L/14 (the highest-performing model in \cref{tab:1_main_results}) and run the zooming procedure to find ``Unclassifiable images'' (defined in \cref{sec:maxpossibleaccuracy}) from the following six datasets: IN-V2 ~\cite{recht2019imagenet}, IN-ReaL, IN-A, IN-R, IN-S, and ON.
That is, for each image $x$, we generate 324 zoomed versions of $x$ and feed them into \clip-ViT-L/14.
We add $x$ to ImageNet-Hard only if none of the 324 versions are correctly classified.

\subsec{Adding ImageNet-C}
The original IN-C \cite{hendrycks2019benchmarking} are  the original IN images but center-cropped to 224 $\times$ 224 px, which significantly makes the classification task unnecessarily more ill-posed (\eg, by adding Gaussian noise to a crop where the main object is already removed).

To find a subset of IN-C images for adding into ImageNet-Hard, we first re-generate ImageNet-C by adding the 19 types of corruption noise to IN without resizing the original IN images.
Second, we run \clip-ViT-L/14 on all 19 corruption types and manually select a subset of six diverse and lowest-accuracy corruption groups: {Impulse Noise}, {Frost}, {Fog}, {Snow}, {Brightness}, and {Zoom Blur}.
We repeat the initial data collection process for these 6 image sets of IN-C.

\subsec{Groundtruth labels} After the above procedure, our dataset contains 13,925 images collected from IN+ReaL, IN-V2, IN-A, IN-C, IN-R, IN-S, and ON
(see the distribution in \cref{supp:imagenet_hard_section_dist}).
ImageNet-Hard presents a 1000-way classification task where the 1000 classes are from ImageNet.
We manually inspect all images and remove 295 samples that are obviously ill-posed (e.g. an entirely black image but labeled \class{great~white~shark} in IN-S~\cref{fig:blcakimages_sketch}), arriving at a total of 13,630 ImageNet-Hard images.
A sample contains only one groundtruth label from its original datasets except for IN and IN-C images, which have a set of IN+ReaL labels.
Each IN or IN-C image is considered correctly labeled by a model if its top-1 predicted label is among the groundtruth labels.

\subsec{Refining groundtruth labels via human feedback}
Label noise is still present in IN and OOD benchmarks despite cleaning efforts \cite{beyer2020we,recht2019imagenet,yun2021re}.
Since ImageNet-Hard contains images misclassified by \clip-ViT-L/14, our manual inspection confirms many misclassified images have debatable labels.

To ameliorate the issue, we orchestrate a human feedback study for eliminating images with inaccurate labels.
First, the first author examine every image and flag 3,133 images as ambiguous and needs verification.
Then, we have two groups of annotators to help verify the labels (by choosing Accept, Reject, or Not Sure).
Group A is composed of three students, each examine all 3,133 images where Group B is composed of 38 students, each examine 50 randomly-selected images.
Our inter-annotator aggregation procedure merges labels from both groups and results in 2,280 images removed (out of 3,133 originally flagged), leaving ImageNet-Hard at a total of 11,350 images.

That is, we accept an image $x$ if one of the two conditions is satisfied: (1) when all 3/3 group-A annotators accept $x$; or (2) when 2/3 group-A annotators accept $x$ and all group-B reviewers of $x$ accept $x$ (assuming at least 1 group-B annotator reviews $x$; otherwise $x$ will be rejected).


Inspired by IN-ReaL \cite{beyer2020we}, we further clean up the labels by eliminating 370 images associated with the labels \class{sunglass}, \class{sunglasses}, \class{tub}, \class{bathtub}, \class{cradle}, \class{bassinet}, \class{projectile}, and \class{missile}, \ie, the classes that often contain similar images that belong to more than one class.
After this refinement, the final ImageNet-Hard dataset contains a total of \textbf{10,980 images}.


\subsec{4K version} 
We utilize GigaGAN~\cite{kang2023scaling} to upscale every image in our final dataset and construct ImageNet-Hard-4K, which is aimed to facilitate future research into how a super-resolution step may improve image classification results (\eg, to classify an object when the image is blurry).


\subsec{Release}
\label{sec:imagenet_hard_release}
ImageNet-Hard and ImageNet-Hard-4K are released on \href{https://huggingface.co/datasets/taesiri/imagenet-hard}{HuggingFace} (see samples in \cref{fig:imagenet_hard_sample_images}) under MIT License.
Code for evaluating models on ImageNet-Hard is on \href{https://github.com/taesiri/ZoomIsAllYouNeed}{GitHub}.


\subsubsection{ImageNet-Hard challenges SOTA classifiers, especially those operating at 224$\times$224}
\label{sec:imagenet-hard}

Here, we evaluate the standard 1-crop, top-1 accuracy of SOTA classifiers on ImageNet-Hard. 
We use the image pre-processing function defined by each classifier.
In addition to the 6 models in \cref{sec:maxpossibleaccuracy}, we also test CLIP-ViT-L/14@336px~\cite{radford2021learning}, EfficientNet (B0@224px and B7@600px)~\cite{tan2019efficientnet}, and EfficientNet-L2@800px~\cite{EfficientNetL2_model}. 
CLIP-ViT-L/14@336px, EfficientNet-B7@600px, and EfficientNet-L2@800px are state-of-the-art models that operate at high resolutions of 336$\times$336, 600$\times$600, and 800$\times$800 respectively.
In addition, our evaluation includes models from the OpenCLIP family~\cite{ilharco_gabriel_2021_5143773}.

\subsec{Results}
\cref{tab:imagenet_hard_performance} shows fairly low top-1 accuracy by various classifiers on ImageNet-Hard.
First, all well-known \textcolor{specialcolor}{IN-trained} classifiers that operate at 224$\times$224 perform poorly between 7.34\% (AlexNet) and 18.52\% accuracy (ViT-B/32).


Since ImageNet-Hard is based on a collection of images that OpenAI's CLIP ViT-L/14@224px mislabels, this classifier's accuracy on our dataset is only 1.86\%.
Yet, interestingly, CLIP-ViT-L/14@336px also performs poorly at 2.02\% (\cref{tab:imagenet_hard_performance}).
Furthermore, all 68 tested OpenCLIP models perform poorly, with an accuracy below 16\% (see details in~\cref{supp:imagenet_hard_openclip}).


Separately, we observe a trend that models operating at a higher resolution tend to perform better on ImageNet-Hard with EfficientNet-L2@800px scoring highest at 39.00\% (compared to 88.40\%~\cite{xie2020self} on the original ImageNet).
Overall, all models perform substantially worse on ImageNet-Hard (\cref{tab:imagenet_hard_performance}) than on other ImageNet-scale datasets (see \cref{supptab:aggregation_methods_extended}; 1-crop).
This result is expected because ImageNet-Hard is a set of hard cases collected from those OOD benchmarks.

\subsec{ImageNet-Hard-4K}
We find that when upsampling images to 4K using GigaGAN~\cite{kang2023scaling} and downsampling them back to the resolution of each classifier does not help but even hurt the accuracy slightly (\cref{tab:imagenet_hard4k_perforamnce}).
Given that GigaGAN performs remarkably well, this result suggests ImageNet-Hard is different from typical fine-grained animal classification where improving the texture details increases classification accuracy~\cite{wei2021fine}.
The next section (\cref{sec:analysis_errors}) sheds light on model failures on ImageNet-Hard, revealing challenges posed to future SOTA models.

\begin{table}[ht]
\centering
\caption{Top-1 accuracy (\%) on ImageNet-Hard of \textcolor{specialcolor}{IN-trained models} and those trained on larger, non-ImageNet datasets (black). 
All models operate at 224$\times$224 unless otherwise specify.
}
\label{tab:imagenet_hard_performance}
\resizebox{0.9\textwidth}{!}{
    \begin{tabular}{@{}lrlrlr@{}}
    \toprule
    \multicolumn{1}{l}{Classifier} & \multicolumn{1}{c}{{Accuracy}} & \multicolumn{1}{l}{{Classifier}} & \multicolumn{1}{c}{{Accuracy}} & \multicolumn{1}{l}{{Classifier}} & \multicolumn{1}{c}{{Accuracy}} \\ \midrule
    \textcolor{specialcolor}{AlexNet} & 7.34  & \textcolor{specialcolor}{ViT-B/32} & 18.52 & \clip-ViT-L/14@224px & 1.86 \\
    \textcolor{specialcolor}{VGG-16} & 12.00 & \textcolor{specialcolor}{EfficientNet-B0@224px} & 16.57 & \clip-ViT-L/14@336px & 2.02 \\
    \textcolor{specialcolor}{ResNet-18} & 10.86 & \textcolor{specialcolor}{EfficientNet-B7@600px} & 23.20 & OpenCLIP-ViT-bigG-14 & 15.93 \\
    \textcolor{specialcolor}{ResNet-50} & 14.74 & EfficientNet-L2@800px & 39.00 & OpenCLIP-ViT-L-14 & 15.60 \\ \bottomrule
    \end{tabular}%
}
\end{table}

\subsection{Analysis of Image Classification Errors}
\label{sec:analysis_errors}

Motivated by the fact that EfficientNet-L2 is the best classifier on ImageNet-Hard, we qualitatively analyze its failure cases to characterize the challenge posed by our benchmark.
Specifically, we provide \texttt{gpt-3.5-turbo}~\cite{openai2023gpt35turbo} with a pair of EfficientNet-L2's top-1 (incorrect) label and the groundtruth label and ask it to categorize the error into ``common'' or ``rare'' based on the labels' semantic similarity (see \cref{supp:analysis_of_gpt} for full details).
For instance, mislabeling \class{bucket} into \class{barrel} is common (as two objects are quite related) while mislabeling \class{cloak} into \class{jigsaw puzzle} is rare.

\begin{figure}[h]
    \centering
    \begin{subfigure}[b]{0.495\textwidth}
        \includegraphics[width=\textwidth]{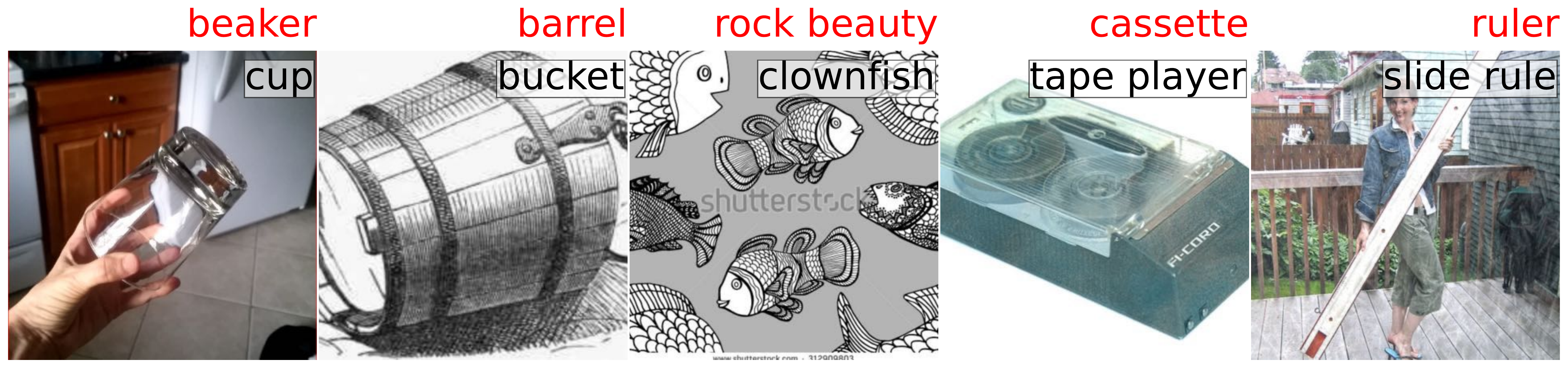}
        \caption{Common}
        \label{fig:commonmisclassification}
    \end{subfigure}
    \hfill
    \begin{subfigure}[b]{0.495\textwidth}
        \includegraphics[width=\textwidth]{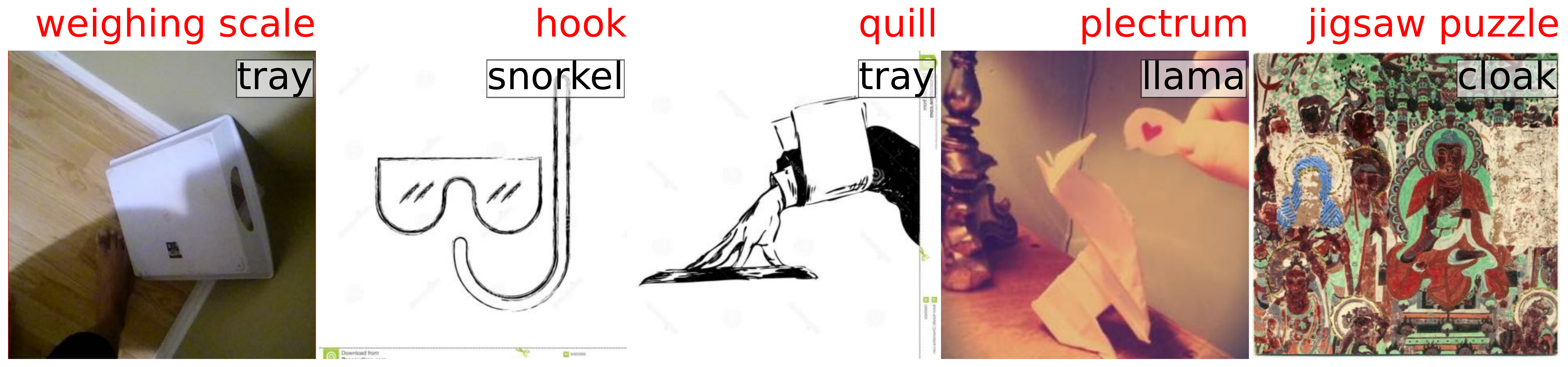}
        \caption{Rare}
        \label{fig:raremisclassifications}
    \end{subfigure}
    \caption{
    ImageNet-Hard samples \textcolor{red}{misclassified} by EfficientNet-L2@800px can be categorized into two groups: (a) \textbf{Common}: the top-1 label is related to the groundtruth label; and (b) \textbf{Rare}: the top-1 label is semantically far from the groundtruth label.
    See \cref{suppfig:commonmisclassification,suppfig:raremisclassifications} for more samples.
    }
    \label{fig:images}
\end{figure}

\subsec{Results} See \cref{supp:analysis_confusing_classes} for samples of wrong labels that EfficientNet-L2 most frequently misclassifies into.
We find that 39.4\% of EfficientNet-L2's misclassifications on the ImageNet-Hard dataset are ``common'', while 60.6\% are ``rare''. 

\textbf{A. Common} group captures model confusion between two related classes (e.g. two fish species: \class{clownfish} and \class{rock beauty}; \cref{fig:commonmisclassification}).
Yet, another source of problem for these ``errors'' is the debatable groundtruth labels, which may require domain-expert annotators to verify and rectify~\cite{luccioni2022bugs}.


\textbf{B. Rare} group captures errors where the model confusion is between two semantically distant classes (\eg, \class{llama} $\to$ \class{plectrum}; \cref{fig:raremisclassifications}).
This often happens with abstract images or objects in unusual poses \cite{alcorn2019strike} or forms \cite{de2019does}.
Classifying this group of images is challenging and sometimes requires a strong understanding of context and reasoning capabilities.






\section{Discussion and Conclusion}

\subsec{Limitations} By manual inspection, we estimate 14.7\% of labeling noise, which ImageNet-Hard inherits from the source datasets.

Our study rigorously analyzed the zooming effect on six known classifiers and image classification benchmarks.
We first demonstrate that previous state-of-the-art classifiers, as old as AlexNet~\cite{krizhevsky2012imagenet}, could potentially achieve near $90\%$ accuracy with optimal zooming.
This sparks the intriguing question of whether image classifiers' evolution over the past ten years is about mastering where and at what scale to zoom (instead of enhancing feature extractors, a.k.a. representation learning \cite{iclr}).
\rebuttal{Through another lens, we probe the evolution by analyzing the implicit zooming mechanisms that deep classifiers apply to input images. This perspective diverges from \cite{raghu2021vision}, which studied the progression of representation learning from CNNs to ViTs.}


We are the first to document the spatial biases of existing benchmarks.
Notably, IN-A and ON contain a large center bias and simply zooming to the center will de-clutter the scene and yield a high accuracy ($24.69\%$ for ViT-B/32 on IN-A; \cref{tab:aggregation_main_text}), which is competitive with state-of-the-art trained models (\eg $24.1\%$ of Robust ViT \cite{chefer2022optimizing}) and much higher than state-of-the-art TTA techniques (\eg $11.21\%$ of MEMO \cite{zhang2021memo}; \cref{tab:memo_augmix_randomcropresize}).
Our simple, but strong zoom-in baselines on IN-A and ON motivate future research into better-controlled benchmarks that more explicitly test models on a set of pre-defined properties.
Our proposed TTA method with zoom-in transforms (MEMO + \rrc) is not only more accurate but also more interpretable and faster to run (\cref{supptab:memo_runtime}) than the original MEMO.

Finally, we introduce ImageNet-Hard (\cref{sec:imagenet_hard}), a new challenging dataset for SOTA IN-trained and vision-language classifiers. 

\section*{Acknowledgement}
GN is supported by Auburn University PGRF Fellowship. AN was supported by a NSF CAREER award No. 2145767,
and donations from NaphCare Foundation, and Adobe Research.
We greatly appreciate David Seunghyun Yoon's help in generating the ImageNet-Hard-4K (\cref{sec:imagenet-hard}) version using GigaGAN.
We also thank those students at Auburn University and Alberta University who participated in our experiment for cleaning up the labels of ImageNet-Hard images.

\clearpage
{\small
\bibliographystyle{icml2021}
\bibliography{references.bib}

\begin{thebibliography}{86}
\providecommand{\natexlab}[1]{#1}
\providecommand{\url}[1]{\texttt{#1}}
\expandafter\ifx\csname urlstyle\endcsname\relax
  \providecommand{\doi}[1]{doi: #1}\else
  \providecommand{\doi}{doi: \begingroup \urlstyle{rm}\Url}\fi

\bibitem[Eff()]{EfficientNetL2_model}
timm/tf\_efficientnet\_l2.ns\_jft\_in1k\_475 · hugging face.
\newblock \url{https://huggingface.co/timm/tf_efficientnet_l2.ns_jft_in1k_475}.
\newblock (Accessed on 06/05/2023).

\bibitem[cen()]{center-crop}
Centercrop — torchvision main documentation.
\newblock
  \url{https://pytorch.org/vision/main/generated/torchvision.transforms.CenterCrop.html}.
\newblock (Accessed on 03/07/2023).

\bibitem[icl()]{iclr}
Iclr 2023.
\newblock \url{https://iclr.cc/}.
\newblock (Accessed on 11/10/2022).

\bibitem[ran()]{randomresizecrop}
Randomresizedcrop — torchvision main documentation.
\newblock
  \url{https://pytorch.org/vision/main/generated/torchvision.transforms.RandomResizedCrop.html}.
\newblock (Accessed on 03/08/2023).

\bibitem[Alcorn et~al.(2019)Alcorn, Li, Gong, Wang, Mai, Ku, and
  Nguyen]{alcorn2019strike}
Alcorn, M.~A., Li, Q., Gong, Z., Wang, C., Mai, L., Ku, W.-S., and Nguyen, A.
\newblock Strike (with) a pose: Neural networks are easily fooled by strange
  poses of familiar objects.
\newblock In \emph{Proceedings of the IEEE/CVF conference on computer vision
  and pattern recognition}, pp.\  4845--4854, 2019.

\bibitem[Ayhan \& Berens(2018)Ayhan and Berens]{ayhan2018test}
Ayhan, M.~S. and Berens, P.
\newblock Test-time data augmentation for estimation of heteroscedastic
  aleatoric uncertainty in deep neural networks.
\newblock In \emph{Medical Imaging with Deep Learning}, 2018.

\bibitem[Bahat \& Shakhnarovich(2020)Bahat and
  Shakhnarovich]{bahat2020classification}
Bahat, Y. and Shakhnarovich, G.
\newblock Classification confidence estimation with test-time
  data-augmentation.
\newblock \emph{arXiv e-prints}, pp.\  arXiv--2006, 2020.

\bibitem[Barbu et~al.(2019)Barbu, Mayo, Alverio, Luo, Wang, Gutfreund,
  Tenenbaum, and Katz]{barbu2019objectnet}
Barbu, A., Mayo, D., Alverio, J., Luo, W., Wang, C., Gutfreund, D., Tenenbaum,
  J., and Katz, B.
\newblock Objectnet: A large-scale bias-controlled dataset for pushing the
  limits of object recognition models.
\newblock \emph{Advances in neural information processing systems}, 32, 2019.

\bibitem[Bau et~al.(2017)Bau, Zhou, Khosla, Oliva, and
  Torralba]{bau2017network}
Bau, D., Zhou, B., Khosla, A., Oliva, A., and Torralba, A.
\newblock Network dissection: Quantifying interpretability of deep visual
  representations.
\newblock In \emph{Proceedings of the IEEE conference on computer vision and
  pattern recognition}, pp.\  6541--6549, 2017.

\bibitem[Beyer et~al.(2020)Beyer, H{\'e}naff, Kolesnikov, Zhai, and
  Oord]{beyer2020we}
Beyer, L., H{\'e}naff, O.~J., Kolesnikov, A., Zhai, X., and Oord, A. v.~d.
\newblock Are we done with imagenet?
\newblock \emph{arXiv preprint arXiv:2006.07159}, 2020.

\bibitem[Chefer et~al.(2022)Chefer, Schwartz, and Wolf]{chefer2022optimizing}
Chefer, H., Schwartz, I., and Wolf, L.
\newblock Optimizing relevance maps of vision transformers improves robustness.
\newblock In Oh, A.~H., Agarwal, A., Belgrave, D., and Cho, K. (eds.),
  \emph{Advances in Neural Information Processing Systems}, 2022.
\newblock URL \url{https://openreview.net/forum?id=upuYKQiyxa_}.

\bibitem[Chen et~al.(2020)Chen, Agarwal, and Nguyen]{chen2020shape}
Chen, P., Agarwal, C., and Nguyen, A.
\newblock The shape and simplicity biases of adversarially robust
  imagenet-trained cnns.
\newblock \emph{arXiv preprint arXiv:2006.09373}, 2020.

\bibitem[Chen et~al.(2019)Chen, Huang, He, and Sun]{chen2019damagenet}
Chen, S., Huang, X., He, Z., and Sun, C.
\newblock Damagenet: A universal adversarial dataset.
\newblock \emph{arXiv preprint arXiv:1912.07160}, 2019.

\bibitem[Chun et~al.(2022)Chun, Lee, and Kim]{chun2022cyclic}
Chun, S., Lee, J.~Y., and Kim, J.
\newblock Cyclic test time augmentation with entropy weight method.
\newblock In \emph{Uncertainty in Artificial Intelligence}, pp.\  433--442.
  PMLR, 2022.

\bibitem[De~Vries et~al.(2019)De~Vries, Misra, Wang, and Van~der
  Maaten]{de2019does}
De~Vries, T., Misra, I., Wang, C., and Van~der Maaten, L.
\newblock Does object recognition work for everyone?
\newblock In \emph{Proceedings of the IEEE/CVF conference on computer vision
  and pattern recognition workshops}, pp.\  52--59, 2019.

\bibitem[Donnelly et~al.(2022)Donnelly, Barnett, and
  Chen]{donnelly2022deformable}
Donnelly, J., Barnett, A.~J., and Chen, C.
\newblock Deformable protopnet: An interpretable image classifier using
  deformable prototypes.
\newblock In \emph{Proceedings of the IEEE/CVF Conference on Computer Vision
  and Pattern Recognition}, pp.\  10265--10275, 2022.

\bibitem[Dosovitskiy et~al.(2020)Dosovitskiy, Beyer, Kolesnikov, Weissenborn,
  Zhai, Unterthiner, Dehghani, Minderer, Heigold, Gelly,
  et~al.]{dosovitskiy2020image}
Dosovitskiy, A., Beyer, L., Kolesnikov, A., Weissenborn, D., Zhai, X.,
  Unterthiner, T., Dehghani, M., Minderer, M., Heigold, G., Gelly, S., et~al.
\newblock An image is worth 16x16 words: Transformers for image recognition at
  scale.
\newblock In \emph{International Conference on Learning Representations}, 2020.

\bibitem[Fu et~al.(2017)Fu, Zheng, and Mei]{fu2017look}
Fu, J., Zheng, H., and Mei, T.
\newblock Look closer to see better: Recurrent attention convolutional neural
  network for fine-grained image recognition.
\newblock In \emph{Proceedings of the IEEE conference on computer vision and
  pattern recognition}, pp.\  4438--4446, 2017.

\bibitem[Geirhos et~al.(2019)Geirhos, Rubisch, Michaelis, Bethge, Wichmann, and
  Brendel]{geirhos2018imagenettrained}
Geirhos, R., Rubisch, P., Michaelis, C., Bethge, M., Wichmann, F.~A., and
  Brendel, W.
\newblock Imagenet-trained {CNN}s are biased towards texture; increasing shape
  bias improves accuracy and robustness.
\newblock In \emph{International Conference on Learning Representations}, 2019.
\newblock URL \url{https://openreview.net/forum?id=Bygh9j09KX}.

\bibitem[Geirhos et~al.(2020)Geirhos, Jacobsen, Michaelis, Zemel, Brendel,
  Bethge, and Wichmann]{geirhos2020shortcut}
Geirhos, R., Jacobsen, J.-H., Michaelis, C., Zemel, R., Brendel, W., Bethge,
  M., and Wichmann, F.~A.
\newblock Shortcut learning in deep neural networks.
\newblock \emph{Nature Machine Intelligence}, 2\penalty0 (11):\penalty0
  665--673, 2020.

\bibitem[Goh et~al.(2021)Goh, Cammarata, Voss, Carter, Petrov, Schubert,
  Radford, and Olah]{goh2021multimodal}
Goh, G., Cammarata, N., Voss, C., Carter, S., Petrov, M., Schubert, L.,
  Radford, A., and Olah, C.
\newblock Multimodal neurons in artificial neural networks.
\newblock \emph{Distill}, 6\penalty0 (3):\penalty0 e30, 2021.

\bibitem[Gupta et~al.(2019)Gupta, Dollar, and Girshick]{gupta2019lvis}
Gupta, A., Dollar, P., and Girshick, R.
\newblock Lvis: A dataset for large vocabulary instance segmentation.
\newblock In \emph{Proceedings of the IEEE/CVF conference on computer vision
  and pattern recognition}, pp.\  5356--5364, 2019.

\bibitem[He et~al.(2016)He, Zhang, Ren, and Sun]{he2016deep}
He, K., Zhang, X., Ren, S., and Sun, J.
\newblock Deep residual learning for image recognition.
\newblock In \emph{Proceedings of the IEEE conference on computer vision and
  pattern recognition}, pp.\  770--778, 2016.

\bibitem[Hendrycks \& Dietterich(2019)Hendrycks and
  Dietterich]{hendrycks2019benchmarking}
Hendrycks, D. and Dietterich, T.
\newblock Benchmarking neural network robustness to common corruptions and
  perturbations.
\newblock \emph{arXiv preprint arXiv:1903.12261}, 2019.

\bibitem[Hendrycks et~al.(2019)Hendrycks, Mu, Cubuk, Zoph, Gilmer, and
  Lakshminarayanan]{hendrycks2019augmix}
Hendrycks, D., Mu, N., Cubuk, E.~D., Zoph, B., Gilmer, J., and
  Lakshminarayanan, B.
\newblock Augmix: A simple data processing method to improve robustness and
  uncertainty.
\newblock \emph{arXiv preprint arXiv:1912.02781}, 2019.

\bibitem[Hendrycks et~al.(2021{\natexlab{a}})Hendrycks, Basart, Mu, Kadavath,
  Wang, Dorundo, Desai, Zhu, Parajuli, Guo, et~al.]{hendrycks2021many}
Hendrycks, D., Basart, S., Mu, N., Kadavath, S., Wang, F., Dorundo, E., Desai,
  R., Zhu, T., Parajuli, S., Guo, M., et~al.
\newblock The many faces of robustness: A critical analysis of
  out-of-distribution generalization.
\newblock In \emph{Proceedings of the IEEE/CVF International Conference on
  Computer Vision}, pp.\  8340--8349, 2021{\natexlab{a}}.

\bibitem[Hendrycks et~al.(2021{\natexlab{b}})Hendrycks, Zhao, Basart,
  Steinhardt, and Song]{hendrycks2021natural}
Hendrycks, D., Zhao, K., Basart, S., Steinhardt, J., and Song, D.
\newblock Natural adversarial examples.
\newblock In \emph{Proceedings of the IEEE/CVF Conference on Computer Vision
  and Pattern Recognition}, pp.\  15262--15271, 2021{\natexlab{b}}.

\bibitem[Howard et~al.(2019)Howard, Sandler, Chu, Chen, Chen, Tan, Wang, Zhu,
  Pang, Vasudevan, et~al.]{howard2019searching}
Howard, A., Sandler, M., Chu, G., Chen, L.-C., Chen, B., Tan, M., Wang, W.,
  Zhu, Y., Pang, R., Vasudevan, V., et~al.
\newblock Searching for mobilenetv3.
\newblock In \emph{Proceedings of the IEEE/CVF international conference on
  computer vision}, pp.\  1314--1324, 2019.

\bibitem[Huang et~al.(2017)Huang, Liu, Van Der~Maaten, and
  Weinberger]{huang2017densely}
Huang, G., Liu, Z., Van Der~Maaten, L., and Weinberger, K.~Q.
\newblock Densely connected convolutional networks.
\newblock In \emph{Proceedings of the IEEE conference on computer vision and
  pattern recognition}, pp.\  4700--4708, 2017.

\bibitem[Ilharco et~al.(2021)Ilharco, Wortsman, Wightman, Gordon, Carlini,
  Taori, Dave, Shankar, Namkoong, Miller, Hajishirzi, Farhadi, and
  Schmidt]{ilharco_gabriel_2021_5143773}
Ilharco, G., Wortsman, M., Wightman, R., Gordon, C., Carlini, N., Taori, R.,
  Dave, A., Shankar, V., Namkoong, H., Miller, J., Hajishirzi, H., Farhadi, A.,
  and Schmidt, L.
\newblock Openclip, July 2021.
\newblock URL \url{https://doi.org/10.5281/zenodo.5143773}.
\newblock If you use this software, please cite it as below.

\bibitem[Jaderberg et~al.(2015)Jaderberg, Simonyan, Zisserman,
  et~al.]{jaderberg2015spatial}
Jaderberg, M., Simonyan, K., Zisserman, A., et~al.
\newblock Spatial transformer networks.
\newblock \emph{Advances in neural information processing systems}, 28, 2015.

\bibitem[Jin et~al.(2021)Jin, Tanno, Mertzanidou, Panagiotaki, and
  Alexander]{jin2021learning}
Jin, C., Tanno, R., Mertzanidou, T., Panagiotaki, E., and Alexander, D.~C.
\newblock Learning to downsample for segmentation of ultra-high resolution
  images.
\newblock \emph{arXiv preprint arXiv:2109.11071}, 2021.

\bibitem[Kang et~al.(2023)Kang, Zhu, Zhang, Park, Shechtman, Paris, and
  Park]{kang2023scaling}
Kang, M., Zhu, J.-Y., Zhang, R., Park, J., Shechtman, E., Paris, S., and Park,
  T.
\newblock Scaling up gans for text-to-image synthesis.
\newblock \emph{arXiv preprint arXiv:2303.05511}, 2, 2023.

\bibitem[Kim et~al.(2020)Kim, Kim, and Kim]{kim2020learning}
Kim, I., Kim, Y., and Kim, S.
\newblock Learning loss for test-time augmentation.
\newblock \emph{Advances in Neural Information Processing Systems},
  33:\penalty0 4163--4174, 2020.

\bibitem[Kleinberg \& Tardos(2005)Kleinberg and Tardos]{algo-design}
Kleinberg, J. and Tardos, E.
\newblock \emph{Algorithm Design}.
\newblock Addison-Wesley Longman Publishing Co., Inc., USA, 2005.
\newblock ISBN 0321295358.

\bibitem[Kong \& Henao(2022)Kong and Henao]{kong2022efficient}
Kong, F. and Henao, R.
\newblock Efficient classification of very large images with tiny objects.
\newblock In \emph{Proceedings of the IEEE/CVF Conference on Computer Vision
  and Pattern Recognition}, pp.\  2384--2394, 2022.

\bibitem[Krause et~al.(2015)Krause, Jin, Yang, and Fei-Fei]{krause2015fine}
Krause, J., Jin, H., Yang, J., and Fei-Fei, L.
\newblock Fine-grained recognition without part annotations.
\newblock In \emph{Proceedings of the IEEE conference on computer vision and
  pattern recognition}, pp.\  5546--5555, 2015.

\bibitem[Krizhevsky et~al.(2012)Krizhevsky, Sutskever, and
  Hinton]{krizhevsky2012imagenet}
Krizhevsky, A., Sutskever, I., and Hinton, G.~E.
\newblock Imagenet classification with deep convolutional neural networks.
\newblock \emph{Advances in neural information processing systems}, 25, 2012.

\bibitem[Leung et~al.(2021)Leung, Ho, Persekian, Orozco, Chang, Sandstrom, Liu,
  and Vasconcelos]{leung2021oowl500}
Leung, B., Ho, C.-H., Persekian, A., Orozco, D., Chang, Y., Sandstrom, E., Liu,
  B., and Vasconcelos, N.
\newblock Oowl500: Overcoming dataset collection bias in the wild.
\newblock \emph{arXiv preprint arXiv:2108.10992}, 2021.

\bibitem[Li et~al.(2021{\natexlab{a}})Li, Wu, Lim, Belongie, and
  Weinberger]{li2021feature}
Li, B., Wu, F., Lim, S.-N., Belongie, S., and Weinberger, K.~Q.
\newblock On feature normalization and data augmentation.
\newblock In \emph{Proceedings of the IEEE/CVF Conference on Computer Vision
  and Pattern Recognition}, pp.\  12383--12392, 2021{\natexlab{a}}.

\bibitem[Li et~al.(2021{\natexlab{b}})Li, Xiong, and Hoi]{li2021learning}
Li, J., Xiong, C., and Hoi, S.~C.
\newblock Learning from noisy data with robust representation learning.
\newblock In \emph{Proceedings of the IEEE/CVF International Conference on
  Computer Vision}, pp.\  9485--9494, 2021{\natexlab{b}}.

\bibitem[Li et~al.(2021{\natexlab{c}})Li, Li, Dai, Shi, Zhu, and
  Hu]{li2021rethinking}
Li, X., Li, J., Dai, T., Shi, J., Zhu, J., and Hu, X.
\newblock Rethinking natural adversarial examples for classification models.
\newblock \emph{arXiv preprint arXiv:2102.11731}, 2021{\natexlab{c}}.

\bibitem[Liu et~al.(2021)Liu, Lin, Cao, Hu, Wei, Zhang, Lin, and
  Guo]{liu2021swin}
Liu, Z., Lin, Y., Cao, Y., Hu, H., Wei, Y., Zhang, Z., Lin, S., and Guo, B.
\newblock Swin transformer: Hierarchical vision transformer using shifted
  windows.
\newblock In \emph{Proceedings of the IEEE/CVF international conference on
  computer vision}, pp.\  10012--10022, 2021.

\bibitem[Liu et~al.(2022)Liu, Mao, Wu, Feichtenhofer, Darrell, and
  Xie]{liu2022convnet}
Liu, Z., Mao, H., Wu, C.-Y., Feichtenhofer, C., Darrell, T., and Xie, S.
\newblock A convnet for the 2020s.
\newblock In \emph{Proceedings of the IEEE/CVF Conference on Computer Vision
  and Pattern Recognition}, pp.\  11976--11986, 2022.

\bibitem[Luccioni \& Rolnick(2022)Luccioni and Rolnick]{luccioni2022bugs}
Luccioni, A.~S. and Rolnick, D.
\newblock Bugs in the data: How imagenet misrepresents biodiversity.
\newblock \emph{arXiv preprint arXiv:2208.11695}, 2022.

\bibitem[Lyzhov et~al.(2020)Lyzhov, Molchanova, Ashukha, Molchanov, and
  Vetrov]{lyzhov2020greedy}
Lyzhov, A., Molchanova, Y., Ashukha, A., Molchanov, D., and Vetrov, D.
\newblock Greedy policy search: A simple baseline for learnable test-time
  augmentation.
\newblock In \emph{Conference on Uncertainty in Artificial Intelligence}, pp.\
  1308--1317. PMLR, 2020.

\bibitem[Ma et~al.(2018)Ma, Zhang, Zheng, and Sun]{ma2018shufflenet}
Ma, N., Zhang, X., Zheng, H.-T., and Sun, J.
\newblock Shufflenet v2: Practical guidelines for efficient cnn architecture
  design.
\newblock In \emph{Proceedings of the European conference on computer vision
  (ECCV)}, pp.\  116--131, 2018.

\bibitem[Marcel \& Rodriguez(2010)Marcel and Rodriguez]{marcel2010torchvision}
Marcel, S. and Rodriguez, Y.
\newblock Torchvision the machine-vision package of torch.
\newblock In \emph{Proceedings of the 18th ACM international conference on
  Multimedia}, pp.\  1485--1488, 2010.

\bibitem[Minderer et~al.(2022)Minderer, Gritsenko, Stone, Neumann, Weissenborn,
  Dosovitskiy, Mahendran, Arnab, Dehghani, Shen, et~al.]{minderer2022simple}
Minderer, M., Gritsenko, A., Stone, A., Neumann, M., Weissenborn, D.,
  Dosovitskiy, A., Mahendran, A., Arnab, A., Dehghani, M., Shen, Z., et~al.
\newblock Simple open-vocabulary object detection with vision transformers.
\newblock \emph{arXiv preprint arXiv:2205.06230}, 2022.

\bibitem[Nguyen et~al.(2016)Nguyen, Yosinski, and
  Clune]{nguyen2016multifaceted}
Nguyen, A., Yosinski, J., and Clune, J.
\newblock Multifaceted feature visualization: Uncovering the different types of
  features learned by each neuron in deep neural networks.
\newblock \emph{arXiv preprint arXiv:1602.03616}, 2016.

\bibitem[OpenAI(2023)]{openai2023gpt35turbo}
OpenAI.
\newblock Chatgpt api.
\newblock \url{https://openai.com/blog/chatgpt}, 2023.

\bibitem[Pang et~al.(2019)Pang, Xu, and Zhu]{pang2019mixup}
Pang, T., Xu, K., and Zhu, J.
\newblock Mixup inference: Better exploiting mixup to defend adversarial
  attacks.
\newblock \emph{arXiv preprint arXiv:1909.11515}, 2019.

\bibitem[Radford et~al.(2021)Radford, Kim, Hallacy, Ramesh, Goh, Agarwal,
  Sastry, Askell, Mishkin, Clark, et~al.]{radford2021learning}
Radford, A., Kim, J.~W., Hallacy, C., Ramesh, A., Goh, G., Agarwal, S., Sastry,
  G., Askell, A., Mishkin, P., Clark, J., et~al.
\newblock Learning transferable visual models from natural language
  supervision.
\newblock In \emph{International conference on machine learning}, pp.\
  8748--8763. PMLR, 2021.

\bibitem[Raghu et~al.(2021)Raghu, Unterthiner, Kornblith, Zhang, and
  Dosovitskiy]{raghu2021vision}
Raghu, M., Unterthiner, T., Kornblith, S., Zhang, C., and Dosovitskiy, A.
\newblock Do vision transformers see like convolutional neural networks?
\newblock \emph{Advances in Neural Information Processing Systems},
  34:\penalty0 12116--12128, 2021.

\bibitem[Recasens et~al.(2018)Recasens, Kellnhofer, Stent, Matusik, and
  Torralba]{recasens2018learning}
Recasens, A., Kellnhofer, P., Stent, S., Matusik, W., and Torralba, A.
\newblock Learning to zoom: a saliency-based sampling layer for neural
  networks.
\newblock In \emph{Proceedings of the European Conference on Computer Vision
  (ECCV)}, pp.\  51--66, 2018.

\bibitem[Recht et~al.(2019)Recht, Roelofs, Schmidt, and
  Shankar]{recht2019imagenet}
Recht, B., Roelofs, R., Schmidt, L., and Shankar, V.
\newblock Do imagenet classifiers generalize to imagenet?
\newblock In \emph{International conference on machine learning}, pp.\
  5389--5400. PMLR, 2019.

\bibitem[Robinson et~al.(2021)Robinson, Sun, Yu, Batmanghelich, Jegelka, and
  Sra]{robinson2021can}
Robinson, J., Sun, L., Yu, K., Batmanghelich, K., Jegelka, S., and Sra, S.
\newblock Can contrastive learning avoid shortcut solutions?
\newblock \emph{Advances in neural information processing systems},
  34:\penalty0 4974--4986, 2021.

\bibitem[Rs et~al.(2020)Rs, Cogswell, Das, Vedantam, Parikh, and
  Batra]{Ramprasaath2020GradCAM}
Rs, R., Cogswell, M., Das, A., Vedantam, R., Parikh, D., and Batra, D.
\newblock Grad-cam: Visual explanations from deep networks via gradient-based
  localization.
\newblock \emph{International Journal of Computer Vision}, 128, 02 2020.
\newblock \doi{10.1007/s11263-019-01228-7}.

\bibitem[Russakovsky et~al.(2015)Russakovsky, Deng, Su, Krause, Satheesh, Ma,
  Huang, Karpathy, Khosla, Bernstein, et~al.]{russakovsky2015imagenet}
Russakovsky, O., Deng, J., Su, H., Krause, J., Satheesh, S., Ma, S., Huang, Z.,
  Karpathy, A., Khosla, A., Bernstein, M., et~al.
\newblock Imagenet large scale visual recognition challenge.
\newblock \emph{International Journal of Computer Vision}, 115\penalty0
  (3):\penalty0 211--252, 2015.

\bibitem[Sandler et~al.(2018)Sandler, Howard, Zhu, Zhmoginov, and
  Chen]{sandler2018mobilenetv2}
Sandler, M., Howard, A., Zhu, M., Zhmoginov, A., and Chen, L.-C.
\newblock Mobilenetv2: Inverted residuals and linear bottlenecks.
\newblock In \emph{Proceedings of the IEEE conference on computer vision and
  pattern recognition}, pp.\  4510--4520, 2018.

\bibitem[Sato et~al.(2015)Sato, Nishimura, and Yokoi]{sato2015apac}
Sato, I., Nishimura, H., and Yokoi, K.
\newblock Apac: Augmented pattern classification with neural networks.
\newblock \emph{arXiv preprint arXiv:1505.03229}, 2015.

\bibitem[Shanmugam et~al.(2021)Shanmugam, Blalock, Balakrishnan, and
  Guttag]{shanmugam2021better}
Shanmugam, D., Blalock, D., Balakrishnan, G., and Guttag, J.
\newblock Better aggregation in test-time augmentation.
\newblock In \emph{Proceedings of the IEEE/CVF International Conference on
  Computer Vision}, pp.\  1214--1223, 2021.

\bibitem[Simonyan \& Zisserman(2014)Simonyan and Zisserman]{simonyan2014very}
Simonyan, K. and Zisserman, A.
\newblock Very deep convolutional networks for large-scale image recognition.
\newblock \emph{arXiv preprint arXiv:1409.1556}, 2014.

\bibitem[Slav\'{\i}k(1996)]{petr1996setcover}
Slav\'{\i}k, P.
\newblock A tight analysis of the greedy algorithm for set cover.
\newblock In \emph{Proceedings of the Twenty-Eighth Annual ACM Symposium on
  Theory of Computing}, STOC '96, pp.\  435–441, New York, NY, USA, 1996.
  Association for Computing Machinery.
\newblock ISBN 0897917855.
\newblock \doi{10.1145/237814.237991}.
\newblock URL \url{https://doi.org/10.1145/237814.237991}.

\bibitem[Smith \& Gal(2018)Smith and Gal]{smith2018understanding}
Smith, L. and Gal, Y.
\newblock Understanding measures of uncertainty for adversarial example
  detection.
\newblock \emph{arXiv preprint arXiv:1803.08533}, 2018.

\bibitem[Steiner et~al.(2021)Steiner, Kolesnikov, Zhai, Wightman, Uszkoreit,
  and Beyer]{steiner2021train}
Steiner, A., Kolesnikov, A., Zhai, X., Wightman, R., Uszkoreit, J., and Beyer,
  L.
\newblock How to train your vit? data, augmentation, and regularization in
  vision transformers.
\newblock \emph{arXiv preprint arXiv:2106.10270}, 2021.

\bibitem[Szegedy et~al.(2015)Szegedy, Liu, Jia, Sermanet, Reed, Anguelov,
  Erhan, Vanhoucke, and Rabinovich]{szegedy2015going}
Szegedy, C., Liu, W., Jia, Y., Sermanet, P., Reed, S., Anguelov, D., Erhan, D.,
  Vanhoucke, V., and Rabinovich, A.
\newblock Going deeper with convolutions.
\newblock In \emph{Proceedings of the IEEE conference on computer vision and
  pattern recognition}, pp.\  1--9, 2015.

\bibitem[Taesiri et~al.(2022)Taesiri, Nguyen, and Nguyen]{taesiri2022visual}
Taesiri, M.~R., Nguyen, G., and Nguyen, A.
\newblock Visual correspondence-based explanations improve ai robustness and
  human-ai team accuracy.
\newblock \emph{Advances in Neural Information Processing Systems},
  35:\penalty0 34287--34301, 2022.

\bibitem[Tan \& Le(2019)Tan and Le]{tan2019efficientnet}
Tan, M. and Le, Q.
\newblock Efficientnet: Rethinking model scaling for convolutional neural
  networks.
\newblock In \emph{International conference on machine learning}, pp.\
  6105--6114. PMLR, 2019.

\bibitem[Thavamani et~al.(2023)Thavamani, Li, Ferroni, and
  Ramanan]{thavamani2023learning}
Thavamani, C., Li, M., Ferroni, F., and Ramanan, D.
\newblock Learning to zoom and unzoom.
\newblock In \emph{Proceedings of the IEEE/CVF Conference on Computer Vision
  and Pattern Recognition}, pp.\  5086--5095, 2023.

\bibitem[Tsipras et~al.(2020)Tsipras, Santurkar, Engstrom, Ilyas, and
  Madry]{tsipras2020imagenet}
Tsipras, D., Santurkar, S., Engstrom, L., Ilyas, A., and Madry, A.
\newblock From imagenet to image classification: Contextualizing progress on
  benchmarks.
\newblock In \emph{International Conference on Machine Learning}, pp.\
  9625--9635. PMLR, 2020.

\bibitem[Tu et~al.(2022)Tu, Talebi, Zhang, Yang, Milanfar, Bovik, and
  Li]{tu2022maxvit}
Tu, Z., Talebi, H., Zhang, H., Yang, F., Milanfar, P., Bovik, A., and Li, Y.
\newblock Maxvit: Multi-axis vision transformer.
\newblock In \emph{European conference on computer vision}, pp.\  459--479.
  Springer, 2022.

\bibitem[Uzkent \& Ermon(2020)Uzkent and Ermon]{uzkent2020learning}
Uzkent, B. and Ermon, S.
\newblock Learning when and where to zoom with deep reinforcement learning.
\newblock In \emph{Proceedings of the IEEE/CVF conference on computer vision
  and pattern recognition}, pp.\  12345--12354, 2020.

\bibitem[Wang et~al.(2020)Wang, Shelhamer, Liu, Olshausen, and
  Darrell]{wang2020tent}
Wang, D., Shelhamer, E., Liu, S., Olshausen, B., and Darrell, T.
\newblock Tent: Fully test-time adaptation by entropy minimization.
\newblock In \emph{International Conference on Learning Representations}, 2020.

\bibitem[Wang et~al.(2019)Wang, Ge, Lipton, and Xing]{wang2019learning}
Wang, H., Ge, S., Lipton, Z., and Xing, E.~P.
\newblock Learning robust global representations by penalizing local predictive
  power.
\newblock \emph{Advances in Neural Information Processing Systems}, 32, 2019.

\bibitem[Wei et~al.(2021)Wei, Song, Mac~Aodha, Wu, Peng, Tang, Yang, and
  Belongie]{wei2021fine}
Wei, X.-S., Song, Y.-Z., Mac~Aodha, O., Wu, J., Peng, Y., Tang, J., Yang, J.,
  and Belongie, S.
\newblock Fine-grained image analysis with deep learning: A survey.
\newblock \emph{IEEE transactions on pattern analysis and machine
  intelligence}, 44\penalty0 (12):\penalty0 8927--8948, 2021.

\bibitem[Wightman et~al.(2021)Wightman, Touvron, and
  J{\'e}gou]{wightman2021resnet}
Wightman, R., Touvron, H., and J{\'e}gou, H.
\newblock Resnet strikes back: An improved training procedure in timm.
\newblock \emph{arXiv preprint arXiv:2110.00476}, 2021.

\bibitem[Xiao et~al.(2020)Xiao, Engstrom, Ilyas, and Madry]{xiao2020noise}
Xiao, K.~Y., Engstrom, L., Ilyas, A., and Madry, A.
\newblock Noise or signal: The role of image backgrounds in object recognition.
\newblock In \emph{International Conference on Learning Representations}, 2020.

\bibitem[Xie et~al.(2020)Xie, Luong, Hovy, and Le]{xie2020self}
Xie, Q., Luong, M.-T., Hovy, E., and Le, Q.~V.
\newblock Self-training with noisy student improves imagenet classification.
\newblock In \emph{Proceedings of the IEEE/CVF conference on computer vision
  and pattern recognition}, pp.\  10687--10698, 2020.

\bibitem[Xie et~al.(2017)Xie, Girshick, Doll{\'a}r, Tu, and
  He]{xie2017aggregated}
Xie, S., Girshick, R., Doll{\'a}r, P., Tu, Z., and He, K.
\newblock Aggregated residual transformations for deep neural networks.
\newblock In \emph{Proceedings of the IEEE conference on computer vision and
  pattern recognition}, pp.\  1492--1500, 2017.

\bibitem[Yun et~al.(2021)Yun, Oh, Heo, Han, Choe, and Chun]{yun2021re}
Yun, S., Oh, S.~J., Heo, B., Han, D., Choe, J., and Chun, S.
\newblock Re-labeling imagenet: from single to multi-labels, from global to
  localized labels.
\newblock In \emph{Proceedings of the IEEE/CVF Conference on Computer Vision
  and Pattern Recognition}, pp.\  2340--2350, 2021.

\bibitem[Zagoruyko \& Komodakis(2016)Zagoruyko and
  Komodakis]{zagoruyko2016wide}
Zagoruyko, S. and Komodakis, N.
\newblock Wide residual networks.
\newblock \emph{arXiv preprint arXiv:1605.07146}, 2016.

\bibitem[Zhang et~al.(2021)Zhang, Levine, and Finn]{zhang2021memo}
Zhang, M., Levine, S., and Finn, C.
\newblock Memo: Test time robustness via adaptation and augmentation.
\newblock \emph{arXiv preprint arXiv:2110.09506}, 2021.

\bibitem[Zheng et~al.(2019)Zheng, Fu, Zha, and Luo]{zheng2019looking}
Zheng, H., Fu, J., Zha, Z.-J., and Luo, J.
\newblock Looking for the devil in the details: Learning trilinear attention
  sampling network for fine-grained image recognition.
\newblock In \emph{Proceedings of the IEEE/CVF Conference on Computer Vision
  and Pattern Recognition}, pp.\  5012--5021, 2019.

\bibitem[Zhong et~al.(2022)Zhong, Yang, Zhang, Li, Codella, Li, Zhou, Dai,
  Yuan, Li, et~al.]{zhong2022regionclip}
Zhong, Y., Yang, J., Zhang, P., Li, C., Codella, N., Li, L.~H., Zhou, L., Dai,
  X., Yuan, L., Li, Y., et~al.
\newblock Regionclip: Region-based language-image pretraining.
\newblock In \emph{Proceedings of the IEEE/CVF Conference on Computer Vision
  and Pattern Recognition}, pp.\  16793--16803, 2022.

\bibitem[Zhu et~al.(2016)Zhu, Xie, and Yuille]{zhu2016object}
Zhu, Z., Xie, L., and Yuille, A.~L.
\newblock Object recognition with and without objects.
\newblock \emph{arXiv preprint arXiv:1611.06596}, 2016.

\end{thebibliography}
}

\clearpage

\newcommand{\beginsupplementary}{%
    \setcounter{table}{0}
    \renewcommand{\thetable}{A\arabic{table}}%
    
    \setcounter{figure}{0}
    \renewcommand{\thefigure}{A\arabic{figure}}%
    
    \setcounter{section}{0}
    \renewcommand{\thesection}{A\arabic{section}}
    \renewcommand{\thesubsection}{\thesection.\arabic{subsection}}
}

\beginsupplementary%
\appendix

\newcommand{\toptitlebar}{
    \hrule height 4pt
    \vskip 0.25in
    \vskip -\parskip%
}
\newcommand{\bottomtitlebar}{
    \vskip 0.29in
    \vskip -\parskip%
    \hrule height 1pt
    \vskip 0.09in%
}

\newcommand{\suptitle}{Appendix for:\\\papertitle}

\newcommand{\maketitlesupp}{
    \newpage
    \onecolumn
        \null
        \vskip .375in
        \begin{center}
            \toptitlebar
            {\Large \bf \suptitle\par}
            \bottomtitlebar
            \vspace*{24pt}
            {
                \large
                \lineskip=.5em
                \par
            }
            \vskip .5em
            \vspace*{12pt}
        \end{center}
}

\maketitlesupp%

\section{Implementation details}

In this section, we provide a detailed description of our experimental setup, including the Python code for our zoom transform, the classifiers we employed, and the setup we used for zero-shot classification.

\FloatBarrier

\subsection{Sample Python code for zoom-based transform}
\label{sec:sample_python}

\begin{adjustbox}{width=\textwidth}
    \centering
    \begin{lstlisting}[language=Python]
from PIL import Image
import torchvision.transforms.functional as fv
import torchvision.transforms as transforms
from functools import partial

def crop_at(size, slice_x, slice_y):
    def slice_crop(image, size, slice_x, slice_y):
        width, height = image.size
        tile_size_x = width // 3
        tile_size_y = height // 3
        anchor_x = (slice_y * tile_size_x) + (tile_size_x // 2)
        anchor_y = (slice_x * tile_size_y) + (tile_size_y // 2)
        return fv.crop(
            image,
            anchor_y - (size // 2),
            anchor_x - (size // 2),
            size,
            size,
        )
    return partial(slice_crop, size=size, slice_x=slice_x, slice_y=slice_y)
    
zoom_scale = 255
zoom_transform = transforms.Compose(
                    [
                        transforms.Resize(
                            zoom_scale,
                            interpolation=transforms.InterpolationMode.BICUBIC,
                            max_size=None,
                            antialias=None,
                        ),
                        crop_at(224, i, j),
                    ]
                )
\end{lstlisting}
\end{adjustbox}
\captionof{figure}{Sample python code.}
\label{fig:example_code}

\clearpage
\subsection{Datasets' licenses}

\begin{table}[!htb]
\centering
\begin{tabular}{l|l}
\toprule
\textbf{Dataset Name} & \textbf{License} \\
\midrule
ImageNet & \href{https://image-net.org/accessagreement}{Custom license, non-commercial} \\
ImageNet-A & \href{https://github.com/hendrycks/natural-adv-examples/blob/master/LICENSE}{ License} \\
ImageNet-R & \href{https://github.com/hendrycks/imagenet-r/blob/master/LICENSE}{MIT License} \\
ImageNet-Sketch & \href{https://github.com/HaohanWang/ImageNet-Sketch/blob/master/LICENSE}{MIT License} \\
ImageNet-C & \href{https://github.com/hendrycks/robustness/blob/master/LICENSE}{MIT License} \\
ObjectNet & \href{https://objectnet.dev/download.html#License}{Custom license derived from Creative Commons Attribution 4.0} \\
ImageNet-V2 & \href{https://github.com/modestyachts/ImageNetV2/blob/master/LICENSE}{MIT License} \\
\bottomrule
\end{tabular}
\caption{Dataset Licenses}
\label{table:dataset_licenses}
\end{table}

\FloatBarrier
\subsection{Zoom Scales used}

In our experiments, we tried the following zoom scales:

\begin{align*}
&10, 16, 32, 48, 64, 96, 122, 128, 192, 224, 235, 240, 256, 288, 320, 348, 384, 448, 460, 512,\\
&573, 576, 640, 664, 672, 680, 686, 690, 700, 720, 768, 798, 832, 896, 911, 1024.
\end{align*}

\FloatBarrier
\subsection{Model selection}

We use the official \href{https://github.com/openai/CLIP}{{OpenAI's official CLIP}} for all CLIP-related experiments. All \textcolor{specialcolor}{IN-trained} models are retrieved from the \texttt{torchvision} \cite{marcel2010torchvision} library. For models from the OpenCLIP family, we utilize the \href{https://github.com/mlfoundations/open_clip}{{OpenCLIP}} library version \texttt{2.20.0}. In the case of the EfficientNet-B family, we use the \href{https://huggingface.co/google/efficientnet-b0}{{Hugging Face Transformers}} library. Lastly, for EfficientNet-L2, we use the implementation from the \href{https://huggingface.co/timm/tf_efficientnet_l2.ns_jft_in1k/discussions}{{timm}} library.

\clearpage
\subsection{Zero-shot classification using CLIP}
\label{suppsec:zeroshot_clip_setup}
For CLIP, we follow the standard zero-shot classification.
This involves creating a text template for each class in the dataset, which contains a generic description of an image featuring an object from that class.
Then, we use CLIP's text encoder to obtain embeddings for these templates and then average them to obtain a final vector that represents the class.
To classify an image, we calculate the cosine similarity between its embedding and the text vectors for each class and then select the class with the highest value.

\FloatBarrier
\subsection{Zoom-based transform}
\label{supp:anchorpoint_instruction}

\begin{figure}[!hbt]
    \centering
    \subfloat[\centering]{{\includegraphics[width=5cm]{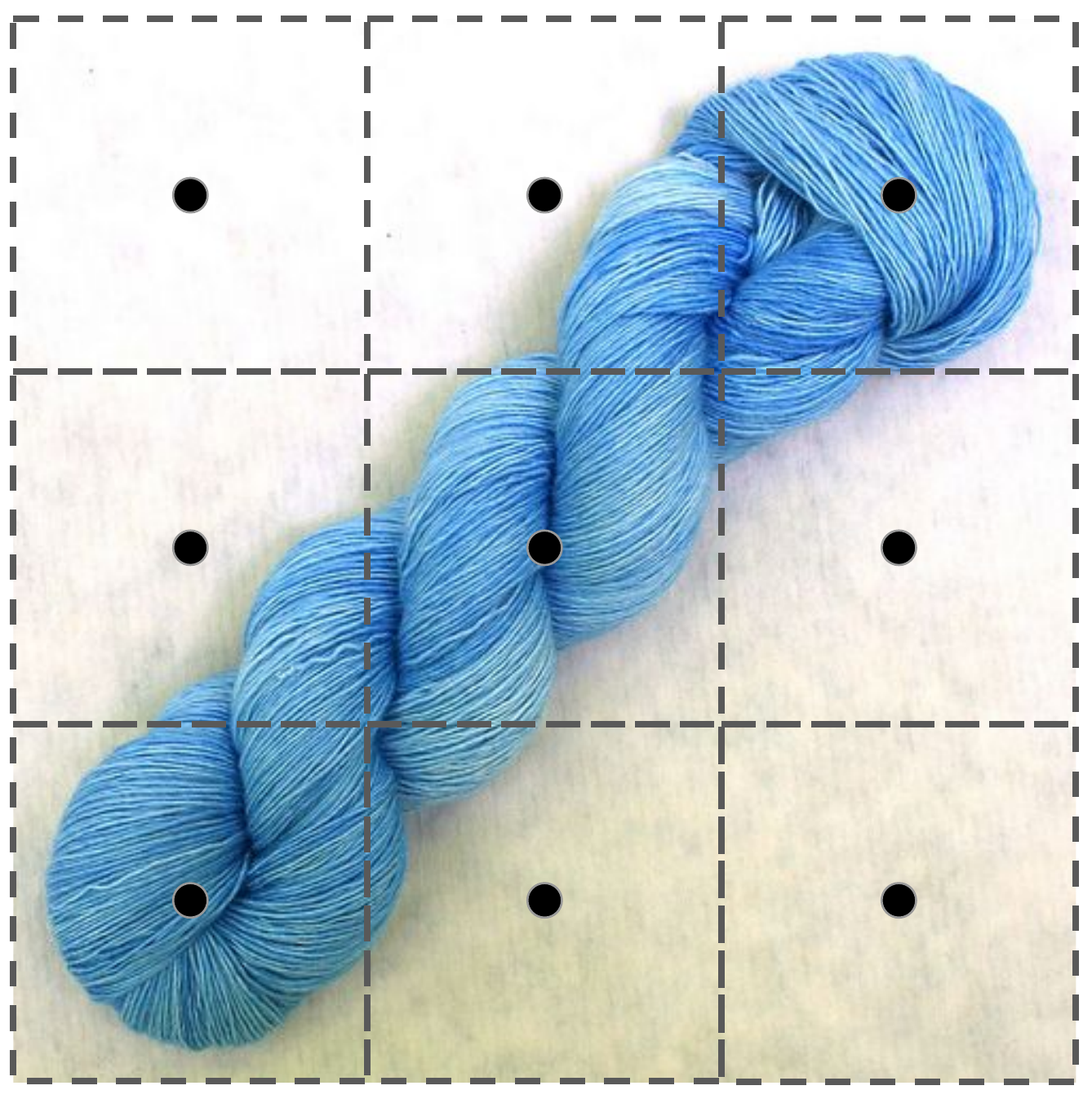} }}%
    \qquad
    \subfloat[\centering]{{\includegraphics[width=10cm]{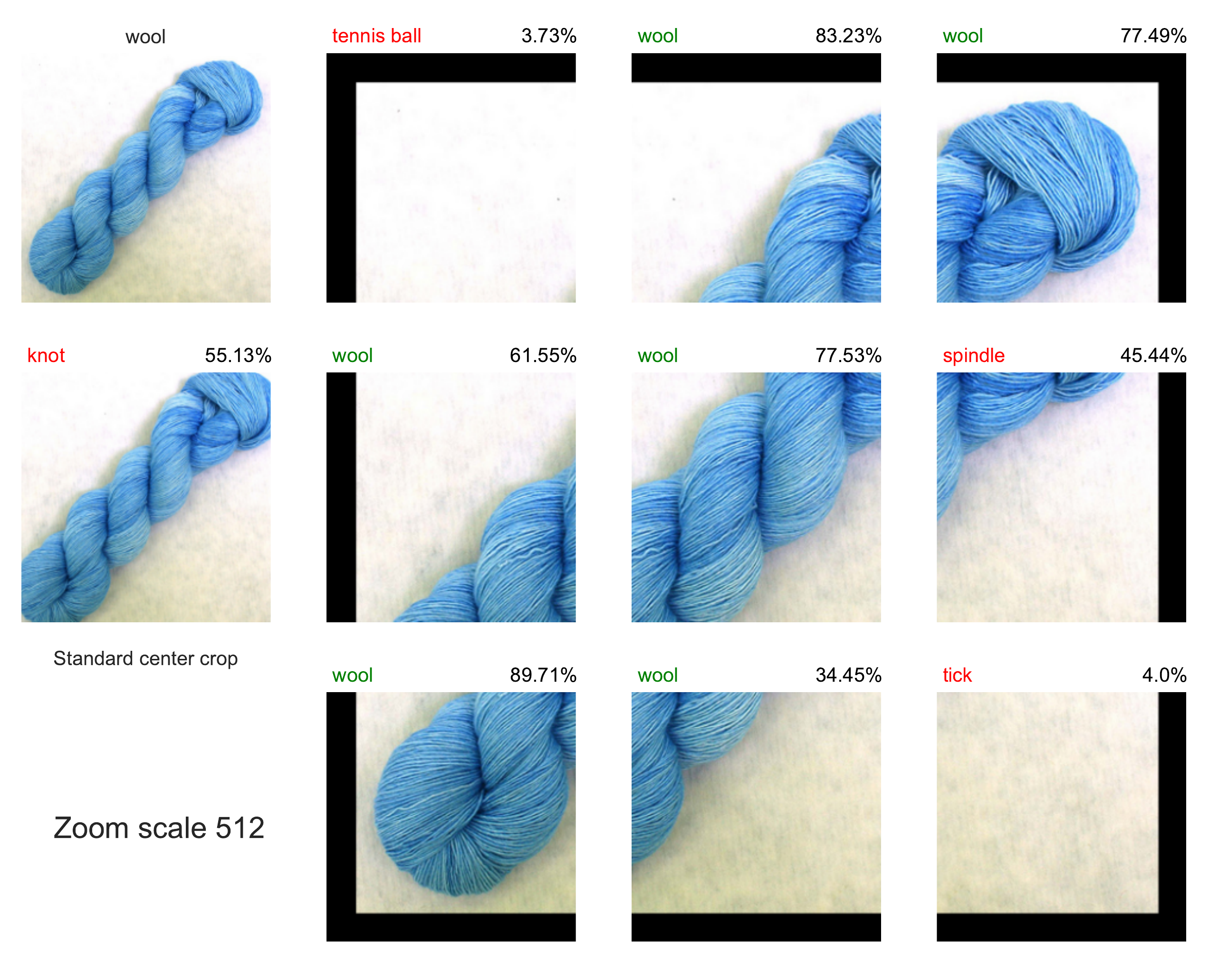} }}%
    \caption{(a) Making a 3-by-3 uniform grid out of the image. We pick the center point in each region as the anchor. (b) Sample image showing how our zoom transform is applied to an image.}%
    \label{fig:example}%
\end{figure}

\clearpage
\section{Additional Results}

In this section, we provide additional results for our experiments.

\subsection{Zooming out is needed for a small portion of the datasets}
\label{sec:disentagling_zoom_appendix}

In our approach, we leverage the power of both zoom-in and zoom-out transforms, and \cref{tab:1_main_results} results indicate that this combined zooming approach can be effective in classifying images from diverse datasets. 
Zooming in enhances texture patterns while zooming out provides a better perspective of the object's shape.
The question we aim to answer is which dataset and model pairs require which type of zoom, and whether zooming is always necessary. Additionally, we investigate which types of networks are less reliant on explicit zooming, as they implicitly focus on the main object in the image.

\paragraph{Experiment} 

We separate zoom transforms into three groups and report the maximum possible accuracy as defined in \cref{sec:method}.
We use transforms in the minimum set covers (as shown in \cref{tab:min_cover_size}) for each dataset and classifier pair.
We then report the number of images that can only be classified using transforms in each group separately.

\paragraph{Results}
    
\begin{figure}[htb!]
\centering
    \includegraphics[width=0.5\textwidth]{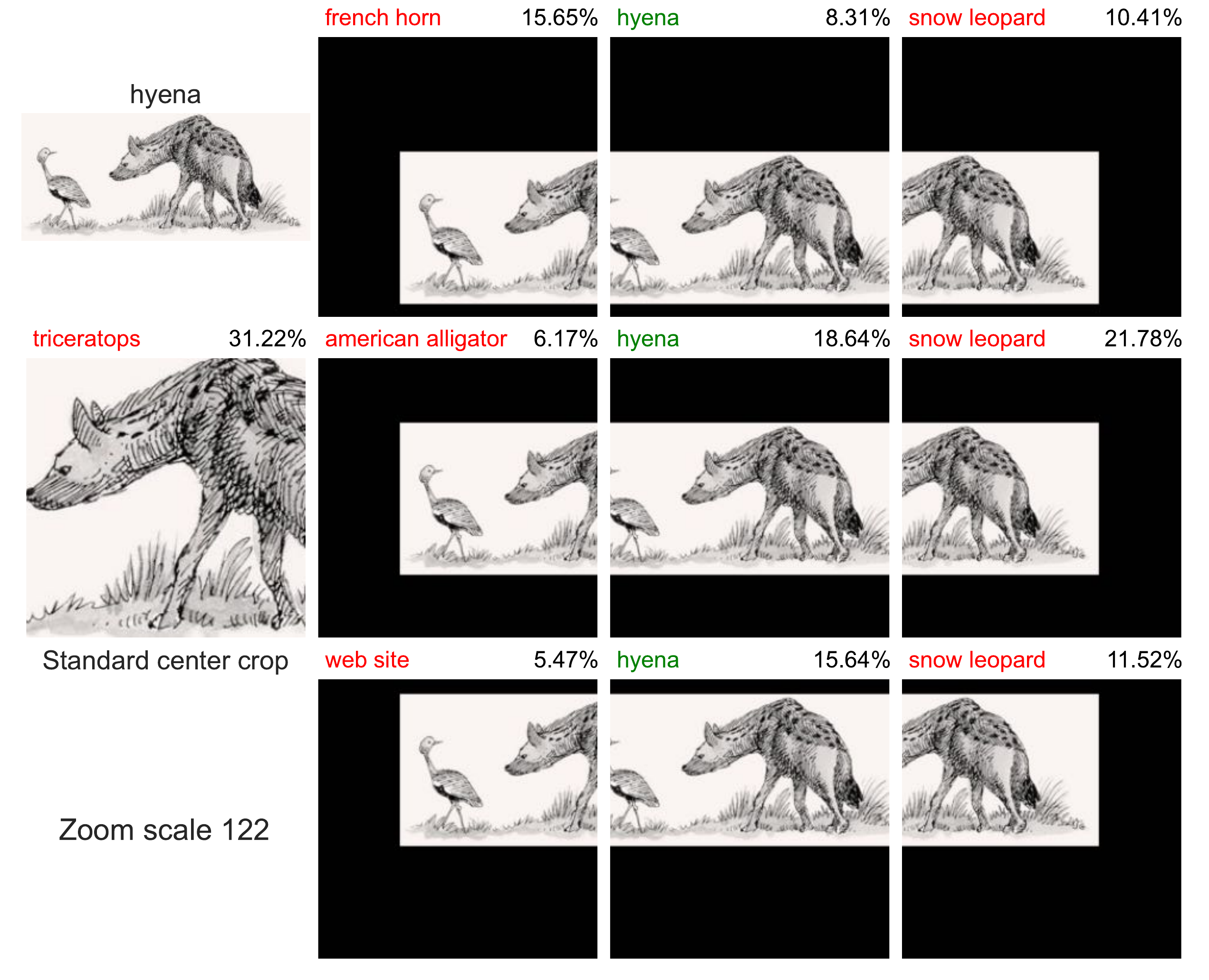}
    \caption{A sample image from the ImageNet-Sketch dataset that can only be solved by zooming out. For this image, with the standard ImageNet transform, the entire body of the animal is not visible. Instead, zooming out of the image helps you see the whole body of the animal. More samples can be found in \cref{supp:only_zoomout_sample_images}.
    }
     \label{fig:zoomout_is_needed}
\end{figure}

In general, we find that zooming in is more effective than zooming out.
Zooming in provides two benefits: (1) it helps the model to focus on the key region where the target object is located, and (2) the model can extract features from the target object at a higher resolution. 
Across all methods and datasets, we can see a certain percentage of images are only classifiable using transforms of the \zoomout group. In particular, for ImageNet-R and ImageNet-Sketch, between $1.2\% - 3$\% (Table~\ref{supptab:zoomgroup}) of the entire dataset can only be solved using a transform in the \zoomout group. 
This is especially true for drawings, where the texture may lack distinguishable features, and zooming out allows us to better perceive the shape.

\begin{table}[htb!]
\centering
\caption{Breakdown of maximum possible accuracy by different zoom groups. In each dataset, certain images necessitate a specific zoom group for correct classification regardless of the model being used. However, CLIP performs well overall without depending heavily on a particular zoom level. On average, the percentage of datasets that can only be solved with a specific zoom group is very small for this model.}
\label{supptab:zoomgroup}
\resizebox{1\textwidth}{!}{
\begin{tabular}{llrrrrrr}
\hline
\multicolumn{1}{c}{\textbf{Dataset}} &
  \multicolumn{1}{c}{\textbf{Model}} &
  \multicolumn{1}{c}{\textbf{\begin{tabular}[c]{@{}c@{}}\zoomin \\ Solve\end{tabular}}} &
  \multicolumn{1}{c}{\textbf{\begin{tabular}[c]{@{}c@{}}\zoomout\\ Solves\end{tabular}}} &
  \multicolumn{1}{c}{\textbf{\begin{tabular}[c]{@{}c@{}}\zoomless\\ Solves\end{tabular}}} &
  \multicolumn{1}{c}{\textbf{\begin{tabular}[c]{@{}c@{}}Only \zoomin\\ Solves\end{tabular}}} &
  \multicolumn{1}{c}{\textbf{\begin{tabular}[c]{@{}c@{}}Only \zoomout\\ Solves\end{tabular}}} &
  \multicolumn{1}{c}{\textbf{\begin{tabular}[c]{@{}c@{}}Only \zoomless\\ Solves\end{tabular}}} \\ \hline
\multirow{6}{*}{ImageNet}        & ResNet-18       &94.57& 79.49& 81.16& 10.59& 0.43& 0.08    \\
                                 & ResNet-50       &96.30& 85.84& 86.39& 7.59& 0.40& 0.04    \\
                                 & ViT-B/32        &96.83& 86.18& 85.12& 7.59& 0.30& 0.02    \\
                                 & VGG-16          &94.60& 82.11& 83.08& 8.92& 0.58& 0.07    \\
                                 & AlexNet         &89.17& 62.92& 67.98& 18.01& 0.65& 0.18    \\
                                 & \clip-ViT-L/14  &95.82& 90.80& 87.04& 4.81& 0.83& 0.05    \\ \hline
\multirow{6}{*}{ImageNet ReaL}   & ResNet-18       &97.37& 86.10& 87.62& 7.38& 0.27& 0.07    \\
                                 & ResNet-50       &98.22& 91.07& 91.87& 4.65& 0.25& 0.04    \\
                                 & ViT-B/32        &98.50& 90.79& 88.06& 4.92& 0.18& 0.03    \\
                                 & VGG-16          &97.38& 88.43& 89.40& 6.02& 0.38& 0.07    \\
                                 & AlexNet         &93.15& 69.58& 74.85& 15.47& 0.45& 0.19    \\
                                 & \clip-ViT-L/14  &98.05& 94.44& 91.69& 3.20& 0.55& 0.04    \\ \hline                                 
\multirow{6}{*}{ImageNet+ReaL}   & ResNet-18       &97.16& 85.51& 86.77& 7.72& 0.28& 0.05    \\
                                 & ResNet-50       &98.25& 91.10& 91.77& 4.60& 0.24& 0.03    \\
                                 & ViT-B/32        &98.70& 91.00& 90.95& 4.92& 0.14& 0.02    \\
                                 & VGG-16          &97.12& 87.88& 89.09& 6.25& 0.42& 0.06    \\
                                 & AlexNet         &92.79& 68.65& 73.93& 16.25& 0.47& 0.16    \\
                                 & \clip-ViT-L/14  &98.24& 95.09& 92.41& 2.75& 0.47& 0.04    \\ \hline
\multirow{6}{*}{ImageNet-A}      & ResNet-18       &63.66& 47.95& 45.37& 13.97& 2.75& 0.21    \\
                                 & ResNet-50       &65.28& 52.36& 48.59& 12.05& 3.13& 0.22    \\
                                 & ViT-B/32        &73.07& 56.34& 54.84& 14.20& 2.04& 0.27    \\
                                 & VGG-16          &56.67& 44.95& 39.35& 11.80& 3.85& 0.24    \\
                                 & AlexNet         &52.69& 32.86& 31.95& 17.15& 2.34& 0.30    \\
                                 & \clip-ViT-L/14  &98.35& 96.71& 93.57& 1.70& 0.69& 0.04    \\ \hline
\multirow{6}{*}{ImageNet-R}      & ResNet-18       &57.07& 12.19& 10.07& 40.67& 0.92& 0.19    \\
                                 & ResNet-50       &64.52& 12.95& 10.36& 48.72& 1.00& 0.23    \\
                                 & ViT-B/32        &76.71& 18.57& 21.92& 51.75& 0.85& 0.15    \\
                                 & VGG-16          &56.59& 13.15& 13.27& 38.24& 0.93& 0.29    \\
                                 & AlexNet         &39.91& 10.39& 9.11& 26.27& 1.08& 0.36    \\
                                 & \clip-ViT-L/14  &97.99& 81.32& 77.03& 12.01& 0.44& 0.05     \\ \hline
\multirow{6}{*}{ImageNet-Sketch} & ResNet-18       &41.14& 27.06& 27.41& 11.83& 1.77& 0.36     \\
                                 & ResNet-50       &44.72& 32.80& 31.45& 10.99& 2.23& 0.24     \\
                                 & ViT-B/32        &53.45& 37.43& 37.38& 13.11& 1.83& 0.36     \\
                                 & VGG-16          &36.20& 27.20& 24.59& 9.47& 2.97& 0.28     \\
                                 & AlexNet         &27.71& 13.84& 15.11& 11.26& 1.22& 0.33    \\
                                 & \clip-ViT-L/14  &86.20& 80.67& 73.94& 6.64& 2.38& 0.12    \\ \hline
\multirow{6}{*}{ObjectNet}       & ResNet-18       &68.98& 38.52& 37.23& 25.76& 1.93& 0.25   \\
                                 & ResNet-50       &74.16& 51.56& 47.79& 19.68& 2.16& 0.30   \\
                                 & ViT-B/32        &77.66& 44.49& 42.65& 27.43& 1.34& 0.20   \\
                                 & VGG-16          &69.19& 41.72& 39.49& 23.34& 2.27& 0.31   \\
                                 & AlexNet         &56.76& 23.45& 22.59& 28.85& 2.27& 0.33   \\
                                 & \clip-ViT-L/14  &91.28& 82.22& 77.60& 8.37& 1.38& 0.15   \\ \hline
\multirow{6}{*}{Average}         & ResNet-18       &74.28& 53.83& 53.66& 16.85& 1.19& 0.17   \\
                                 & ResNet-50       &77.35& 59.67& 58.32& 15.47& 1.34& 0.16   \\
                                 & ViT-B/32        &82.13& 60.69& 60.13& 17.70& 0.95& 0.15   \\
                                 & VGG-16          &72.54& 55.06& 54.04& 14.86& 1.63& 0.19   \\
                                 & AlexNet         &64.60& 40.24& 42.22& 19.04& 1.21& 0.26   \\
                                 & \clip-ViT-L/14  &95.13& 88.75& 84.75& 5.64& 0.95 & 0.07   \\ \hline
\end{tabular}
}
\end{table}


\clearpage
\subsection{Anchor-based analysis of Center bias in ImageNet and OOD datasets}
\label{supp:anchor_based_analys}

\begin{figure}[htb!]
\centering
\begin{subfigure}{0.19\columnwidth}
\resizebox{\columnwidth}{!}{
\includegraphics[]{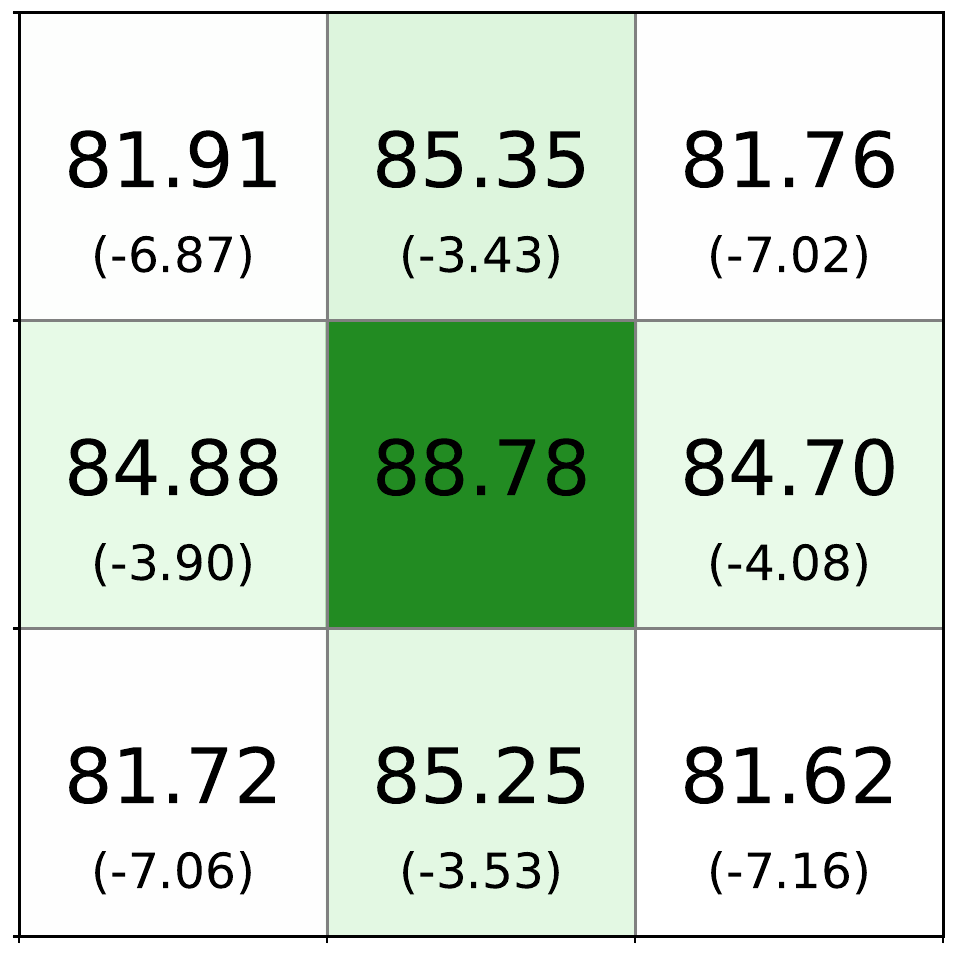}
}
\caption{\scriptsize{ImageNet-ReaL}}
\label{fig:3x3_REAL_5}
\end{subfigure}
\hfill
\begin{subfigure}{0.19\columnwidth}
\resizebox{\columnwidth}{!}{
\includegraphics[]{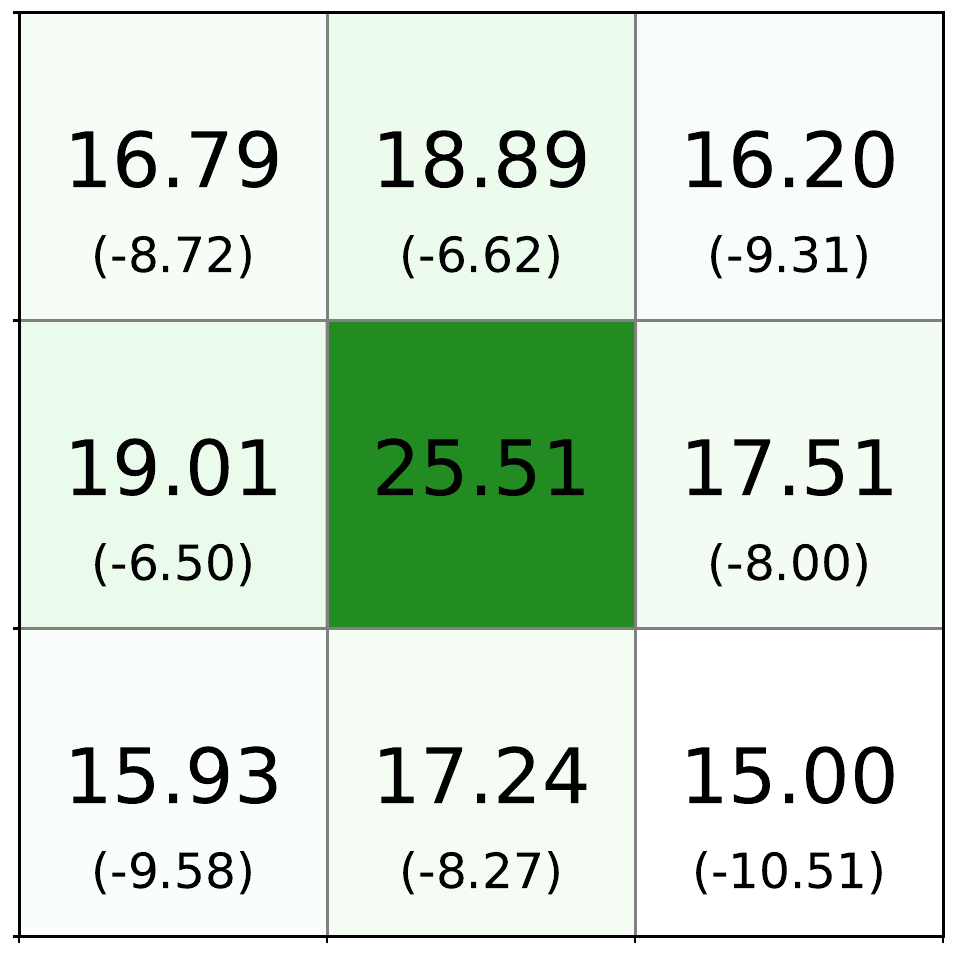}
}
\caption{\scriptsize{ImageNet-A}}
\label{fig:3x3_INA_5}
\end{subfigure}
\hfill
\begin{subfigure}{0.19\columnwidth}
\resizebox{\columnwidth}{!}{
\includegraphics[]{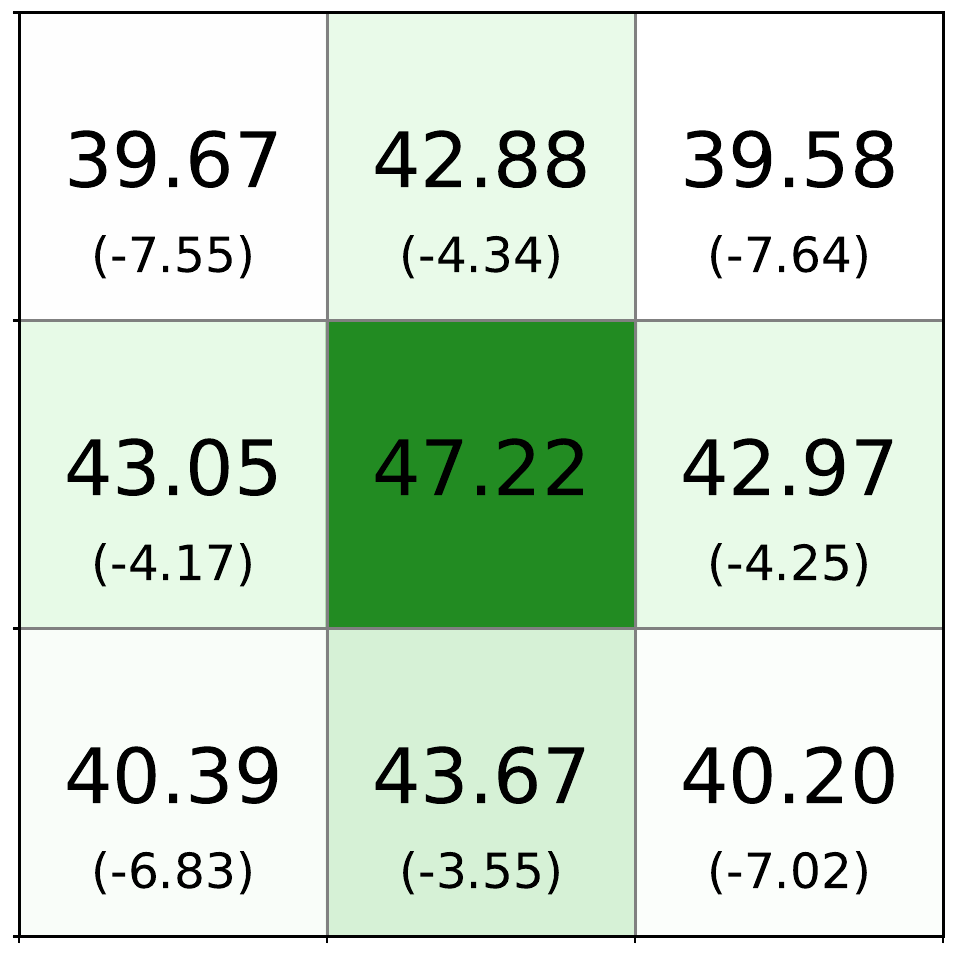}
}
\caption{\scriptsize{ImageNet-R}}
\label{fig:3x3_INR_5}
\end{subfigure}
\hfill
\begin{subfigure}{0.19\columnwidth}
\resizebox{\columnwidth}{!}{
\includegraphics[]{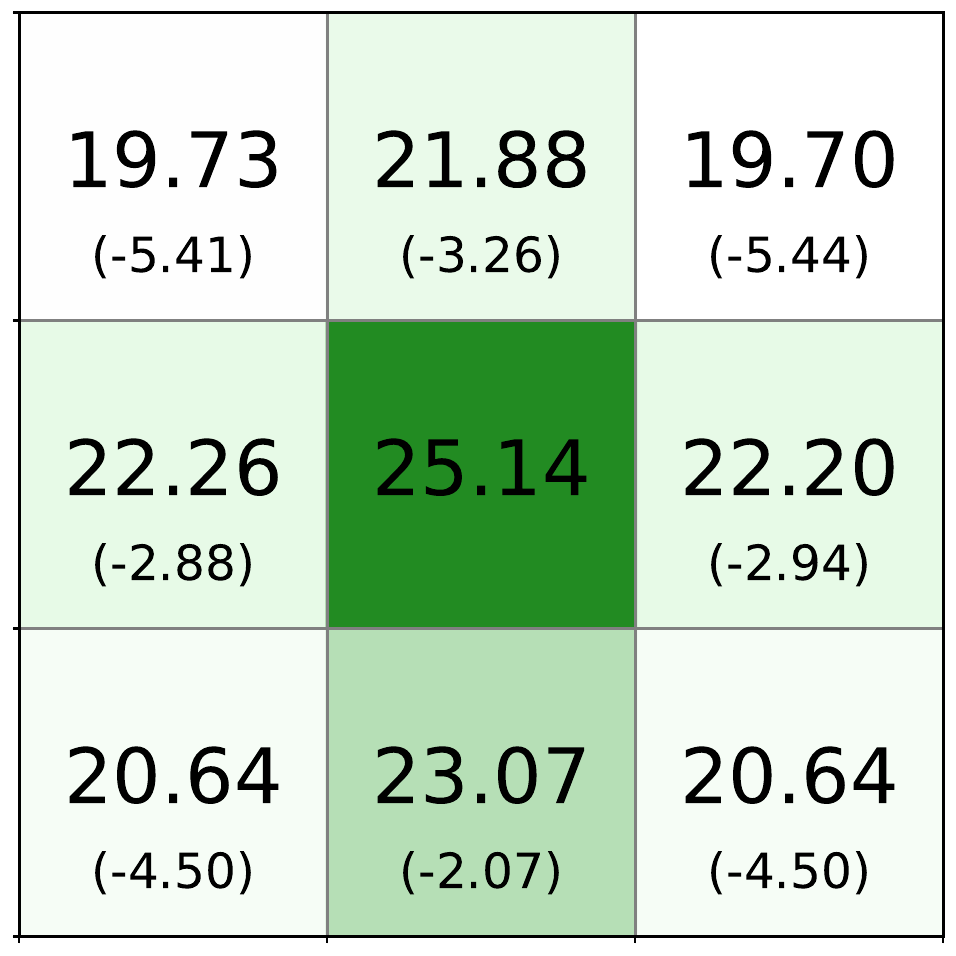}
}
\caption{\scriptsize{ImageNet-Sketch}}
\label{fig:3x3_INS_5}
\end{subfigure}
\hfill
\begin{subfigure}{0.19\columnwidth}
\resizebox{\columnwidth}{!}{
\includegraphics[]{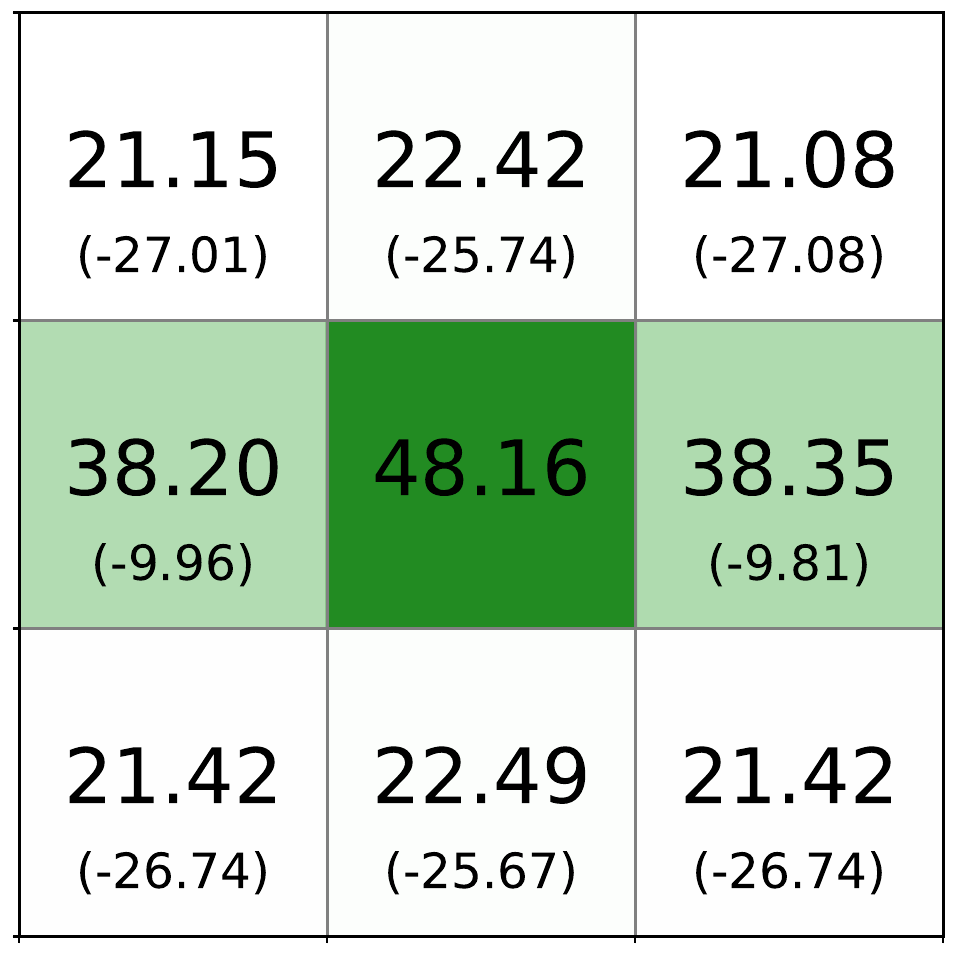}
}
\caption{\scriptsize{ObjectNet}}
\label{fig:3x3_ON_5}
\end{subfigure}
\caption{ AlexNet }
\label{fig:anchor_based_analysis_all_alexnet}
\end{figure}

\begin{figure}[htb!]
\centering
\begin{subfigure}{0.19\columnwidth}
\resizebox{\columnwidth}{!}{
\includegraphics[]{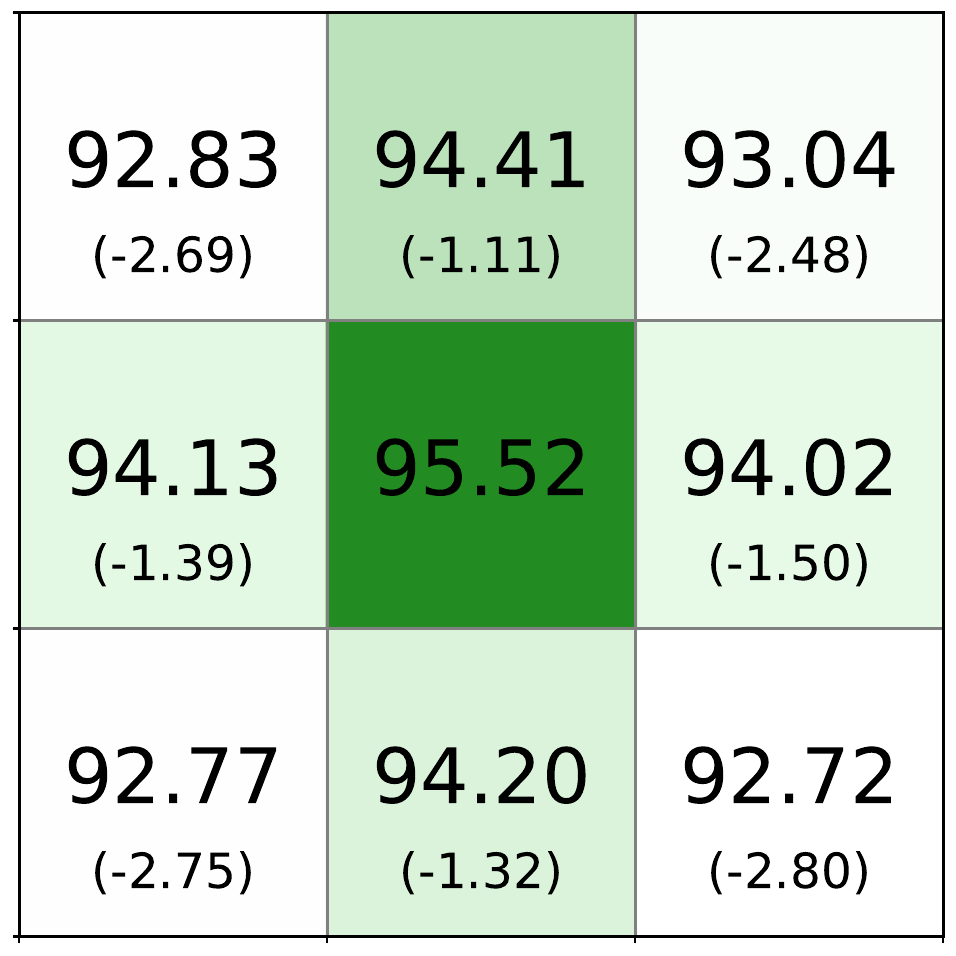}
}
\caption{\scriptsize{ImageNet-ReaL}}
\label{fig:3x3_REAL_4}
\end{subfigure}
\hfill
\begin{subfigure}{0.19\columnwidth}
\resizebox{\columnwidth}{!}{
\includegraphics[]{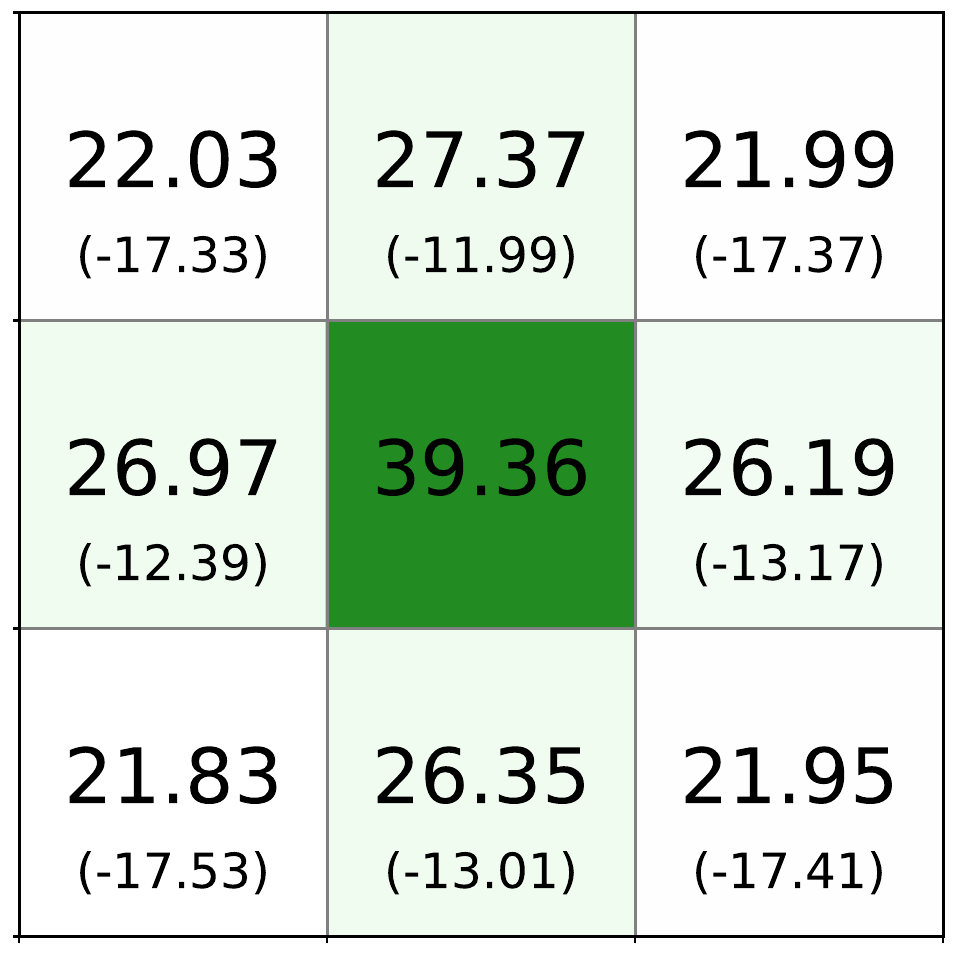}
}
\caption{\scriptsize{ImageNet-A}}
\label{fig:3x3_INA_4}
\end{subfigure}
\hfill
\begin{subfigure}{0.19\columnwidth}
\resizebox{\columnwidth}{!}{
\includegraphics[]{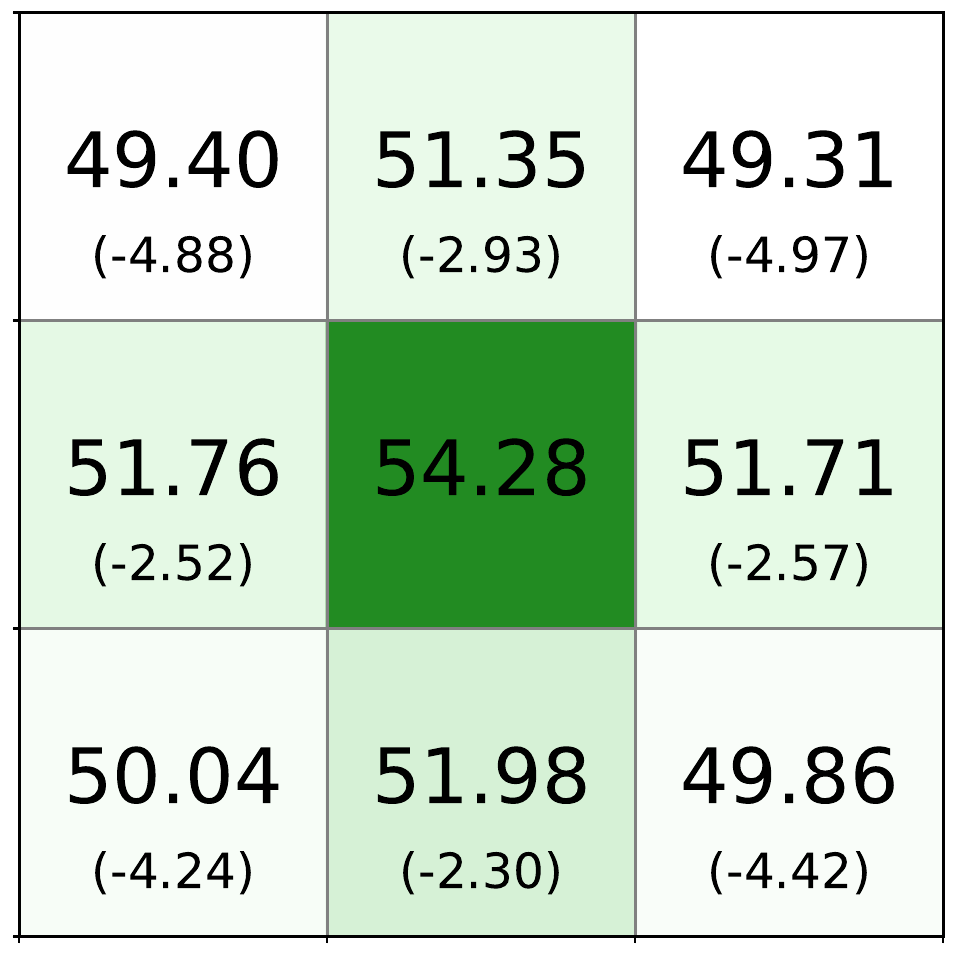}
}
\caption{\scriptsize{ImageNet-R}}
\label{fig:3x3_INR_4}
\end{subfigure}
\hfill
\begin{subfigure}{0.19\columnwidth}
\resizebox{\columnwidth}{!}{
\includegraphics[]{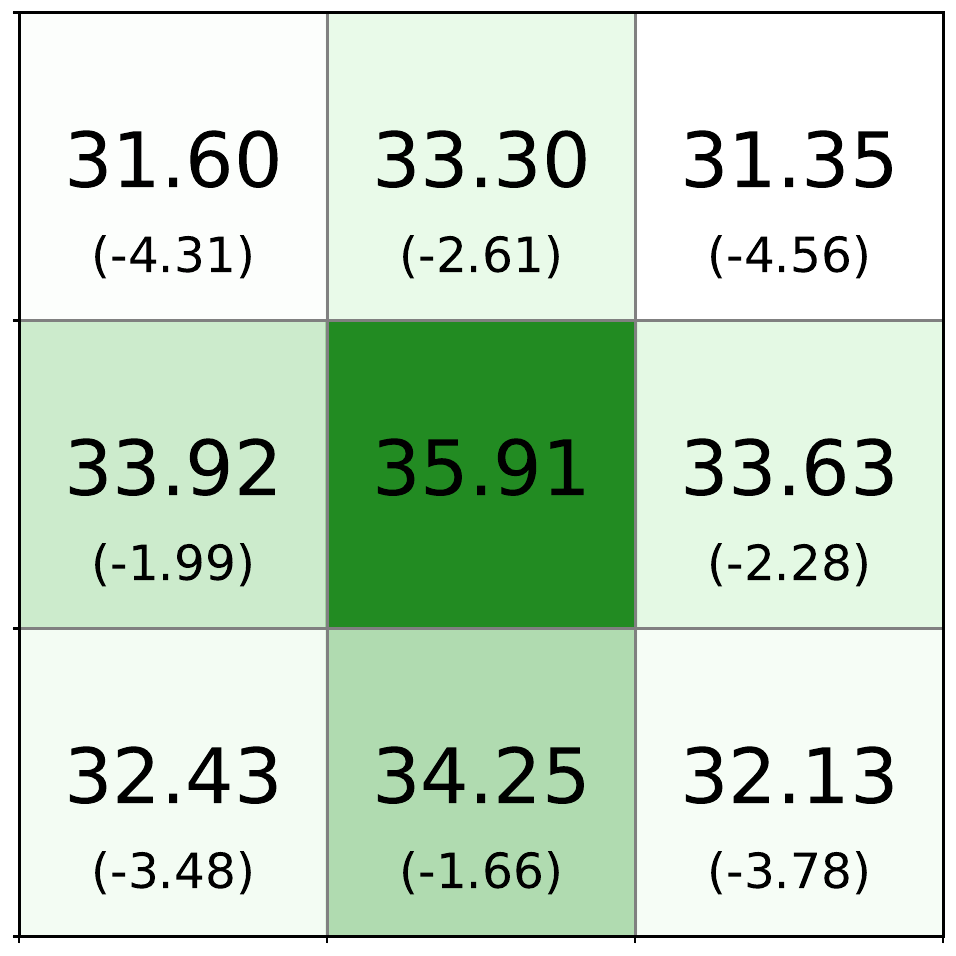}
}
\caption{\scriptsize{ImageNet-Sketch}}
\label{fig:3x3_INS_4}
\end{subfigure}
\hfill
\begin{subfigure}{0.19\columnwidth}
\resizebox{\columnwidth}{!}{
\includegraphics[]{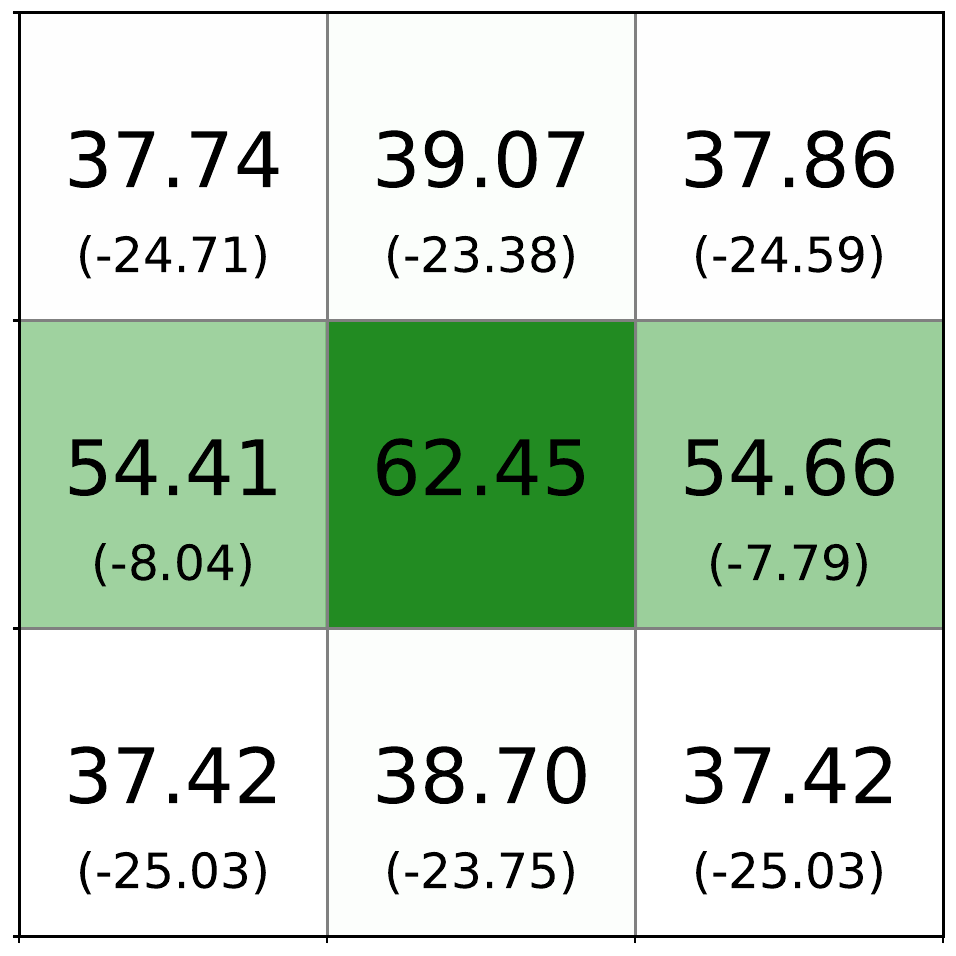}
}
\caption{\scriptsize{ObjectNet}}
\label{fig:3x3_ON_4}
\end{subfigure}
\caption{  VGG16 }
\label{fig:anchor_based_analysis_all_vgg16}
\end{figure}

\begin{figure}[htb!]
\centering
\begin{subfigure}{0.19\columnwidth}
\resizebox{\columnwidth}{!}{
\includegraphics[]{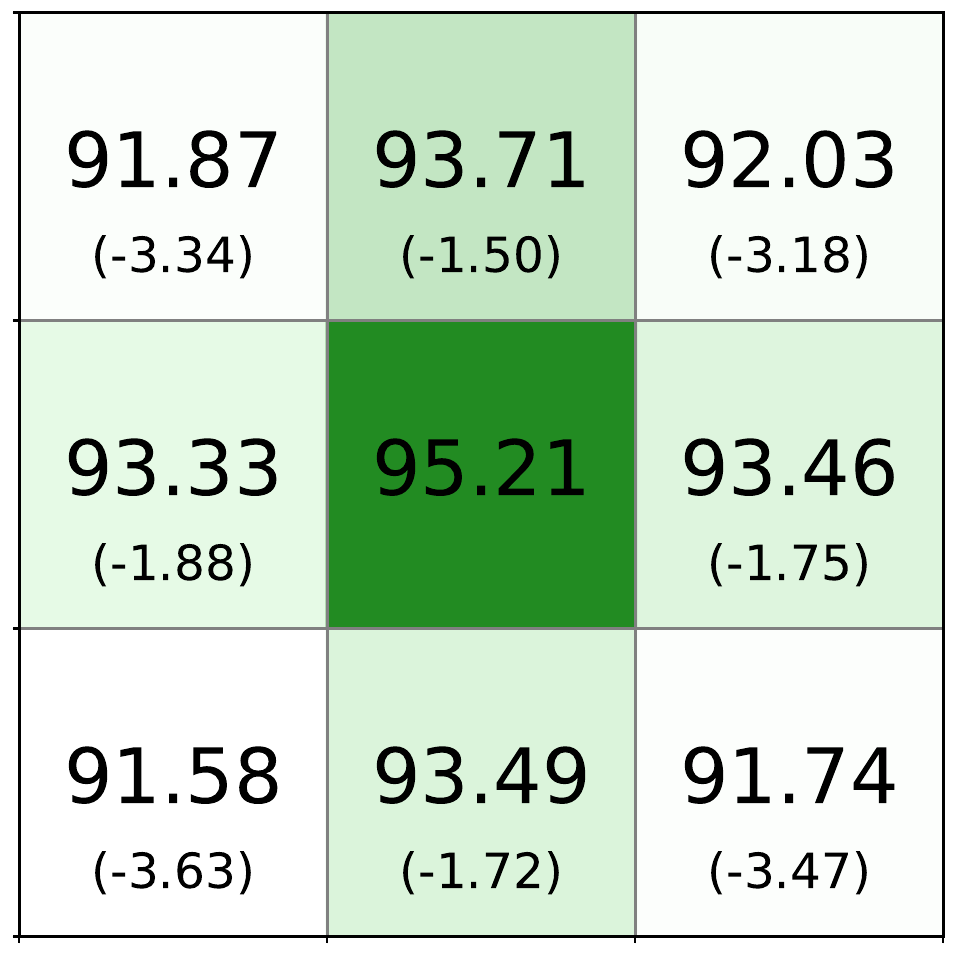}
}
\caption{\scriptsize{ImageNet-ReaL}}
\label{fig:3x3_REAL_1}
\end{subfigure}
\hfill
\begin{subfigure}{0.19\columnwidth}
\resizebox{\columnwidth}{!}{
\includegraphics[]{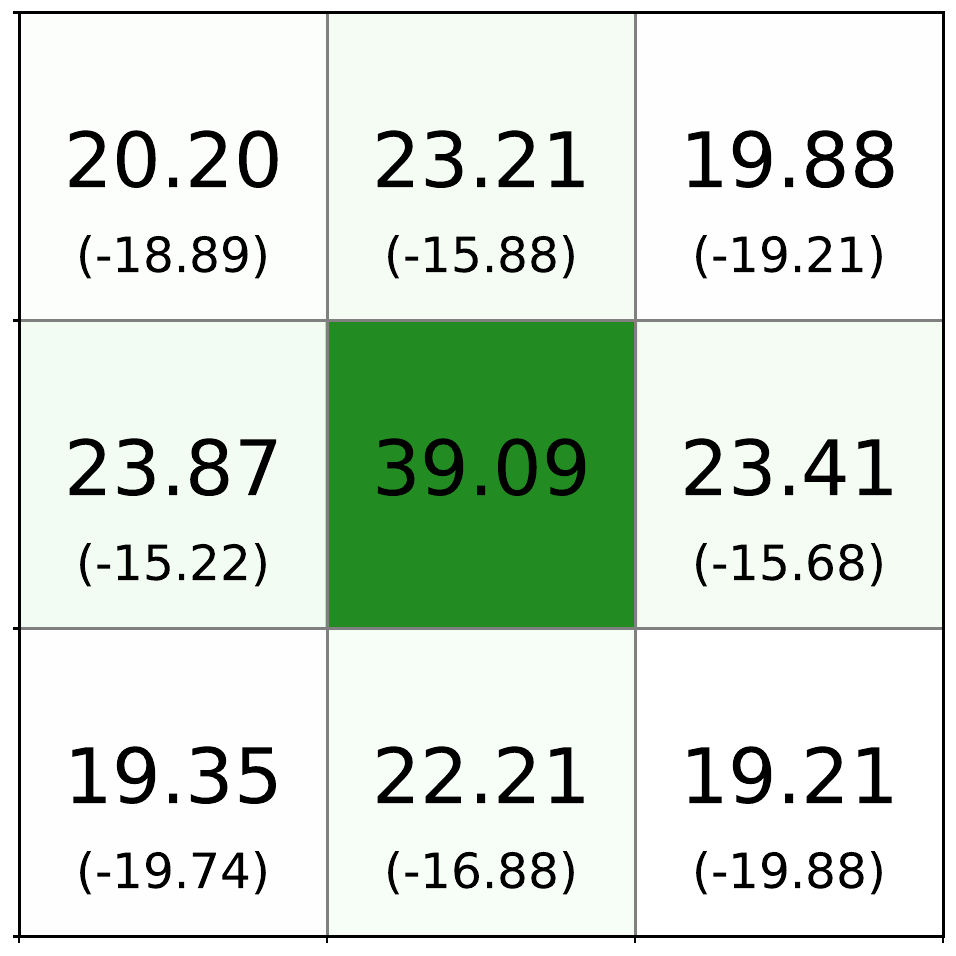}
}
\caption{\scriptsize{ImageNet-A}}
\label{fig:3x3_INA_1}
\end{subfigure}
\hfill
\begin{subfigure}{0.19\columnwidth}
\resizebox{\columnwidth}{!}{
\includegraphics[]{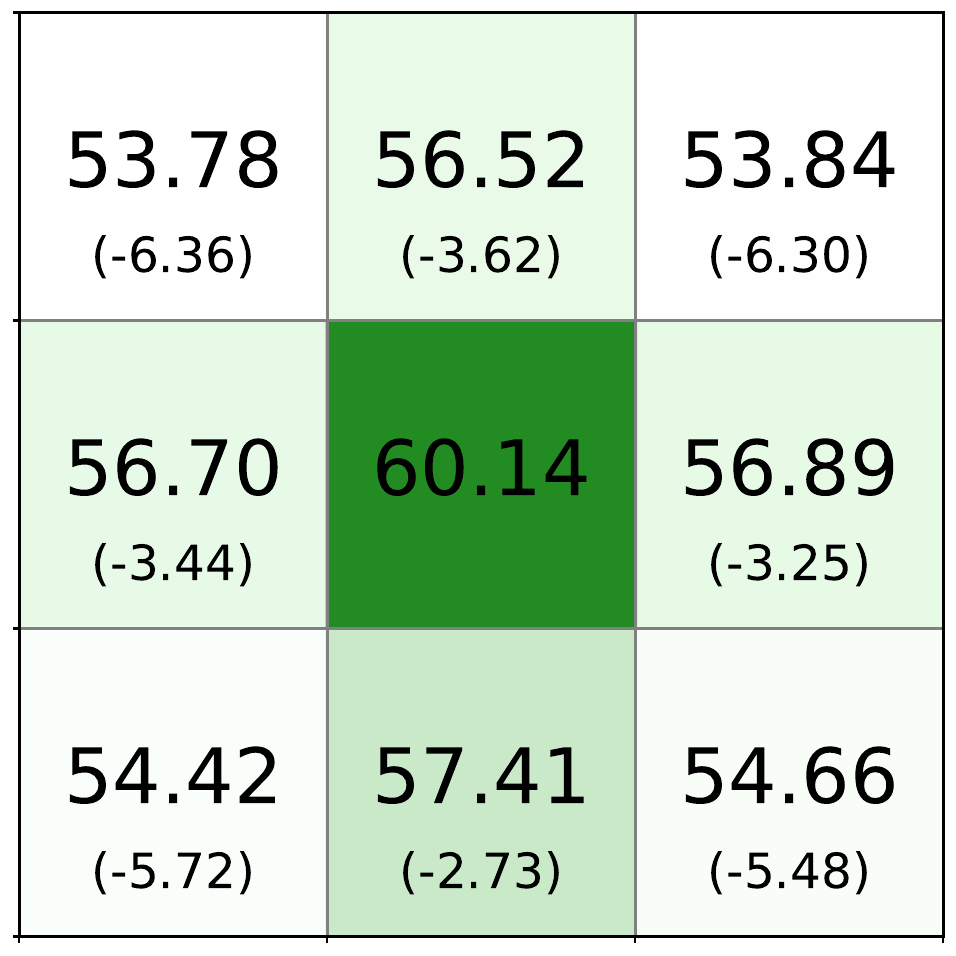}
}
\caption{\scriptsize{ImageNet-R}}
\label{fig:3x3_INR_1}
\end{subfigure}
\hfill
\begin{subfigure}{0.19\columnwidth}
\resizebox{\columnwidth}{!}{
\includegraphics[]{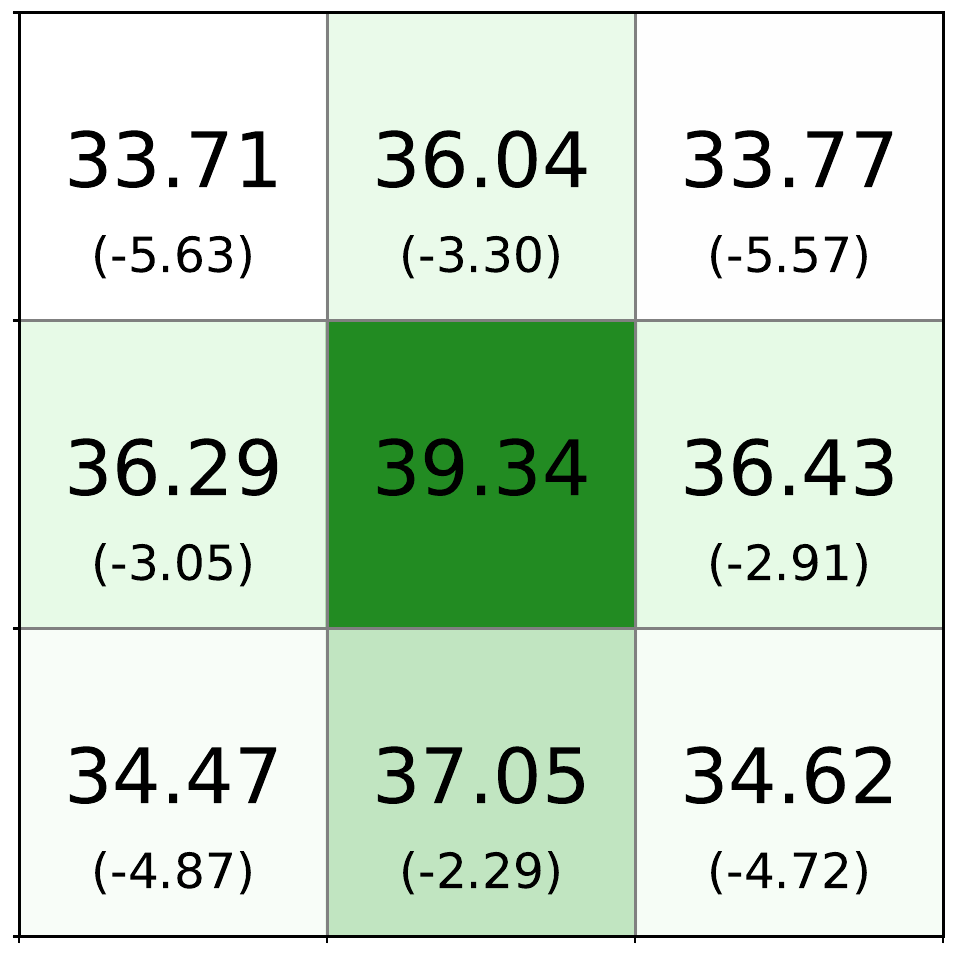}
}
\caption{\scriptsize{ImageNet-Sketch}}
\label{fig:3x3_INS_1}
\end{subfigure}
\hfill
\begin{subfigure}{0.19\columnwidth}
\resizebox{\columnwidth}{!}{
\includegraphics[]{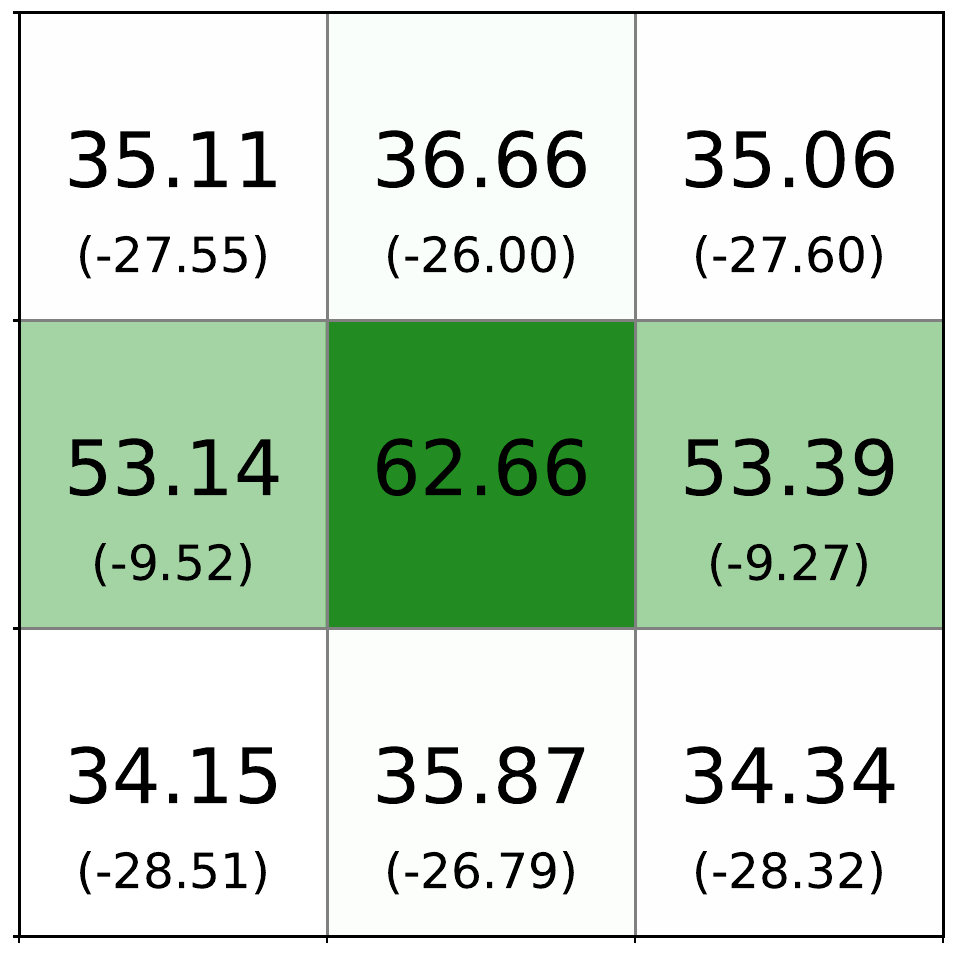}
}
\caption{\scriptsize{ObjectNet}}
\label{fig:3x3_ON_1}
\end{subfigure}
\caption{ ResNet-18  }
\label{fig:anchor_based_analysis_all_resnet_18}
\end{figure}

\begin{figure}[htb!]
\centering
\begin{subfigure}{0.19\columnwidth}
\resizebox{\columnwidth}{!}{
\includegraphics[]{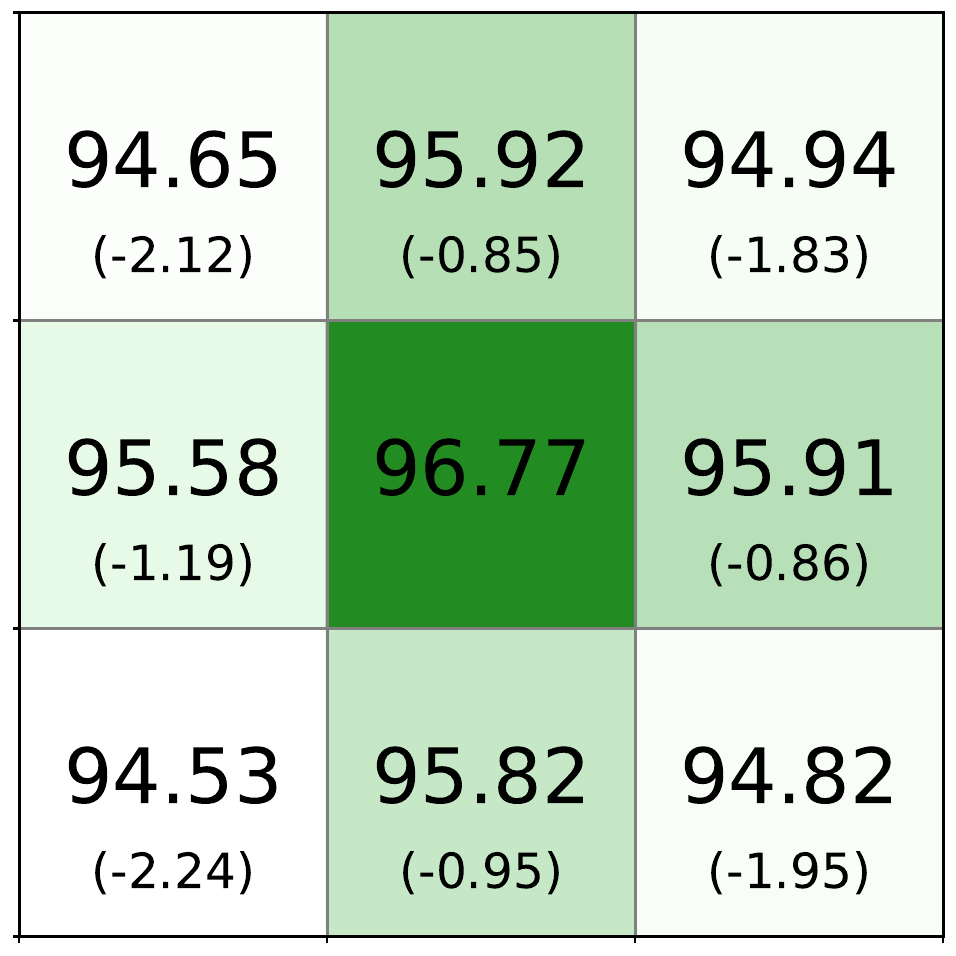}

}
\caption{\scriptsize{ImageNet-ReaL}}
\label{fig:3x3_REAL_2}
\end{subfigure}
\hfill
\begin{subfigure}{0.19\columnwidth}
\resizebox{\columnwidth}{!}{
\includegraphics[]{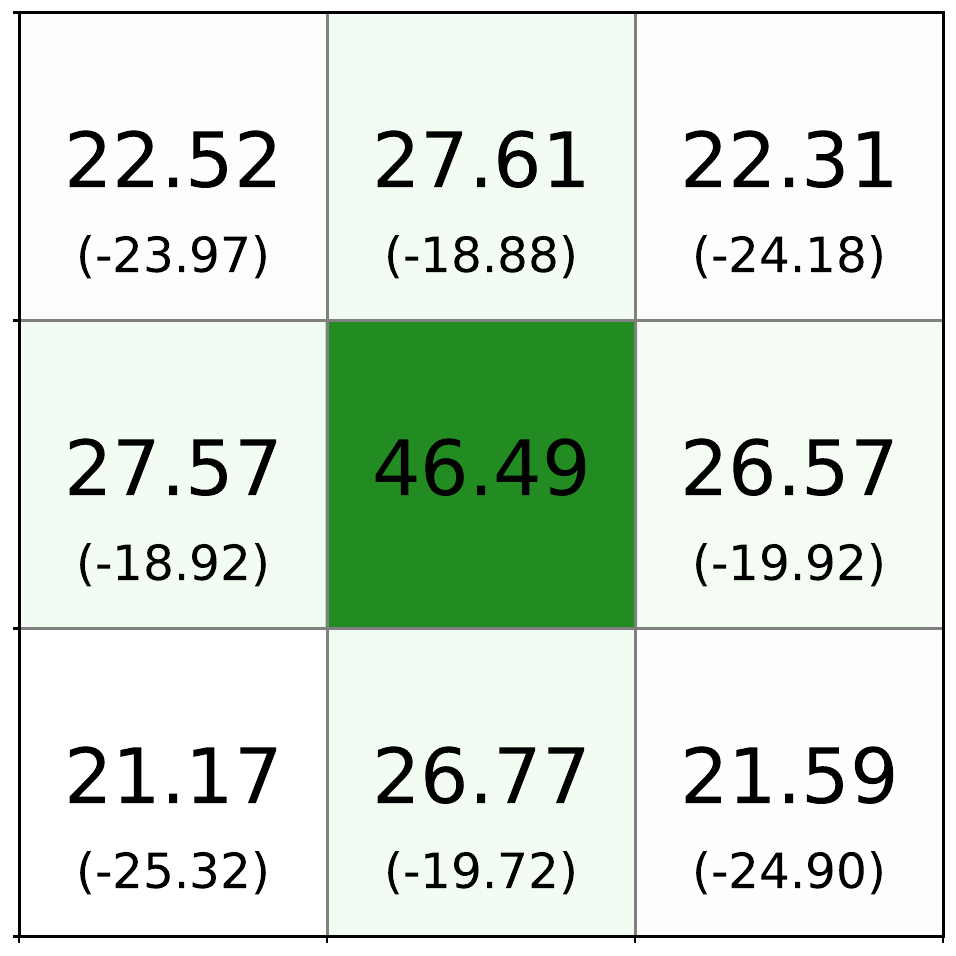}
}
\caption{\scriptsize{ImageNet-A}}
\label{fig:3x3_INA_2}
\end{subfigure}
\hfill
\begin{subfigure}{0.19\columnwidth}
    \resizebox{\columnwidth}{!}{
\includegraphics[]{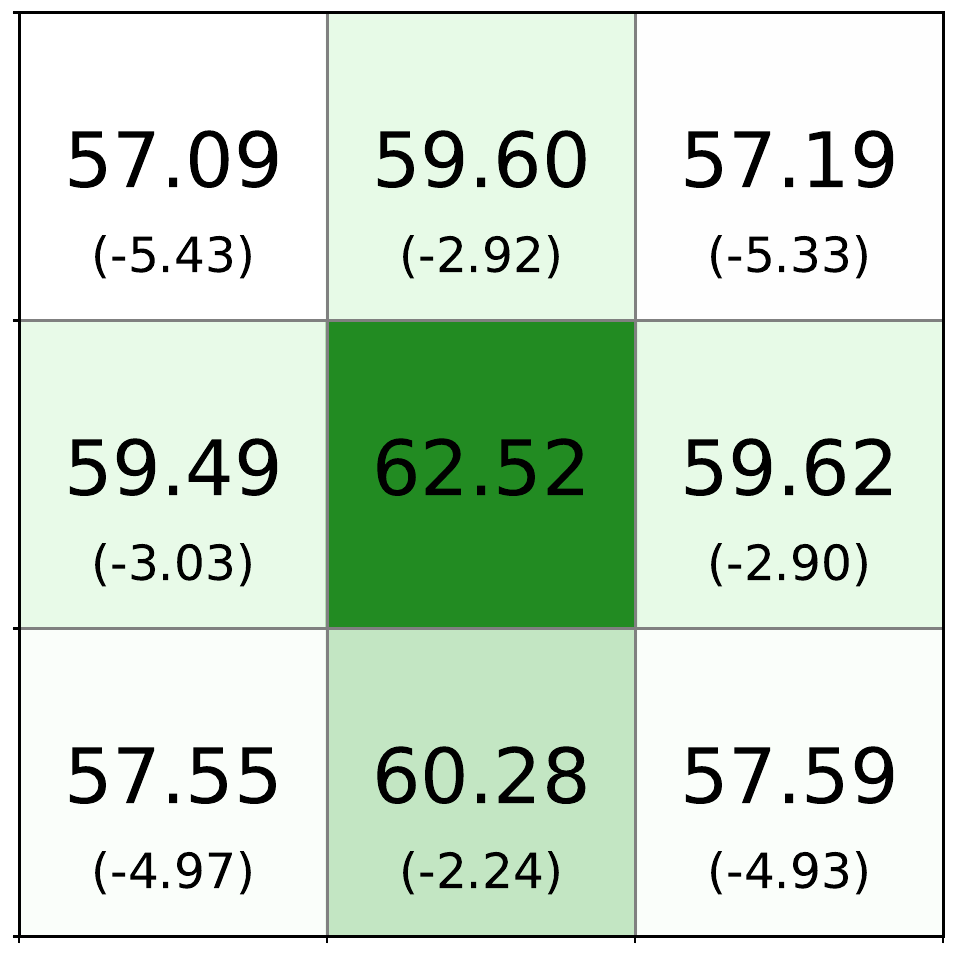}
}
\caption{\scriptsize{ImageNet-R}}
\label{fig:3x3_INR_2}
\end{subfigure}
\hfill
\begin{subfigure}{0.19\columnwidth}
    \resizebox{\columnwidth}{!}{
\includegraphics[]{rebuttal/heatmaps/resnet50/ImageNet_Sketch.pdf}
}
\caption{\scriptsize{ImageNet-Sketch}}
\label{fig:3x3_INS_2}
\end{subfigure}
\hfill
\begin{subfigure}{0.19\columnwidth}
    \resizebox{\columnwidth}{!}{
\includegraphics[]{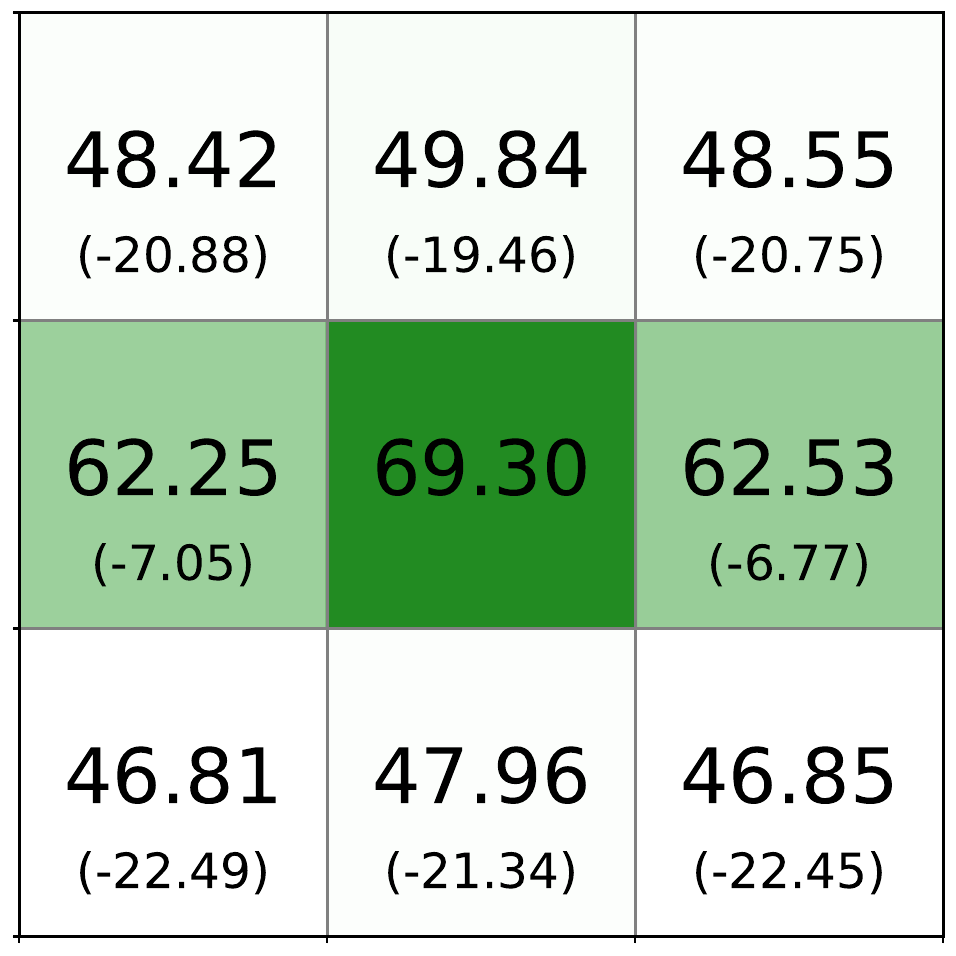}
}
\caption{\scriptsize{ObjectNet}}
\label{fig:3x3_ON_2}
\end{subfigure}
\caption{ ResNet-50  }
\label{fig:anchor_based_analysis_all_resnet_50}
\end{figure}

\begin{figure}[htb!]
\centering
\begin{subfigure}{0.19\columnwidth}
\resizebox{\columnwidth}{!}{
\includegraphics[]{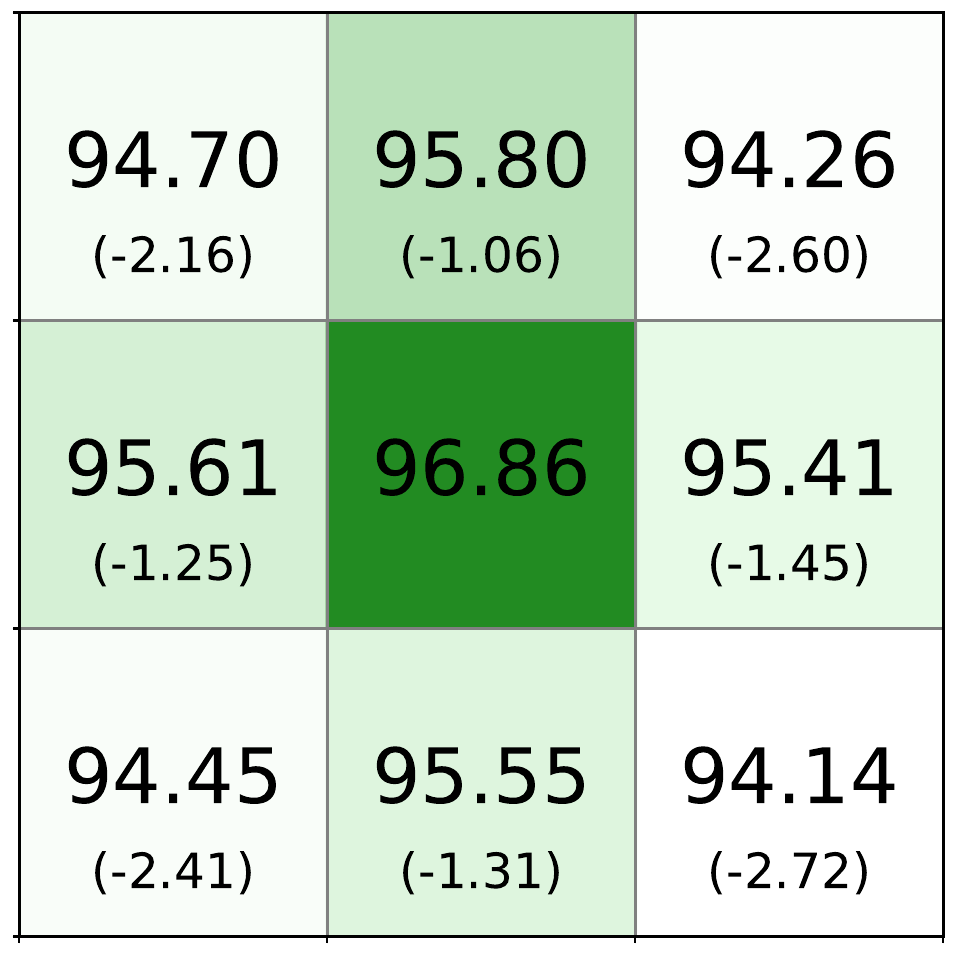}
}
\caption{\scriptsize{ImageNet-ReaL}}
\label{fig:3x3_REAL_3}
\end{subfigure}
\hfill
\begin{subfigure}{0.19\columnwidth}
\resizebox{\columnwidth}{!}{
\includegraphics[]{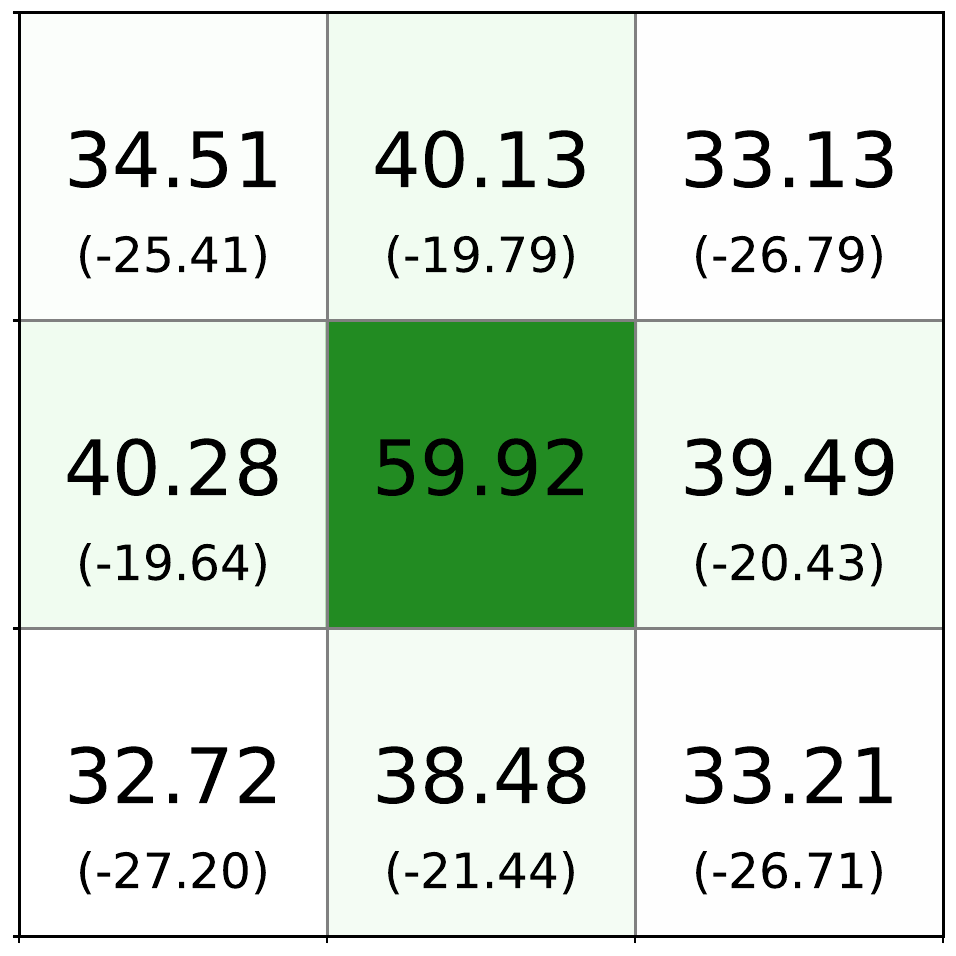}

}
\caption{\scriptsize{ImageNet-A}}
\label{fig:3x3_INA_3}
\end{subfigure}
\hfill
\begin{subfigure}{0.19\columnwidth}
\resizebox{\columnwidth}{!}{
\includegraphics[]{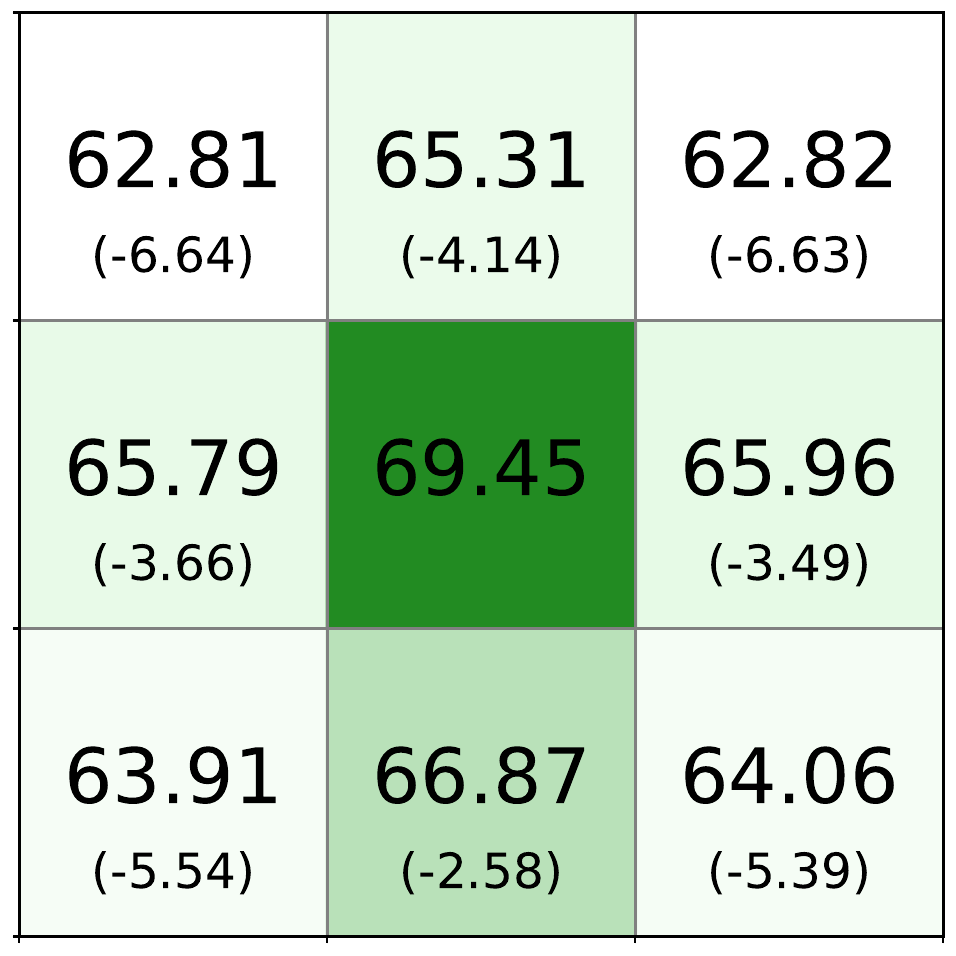}
}
\caption{\scriptsize{ImageNet-R}}
\label{fig:3x3_INR_3}
\end{subfigure}
\hfill
\begin{subfigure}{0.19\columnwidth}
\resizebox{\columnwidth}{!}{
\includegraphics[]{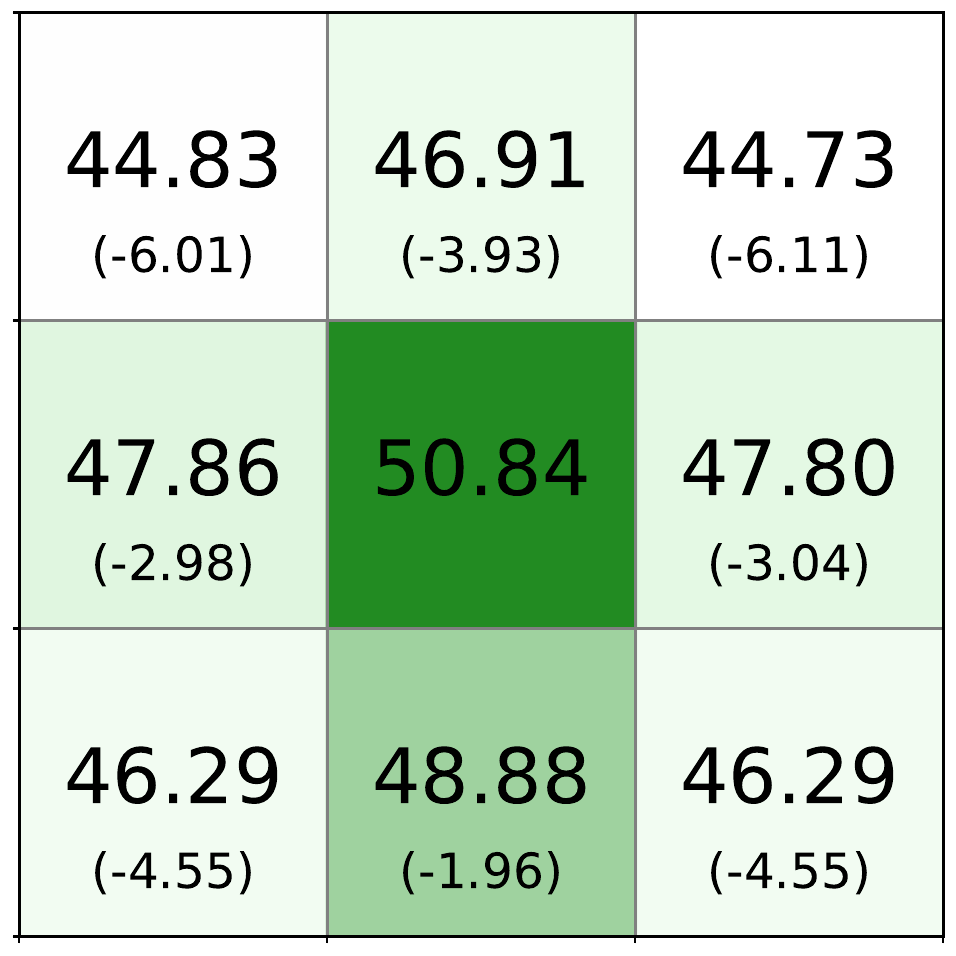}
}
\caption{\scriptsize{ImageNet-Sketch}}
\label{fig:3x3_INS_3}
\end{subfigure}
\hfill
\begin{subfigure}{0.19\columnwidth}
\resizebox{\columnwidth}{!}{
\includegraphics[]{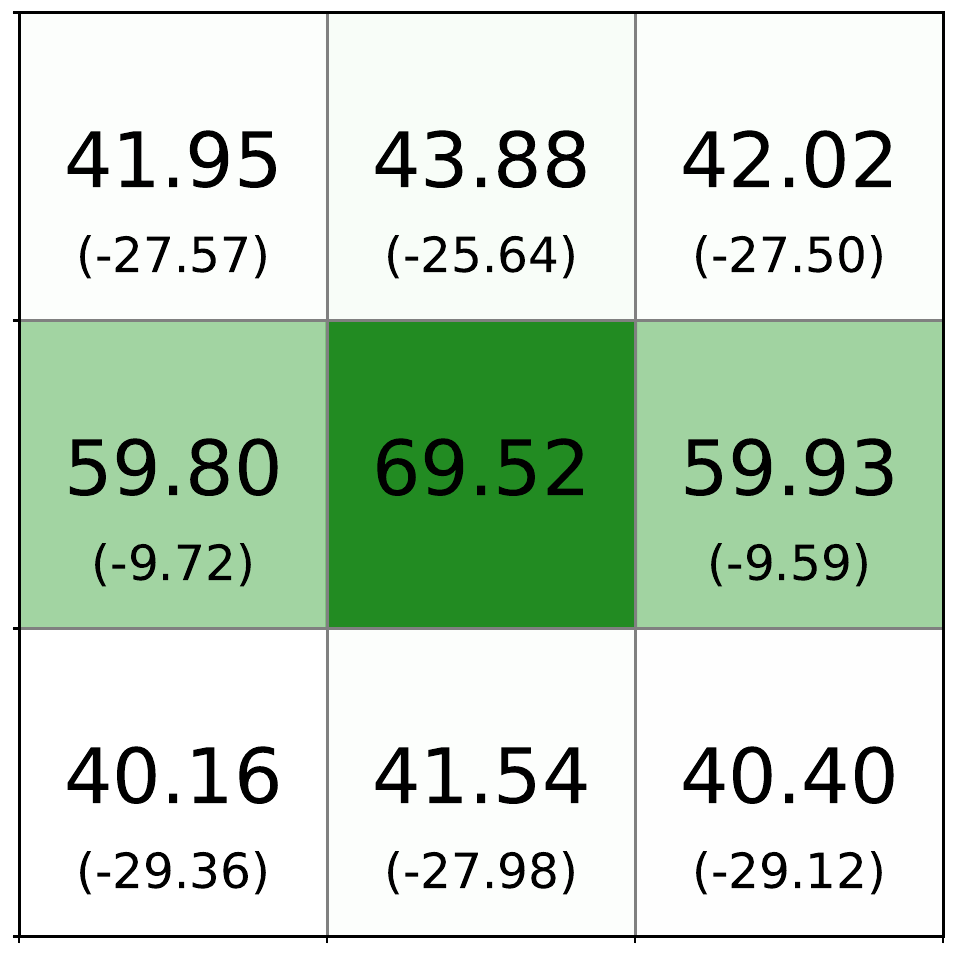}
}
\caption{\scriptsize{ObjectNet}}
\label{fig:3x3_ON_3}
\end{subfigure}
\caption{  ViT-B/32  }
\label{fig:anchor_based_analysis_all_vit_b_32}
\end{figure}

\begin{figure}[htb!]
\centering
\begin{subfigure}{0.19\columnwidth}
\resizebox{\columnwidth}{!}{
\includegraphics[]{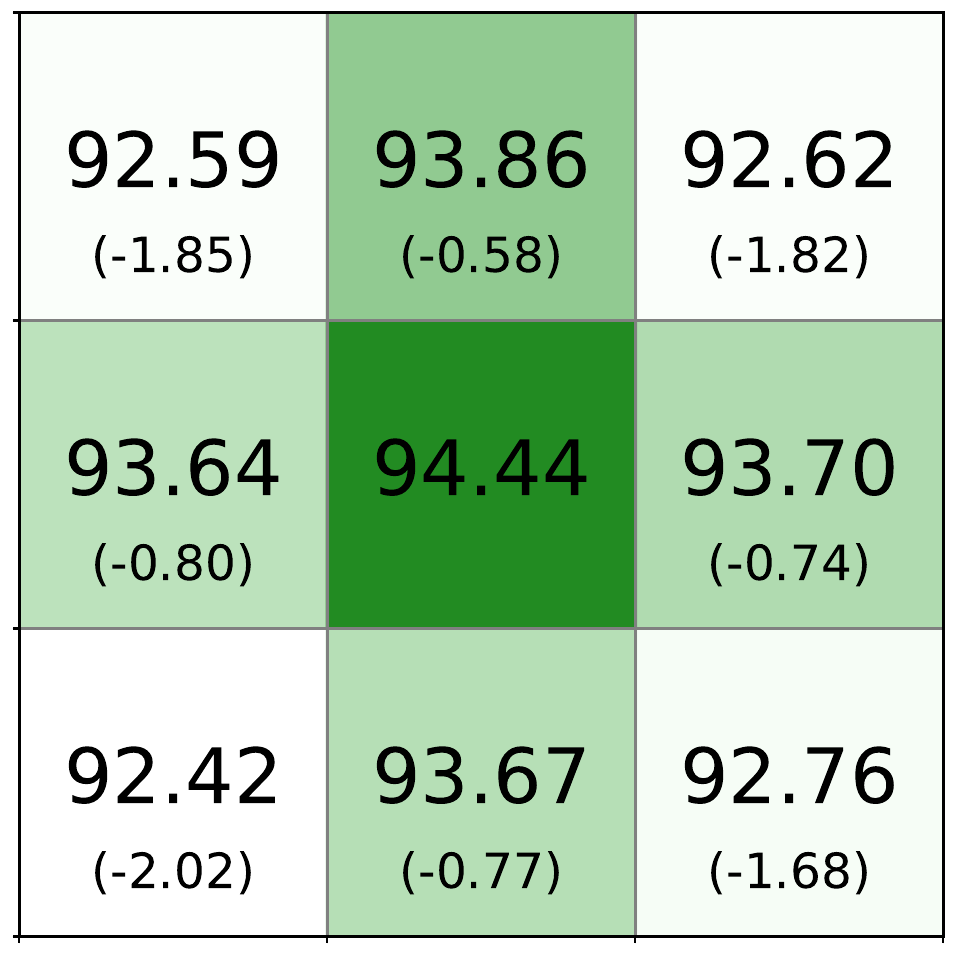}
}
\caption{\scriptsize{ImageNet-ReaL}}
\label{fig:3x3_REAL_6}
\end{subfigure}
\hfill
\begin{subfigure}{0.19\columnwidth}
\resizebox{\columnwidth}{!}{
\includegraphics[]{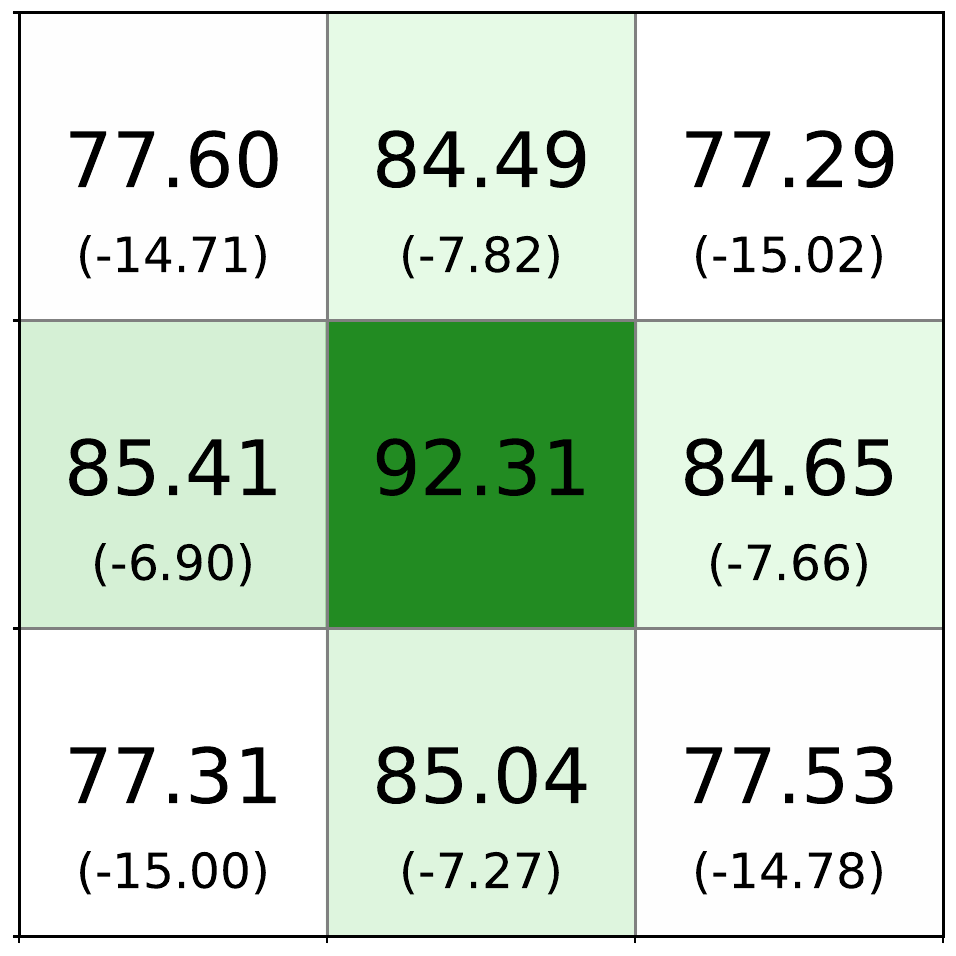}
}
\caption{\scriptsize{ImageNet-A}}
\label{fig:3x3_INA_6}
\end{subfigure}
\hfill
\begin{subfigure}{0.19\columnwidth}
\resizebox{\columnwidth}{!}{
\includegraphics[]{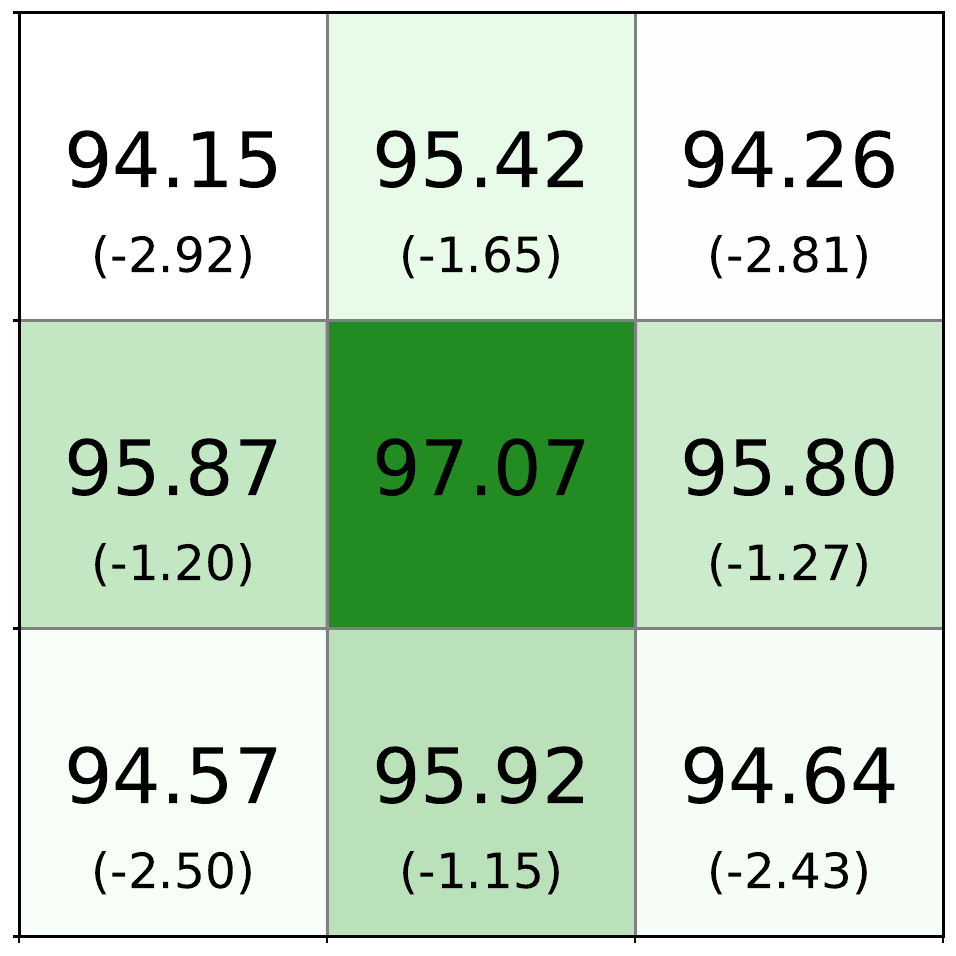}
}
\caption{\scriptsize{ImageNet-R}}
\label{fig:3x3_INR_6}
\end{subfigure}
\hfill
\begin{subfigure}{0.19\columnwidth}
\resizebox{\columnwidth}{!}{
\includegraphics[]{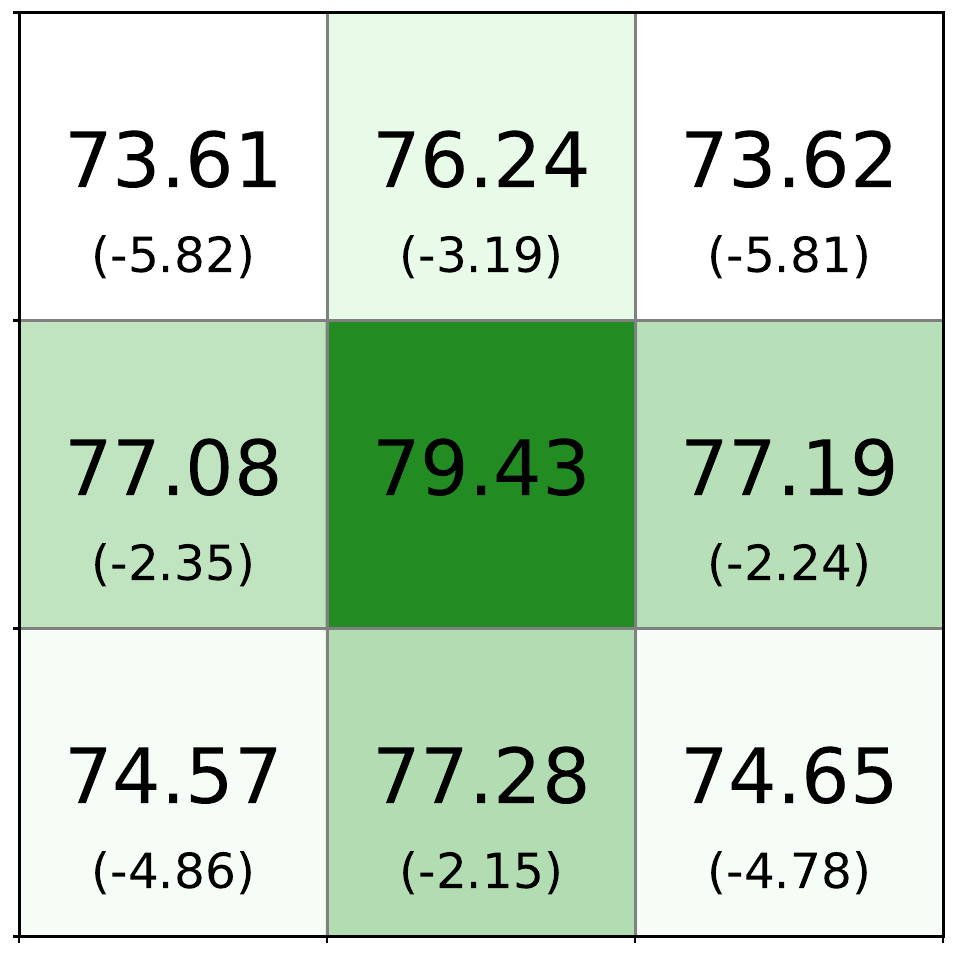}
}
\caption{\scriptsize{ImageNet-Sketch}}
\label{fig:3x3_INS_6}
\end{subfigure}
\hfill
\begin{subfigure}{0.19\columnwidth}
\resizebox{\columnwidth}{!}{
\includegraphics[]{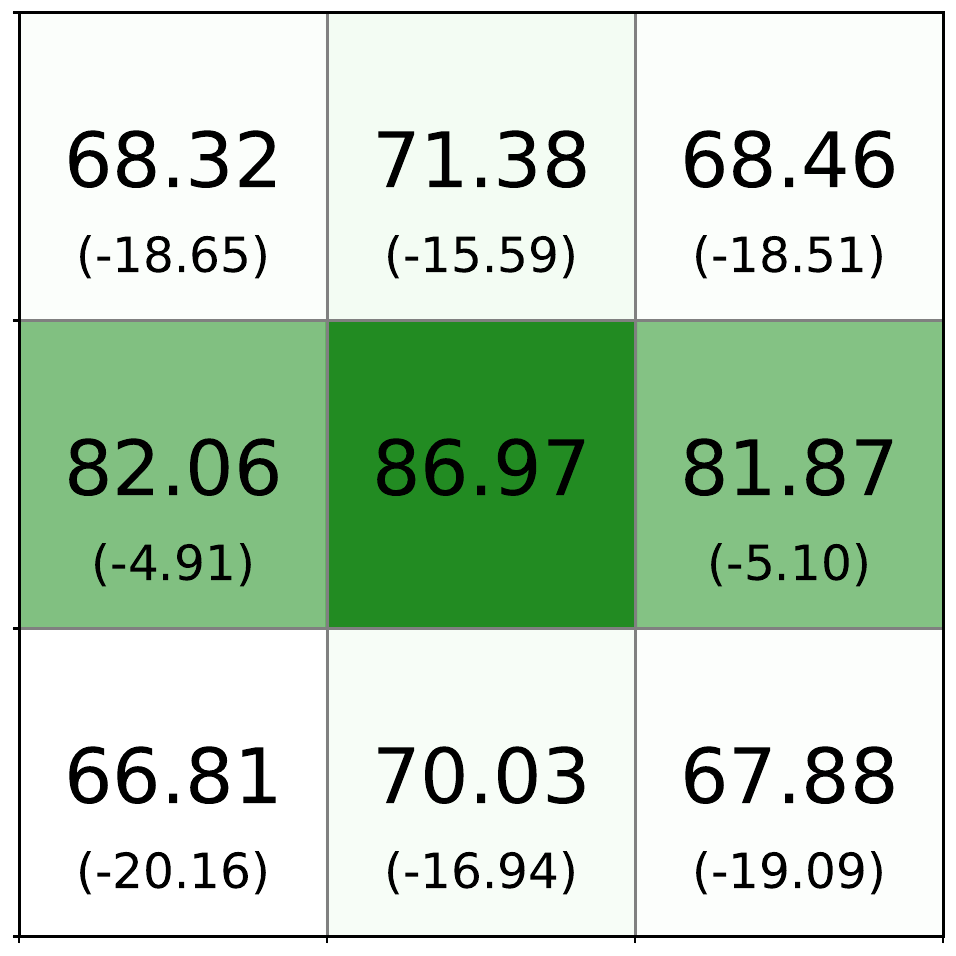}
}
\caption{\scriptsize{ObjectNet}}
\label{fig:3x3_ON_6}
\end{subfigure}
\caption{ CLIP-ViT-L/14 }
\label{fig:anchor_based_analysis_all_clipvit}
\end{figure}

\subsection{Distribution of the minimum set cover per classifier and dataset}

In this section, we provide details on the distribution of minimum set cover size.

\begin{table}[htb!]
\caption{Distribution of the minimum set cover per classifier and dataset. (ZI: \zoomin, ZO: \zoomout, ZL: \zoomless)}
\label{supptab:min_cover_all}
\resizebox{\textwidth}{!}{%
\begin{tabular}{@{}lrrrrrrrrrrrrrrrrrrrr@{}}
\toprule
 &
  \multicolumn{4}{c}{ReaL} &
  \multicolumn{4}{c}{IN-A} &
  \multicolumn{4}{c}{IN-R} &
  \multicolumn{4}{c}{IN-Sketch} &
  \multicolumn{4}{c}{ON} \\ \midrule
 &
  \multicolumn{1}{c}{ZI} &
  \multicolumn{1}{c}{ZO} &
  \multicolumn{1}{c}{ZL} &
  \multicolumn{1}{c}{Total} &
  \multicolumn{1}{c}{ZI} &
  \multicolumn{1}{c}{ZO} &
  \multicolumn{1}{c}{ZL} &
  \multicolumn{1}{c}{Total} &
  \multicolumn{1}{c}{ZI} &
  \multicolumn{1}{c}{ZO} &
  \multicolumn{1}{c}{ZL} &
  \multicolumn{1}{c}{Total} &
  \multicolumn{1}{c}{ZI} &
  \multicolumn{1}{c}{ZO} &
  \multicolumn{1}{c}{ZL} &
  \multicolumn{1}{c}{Total} &
  \multicolumn{1}{c}{ZI} &
  \multicolumn{1}{c}{ZO} &
  \multicolumn{1}{c}{ZL} &
  \multicolumn{1}{c}{Total} \\ \cmidrule(l){2-21} 
ResNet-18 &
  160 &
  33 &
  8 &
  \multicolumn{1}{r|}{201} &
  174 &
  31 &
  6 &
  \multicolumn{1}{r|}{211} &
  204 &
  65 &
  9 &
  \multicolumn{1}{r|}{278} &
  209 &
  51 &
  9 &
  \multicolumn{1}{r|}{269} &
  191 &
  54 &
  9 &
  254 \\
ResNet-50 &
  136 &
  33 &
  9 &
  \multicolumn{1}{r|}{178} &
  165 &
  42 &
  7 &
  \multicolumn{1}{r|}{214} &
  200 &
  62 &
  9 &
  \multicolumn{1}{r|}{271} &
  216 &
  56 &
  9 &
  \multicolumn{1}{r|}{281} &
  187 &
  63 &
  9 &
  259 \\
ViT-B/32 &
  134 &
  30 &
  4 &
  \multicolumn{1}{r|}{168} &
  167 &
  19 &
  7 &
  \multicolumn{1}{r|}{193} &
  196 &
  52 &
  9 &
  \multicolumn{1}{r|}{257} &
  218 &
  46 &
  9 &
  \multicolumn{1}{r|}{273} &
  206 &
  58 &
  9 &
  273 \\
VGG-16 &
  158 &
  34 &
  9 &
  \multicolumn{1}{r|}{201} &
  181 &
  33 &
  8 &
  \multicolumn{1}{r|}{222} &
  214 &
  66 &
  9 &
  \multicolumn{1}{r|}{289} &
  210 &
  54 &
  9 &
  \multicolumn{1}{r|}{273} &
  198 &
  52 &
  9 &
  259 \\
AlexNet &
  191 &
  40 &
  8 &
  \multicolumn{1}{r|}{239} &
  170 &
  33 &
  9 &
  \multicolumn{1}{r|}{212} &
  212 &
  51 &
  9 &
  \multicolumn{1}{r|}{272} &
  217 &
  49 &
  9 &
  \multicolumn{1}{r|}{275} &
  201 &
  58 &
  9 &
  268 \\
\clip-ViT-L/14 &
  141 &
  48 &
  8 &
  \multicolumn{1}{r|}{197} &
  75 &
  14 &
  4 &
  \multicolumn{1}{r|}{93} &
  76 &
  33 &
  5 &
  \multicolumn{1}{r|}{114} &
  142 &
  61 &
  9 &
  \multicolumn{1}{r|}{212} &
  205 &
  66 &
  9 &
  280 \\ \bottomrule
\end{tabular}%
}
\end{table}


\subsection{Only $70\%$ of all transforms are needed to reach maximum possible accuracy}
\label{sec:mincover}

In \cref{sec:maxpossibleaccuracy}, we first pre-define all $324$ zoom transforms and then compute the \emph{maximum} possible accuracy to ensure the predicted labels were the results of models looking at a controlled zoomed region (\ie not because a model was given $324$ arbitrary trials per image).
Here, we aim to compute the minimum number of zoom settings required for a model to reach the same upper-bound accuracy.
Evaluating this minimum set may reveal spatial biases of a dataset (\cref{sec:center_bias}) as well as the implicit zoom operation that a state-of-the-art model (\eg CLIP) may have learned.

\subsec{Experiment}
\rebuttal{Given a (dataset, classifier) pair, we constructed a bipartite graph \( G = (N, E) \), where \( N = A \cup B \), \( A \) represents the set of transforms, and \( B \) represents the set of images. The edges \( E \) are defined as follows:
\[
E = \{(n_i, n_j) \mid n_i \in A, n_j \in B, \text{ and transform } n_i \text{ leads to the correct classification of image } n_j\}
\]
We aim to find a minimum set cover~\cite{petr1996setcover,algo-design} in this graph, synonymous with finding a minimum subset of transforms among the $324$ that lead to the correct prediction for all classifiable images in \cref{sec:experimental_results_main} (\ie those that make up the accuracy scores in \cref{tab:1_main_results}c), without unnecessary transforms.
}

\rebuttal{
The resulting subset of transforms from the process leads to the correct prediction for all classifiable images without sacrificing accuracy. During each iteration of the greedy minimum set cover algorithm, the transform that yields the highest number of correct classifications for the remaining images is selected. This process continues until all of the images have been ``covered,'' \ie all images have connected to a transform with at least one edge. The result aligns with our initial goal to remove unnecessary zoom transforms while maintaining the maximum possible accuracy, as outlined in \cref{sec:experimental_results_main} (i.e., those that make up the accuracy scores in \cref{tab:1_main_results}c).
The outline of the algorithm can be seen in~\cref{algo:mincover}.
}

\begin{algorithm}
\caption{Greedy Minimum Set Cover for Transforms}
\begin{algorithmic}[1]
\STATE \textbf{Initialization:} \( C = \emptyset \) (Covered set of images), \( T = \emptyset \) (Selected transforms)
\WHILE{\( C \neq B \)}
\STATE Find \( n_i \in A \setminus T \) that maximizes \( |n_j \in B \setminus C \mid (n_i, n_j) \in E| \)
\STATE \( C = C \cup \{n_j \mid (n_i, n_j) \in E\} \)
\STATE \( T = T \cup \{n_i\} \)
\ENDWHILE
\STATE \textbf{Result:} The subset of transforms corresponding to \( T \) can classify images without sacrificing accuracy.
\end{algorithmic}
\label{algo:mincover}
\end{algorithm}



\begin{figure}[htb!]
\centering
    \caption{
    The minimum number of zoom transforms (out of 324) required to achieve the maximum possible accuracy scores reported in \cref{tab:1_main_results}c. 
    }
    \label{tab:min_cover_size}
    \setlength{\tabcolsep}{4pt}  
    \begin{tabular}{lrrrrrrr|r}
    \multicolumn{1}{l}{} & \multicolumn{1}{c}{IN} & \multicolumn{1}{c}{ReaL} & \multicolumn{1}{c}{\footnotesize{IN+ReaL}}& \multicolumn{1}{c}{\cellcolor{MyLightGray} IN-A} & \multicolumn{1}{c}{\cellcolor{MyLightGray} IN-R}  & \multicolumn{1}{c}{\cellcolor{MyLightGray} IN-S} & \multicolumn{1}{r}{\cellcolor{MyLightGray} ON} & $\mu$ \\ \midrule
    {\textcolor{specialcolor}{AlexNet}} & 255 & 239 & 246 & 212 & 272 & 275 & 268   &252 \\
    {\textcolor{specialcolor}{VGG-16}} & 242 & 201  & 201 & 222 & 289 & 273 & 259   &241 \\
    {\textcolor{specialcolor}{ResNet-18}} & 250 & 201 & 208 & 211 & 278 & 269 & 254 &239  \\
    {\textcolor{specialcolor}{ResNet-50}} & 234 & 178 & 183 & 214 & 271 & 281 & 259 &231 \\
    {\textcolor{specialcolor}{ViT-B/32}} & 233 & 168 & 173 & 193 & 257 & 273 & 273  &224 \\
    \textbf{\footnotesize{\clip}-ViT-L/14} & 251 & 197 & 186  & 93 & 114 & 280 & 212 & 190 \\ \bottomrule
    \end{tabular}%
\vspace*{-8pt}
\end{figure}

\subsec{Results} \cref{tab:min_cover_size} shows the minimum number of transforms per dataset required to reach the maximum possible accuracy.
Although this number varies depending on the dataset and classifier, on average, the size of the minimum cover is $229$, which is $\sim$70\% of all $324$ pre-defined transforms.

We evaluate the maximum possible accuracy using the top 36 transforms, the same number as the number of zoom scales and report the results in~\cref{tab:1_main_results}b. 
This set of transforms is achieved by stopping the algorithm after 36 iterations, which provided us with 36 high-performing transforms.
The maximum possible accuracy using only 36 crops is only slightly lower than that when using all $324$ crops but is substantially higher than the standard 1-crop, \eg $85.19\%$ vs. $56.16\%$ for AlexNet on IN (\cref{tab:1_main_results}b).
Also, the upper-bound accuracy for 36 crops being much higher than the random baseline (\ie $3.6\%$ for IN) confirms that the pre-defined zoom transforms are important to classification (not because models are given 36 random trials per image).
The top-36 zoom transforms for ResNet-50 on ImageNet contain zooms at various locations in the image (see the visualizations in \cref{supp:viz_36crop}).

Remarkably, CLIP requires 190 transforms on average, which is fewer than every other model (\cref{tab:min_cover_size}; $\mu$ column).
This can be attributed to either the implicit zoom power of CLIP or the fact it has a stronger feature extractor.

\subsection{Center-zooming increases the accuracy of all ImageNet-trained models but not CLIP }  
\label{sec:clip_zoom_invariance}

Previously, we have found that CLIP obtains the best accuracy on all six datasets (\cref{tab:1_main_results}a) and also requires the smallest minimum set of zoom transforms to obtain the upper-bound accuracy (\cref{sec:mincover}).
It is important to understand what classification strategy a CLIP classifier internally performs to classify better.
Here, we test the hypothesis that the state-of-the-art CLIP is already performing an implicit zoom on images.
If that is true, directly zooming to the center, exploiting the strong center bias of ImageNet-A and ObjectNet, will not improve CLIP accuracy.

\begin{figure}[!htb]
     \setkeys{Gin}{width=\linewidth}
     \begin{subfigure}[t]{0.4895\columnwidth}
         \includegraphics{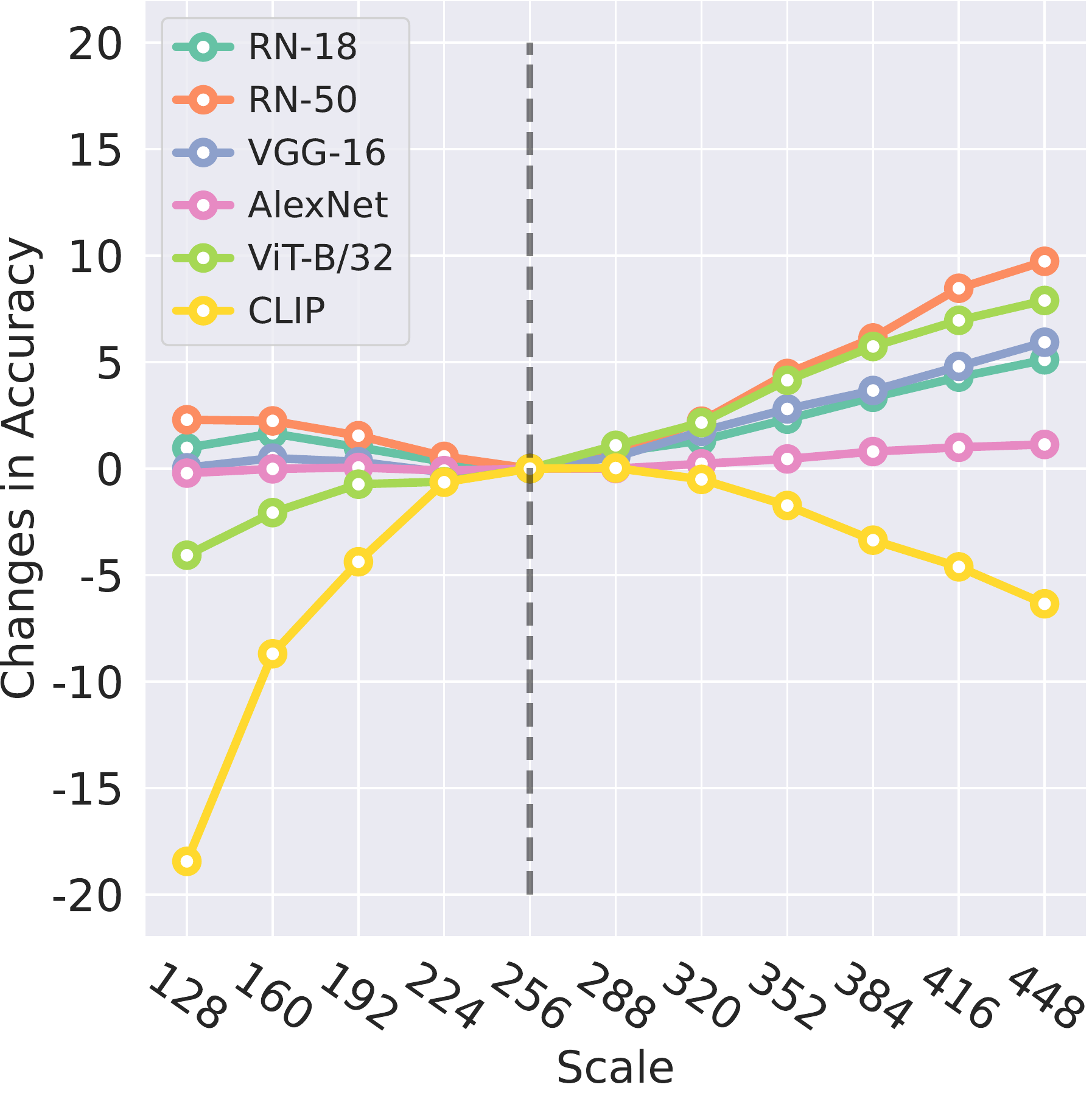}
         \caption{ImageNet-A}
     \end{subfigure}
     \hfill
     \begin{subfigure}[t]{0.4895\columnwidth}
         \centering
         \includegraphics{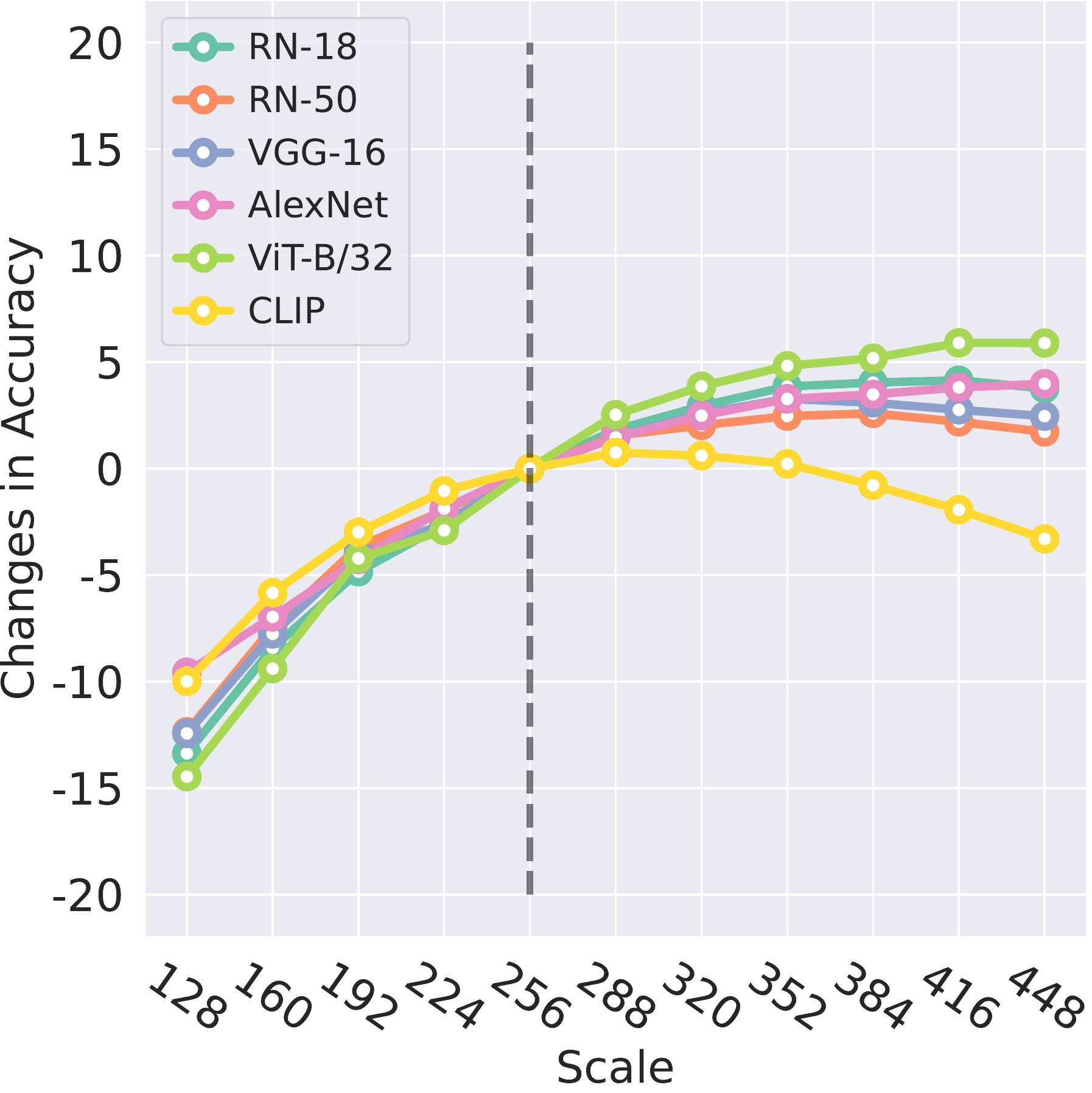}
         \caption{ObjectNet}

     \end{subfigure}
        \caption{
        Absolute changes in the top-1 accuracy (\%) of 6 models on ImageNet-A (a) and ObjectNet (b) when center-zooming images at various scales.
        Interestingly, center-zooming helps IN-trained networks but hurts CLIP.
        }
        \label{fig:clip_is_less_sensetive}
\end{figure}

\subsec{Experiment}
We evaluate the accuracy of all models when center-zooming on IN-A and ON images at 11 different scales $S \in \{128, 160, 192, ..., 448 \}$ (\cref{fig:clip_is_less_sensetive}).
That is, center-zooming at $S$ first resizes the input image so that the smaller dimension becomes $S$ and then takes a 224$\times$224 center crop (zero-padding is applied when necessary).

\subsec{Results} 
In \cref{fig:clip_is_less_sensetive}, we show the changes in the top-1 accuracy (1-crop) when varying the center-zoom scales away from the default ImageNet transform scale ($S=256$) for both ImageNet-A and ObjectNet.
While IN-trained networks exhibit consistent improvement as the zoom scale increases, CLIP shows a monotonic decrease in performance (\cref{fig:clip_is_less_sensetive}; yellow curves decreasing on both sides of $S = 256$).
This result is surprising but consistent with our hypothesis that CLIP internally performs implicit zooming to reach its peak accuracy and therefore manually zooming (either in or out) at the center mostly ruins its performance.

\subsection{Zoom-in is more useful than zoom-out, which is most important to abstract images }
\label{sec:disentagling_zoom}

Zooming in enhances texture patterns while zooming out provides a better perspective of the object's shape, which is known to be useful to image classification \cite{chen2020shape,geirhos2018imagenettrained}. 
Results in \cref{sec:maxpossibleaccuracy,sec:mincover} indicate that this combined zooming approach can be effective in classifying images from diverse datasets. Here, we test which dataset and model pairs require which type of zoom, and whether zooming in or out is always necessary. 

\subsec{Experiment}
To better understand the effectiveness of each zoom group, we calculate the maximum possible accuracy using all nine locations and different zoom scales $S$ to show per-dataset trends.
Additionally, we examined the percentage of images within each dataset that required a specific zoom group to be accurately classified. This analysis allowed us to gain a more comprehensive understanding of the role that each zoom group played in reaching the maximum possible accuracy reported in \cref{tab:1_main_results}.

\begin{figure}[tb]
     \setkeys{Gin}{width=\linewidth}
     \begin{subfigure}[t]{0.495\columnwidth}
         \includegraphics{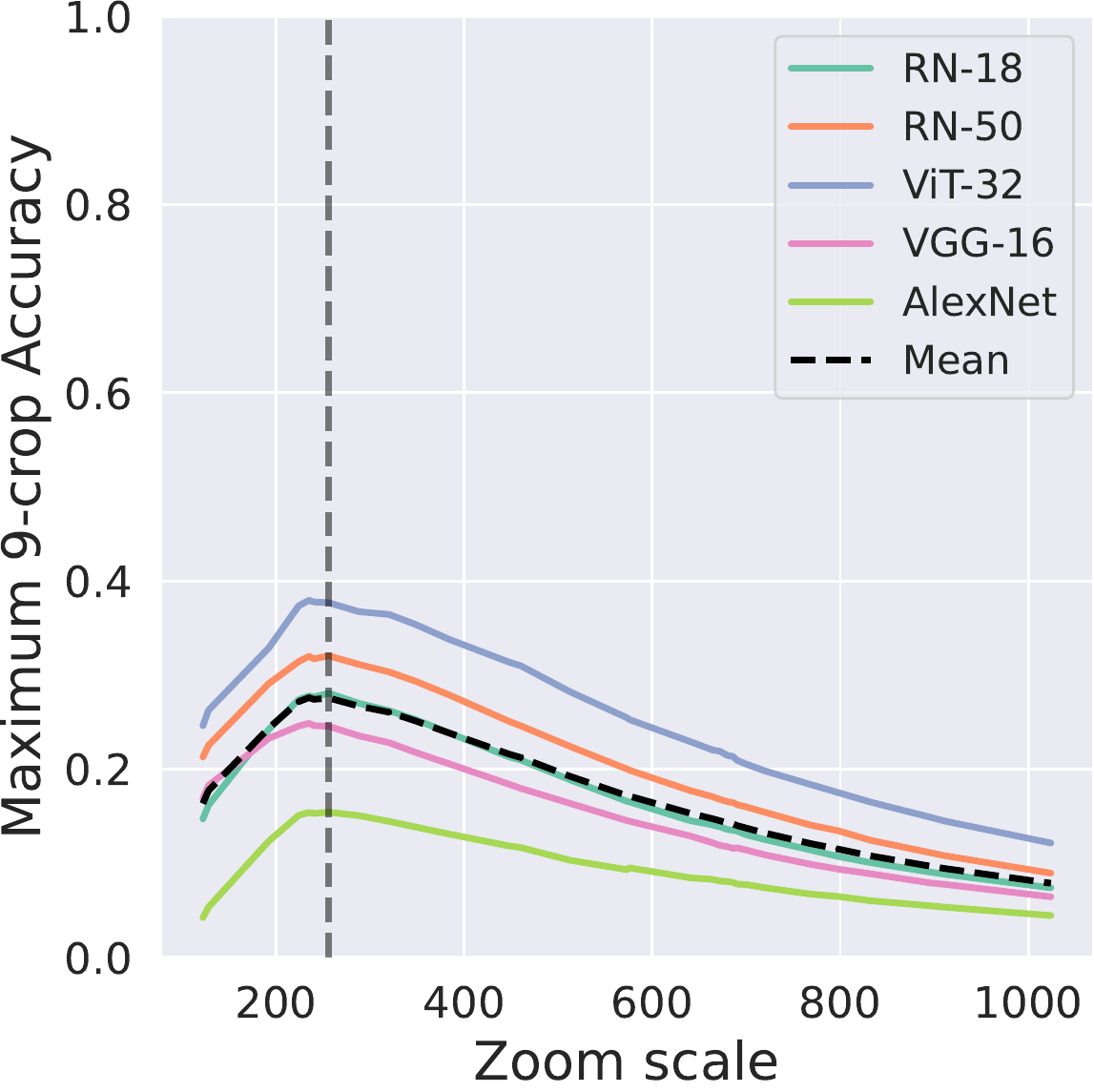}
         \caption{ImageNet-Sketch}
     \end{subfigure}
     \hfill
     \begin{subfigure}[t]{0.495\columnwidth}
         \centering
         \includegraphics{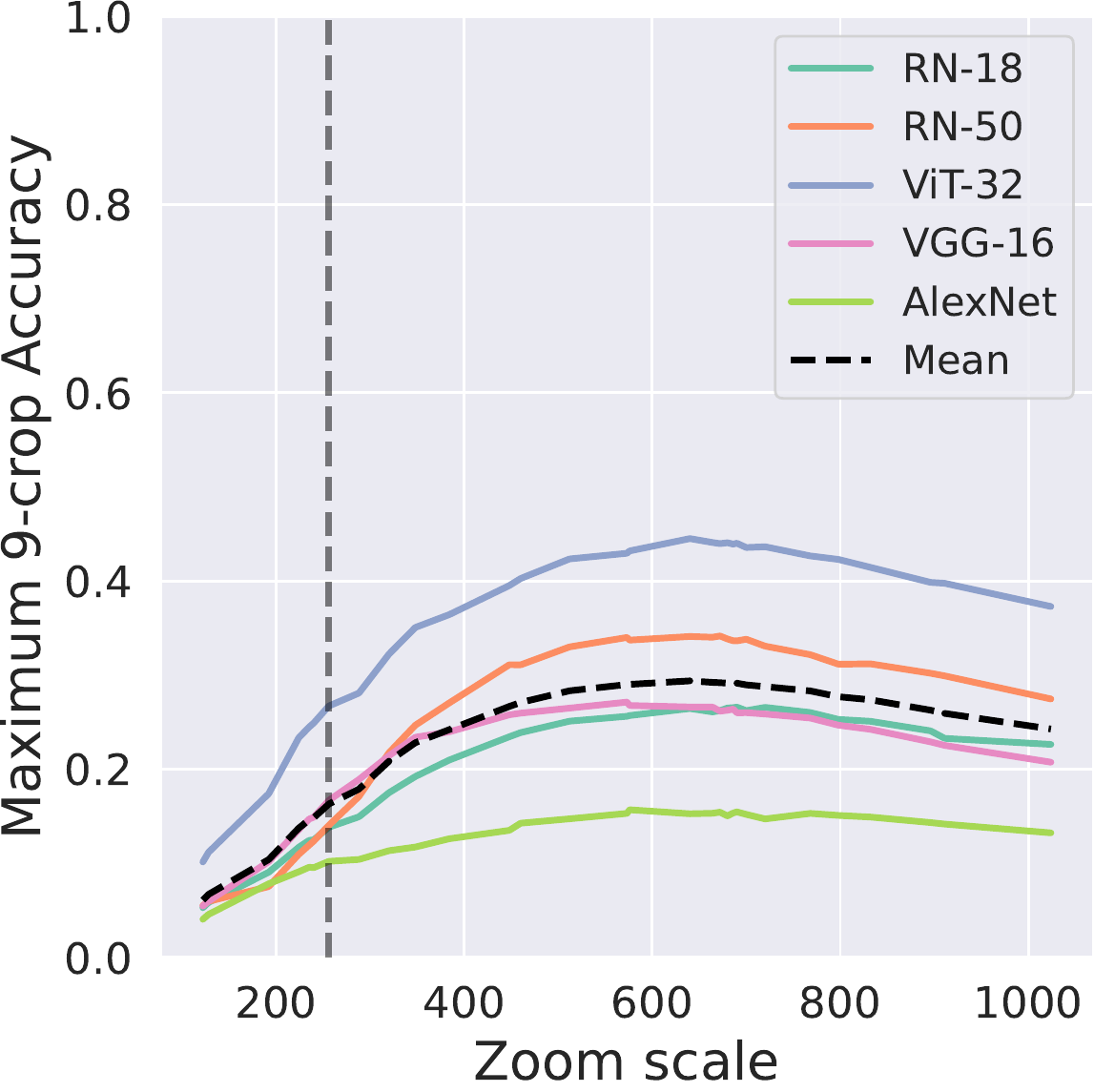}
         \caption{ImageNet-A}

     \end{subfigure}

        \caption{Maximum possible accuracy using nine crops at varying scales. 
        The vertical line  represents the standard ImageNet zoom scale ($S=256$).
        While for ImageNet-Sketch (a), zooming out marginally improves the accuracy, for scale factors larger than $256$, ImageNet-A (b) exhibits an increase in accuracy. 
        See \cref{suppsec:waterfallplots}  for details. 
        }
        \label{fig:waterfall_maintext}
\end{figure}

\subsec{Results}
The maximum possible accuracy for different zoom scales reveals a clear trend for each dataset.
For instance, a slight zoom-\textbf{out} enhances accuracy for abstract image datasets like IN-Sketch (\cref{fig:waterfall_maintext}a).
Conversely, for adversarial image datasets such as IN-A, zooming \textbf{in} improves accuracy (\cref{fig:waterfall_maintext}b)
This pattern is also evident in evaluations using standard 1-crop accuracy (\cref{suppsec:waterfallplots}).

Furthermore, the percentage of images that are \emph{exclusively} classifiable with the \zoomin group is consistently higher than the other two groups, \ie using ViT-B/32 $51.75\%$ on IN-A, and $13.11\%$ on IN-S (\cref{tab:average_zoom_in_out_over_datasets}a).
This shows that most datasets necessitate focusing on the object of interest in the image to both see texture patterns better and reduce background clutter (see \cref{supptab:zoomgroup} for full results).

However, we also find that the \zoomout group is also necessary for the correct classification of a small portion of each dataset. 
For instance, $1.22\% - 2.97$\% of IN-S images (\cref{tab:average_zoom_in_out_over_datasets}b) require a \zoomout transform to be correctly labeled (\ie \zoomin does not help at all).

\begin{table}[htb!]
\centering
\caption{\% of images in the entire dataset that require a particular zoom group to be classified correctly. See \cref{supptab:zoomgroup} for full results.}
\label{tab:average_zoom_in_out_over_datasets}
\begin{tabular}{@{}lrrrrrr@{}}
\cmidrule(l){2-7}
 & \multicolumn{2}{c}{\textbf{\zoomin} (a)}                & \multicolumn{2}{c}{\textbf{\zoomout} (b)}               & \multicolumn{2}{c}{\textbf{\zoomless} (d)}              \\ \cmidrule(l){2-7} 
 & \multicolumn{1}{c}{IN-A} & \multicolumn{1}{c}{IN-S} & \multicolumn{1}{c}{IN-A} & \multicolumn{1}{c}{IN-S} & \multicolumn{1}{c}{IN-A} & \multicolumn{1}{c}{IN-S} \\ \midrule
{ResNet-18}    & 40.67& 11.83& 0.92& 1.77& 0.19& 0.36  \\
{ResNet-50}    & 48.72& 10.99& 1.00& 2.23& 0.23& 0.24  \\
{ViT-B/32 }    & 51.75& 13.11& 0.85& 1.83& 0.15& 0.36  \\
{VGG-16   }    & 38.24& 9.47& 0.93& 2.97& 0.29& 0.28  \\
{AlexNet  }    & 26.27& 11.26& 1.08& 1.22& 0.36& 0.33  \\
\clip-ViT-L/14 & 12.01& 6.64& 0.44& 2.38& 0.05& 0.12  \\ \bottomrule
\end{tabular}%
\end{table}

\subsection{Simple aggregation of the zoom transforms can improve accuracy on some datasets but not all}
\label{supsec:naive_aggregation}

\cref{sec:maxpossibleaccuracy,sec:clip_zoom_invariance} show that using the same feature extractors (even as old as AlexNet), it is possible to achieve higher image classification accuracy if we know where to zoom and at which scale.
A practical follow-up question is: How to build a classifier that knows how to zoom given a test image?
In this section, we establish simple baselines that aggregate predictions over a set of zoom transforms.


\subsec{Experiment} 
We employ the \textbf{\layer{mean}} method from prior work~\cite{sato2015apac, lyzhov2020greedy}, and the \textbf{\layer{max}} method to aggregate output marginal distributions.  
For a given image, we get $N$ output distributions over classes from a classifier, in which $N$ is the total number of used transforms.
The aggregation process combines these $N$ distributions and outputs a final prediction for the given image.
In the aggregation step, we use the \layer{mean} or \layer{max} method to infer the final confidence for each class along $N$ distributions.
Finally, we select the class that has the highest confidence score.
Additionally, we test 5-crop and 10-crop evaluation \cite{krizhevsky2012imagenet, simonyan2014very, he2016deep} and compare them with our methods.
We use the transforms in the minimum set found for IN-ReaL to evaluate the remaining datasets.
The purpose is to reduce the number of augmentations and prevent training on OOD benchmarks.

\subsec{Results} 
\layer{max} aggregation of zoom-in transforms results in the largest improvements on ImageNet-A.
That is, on IN-A, ViT-B/32 reaches a top-1 accuracy of $24.69\%$ \increase{15.05} (\cref{tab:aggregation_main_text,supptab:aggregation_methods_extended}) and a ResNet-50 accuracy increases by \increasenoparent{13.03} points from $16.62\%$ to $29.65\%$ (\cref{sec:zooming_vs_imagenet_a})--a surprisingly strong baseline for future studies.
On ObjectNet, \layer{max} aggregation of zoom-in transforms also yields \increasenoparent{1.99} improvement over the 1-crop ViT-B/32 baseline.

On the other hand, \layer{mean} aggregation results in smaller but more consistent improvements over the 1-crop baseline for many datasets (\increasenoparent{3.56} on IN, \increasenoparent{4.08} on ReaL, \increasenoparent{4.65} on IN-A, and \increasenoparent{3.03} on ON; \cref{tab:aggregation_main_text}).
\layer{mean} aggregation (\cref{tab:aggregation_main_text}b) also outperforms the standard 5-crop and 10-crop \cite{krizhevsky2012imagenet,he2016deep} aggregation on these four datasets (\cref{tab:aggregation_main_text}e--f).

In contrast, for all 6 datasets, aggregating zoom-out and \zoomless transforms consistently worsen the performance over the 1-crop baseline (\cref{tab:aggregation_main_text}c--d).
That is, we find that for a few dozen images (\eg  sketches and abstract visuals; \cref{fig:teaser}c), interestingly, only zooming out can lead to a correct classification (\cref{sec:disentagling_zoom}), yet for most images in these 6 benchmarks, zooming out hurts the accuracy.

In summary, based on the insights from \cref{sec:maxpossibleaccuracy}, showing that zooming could help classification, we find that simple methods for aggregating zoom-in transforms at test-time can directly improve model accuracy over the 1-crop and \zoomless baselines on four benchmarks, \ie all except IN-R and IN-S, which contain abstract images.

\begin{table}[!htb]
\caption{Top-1 accuracy (\%) of aggregation methods on an IN-trained ViT-B/32 model.
Compared to the 1-crop baseline, aggregating zoom-in transforms consistently yields \textcolor{ForestGreen}{improved} accuracy on IN-A, ON but \textcolor{red}{worse} accuracy on IN-R and IN-S.
\zoomless refers to the set of zoom transforms at $S = 224$.
See \cref{supptab:aggregation_methods_extended} for more results.}
\label{tab:aggregation_main_text}
\resizebox{1\textwidth}{!}{%
\begin{tabular}{lrrrrrrrrrrr}
\toprule
& \multicolumn{1}{c}{(a)} & \multicolumn{2}{c}{(b) \zoomin \faSearchPluss\xspace} & \multicolumn{2}{c}{(c) \zoomout \faSearchMinuss\xspace} & \multicolumn{2}{c}{(d) \zoomless} & \multicolumn{2}{c}{(e) 5-crop} & \multicolumn{2}{c}{(f) 10-crop \cite{krizhevsky2012imagenet}} \\ \midrule
\textbf{Dataset} & 1-crop & \multicolumn{1}{c}{\layer{max}} & \multicolumn{1}{c}{\layer{mean}} & \multicolumn{1}{c}{\layer{max}} & \multicolumn{1}{c}{\layer{mean}} & \multicolumn{1}{c}{\layer{max}} & \multicolumn{1}{c}{\layer{mean}} & \multicolumn{1}{c}{\layer{max}} & \multicolumn{1}{c|}{\layer{mean}} & \multicolumn{1}{c}{\layer{max}} & \multicolumn{1}{c}{\layer{mean}} \\ \midrule
{IN} & 75.75 & 74.35 \decrease{1.40} & \textbf{79.31} \increase{3.56} & 71.48 & 69.47 & 72.66 & 73.67 & 77.33 & \multicolumn{1}{r|}{77.73} & 77.30 & 77.87 \\
{ReaL} & 81.89 & 80.22 \decrease{1.67} & \textbf{85.97} \increase{4.08} & 77.95 & 76.28 & 79.25 & 80.31 & 83.24 & \multicolumn{1}{r|}{83.80} & 83.17 & 83.87 \\
\cellcolor{MyLightGray}{IN-A} & 9.64 & \textbf{24.69} \increase{15.05} & 14.29 \increase{4.65} & 7.79 & 5.48 & 8.12 & 7.39 & 12.19 & \multicolumn{1}{r|}{9.88} & 12.32 & 9.67 \\
\cellcolor{MyLightGray}{IN-R} & 41.29 & 39.90 \decrease{1.39} & 40.06 \decrease{1.23} & 39.05 & 36.21 & 39.52 & 39.28 & 43.90 & \multicolumn{1}{r|}{43.17} & \textbf{44.31} & 43.28 \\
\cellcolor{MyLightGray}{IN-S} & 26.83 & 19.74 \decrease{7.09} & 20.89 \decrease{5.94} & 22.37 & 19.25 & 25.06 & 25.21 & 28.72 & \multicolumn{1}{r|}{28.66} & \textbf{28.94} & 28.76 \\
\cellcolor{MyLightGray}{ON} & 30.89 & 32.88 \increase{1.99} & \textbf{33.92} \increase{3.03} & 22.56 & 19.51 & 22.75 & 22.72 & 26.96 & \multicolumn{1}{r|}{24.98} & 27.14 & 24.97 \\ \bottomrule
\end{tabular}%
}
\end{table}

\begin{table}[htb!]
\centering
\caption{Performance of various aggregating methods (\%) -- The bold numbers show maximum accuracy per model/dataset. CLIP strongly and consistently favors 10-crop over other settings.}
\label{supptab:aggregation_methods_extended}
\resizebox{\textwidth}{!}{%
\begin{tabular}{clrrrrrrrrrrr}
\cline{3-13}
\multicolumn{2}{l}{\textbf{}} &
  \multicolumn{1}{c}{(a)} &
  \multicolumn{2}{c}{(b) \zoomin \faSearchPluss\xspace} &
  \multicolumn{2}{c}{(c) \zoomout \faSearchMinuss\xspace} &
  \multicolumn{2}{c}{(d) \zoomless} &
  \multicolumn{2}{c}{(e) 5-crop} &
  \multicolumn{2}{c}{(f) 10-crop~\cite{krizhevsky2012imagenet}} \\ \hline
\multicolumn{1}{l}{} &
  \textbf{Dataset} &
  \textbf{1-crop} &
  \multicolumn{1}{c}{\textit{\textbf{Max}}} &
  \multicolumn{1}{c}{\textit{\textbf{Mean}}} &
  \multicolumn{1}{c}{\textit{\textbf{Max}}} &
  \multicolumn{1}{c}{\textit{\textbf{Mean}}} &
  \multicolumn{1}{c}{\textit{\textbf{Max}}} &
  \multicolumn{1}{c}{\textit{\textbf{Mean}}} &
  \multicolumn{1}{c}{\textit{\textbf{Max}}} &
  \multicolumn{1}{c|}{\textit{\textbf{Mean}}} &
  \multicolumn{1}{c}{\textit{\textbf{Max}}} &
  \multicolumn{1}{c}{\textit{\textbf{Mean}}} \\ \hline
\multirow{6}{*}{\textbf{\textbf{\rotatebox[origin=c]{90}{ResNet-18}}}} &
  \textbf{IN} &
  69.45 &
  68.45 \decrease{1.00} &
  71.45 \increase{2.00} &
  60.33 &
  56.79 &
  67.85 &
  68.70 &
  70.61 &
  \multicolumn{1}{r|}{71.32} &
  70.83 &
  \textbf{71.85} \\
 &
  \textbf{ReaL} &
  76.94 &
  76.33 \decrease{0.61} &
  \textbf{79.94} \increase{3.00} &
  67.64 &
  63.92 &
  75.73 &
  76.74 &
  78.26 &
  \multicolumn{1}{r|}{79.01} &
  78.42 &
  79.46 \\
 &
  \textbf{IN-A} &
  1.37 &
  \textbf{11.68} \increase{10.31} &
  5.48 \increase{4.11} &
  2.44 &
  2.19 &
  3.41 &
  2.69 &
  3.16 &
  \multicolumn{1}{r|}{2.13} &
  3.28 &
  1.87 \\
 &
  \textbf{IN-R} &
  32.14 &
  30.60 \decrease{1.54} &
  28.95 \decrease{3.19} &
  29.08 &
  27.28 &
  32.29 &
  32.54 &
  33.99 &
  \multicolumn{1}{r|}{33.38} &
  \textbf{34.59} &
  33.83 \\
 &
  \textbf{IN-S} &
  19.41 &
  14.86 \decrease{4.55} &
  14.34 \decrease{5.07} &
  14.48 &
  11.49 &
  17.80 &
  17.83 &
  20.83 &
  \multicolumn{1}{r|}{20.70} &
  \textbf{21.39} &
  21.06 \\
 &
  \textbf{ON} &
  27.59 &
  \textbf{28.21} \increase{0.62} &
  25.92 \decrease{1.67} &
  16.11 &
  14.10 &
  22.82 &
  22.86 &
  24.77 &
  \multicolumn{1}{r|}{20.91} &
  25.47 &
  21.03 \\ \hline
\multirow{6}{*}{\textbf{\textbf{\rotatebox[origin=c]{90}{ResNet-50}}}} &
  \textbf{IN} &
  75.75 &
  73.24 \decrease{2.51} &
  77.30 \increase{1.55} &
  69.06 &
  66.42 &
  74.45 &
  75.39 &
  76.67 &
  \multicolumn{1}{r|}{77.13} &
  76.89 &
  \textbf{77.43} \\
 &
  \textbf{ReaL} &
  82.63 &
  80.36 \decrease{2.27} &
  \textbf{84.68} \increase{2.05} &
  76.35 &
  73.85 &
  81.96 &
  82.85 &
  83.67 &
  \multicolumn{1}{r|}{84.06} &
  83.82 &
  84.31 \\
 &
  \textbf{IN-A} &
  0.21 &
  \textbf{16.11} \increase{15.9} &
  6.23 \increase{6.02} &
  2.79 &
  2.19 &
  3.04 &
  2.11 &
  2.28 &
  \multicolumn{1}{r|}{0.95} &
  2.43 &
  1.00 \\
 &
  \textbf{IN-R} &
  35.39 &
  33.58 \decrease{1.81} &
  32.73 \decrease{2.66} &
  35.85 &
  33.22 &
  36.64 &
  36.44 &
  37.47 &
  \multicolumn{1}{r|}{36.50} &
  \textbf{38.23} &
  36.86 \\
 &
  \textbf{IN-S} &
  22.91 &
  16.89 \decrease{6.02} &
  17.80 \decrease{5.11} &
  19.51 &
  17.12 &
  21.60 &
  21.66 &
  24.71 &
  \multicolumn{1}{r|}{24.51} &
  \textbf{24.94} &
  24.74 \\
 &
  \textbf{ON} &
  \textbf{36.18} &
  34.56 \decrease{1.62} &
  34.22 \decrease{1.96} &
  27.10 &
  25.32 &
  31.78 &
  31.98 &
  33.34 &
  \multicolumn{1}{r|}{29.58} &
  33.93 &
  29.86 \\ \hline
\multirow{6}{*}{\textbf{\textbf{\rotatebox[origin=c]{90}{ViT-B/32}}}} &
  \textbf{IN} &
  75.75 &
  74.35 \decrease{1.40} &
  \textbf{79.31} \increase{3.56} &
  71.48 &
  69.47 &
  72.66 &
  73.67 &
  77.33 &
  \multicolumn{1}{r|}{77.73} &
  77.30 &
  77.87 \\
 &
  \textbf{ReaL} &
  81.89 &
  80.22 \decrease{1.67} &
  \textbf{85.97} \increase{4.08} &
  77.95 &
  76.28 &
  79.25 &
  80.31 &
  83.24 &
  \multicolumn{1}{r|}{83.80} &
  83.17 &
  83.87 \\
 &
  \textbf{IN-A} &
  9.64 &
  \textbf{24.69} \increase{15.05} &
  14.29 \increase{4.65} &
  7.79 &
  5.48 &
  8.12 &
  7.39 &
  12.19 &
  \multicolumn{1}{r|}{9.88} &
  12.32 &
  9.67 \\
 &
  \textbf{IN-R} &
  41.29 &
  39.90 \decrease{1.39} &
  40.06 \decrease{1.23} &
  39.05 &
  36.21 &
  39.52 &
  39.28 &
  43.90 &
  \multicolumn{1}{r|}{43.17} &
  \textbf{44.31} &
  43.28 \\
 &
  \textbf{IN-S} &
  26.83 &
  19.74 \decrease{7.09} &
  20.89 \decrease{5.94} &
  22.37 &
  19.25 &
  25.06 &
  25.21 &
  28.72 &
  \multicolumn{1}{r|}{28.66} &
  \textbf{28.94} &
  28.76 \\
 &
  \textbf{ON} &
  30.89 &
  32.88 \increase{1.99} &
  \textbf{33.92} \increase{3.03} &
  22.56 &
  19.51 &
  22.75 &
  22.72 &
  26.96 &
  \multicolumn{1}{r|}{24.98} &
  27.14 &
  24.97 \\ \hline
\multirow{6}{*}{\textbf{\textbf{\rotatebox[origin=c]{90}{VGG-16}}}} &
  \textbf{IN} &
  71.37 &
  69.60 \decrease{1.77} &
  72.46 \increase{1.09} &
  64.75 &
  59.95 &
  69.51 &
  70.48 &
  72.31 &
  \multicolumn{1}{r|}{73.09} &
  72.67 &
  \textbf{73.53} \\
 &
  \textbf{ReaL} &
  78.90 &
  77.23 \decrease{1.67} &
  80.59 \increase{1.69} &
  72.55 &
  67.68 &
  77.48 &
  78.58 &
  79.80 &
  \multicolumn{1}{r|}{80.42} &
  80.13 &
  \textbf{80.80} \\
 &
  \textbf{IN-A} &
  2.69 &
  \textbf{11.55} \increase{8.86} &
  6.24 \increase{3.55} &
  3.33 &
  2.77 &
  4.69 &
  3.87 &
  4.87 &
  \multicolumn{1}{r|}{3.19} &
  5.09 &
  3.19 \\
 &
  \textbf{IN-R} &
  26.98 &
  26.18 \decrease{0.80} &
  24.74 \decrease{2.24} &
  28.01 &
  25.62 &
  27.76 &
  27.78 &
  28.75 &
  \multicolumn{1}{r|}{27.95} &
  \textbf{29.23} &
  28.35 \\
 &
  \textbf{IN-S} &
  16.78 &
  13.30 \decrease{3.48} &
  13.05 \decrease{3.73} &
  15.18 &
  13.37 &
  15.82 &
  15.97 &
  17.80 &
  \multicolumn{1}{r|}{17.63} &
  \textbf{18.28} &
  17.92 \\
 &
  \textbf{ON} &
  \textbf{28.32} &
  26.96 \decrease{1.36} &
  26.15 \decrease{2.17} &
  19.88 &
  16.42 &
  23.47 &
  23.60 &
  26.21 &
  \multicolumn{1}{r|}{21.65} &
  26.52 &
  21.80 \\ \hline
\multirow{6}{*}{\textbf{\textbf{\rotatebox[origin=c]{90}{AlexNet}}}} &
  \textbf{IN} &
  56.16 &
  54.74 \decrease{1.42} &
  56.98 \increase{0.82} &
  40.78 &
  27.09 &
  51.80 &
  51.50 &
  57.86 &
  \multicolumn{1}{r|}{58.60} &
  58.26 &
  \textbf{59.11} \\
 &
  \textbf{ReaL} &
  62.67 &
  61.46 \decrease{1.21} &
  64.35 \increase{1.68} &
  45.84 &
  30.58 &
  58.25 &
  58.16 &
  64.53 &
  \multicolumn{1}{r|}{65.39} &
  64.98 &
  \textbf{65.94} \\
 &
  \textbf{IN-A} &
  1.75 &
  \textbf{4.65} \increase{2.90} &
  3.27 \increase{1.52} &
  1.56 &
  1.23 &
  2.31 &
  1.97 &
  2.53 &
  \multicolumn{1}{r|}{2.04} &
  2.64 &
  2.03 \\
 &
  \textbf{IN-R} &
  21.10 &
  20.65 \decrease{0.45} &
  17.97 \decrease{3.13} &
  15.72 &
  11.25 &
  19.91 &
  19.55 &
  22.79 &
  \multicolumn{1}{r|}{21.86} &
  \textbf{23.26} &
  22.16 \\
 &
  \textbf{IN-S} &
  10.05 &
  7.94 \decrease{2.11} &
  6.54 \decrease{3.51} &
  5.82 &
  2.72 &
  8.29 &
  7.39 &
  10.84 &
  \multicolumn{1}{r|}{10.65} &
  \textbf{11.20} &
  10.80 \\
 &
  \textbf{ON} &
  14.23 &
  \textbf{14.91} \increase{0.68} &
  11.80 \decrease{2.43} &
  6.11 &
  3.75 &
  9.65 &
  9.01 &
  12.63 &
  \multicolumn{1}{r|}{9.57} &
  12.84 &
  9.58 \\ \hline
\multirow{6}{*}{\textbf{\textbf{\rotatebox[origin=c]{90}{\clip-ViT-L/14}}}} &
  \textbf{IN} &
  75.03 &
  70.01 \decrease{5.02} &
  74.45 \decrease{0.58} &
  72.01 &
  72.21 &
  74.45 &
  76.04 &
   76.77 &
  \multicolumn{1}{r|}{76.91} &
   76.72 &
   \textbf{77.00}
   \\
 &
  \textbf{ReaL} &
  80.68 &
  76.37 \decrease{4.31} &
  81.31 \increase{0.63} &
  78.28 &
  78.93 &
  81.45 &
  82.05 &
  82.26 &
  \multicolumn{1}{r|}{\textbf{82.55}} &
   82.26 &
   \textbf{82.55} \\
 &
  \textbf{IN-A} &
  71.28 &
  76.57 \increase{5.29} &
  68.16 \decrease{3.12} &
  60.71 &
  49.51 &
  71.69 &
  70.04 &
  77.80 &
  \multicolumn{1}{r|}{76.61} &
   \textbf{78.25} &
   76.83
   \\
 &
  \textbf{IN-R} &
  87.74 &
  84.12 \decrease{3.62} &
  83.54 \decrease{4.20} &
  86.84 &
  86.29 &
  88.12 &
  88.24 &
  89.64 &
  \multicolumn{1}{r|}{89.66} &
   \textbf{90.01} &
   89.94 \\
 &
  \textbf{IN-S} &
  58.23 &
  51.88 \decrease{6.35} &
  56.06 \decrease{2.17} &
  57.14 &
  57.43 &
  59.00 &
  59.90 &
  \multicolumn{1}{l}{61.28} &
  \multicolumn{1}{l|}{61.61} &
  \multicolumn{1}{l}{61.59} &
  \multicolumn{1}{l}{\textbf{62.07}} \\
 &
  \textbf{ON} &
  66.32 &
  60.20 \decrease{6.12} &
  58.10 \decrease{8.22} &
  56.57 &
  58.11 &
  62.44 &
  62.65 &
  66.70 &
  \multicolumn{1}{r|}{64.88} &
   \textbf{66.87} &
   64.97 \\ \hline
\end{tabular}%
}
\end{table}

\FloatBarrier
\subsection{Runtime analysis of MEMO}

Another benefit of  \texttt{RRC} compared to AugMix is faster inference time.
Table~\ref{supptab:memo_runtime} shows the runtime analysis of MEMO.
Typically, TTA methods suffer from slow runtime due to augmentation and test-time training processes.
We find that MEMO + \texttt{RRC} consistently leads to an average \textcolor{ForestGreen}{$1.6\times$} speed-up compared to MEMO + AugMix (\cref{supptab:memo_runtime}; 0.65s / image vs. 1.15s / image), providing more evidence to support this transformation as a viable option for test-time augmentations.

\begin{table}[!hbt]
\centering
\caption{Average runtime per query image (in seconds). Using \texttt{RandomResizedCrop} in MEMO speed ups the runtime by an average factor of \textcolor{ForestGreen}{1.6$\times$}.}
\label{supptab:memo_runtime}
\begin{tabular}{llrrrr}
\hline
 Runtime (in seconds) & IN & \multicolumn{1}{l}{IN-A} & \multicolumn{1}{l}{IN-R} & \multicolumn{1}{l}{IN-S} & \multicolumn{1}{l}{ON} \\ \hline
\multicolumn{4}{l}{MEMO + {AugMix}~\cite{zhang2021memo} }             & \multicolumn{1}{l}{} & \multicolumn{1}{l}{} \\ \hline
~~~ResNet-50~\cite{he2016deep}                   & 1.24 & 1.12 & 1.12 & 1.32                 & 1.51                 \\
~~~DeepAug+AugMix~\cite{hendrycks2021many}              & 1.19 & 1.07 & 1.12 & 1.23                 & 1.55                 \\
~~~MoEx+CutMix~\cite{li2021feature}                 & 1.15 & 1.16 & 1.11 & 1.31                 & 1.53                 \\ \hline
 MEMO + \rrc (\textbf{Ours}) &      &      &      & \multicolumn{1}{l}{} & \multicolumn{1}{l}{} \\ \hline
~~~ResNet-50~\cite{he2016deep}                   & 0.64 & 0.60 & 0.65 & 0.88                 & 1.19                 \\
~~~DeepAug+AugMix~\cite{hendrycks2021many}              & 0.62 & 0.62 & 0.64 & 0.87                 & 1.18                 \\
~~~MoEx+CutMix~\cite{li2021feature}                 & 0.65 & 0.62 & 0.66 & 0.88                 & 1.19                 \\ \hline
\end{tabular}
\end{table}


\subsection{1-crop accuracy with different zoom scales}
\label{suppsec:waterfallplots}

In this section, we demonstrate the performance of various models when zooming in or out of an image. 
In other words, we utilize the standard 1-crop ImageNet transform while altering the initial scale of the image.

In this section, we are conducting experiments using the following models: AlexNet~\cite{krizhevsky2012imagenet}, ConvNext (Base, Large, Small, Tiny)~\cite{liu2022convnet}, DenseNet-161~\cite{huang2017densely}, EfficientNet-B7~\cite{tan2019efficientnet}, MobileNet (V2, V3 Large)~\cite{sandler2018mobilenetv2, howard2019searching}, ResNet (50, 101)~\cite{he2016deep}, ResNeXt-50 (32x4d)~\cite{xie2017aggregated}, ShuffleNet V2 x1.0~\cite{ma2018shufflenet}, VGG-19~\cite{simonyan2014very}, Vision Transformer (ViT-B/16, ViT-B/32, ViT-L/16, ViT-L/32)~\cite{dosovitskiy2020image}, and Wide ResNet-50-2~\cite{zagoruyko2016wide}.

\begin{figure}[!hbt]
    \centering
    \begin{subfigure}[b]{0.9\linewidth}   
        \centering
        \includegraphics[width=\linewidth]{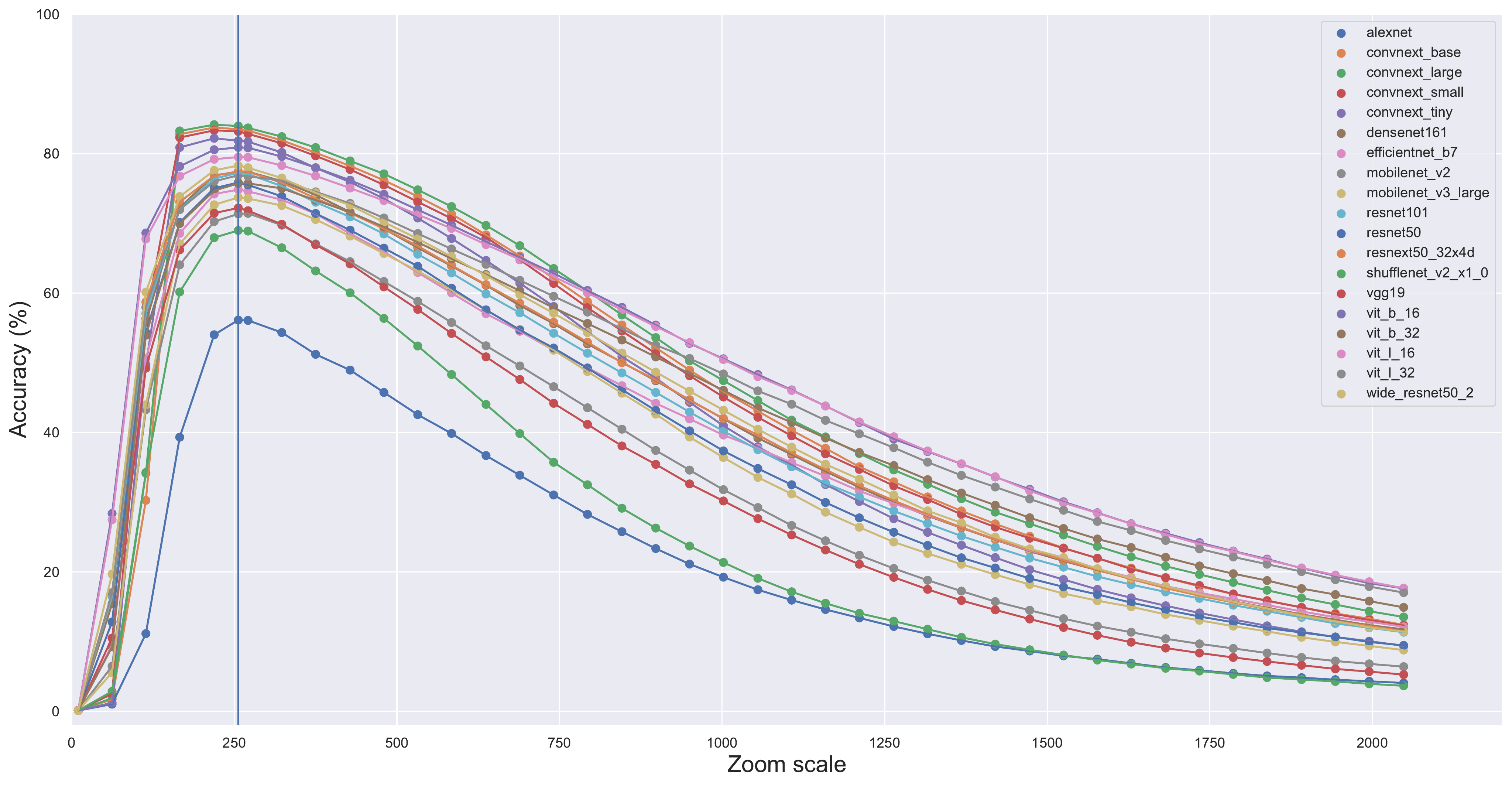}
    \end{subfigure}
    \caption{ImageNet accuracy using a 1-crop transform (the vertical line represents the standard ImageNet transform scale factor).}
    \label{suppfig:waterfall_imagenet}
\end{figure}

\begin{figure}[!hbt]
    \centering
    \begin{subfigure}[b]{0.9\linewidth}   
        \centering
        \includegraphics[width=\linewidth]{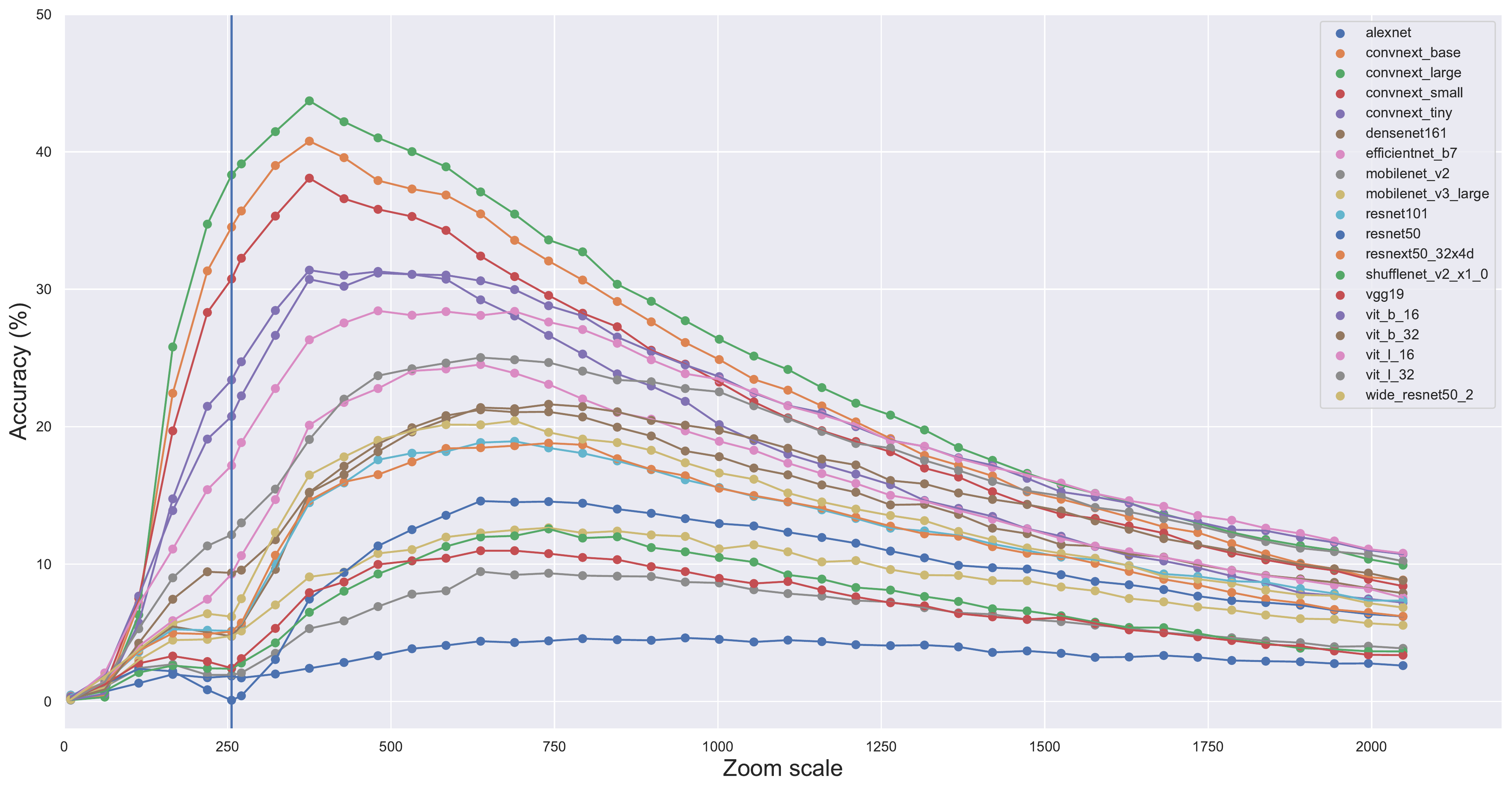}
    \end{subfigure}
    \caption{ImageNet-A accuracy using a 1-crop transform (the vertical line represents the standard ImageNet transform scale factor).}
    \label{suppfig:waterfall_imagenet_a}
\end{figure}

\begin{figure}[!hbt]
    \centering
    \begin{subfigure}[b]{0.9\linewidth}   
        \centering
        \includegraphics[width=\linewidth]{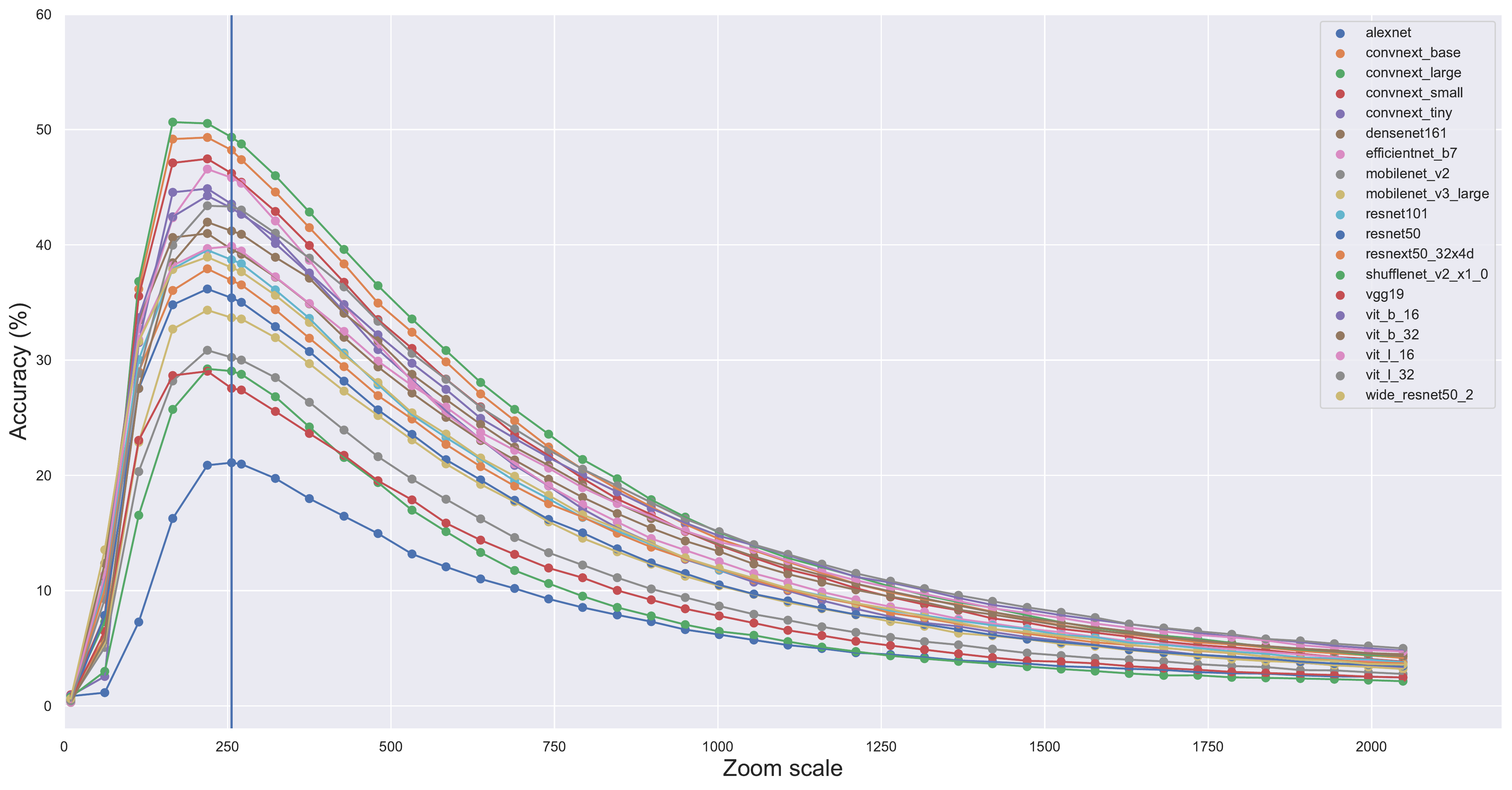}
    \end{subfigure}
    \caption{ImageNet-R accuracy using a 1-crop transform (the vertical line represents the standard ImageNet transform scale factor).}
    \label{suppfig:waterfall_imagenet_r}
\end{figure}

\begin{figure}[!hbt]
    \centering
    \begin{subfigure}[b]{0.9\linewidth}   
        \centering
        \includegraphics[width=\linewidth]{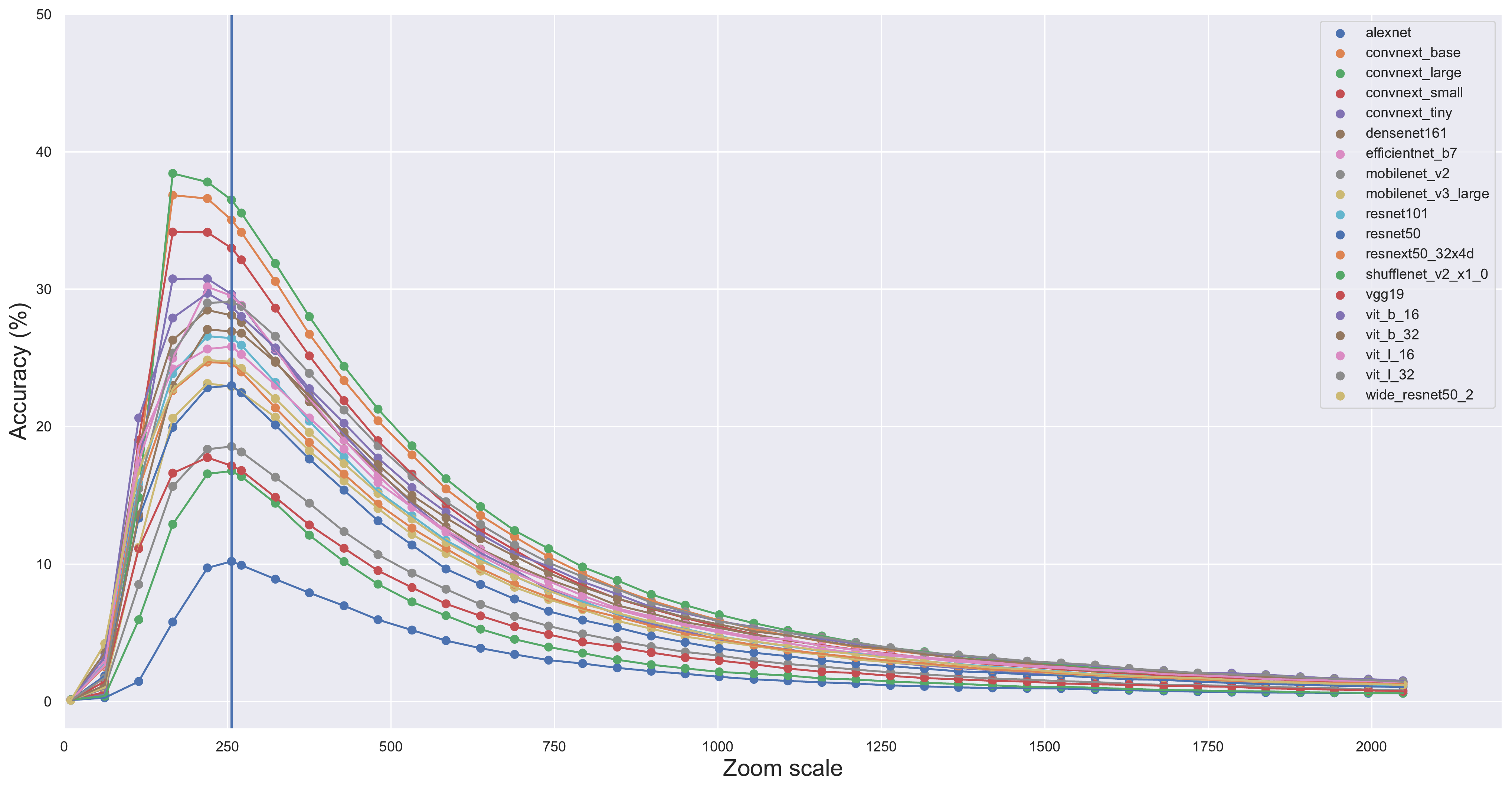}
    \end{subfigure}
    \caption{ImageNet-Sketch accuracy using a 1-crop transform (the vertical line represents the standard ImageNet transform scale factor).}
    \label{suppfig:waterfall_imagenet_sketch}
\end{figure}

\begin{figure}[!hbt]
    \centering
    \begin{subfigure}[b]{0.9\linewidth}   
        \centering
        \includegraphics[width=\linewidth]{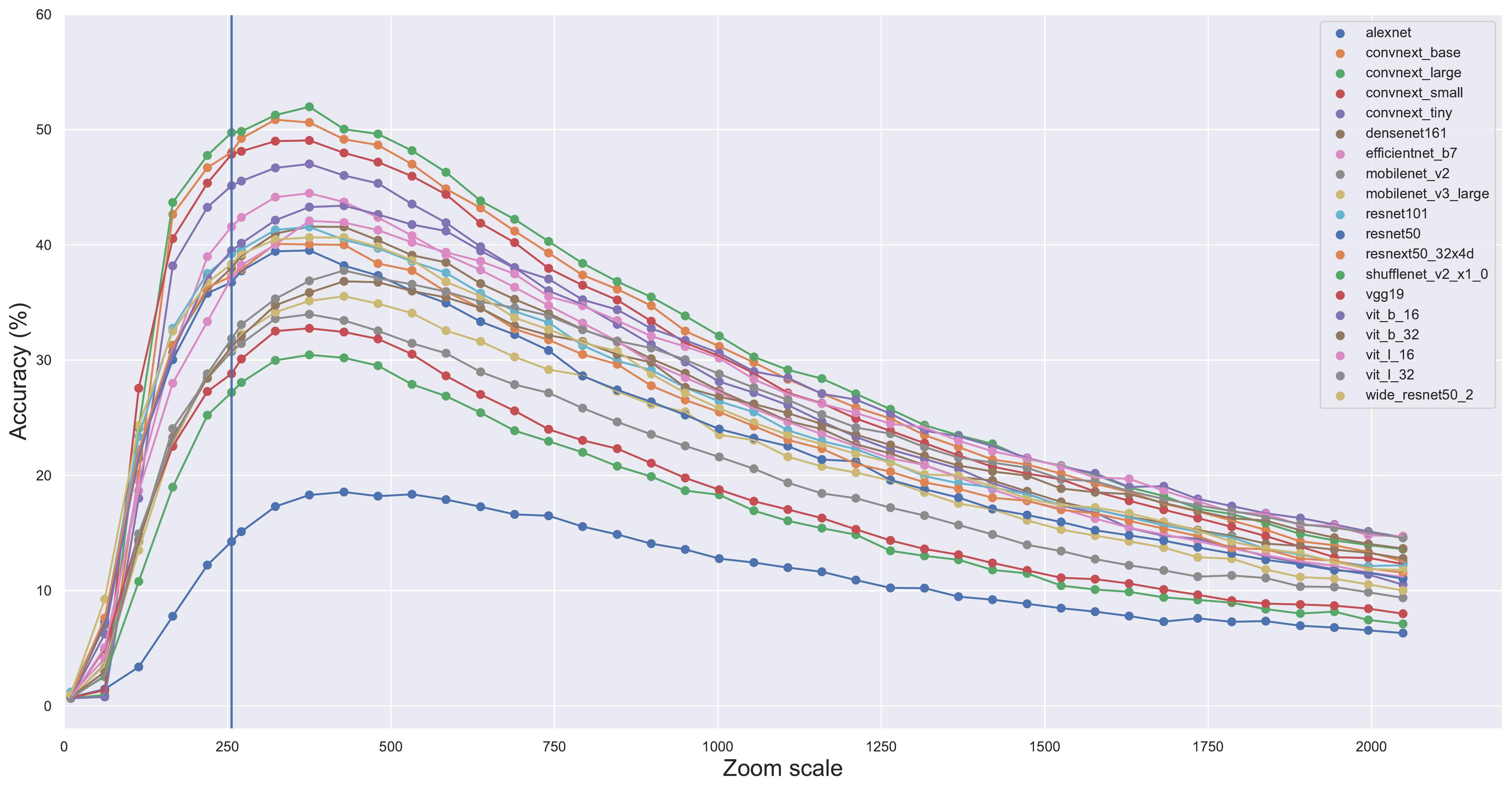}
    \end{subfigure}
    \caption{Accuracy using a 1-crop transform on 5K random images of the ObjectNet dataset (the vertical line represents the standard ImageNet transform scale factor).}
    \label{suppfig:waterfall_objectnet5k}
\end{figure}

\begin{figure*}[!htb]
\centering
{
	\begin{flushleft}
		\hskip -0.1in
        \scriptsize\rotatebox{90}{\kern -12.00pc ON\kern 2.2pc IN-S\kern 2.2pc IN-A\kern 2.2pc ReaL}
	\end{flushleft}
}
\begin{subfigure}[t]{1\textwidth}
     \includegraphics[width=1\textwidth]{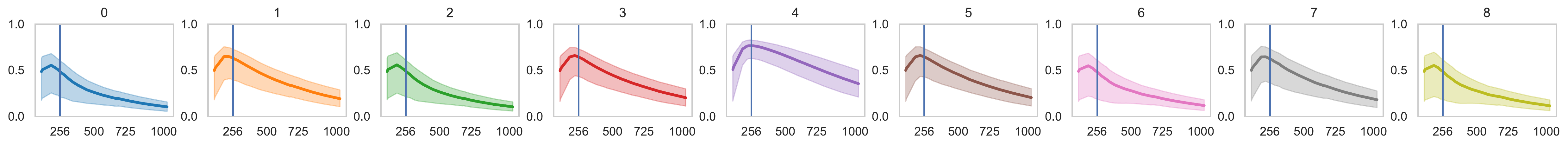}
\end{subfigure}
\begin{subfigure}[t]{1\textwidth}
     \includegraphics[width=1\textwidth]{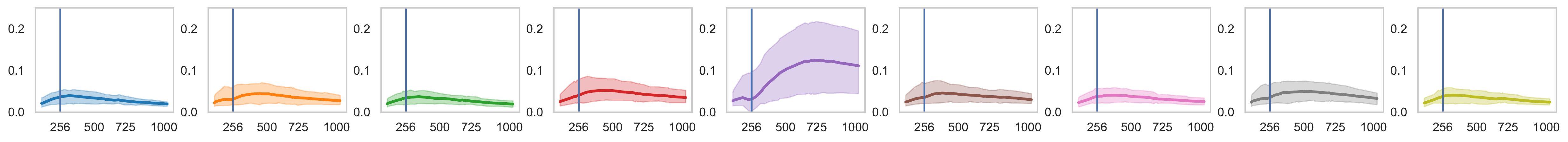}
\end{subfigure}
\begin{subfigure}[t]{1\textwidth}
     \includegraphics[width=1\textwidth]{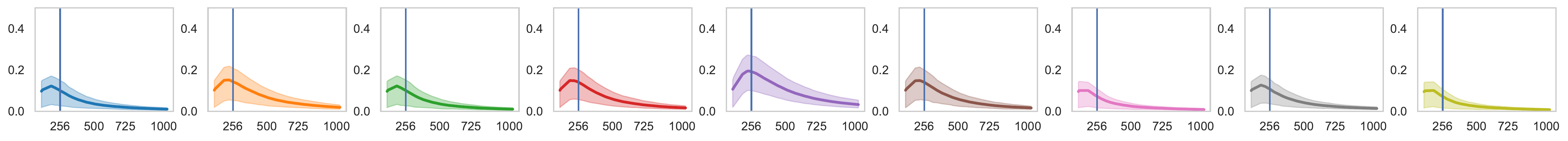}
\end{subfigure}
\begin{subfigure}[t]{1\textwidth}
     \includegraphics[width=1\textwidth]{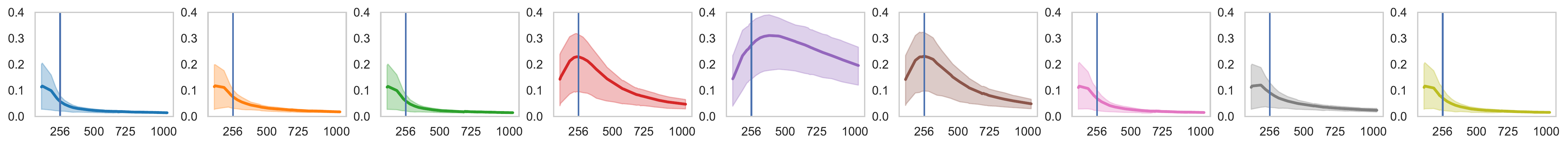}
\end{subfigure}
\caption{Breakdown of the accuracy of IN-trained models at different crop locations and scale size -- Analysis of accuracy across various crop locations and scale sizes reveals that different datasets exhibit distinct optimal conditions. For instance, the IN-A dataset experiences a considerable increase in accuracy when zoomed in, while ImageNet-R yields better results when zoomed out.}
\label{suppfig:9by9_waterfalls}
\end{figure*}

\FloatBarrier
\subsection{Distribution of the Top 36 performing transforms.}

In this section, we provide more details about the distribution of the top-36 performing transforms. Our results suggest that, on average, 26.65\% of all top-36 performing transforms belong to the center at varying scales.

\begin{figure}[!hbt]
    \centering
    \begin{subfigure}[b]{1\linewidth}   
        \centering
        \includegraphics[width=1\linewidth]{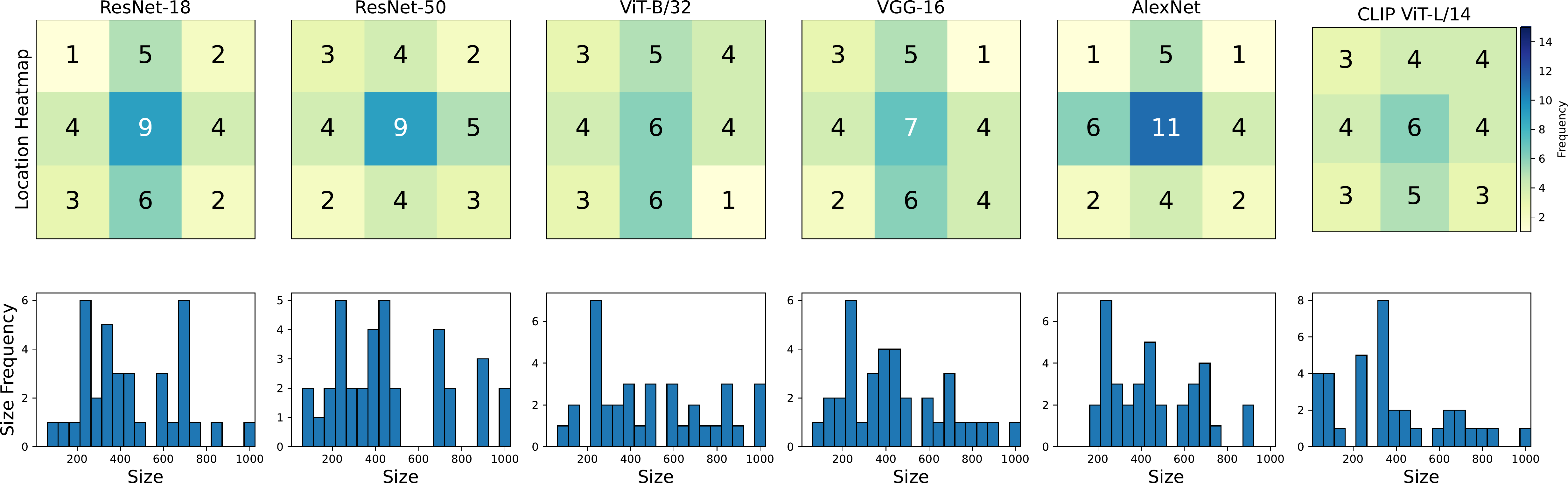}
    \end{subfigure}
    \caption{Distribution of Top-36 performing transforms for ImageNet-ReaL}
    \label{suppfig:dist_top36_in_real}
\end{figure}

\begin{figure}[!hbt]
    \centering
    \begin{subfigure}[b]{1\linewidth}   
        \centering
        \includegraphics[width=1\linewidth]{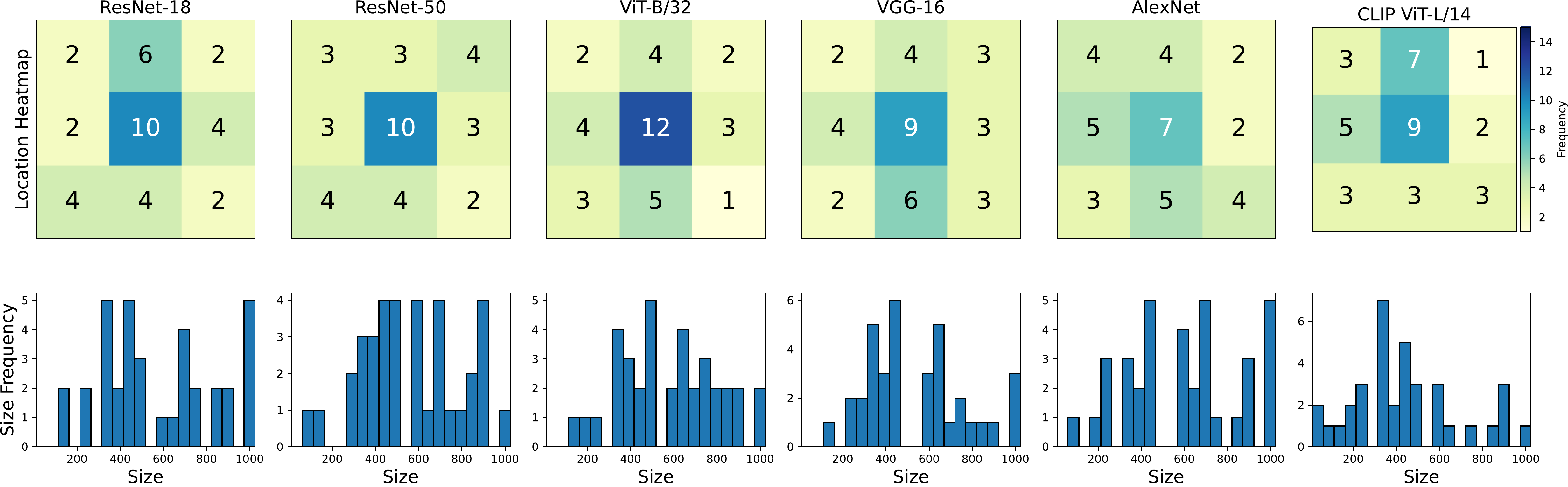}
    \end{subfigure}
    \caption{Distribution of Top-36 performing transforms for ImageNet-A}
    \label{suppfig:dist_top36_in_a}
\end{figure}

\begin{figure}[!hbt]
    \centering
    \begin{subfigure}[b]{1\linewidth}   
        \centering
        \includegraphics[width=1\linewidth]{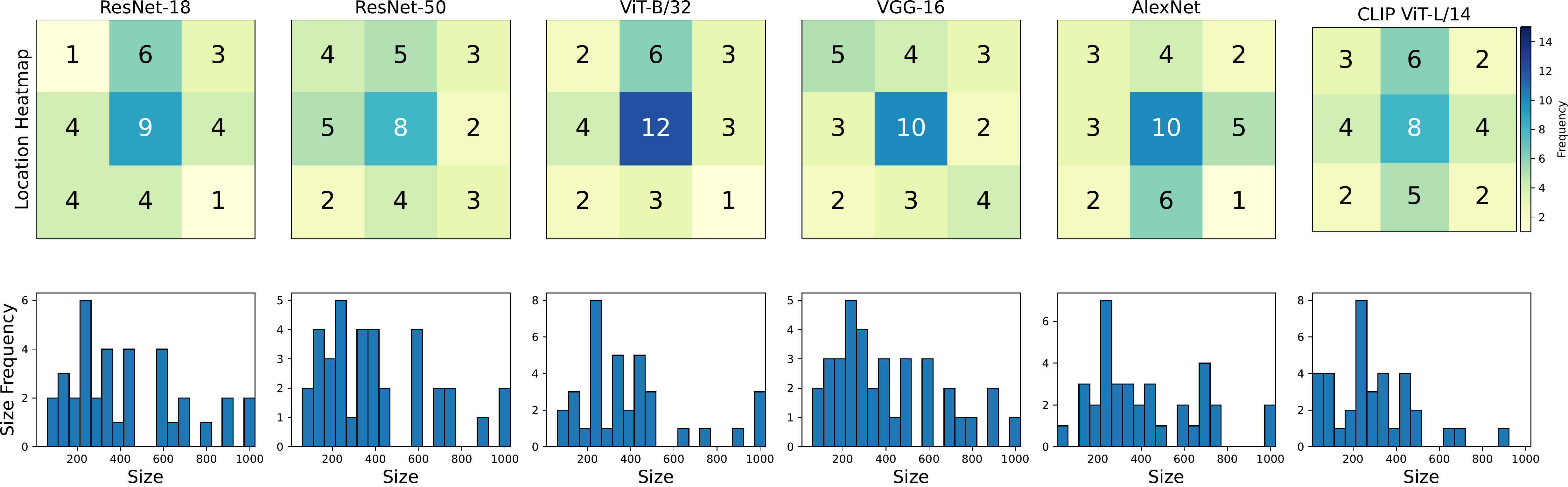}
    \end{subfigure}
    \caption{Distribution of Top-36 performing transforms for ImageNet-R}
    \label{suppfig:dist_top36_in_r}
\end{figure}

\begin{figure}[!hbt]
    \centering
    \begin{subfigure}[b]{1\linewidth}   
        \centering
        \includegraphics[width=1\linewidth]{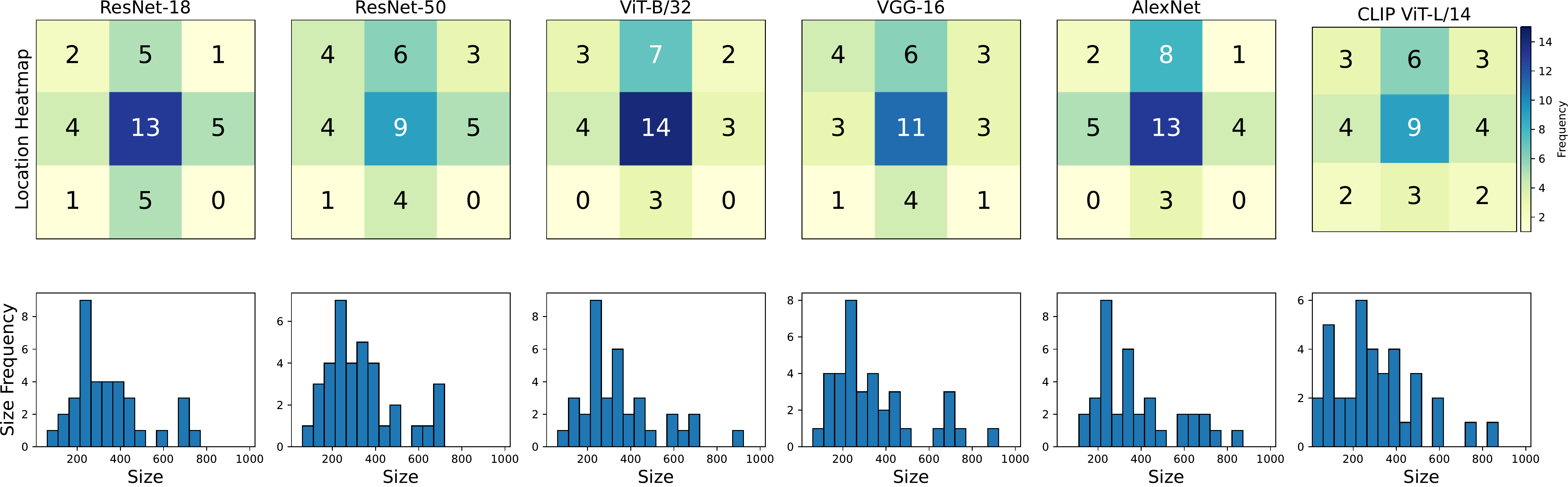}
    \end{subfigure}
    \caption{Distribution of Top-36 performing transforms for ImageNet-Sketch}
    \label{suppfig:dist_top36_in_sketch}
\end{figure}

\begin{figure}[!hbt]
    \centering
    \begin{subfigure}[b]{1\linewidth}   
        \centering
        \includegraphics[width=1\linewidth]{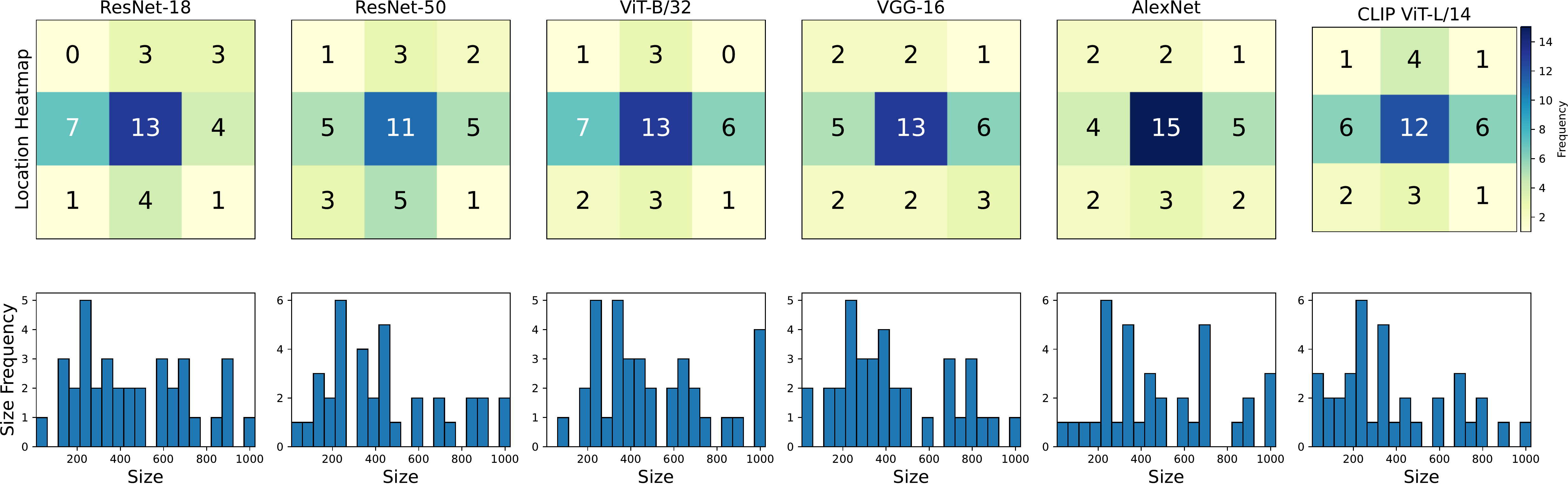}
    \end{subfigure}
    \caption{Distribution of Top-36 performing transforms for ObjectNet}
    \label{suppfig:dist_top36_in_on}
\end{figure}

\FloatBarrier
\subsection{Background occlusion in ImageNet dataset}
\label{suppsec:background_occlusion}

Sample images for images with and without occlusion.

\begin{figure}[!hbt]
    \centering
    \begin{minipage}{0.45\linewidth}
        \begin{subfigure}{\linewidth}   
            \centering
            \includegraphics[width=\linewidth]{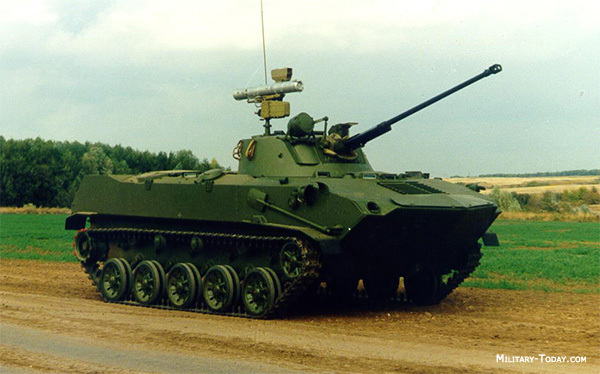}
            \caption{A sample image of the \class{Tank} class without occlusion.}
            \label{fig:Tank_Sample}
        \end{subfigure}
        \par\bigskip 
        \begin{subfigure}{\linewidth}  
            \centering
            \includegraphics[width=\linewidth]{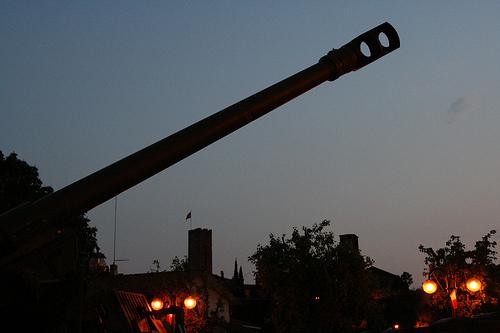}
            \caption{Image with heavy background occlusion.}
            \label{fig:Tank_Occluded}
        \end{subfigure}
        \par\bigskip 
        \begin{subfigure}{\linewidth}   
            \centering
            \includegraphics[width=\linewidth]{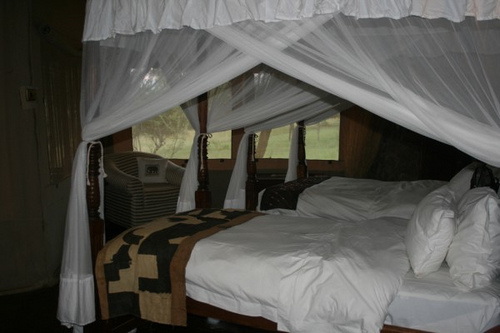}
            \caption{A clean sample image of the \class{Four Poster} class.}
            \label{fig:FourPost_Sample}
        \end{subfigure}
    \end{minipage}
    \hfill
    \begin{minipage}{0.45\linewidth}
        \begin{subfigure}{\linewidth}  
            \centering
            \includegraphics[width=\linewidth]{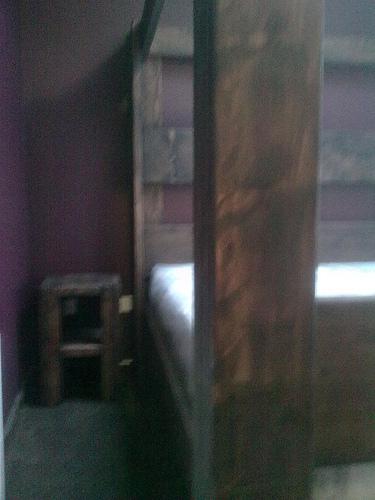}
            \caption{A low-quality image with background occlusion.}
            \label{fig:FourPost_Occluded}
        \end{subfigure}
    \end{minipage}
    \caption{Background occlusion examples.}
    \label{suppfig:background_occlusion}
\end{figure}




\clearpage
\section{Additional Experiments}

In this section, we provide additional experiments with the proposed zoom-based transform.


\subsection{Zooming is similarly important to the foreground and background contents}
\label{sec:foreground_background}

Background pixels, despite often being neglected in image classification, can contain predictive signals \cite{zhu2016object, xiao2020noise, geirhos2020shortcut, robinson2021can}.
It has remained largely unknown how much the image context (background) could contribute to the model performance.
While Zhu et al.~\cite{zhu2016object} disentangle the predictiveness of background (BG) and foreground (FG) via model training, we directly measure how pretrained models perceive these two signals.

\begin{figure}[htb!]
    \centering
\resizebox{0.5\textwidth}{!}{
    
            \begin{minipage}{.3\columnwidth}
            \begin{subfigure}{\columnwidth}
            \centering
            \includegraphics[width=\columnwidth]{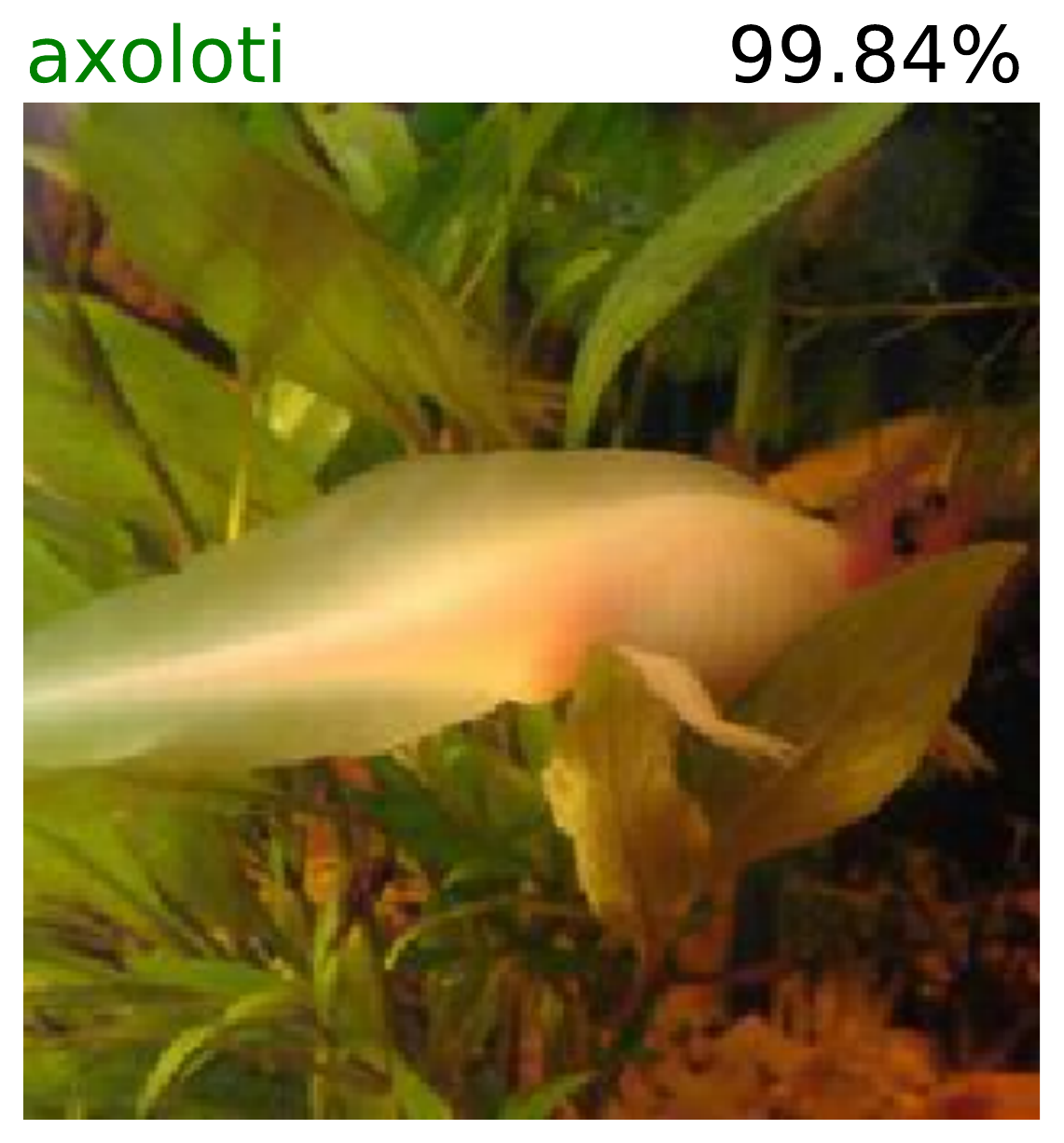}
            \caption{}
            \end{subfigure}\\
            \begin{subfigure}{\columnwidth}
            \centering
            \includegraphics[width=\columnwidth]{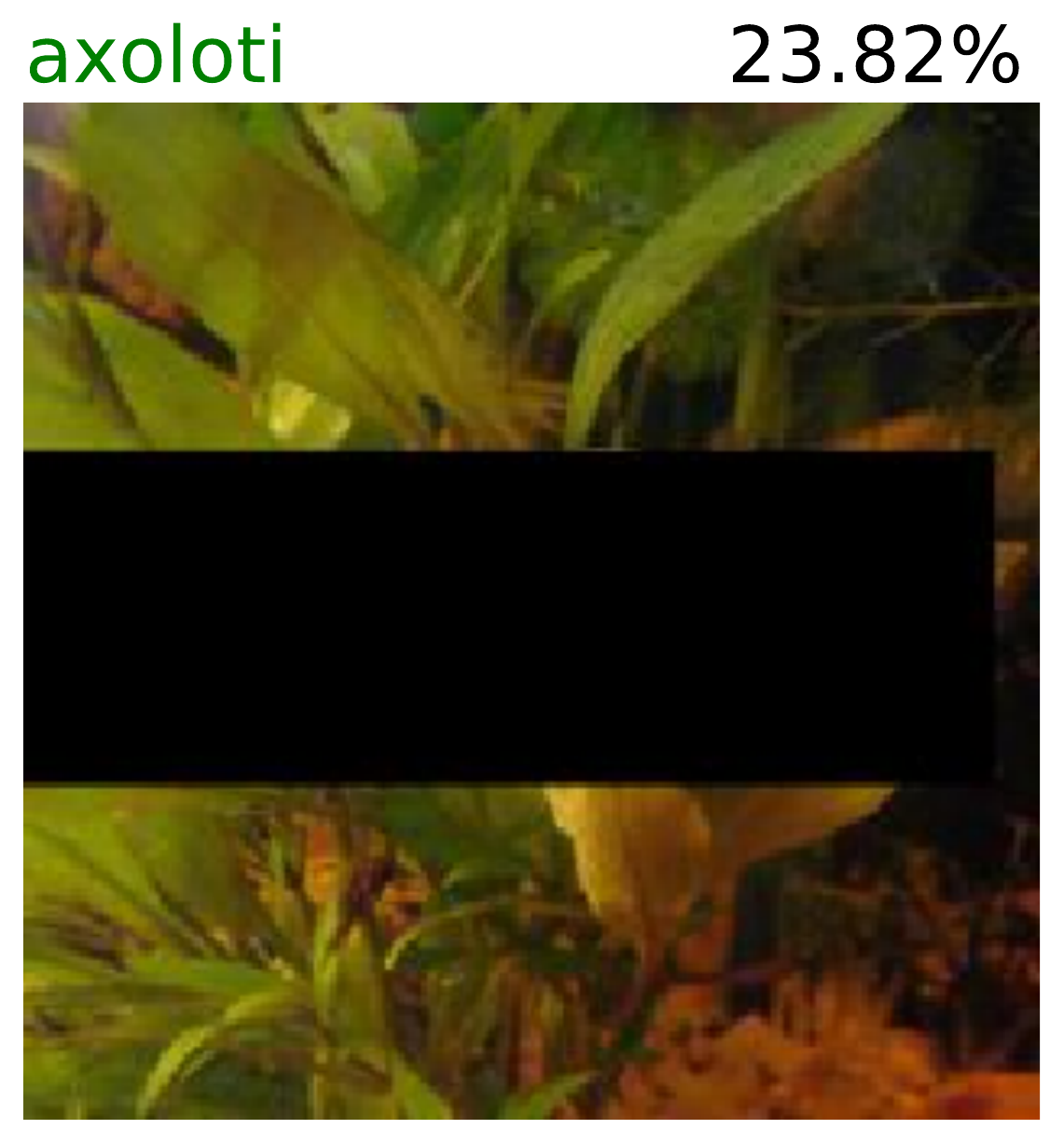}
            \caption{}
            \end{subfigure}%
        \end{minipage}%
         \hfill
        \begin{minipage}{.3\columnwidth}
            \begin{subfigure}{\columnwidth}
            \centering
            \includegraphics[width=\columnwidth]{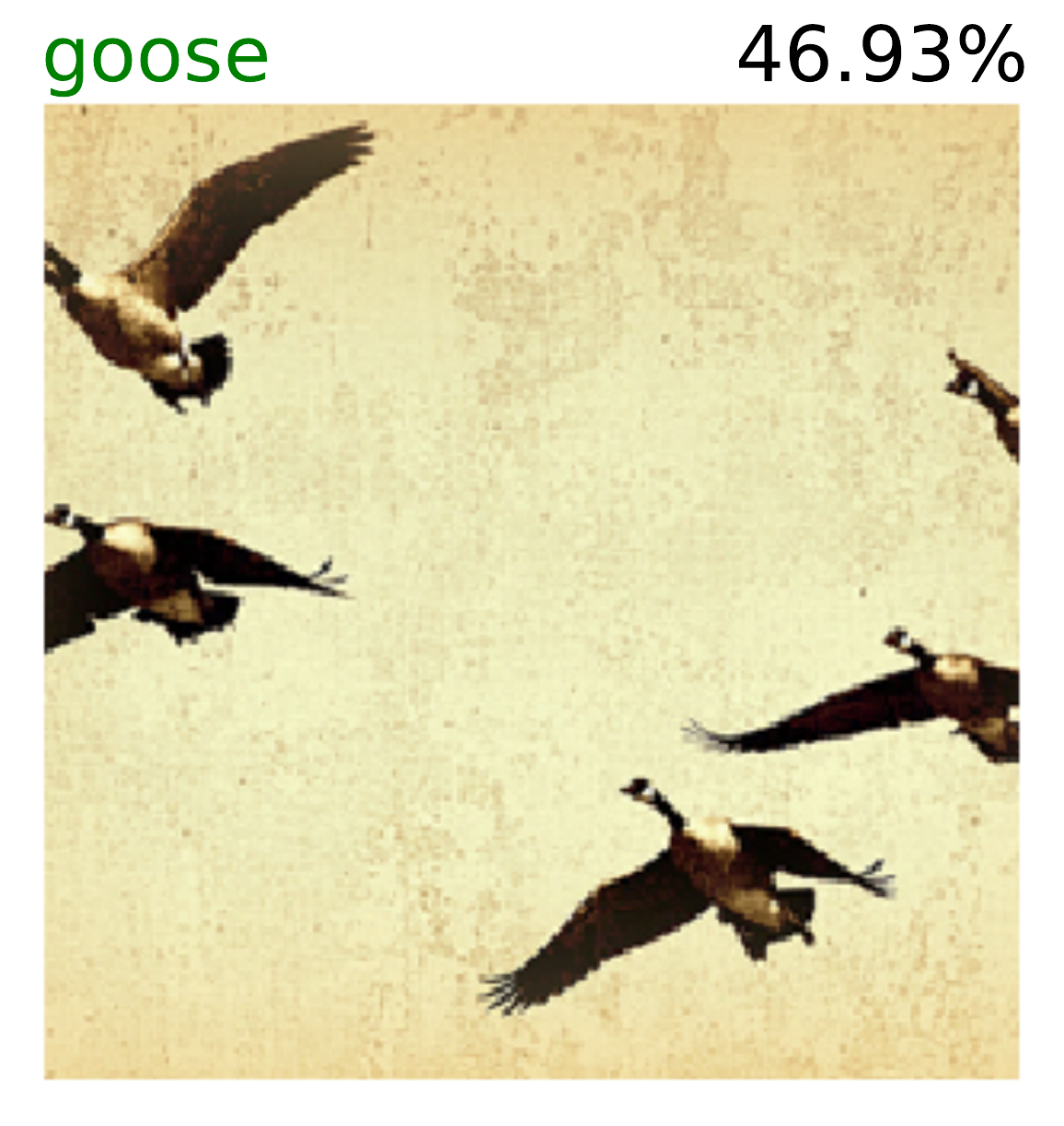}
             \caption{}
            \end{subfigure}\\
            \begin{subfigure}{\columnwidth}
            \centering
            \includegraphics[width=\columnwidth]{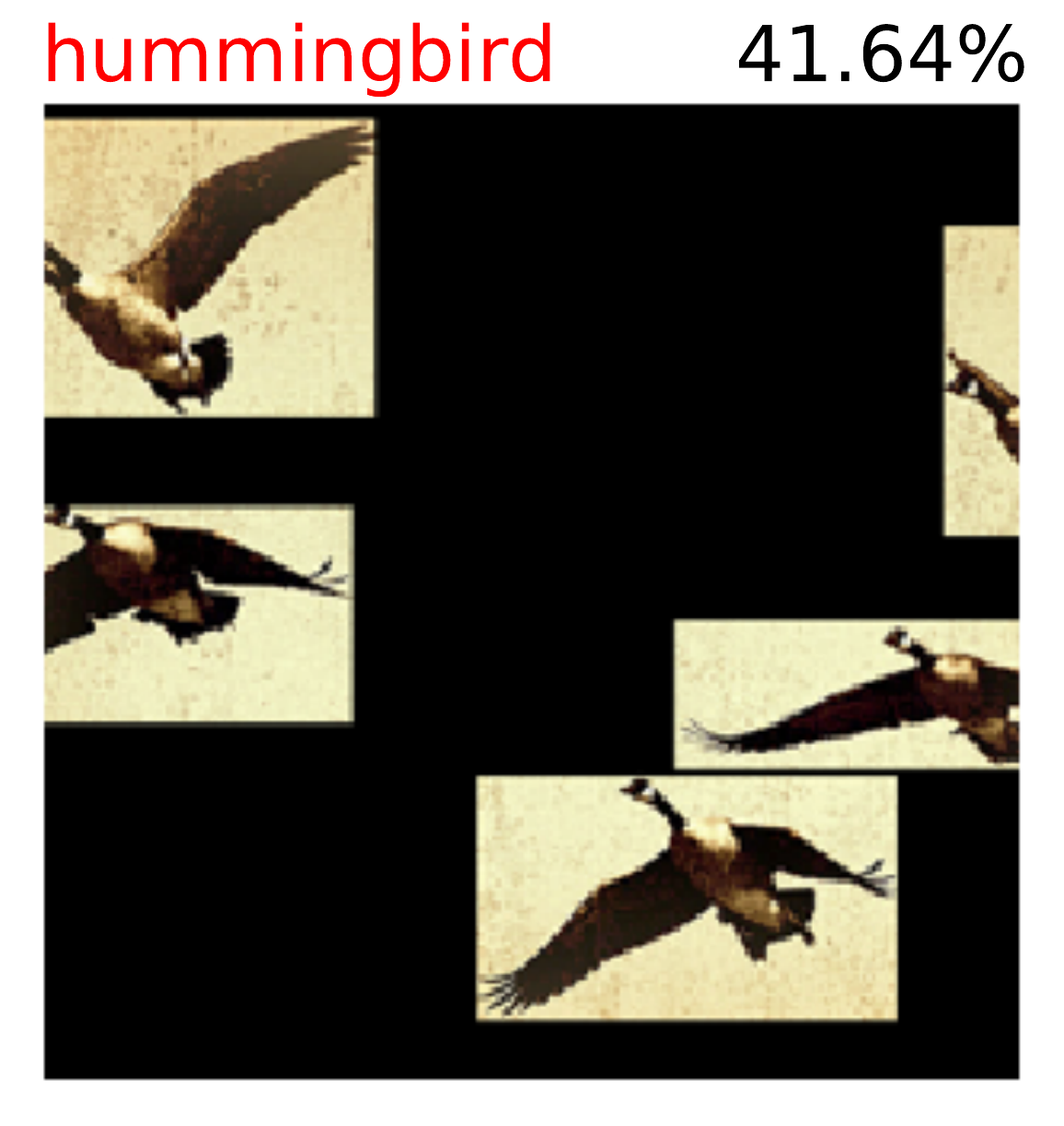}
            \caption{}
            \end{subfigure}%
        \end{minipage}
        \hfill
        \begin{minipage}{.3\columnwidth}
            \begin{subfigure}{\columnwidth}
            \centering
            \includegraphics[width=\columnwidth]{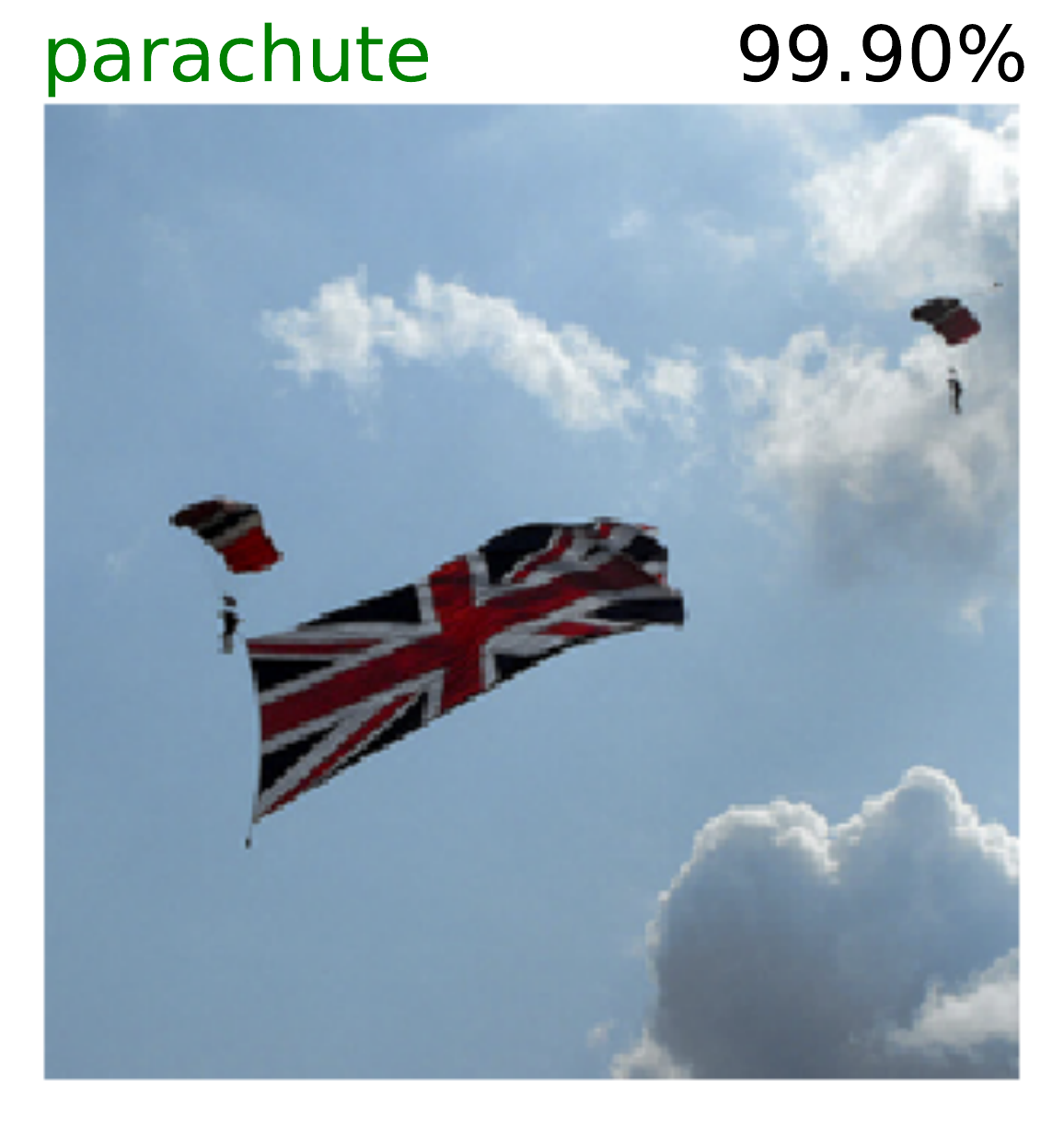}
             \caption{}
            \end{subfigure}\\
            \begin{subfigure}{\columnwidth}
            \centering
            \includegraphics[width=\columnwidth]{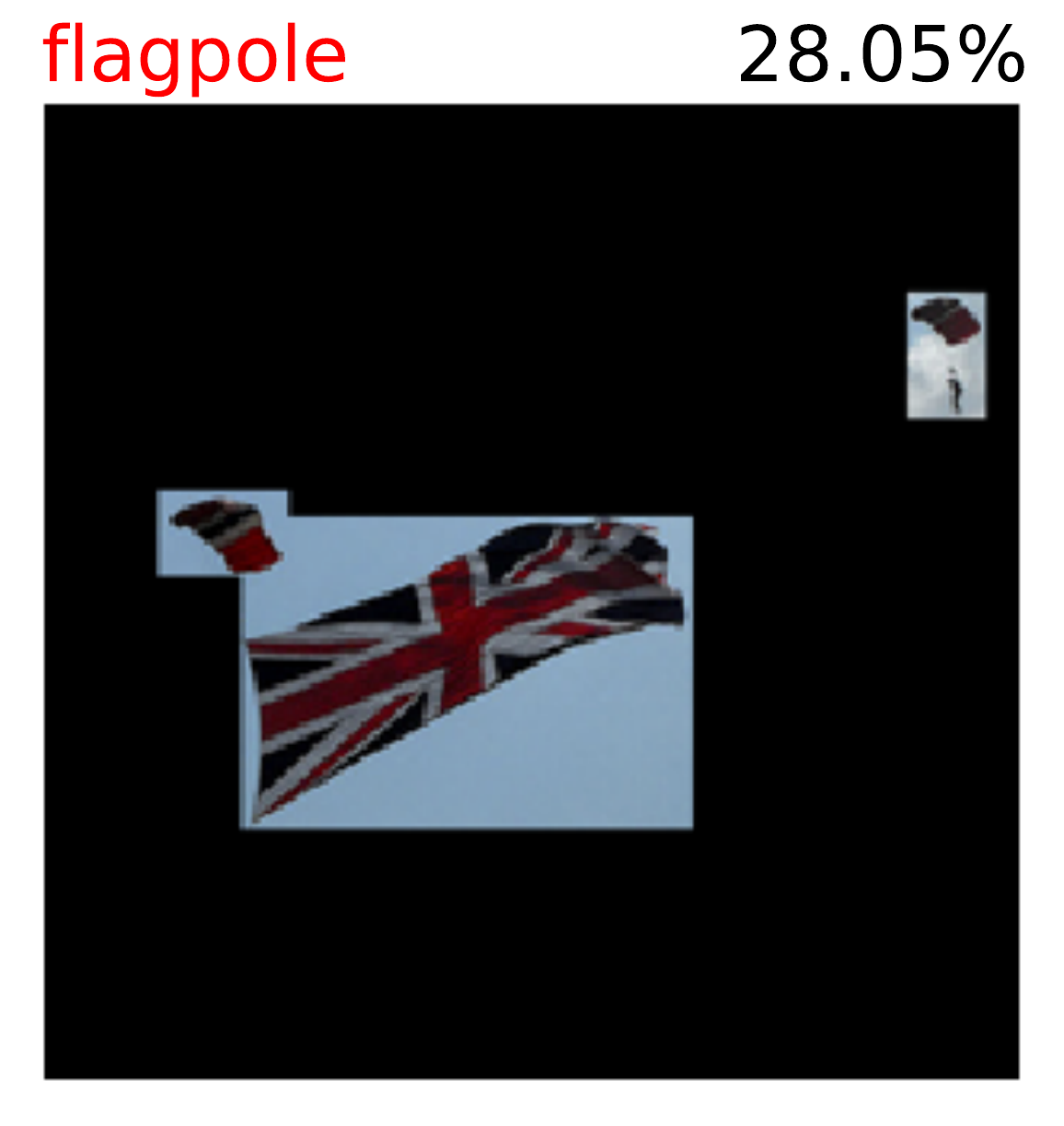}
            \caption{}
            \end{subfigure}%
        \end{minipage}
        \hfill
}

        \caption{The foreground and the background both contain predictive signals. A ResNet-50 classifier can detect \class{axolotl} (a), even when the main object is masked (b). Removing the background from images of `goose' (c) and `parachute' (e) causes misclassification (d, f).}
    \label{fig:bgfgsample}
\end{figure}

\paragraph{Experiment}
Using bounding-box annotations provided by Russakovsky et al.~\cite{russakovsky2015imagenet}, we create two dataset variations of ImageNet: \textit{FGSet} and \textit{BGSet}, following Zhu et al.~\cite{zhu2016object}.
We mask all the background for \textit{FGSet} as in \cref{fig:bgfgsample}b, and for \textit{BGSet} we mask all the main objects, as depicted in \cref{fig:bgfgsample}d \& \cref{fig:bgfgsample}f.
After that, we compute the accuracy of these two sets with all tested classifiers using ImageNet and ImageNet-ReaL labels as in~\cref{tab:bgfgset}.

\paragraph{Results}
Our results suggest that zooming is important to ImageNet regardless of whether foreground or background features are used, with the difference for \textit{FGSet} and \textit{BGSet} on average being similar (\cref{tab:bgfgset}). 
Additionally, when only the background features were available, almost half of ImageNet images ($45.23\%$) could be correctly classified if optimal Zoom was used.
Finally, we found that with only foreground information, ViT-B/32 could achieve a maximum possible accuracy of $95.50\%$ given an optimal zooming method, suggesting that only $98.75\% - 95.50\% = 3.25\% $ of images (\cref{tab:1_main_results}) required the background information.
These findings suggest that both foreground and background features are important for ImageNet classification, but that an optimal zooming method can considerably improve performance even in the absence of one of these feature sets.

\begin{table}[htb!]
\centering
\caption{
ImageNet classification from object-only and background-only signals. 
Numbers show the maximum possible top-1 accuracy (\%) using zoom-based transforms for minimum set covers in~\cref{sec:mincover}. 
We discover that background signals potentially hold significance for image classification. The bold numbers show the highest possible accuracy per dataset and group.
}
\label{tab:bgfgset}
\resizebox{1\textwidth}{!}{
\begin{tabular}{@{}lrrrrrrrr@{}}
\toprule
\multicolumn{1}{l}{} & \multicolumn{4}{c}{1-crop} & \multicolumn{4}{c}{Max possible using zooming} \\ \midrule
\multicolumn{1}{l}{} & \multicolumn{2}{c}{FGSet} & \multicolumn{2}{c}{BGSet} & \multicolumn{2}{c}{FGSet} & \multicolumn{2}{c}{BGSet} \\ \cmidrule(l){2-9} 
\multicolumn{1}{l}{} & \multicolumn{1}{c}{IN} & \multicolumn{1}{c}{ReaL} & \multicolumn{1}{c}{IN} & \multicolumn{1}{c}{ReaL} & \multicolumn{1}{c}{IN} & \multicolumn{1}{c}{ReaL} & \multicolumn{1}{c}{IN} & \multicolumn{1}{c}{ReaL} \\ \cmidrule(l){2-9} 
ResNet-18 & 59.77 & 64.97 & 4.91 & 7.84 & 89.89\increase{30.12} & 92.04\increase{27.07} & 25.81\increase{20.90} & 31.33\increase{23.49} \\
ResNet-50 & 68.02 & 72.90 & 6.18 & 9.83 & 93.45\increase{25.43} & 94.89\increase{21.99} & 30.30\increase{24.12} & 35.98\increase{26.15} \\
ViT-B/32 & 67.46 & 71.78 & \textbf{9.72} & 13.38 & 94.40\increase{26.94} & 95.50\increase{23.72} & \textbf{39.70}\increase{29.98} & \textbf{45.23}\increase{31.85} \\
VGG-16 & 63.78 & 69.09 & 5.36 & 8.59 & 91.01\increase{27.23} & 92.91\increase{23.82} & 26.98\increase{21.62} & 32.62\increase{24.03} \\
AlexNet & 42.38 & 46.54 & 3.66 & 5.46 & 80.20\increase{37.82} & 83.25\increase{36.71} & 22.02\increase{18.36} & 27.04\increase{21.58} \\ 
\clip-ViT-L/14 & \textbf{74.46} & \textbf{78.62} & 9.49 & \textbf{13.80} & \textbf{96.14}\increase{21.68} & \textbf{97.35}\increase{18.73} & 36.85\increase{27.36} & 42.51\increase{28.71} \\ 
\bottomrule
\layer{mean} & 62.65 & 67.32 & 6.55 & 9.82 & 90.85 \increase{28.20} & 92.66 \increase{25.34} & 30.28 \increase{23.73} & 35.79 \increase{25.97} \\ 
\bottomrule
\end{tabular}%
}
\end{table}



\subsection{Adversarial datasets contain more objects compared to ImageNet}
\label{sec:dataset_statistics}

So far, our findings indicate that if we apply the zoom-in operation to the two datasets of ImageNet-A and ObjectNet, the performance of conventional vision models improves consistently up to a certain threshold (\cref{sec:experimental_results_main} and \cref{sec:clip_zoom_invariance}).
This suggests that the initial images contain distracting elements that impede the model from correctly identifying the object of interest.
Both ImageNet-A and ObjectNet are considered out-of-distribution datasets, which are specifically designed to evaluate a vision model's ability to withstand natural adversarial and pose attacks.
We hypothesize that the primary reason that these datasets are hard can be attributed to background clutter, multiple objects, and the presence of a positional bias in these images.

\paragraph{Experiment} We use OWL-ViT~\cite{minderer2022simple}, an open vocabulary object detection model, to quantify the number of objects present in three datasets of ImageNet, ImageNet-A, and ObjecNet.
The OWL-ViT expects an input image with a set of object names and will determine if any object instances are present in the image. 
To specify object names, we use LVIS vocabulary~\cite{gupta2019lvis}, which encompasses a comprehensive list of 1203 distinct objects.
The OWL-ViT model includes a threshold parameter that reflects its confidence level in its predictions. To assess whether different threshold values would affect our results, we conducted our experiment using both $0.1$ and $0.05$ as threshold values.

After calculating the distribution of the number of objects in images, we perform a Mann-Whitney U test to determine whether there is a statistically significant difference in this distribution between datasets.
As each dataset has a different number of classes, we limited our analysis to shared classes between any two datasets.

\paragraph{Results} 

The results of our study reveal a contrast between ImageNet and ImageNet-A, as well as ImageNet and ObjectNet.
This finding implies a dissimilarity between the images in the original ImageNet dataset and its OOD datasets that might arise from the presence of background clutter.
Specifically, on average, images in ImageNet-A and ObjectNet datasets tend to feature more objects, which can pose more significant distractions for image classification models.

The results of the Mann-Whitney U test also reflect this finding, the p-value for both thresholds was found to be less than $0.05$, which is statistically significant at the $95\%$ confidence level (\cref{supptab:u_test_results}).

\begin{figure}[!hbt]
    \centering
    \begin{subfigure}[b]{\linewidth}   
        \centering
        \includegraphics[width=\linewidth]{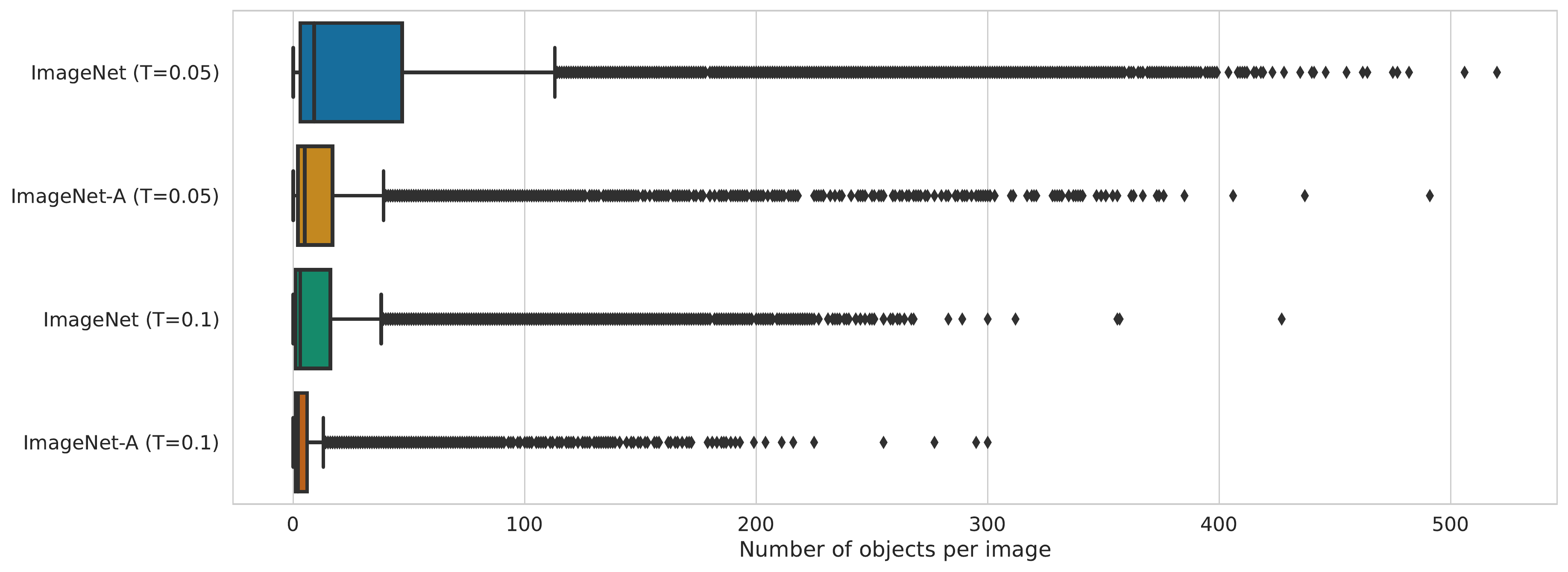}
    \end{subfigure}
    \caption{Comparison of the number of objects in two datasets of ImageNet and ImageNet-A using OWL-ViT~\cite{minderer2022simple}  -- $T$ denotes the classification's threshold }
    \label{suppfig:owl_vit_imagenet_vs_imagenet_a_1}
\end{figure}

\begin{figure}[!hbt]
    \centering
    \begin{subfigure}[b]{\linewidth}   
        \centering
        \includegraphics[width=\linewidth]{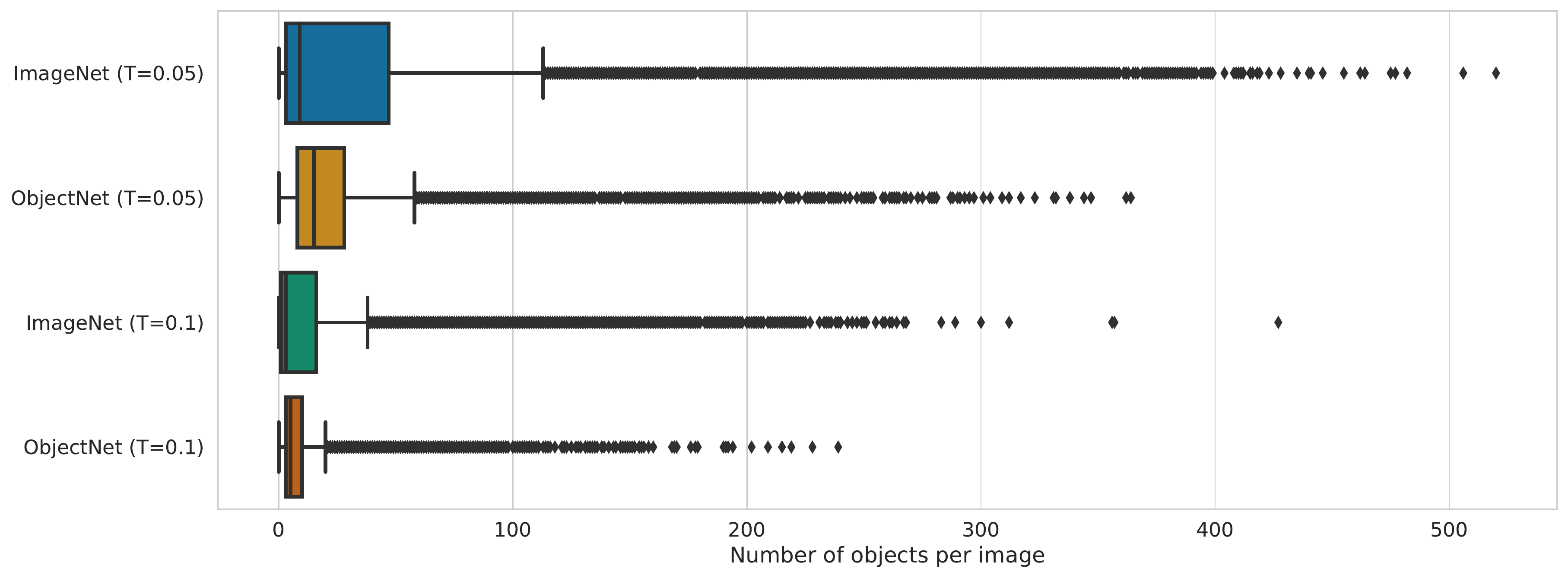}
    \end{subfigure}
    \caption{Comparison of the number of objects in two datasets of ImageNet and ObjectNet using OWL-ViT~\cite{minderer2022simple} -- $T$ denotes the classification's threshold  }
    \label{suppfig:owl_vit_imagenet_vs_objectnet}
\end{figure}

\subsubsection{$p$-values for Mann Whitney U test}

\begin{table}[htb!]
\centering
\caption{The result of the Mann-Whitney U test to compare ImageNet with ImageNet-A and ObjectNet}
\label{supptab:u_test_results}
\begin{tabular}{@{}lrr@{}}
\toprule
 & \multicolumn{1}{l}{$T=0.05$} & \multicolumn{1}{l}{$T=0.01$} \\ \midrule
\textbf{ImageNet-A} & 6.27E-265 & 1.71E-235 \\
\textbf{ObjectNet} & 1.80E-02 & 3.66E-02 \\ \bottomrule
\end{tabular}
\end{table}

\subsection{Zooming further improves robustified models on ImageNet-A}
\label{sec:zooming_vs_imagenet_a}

Intensive data augmentations have been proven to significantly boost CNNs' performance~\cite{wightman2021resnet, steiner2021train} on ImageNet. 
Motivated by these previous successes and the fact that neural networks trained on diverse augmentations are able to learn robust representations \cite{li2021learning}, we want to know if robustified pretrained models (\ie trained with intensive augmentations) could reach higher accuracy on ImageNet-A using zooming in.

\paragraph{Experiment}

We test 4 different ResNet-50 classifier versions that have been trained with different data augmentation procedures. 
From the the \torchvision library, we select two sets of model weights; trained with (V2\footnote{\texttt{ResNet50\_Weights.IMAGENET1K\_V2}}) and without (V1\footnote{\texttt{ResNet50\_Weights.IMAGENET1K\_V1}}) data augmentations. 
We also take two other models trained with DeepAugmentation+AugMix~\cite{hendrycks2021many} and MoEx+CutMix~\cite{li2021feature}.
The second column in~\cref{tab:resnet_50_imagenet_a} represents the accuracy of models using 1-crop.

\paragraph{Results}

Zooming in consistently helps ResNet-50 networks, with improvements varying from~\increasenoparent{13} to \increasenoparent{24} points.
The best-performing network is \torchvision-V2 which uses the max aggregator and achieves 29.65\%.
These results suggest that simple aggregation over the proposed zoom transform is effective for  datasets that have dominant center bias.

\begin{table}[hbt!]
\centering
\caption{The results of different aggregation functions on four ResNet-50 variants when tested on ImageNet-A (\%). Each model has been trained using different training-time augmentation techniques.
\textcolor{ForestGreen}{Improvements} values in parentheses are with respect to the 1-crop baseline.
}
\label{tab:resnet_50_imagenet_a}
\begin{tabular}{@{}llrrr@{}}
\toprule
\multicolumn{2}{l}{ResNet-50} & \multicolumn{1}{c}{Baseline} & \multicolumn{1}{c}{\emph{Max}} & \multicolumn{1}{c}{\emph{Mean}} \\ \midrule
 & \torchvision V1 & 0.21 & 16.11 (\increasenoparent{15.90}) & 6.23 \\
 & MoEx+CutMix~\cite{li2021feature} & 8.60 & 24.72 (\increasenoparent{16.12}) & 15.32 \\
 & DeepAug+AugMix~\cite{hendrycks2021many} & 3.94 & 27.93 (\increasenoparent{23.99}) & 13.16 \\
 & \torchvision V2 & 16.62 & 29.65 (\increasenoparent{13.03}) & 22.08 \\ \bottomrule
\end{tabular}%
\end{table}

\clearpage
\section{Visualization}
\label{supp:visualizations}

In this section, we provide several visualizations of zooming transforms.

\subsection{Visualizations for 36 top performing zoom transforms}
\label{supp:viz_36crop}

\begin{figure}[!hbt]
    \centering
    \includegraphics[width=1\textwidth]{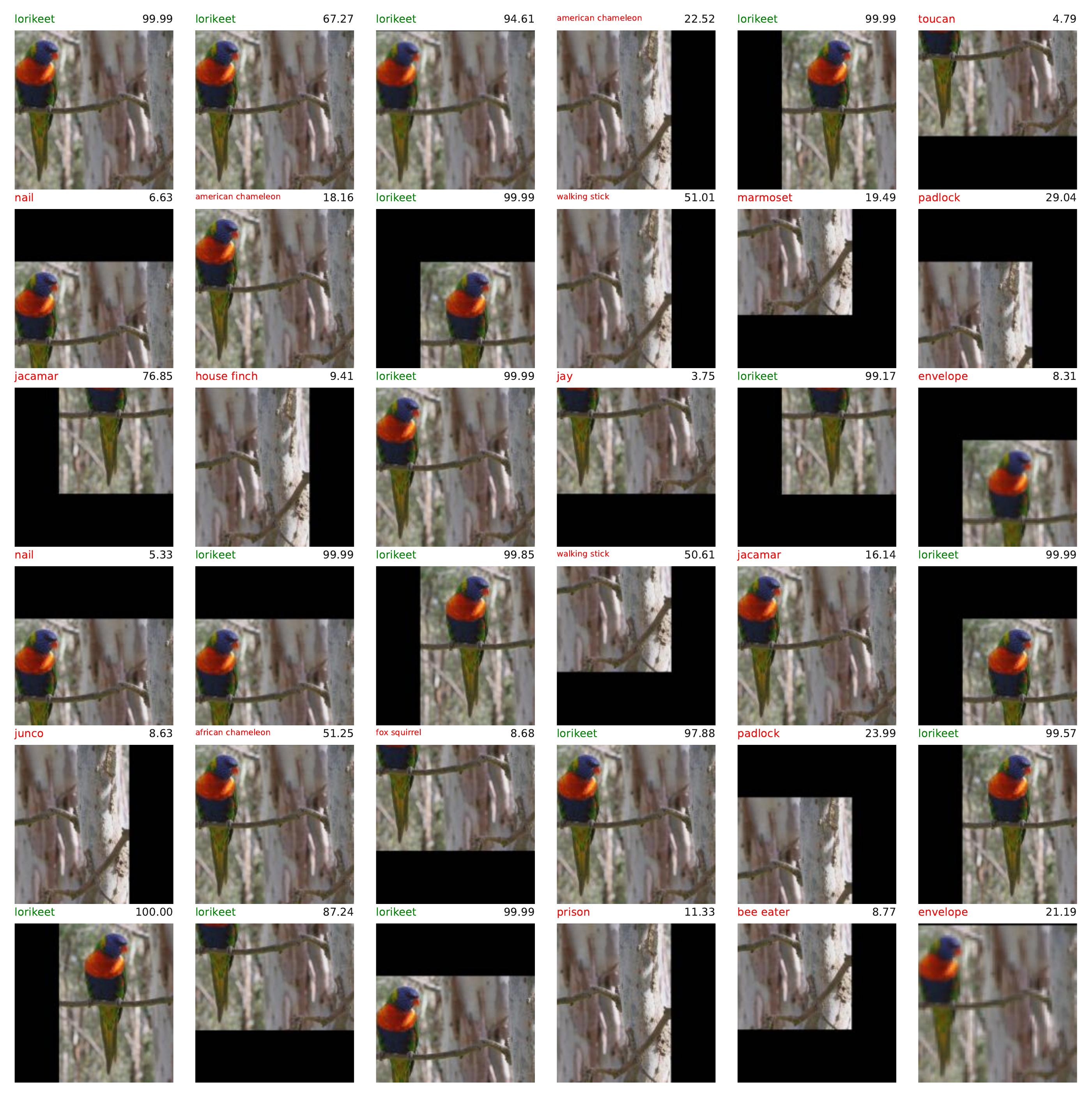}
    \caption{ 
    Different framing of an image of a \class{lorikeet} according to 36 high-performing transforms of a ResNet-50 model
    }
    \label{suppfig:36_crop_1}
\end{figure}

\begin{figure}[!hbt]
    \centering
    \includegraphics[width=1\textwidth]{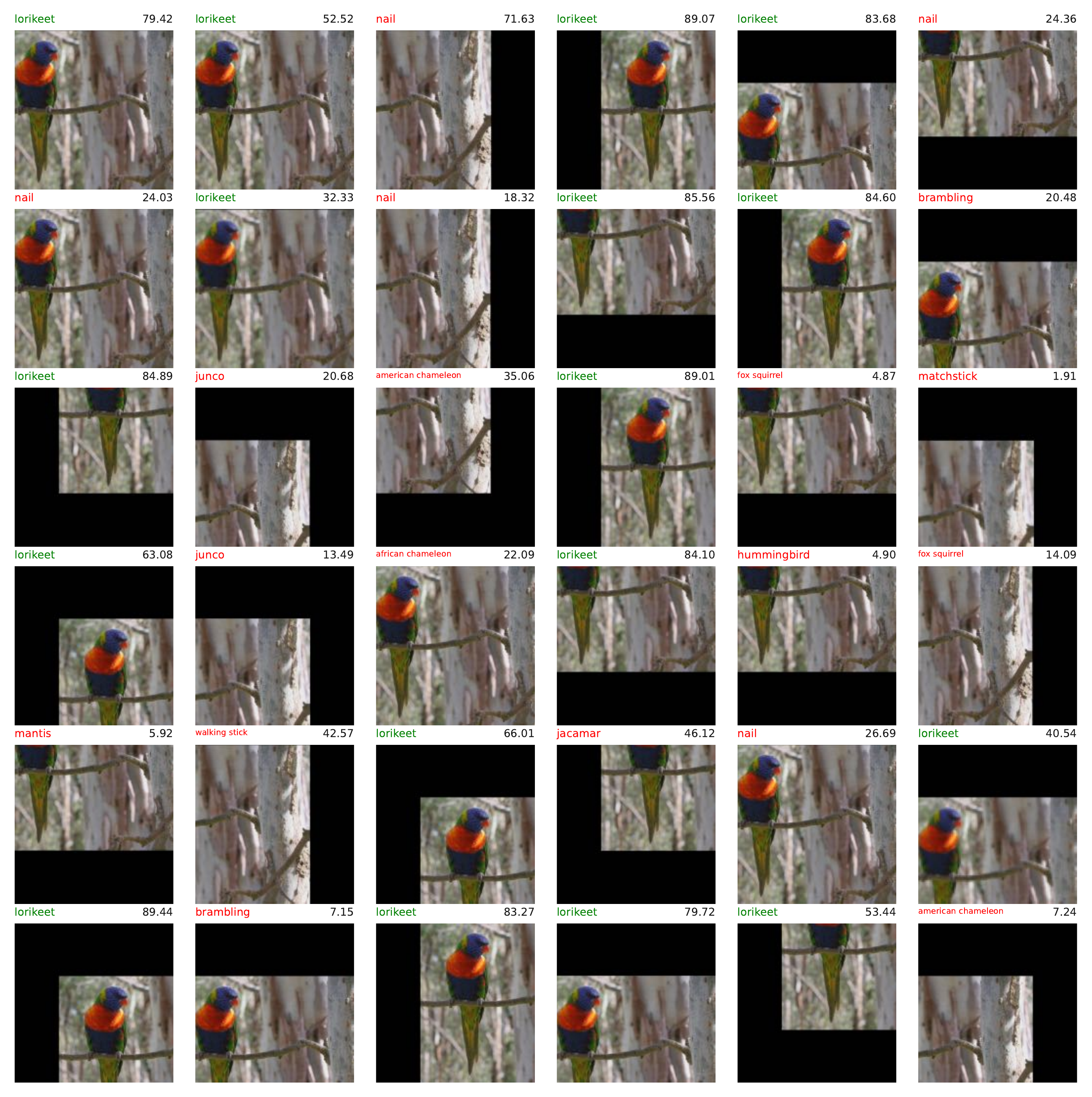}
    \caption{ 
        Different framing of an image of a \class{lorikeet} according to 36 high-performing transforms of a ViT/B-32 model
    }
    \label{suppfig:36_crop_2}
\end{figure}

\begin{figure}[!hbt]
    \centering
    \includegraphics[width=1\textwidth]{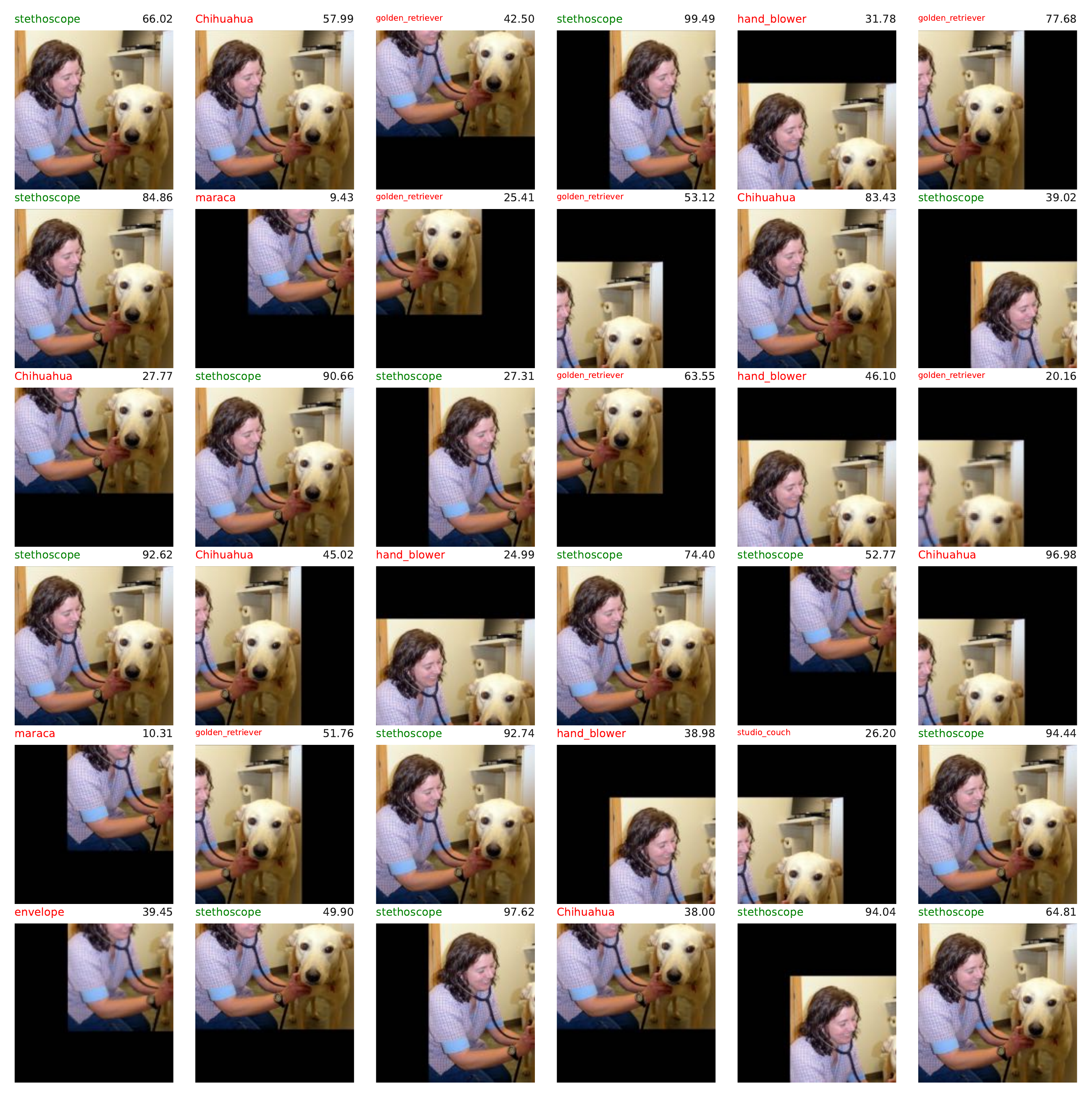}
    \caption{ 
    Different framing of an image of a \class{stethoscope} according to 36 high-performing transforms of a ResNet-50 model
    }
    \label{suppfig:36_crop_3}
\end{figure}

\begin{figure}[!hbt]
    \centering
    \includegraphics[width=1\textwidth]{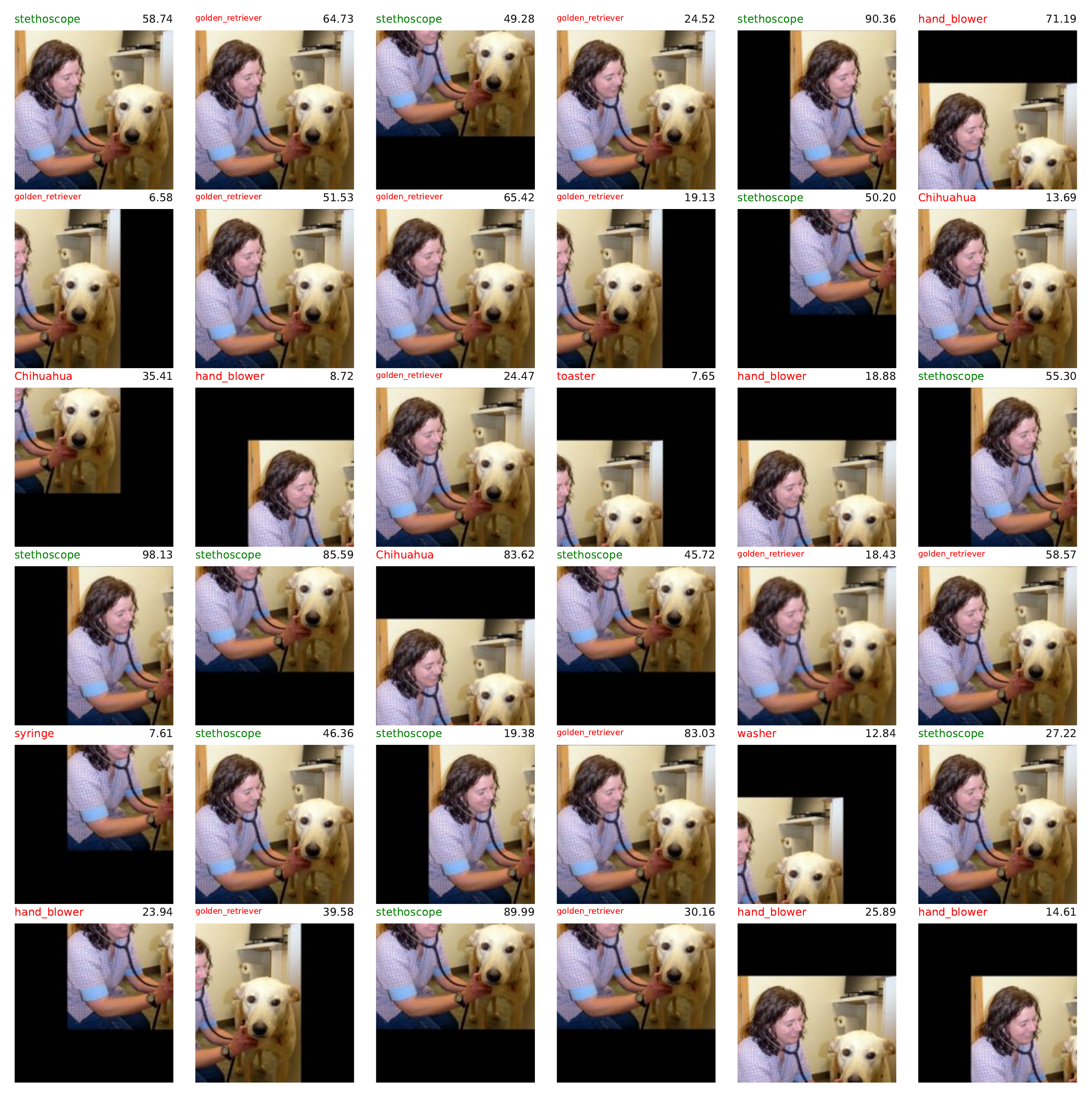}
    \caption{ 
        Different framing of an image of a \class{stethoscope} according to 36 high-performing transforms of a ViT/B-32 model
    }
    \label{suppfig:36_crop_4}
\end{figure}

\clearpage
\subsection{Overview of $324$ transforms}

The visualizations below illustrate the transforms that result in the correct prediction of the query image, using  ViT-B/32~\cite{dosovitskiy2020image} and CLIP-ViT-L/14~\cite{radford2021learning}.
Each circle represents a transform, with the initial zoom scale indicated in the accompanying text.
The green circles represent the transformations that lead to correct classification, while the red circles indicate incorrect ones.

\begin{figure}[!hbt]
    \centering
    \includegraphics[width=1\textwidth]{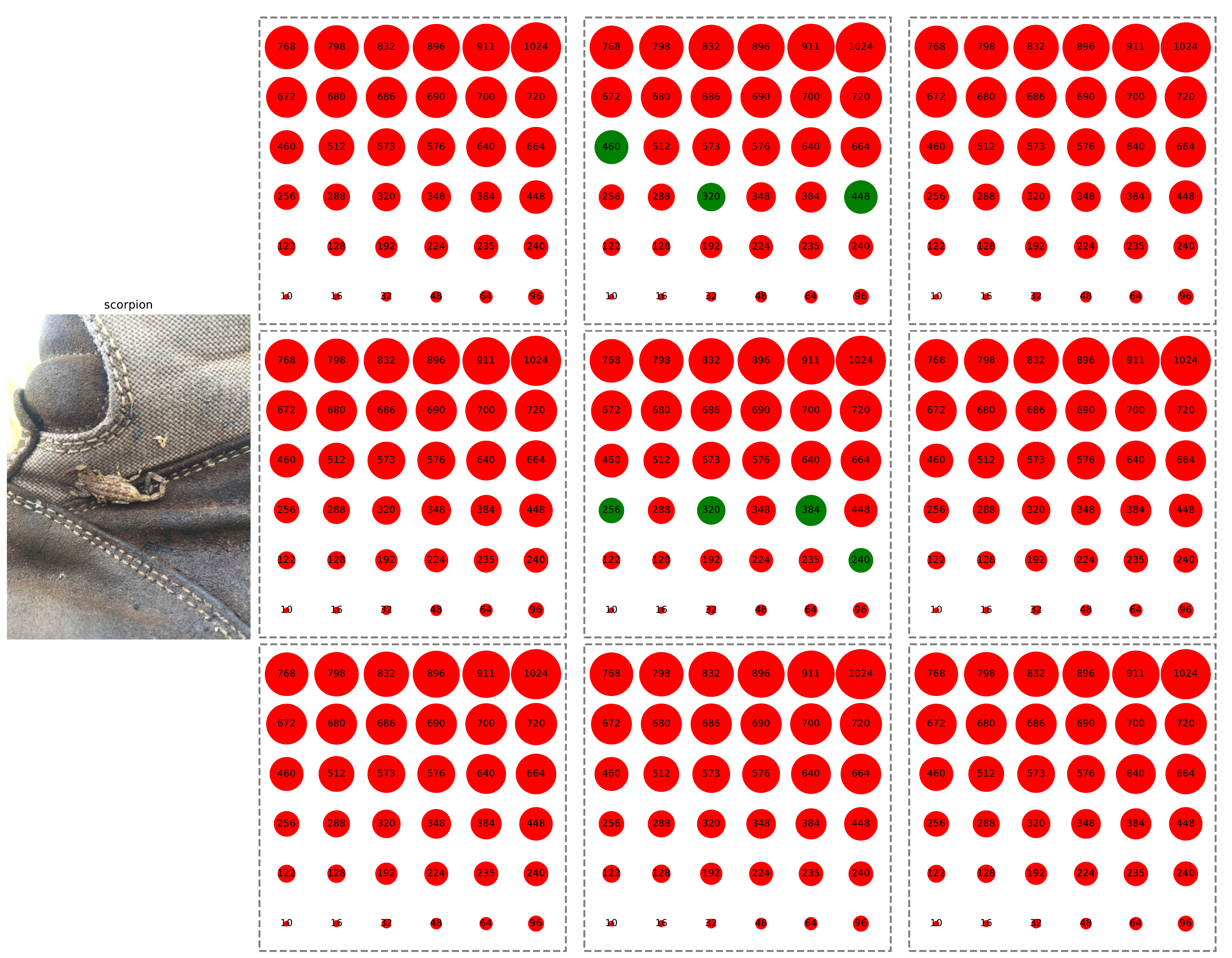}
    \caption{ 
        Visualization of effective transforms that lead to the correct classification of an image containing \class{scorpion}, using a ViT-B/32 model.
    }
    \label{fig:viz1_vit}
\end{figure}

\begin{figure}[!hbt]
    \centering
    \includegraphics[width=1\textwidth]{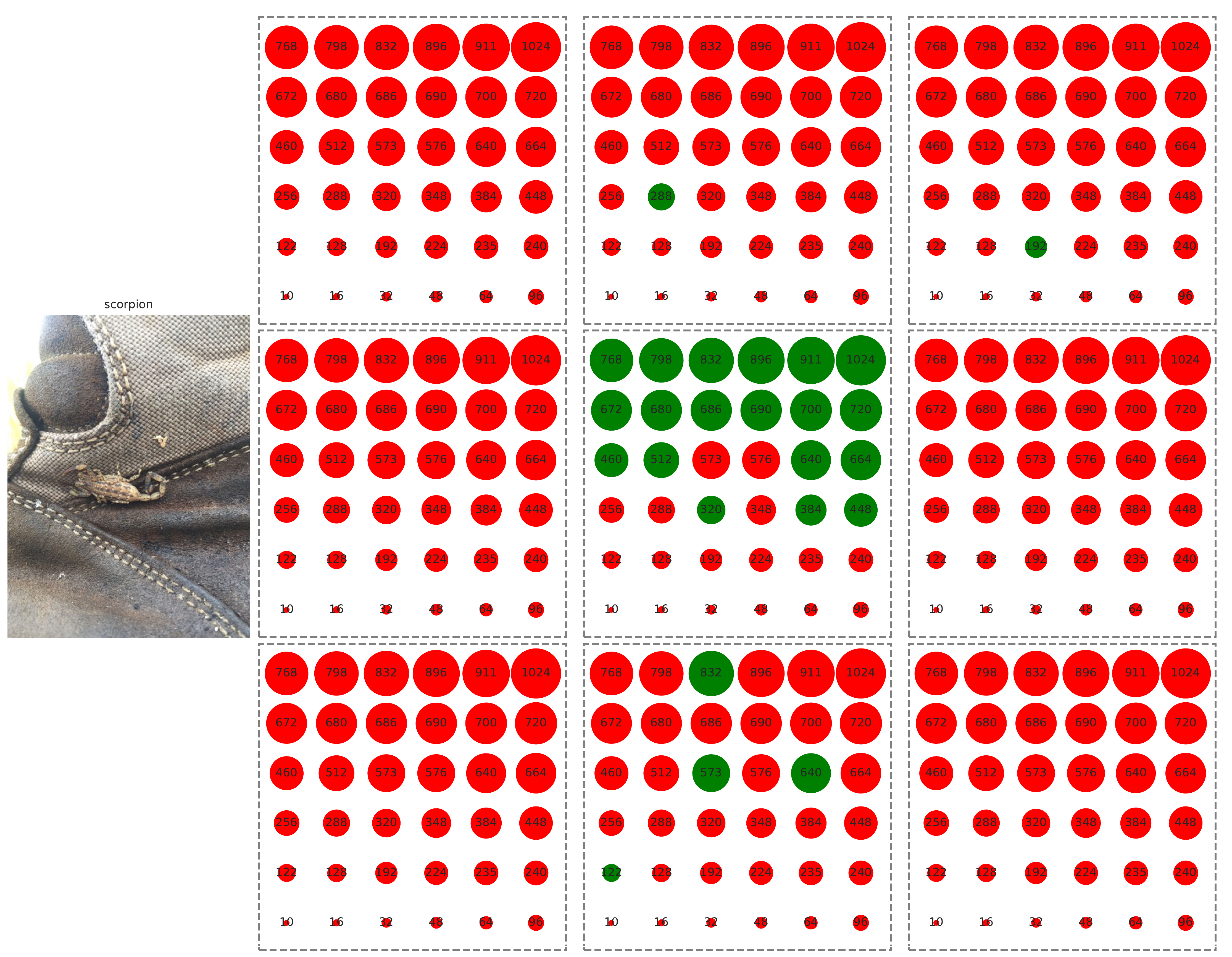}
     \caption{ 
        Visualization of effective transforms that lead to the correct classification of an image containing \class{scorpion}, using a CLIP-ViT-L/14 model.
    }
    \label{fig:viz1_clipvit}
\end{figure}


\begin{figure}[!hbt]
    \centering
    \includegraphics[width=1\textwidth]{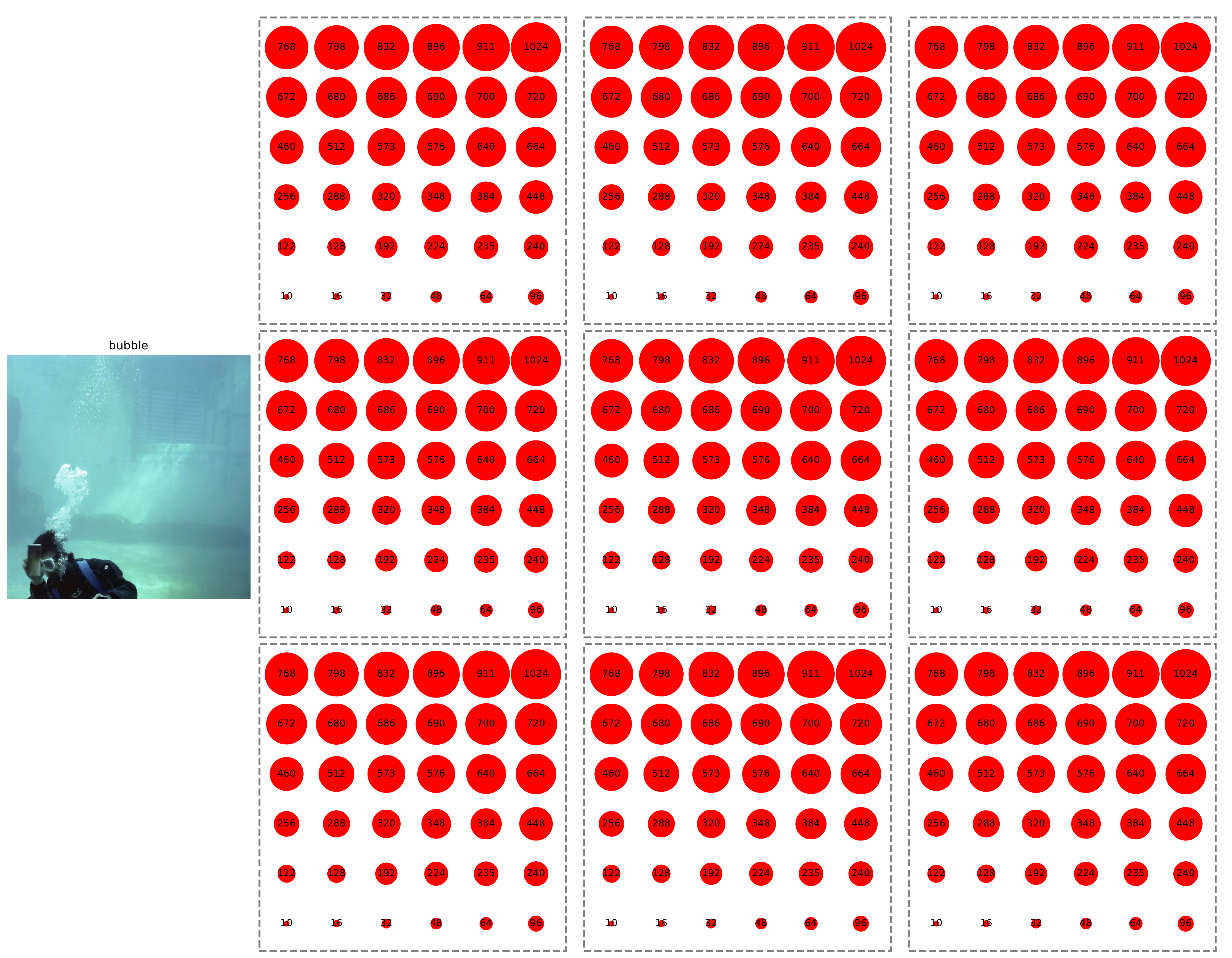}
    \caption{ 
        Visualization of effective transforms that lead to the correct classification of an image containing \class{bubble}, using a ViT-B/32 model.
    }
    \label{fig:viz2_vit}
\end{figure}

\begin{figure}[!hbt]
    \centering
    \includegraphics[width=1\textwidth]{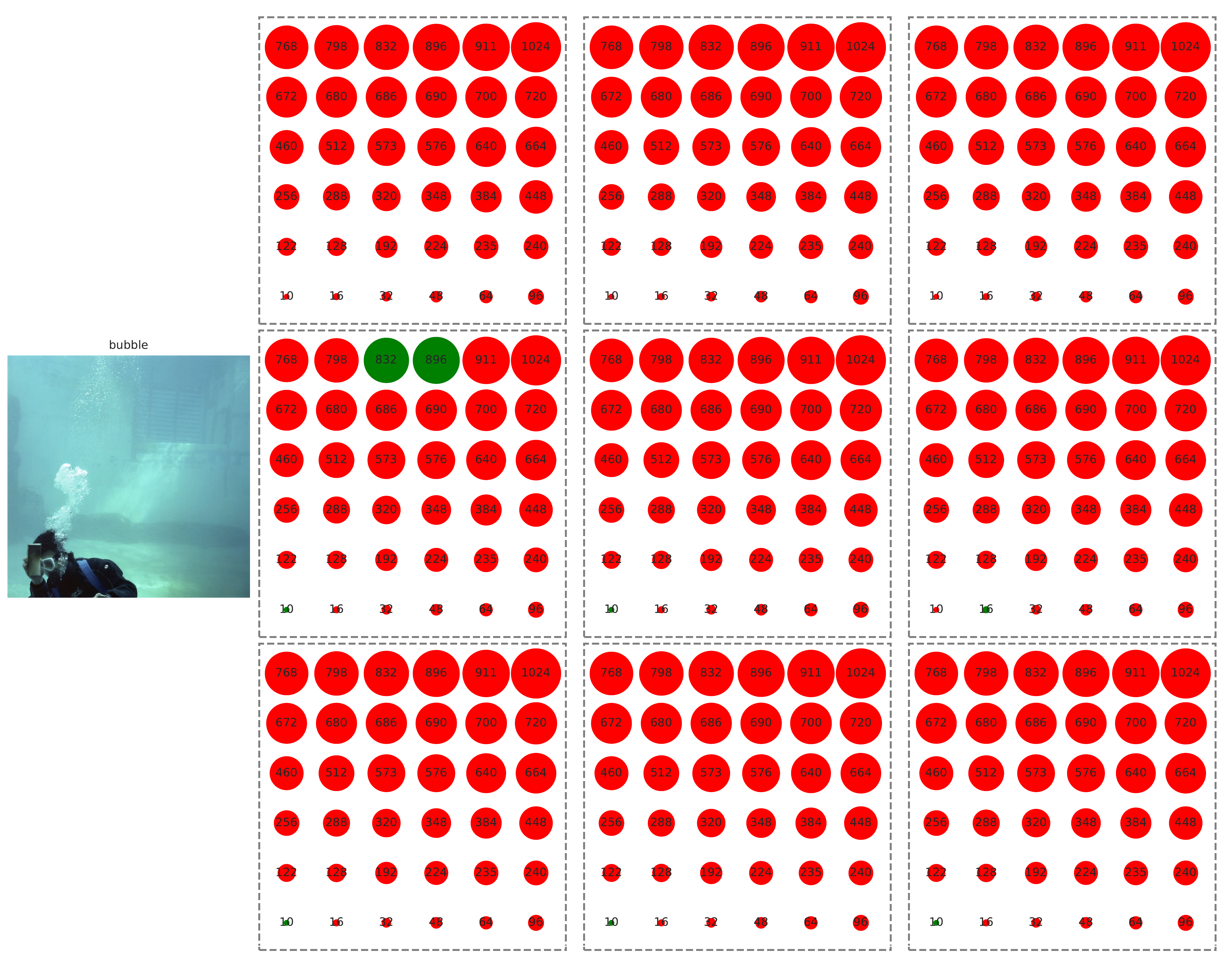}
     \caption{ 
        Visualization of effective transforms that lead to the correct classification of an image containing \class{bubble}, using a CLIP-ViT-L/14 model.
    }
    \label{fig:viz2_clipvit}
\end{figure}


\begin{figure}[!hbt]
    \centering
    \includegraphics[width=1\textwidth]{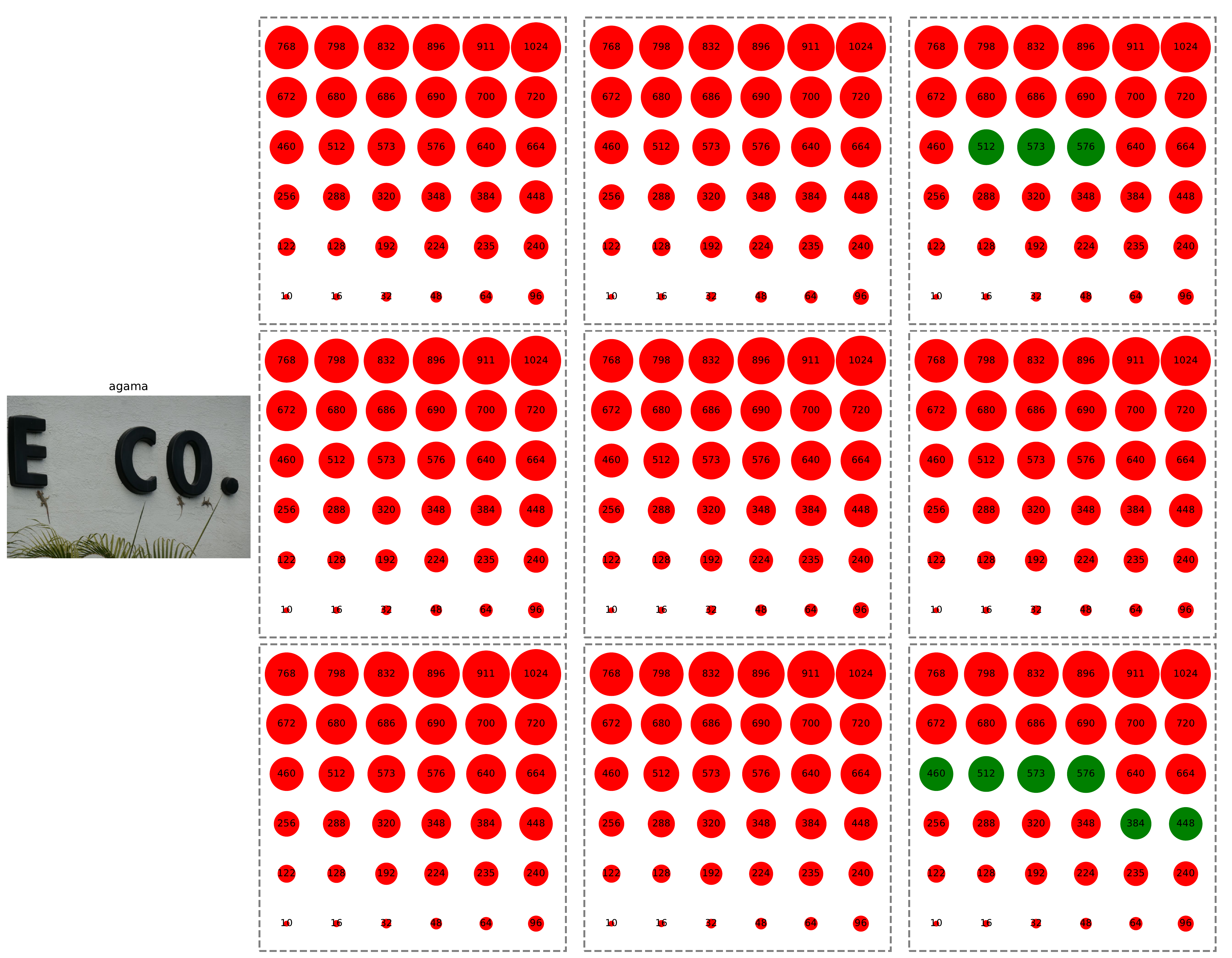}
    \caption{ 
        Visualization of effective transforms that lead to the correct classification of an image containing \class{agama}, using a ViT-B/32 model.
    }
    \label{fig:viz2_vit2}
\end{figure}

\begin{figure}[!hbt]
    \centering
    \includegraphics[width=1\textwidth]{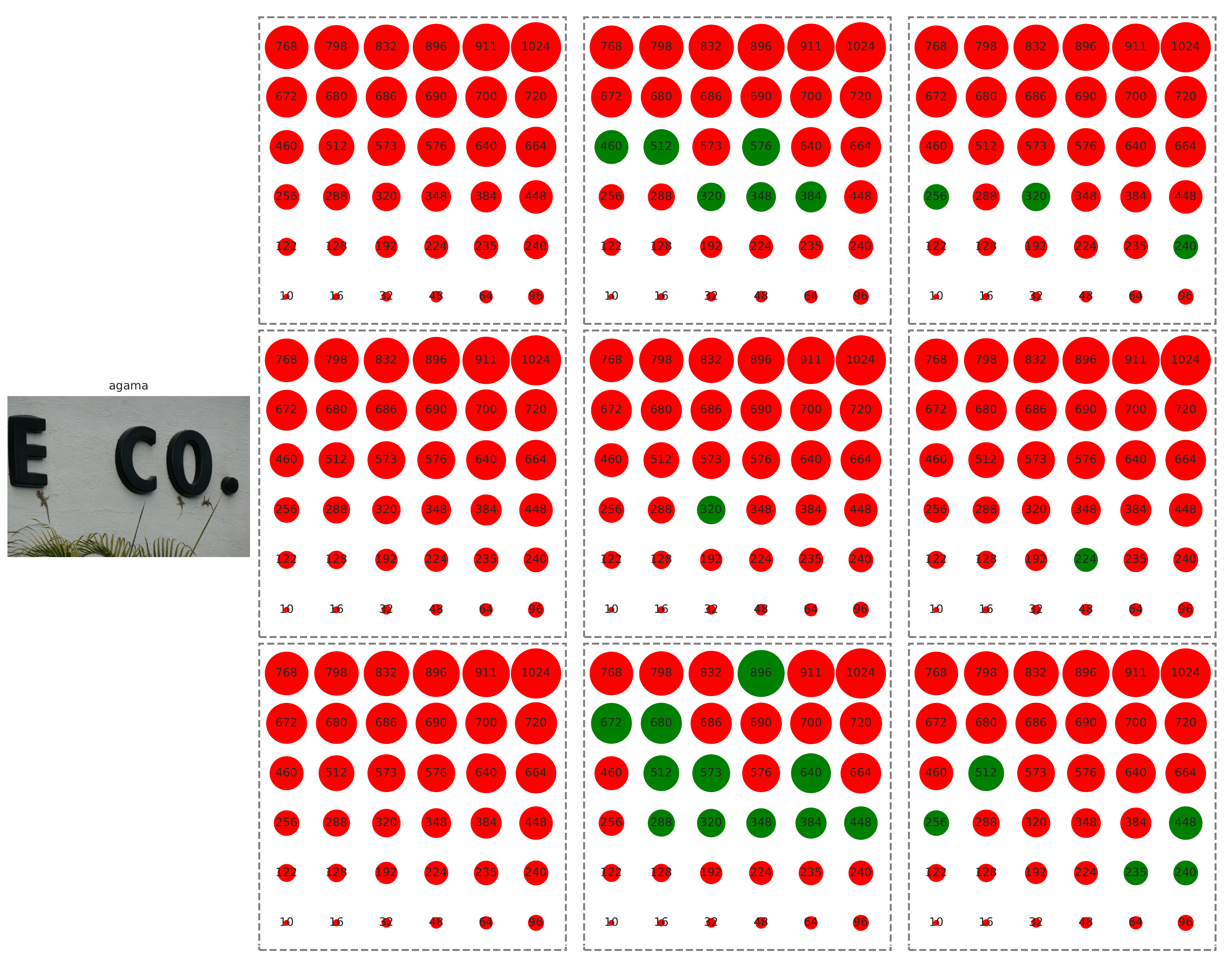}
     \caption{ 
        Visualization of effective transforms that lead to the correct classification of an image containing \class{agama}, using a CLIP-ViT-L/14 model.
    }
    \label{fig:viz2_clipvit2}
\end{figure}


\begin{figure}[!hbt]
    \centering
    \includegraphics[width=1\textwidth]{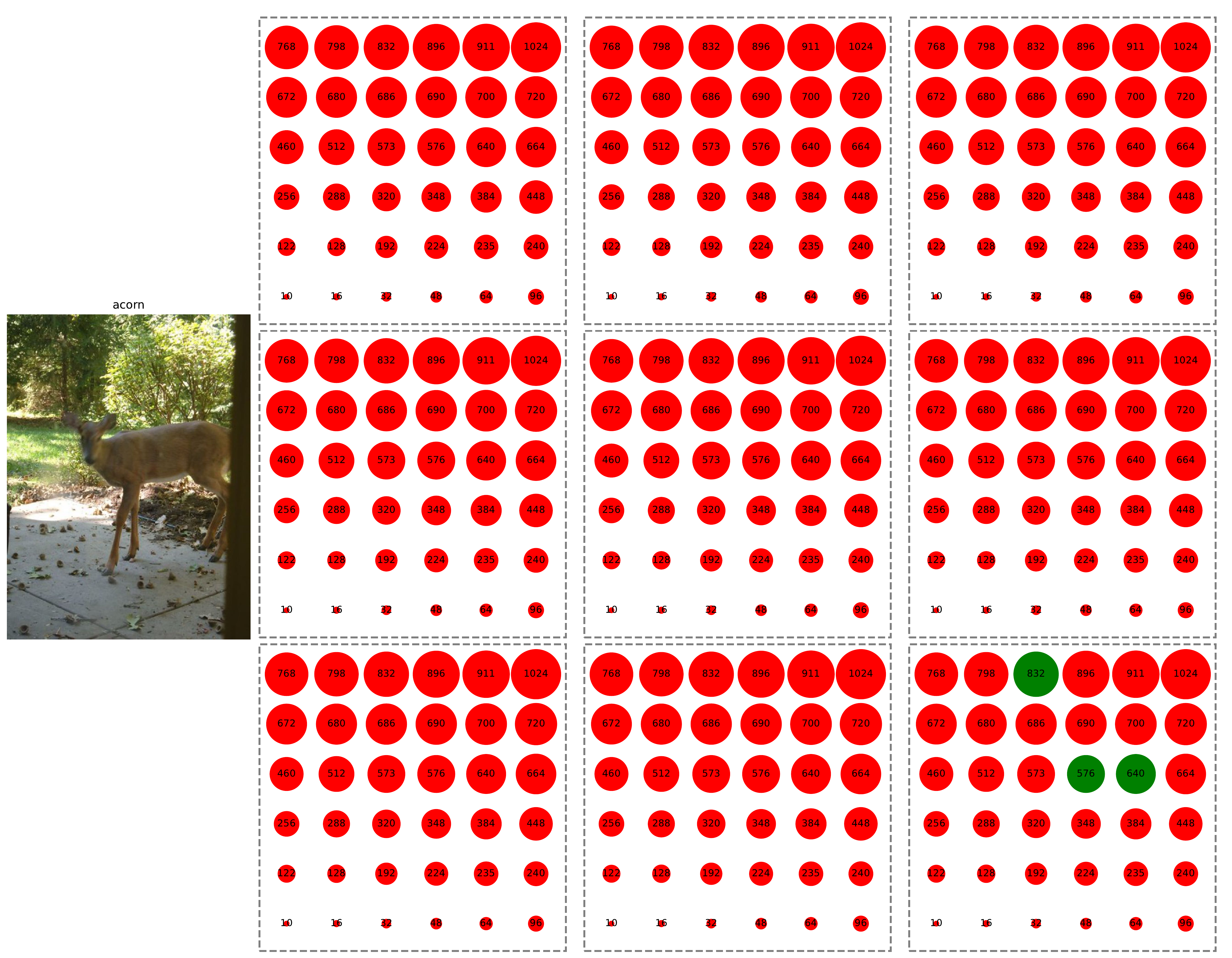}
    \caption{ 
        Visualization of effective transforms that lead to the correct classification of an image containing \class{acorn}, using a ViT-B/32 model.
    }
    \label{fig:viz2_vit3}
\end{figure}

\begin{figure}[!hbt]
    \centering
    \includegraphics[width=1\textwidth]{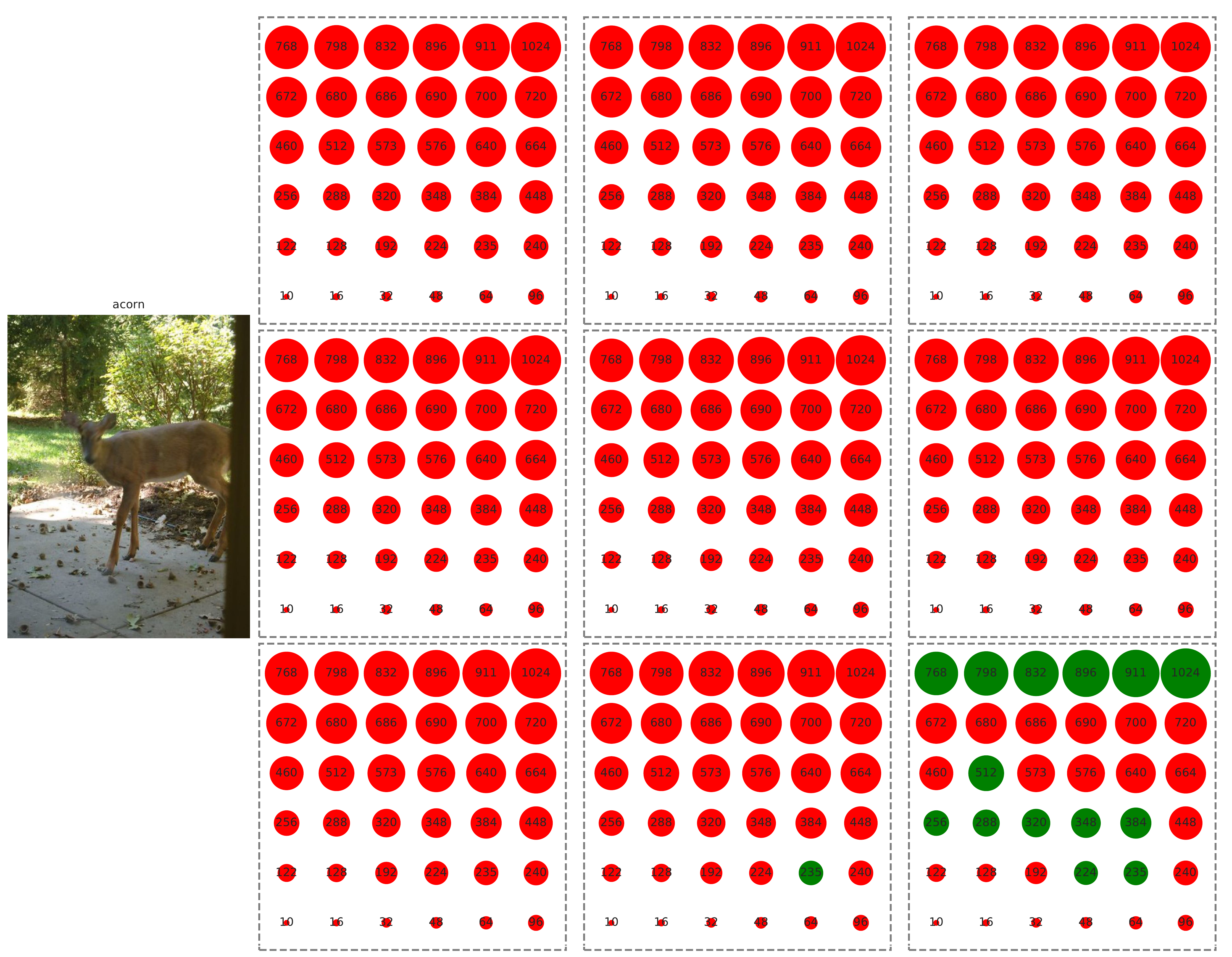}
     \caption{ 
        Visualization of effective transforms that lead to the correct classification of an image containing \class{acorn}, using a CLIP-ViT-L/14 model.
    }
    \label{fig:viz2_clipvit3}
\end{figure}

\clearpage
\subsection{Only \zoomout solves}
\label{supp:only_zoomout_sample_images}
Sample images that required zooming out to be  classified correctly.

\subsubsection{ImageNet-Sketch}
\label{suppsec:imagenet_sketch_zoomout}

\begin{figure}[!hbt]
    \centering
    \begin{subfigure}[b]{0.48\linewidth}   
        \centering
        \includegraphics[width=\linewidth]{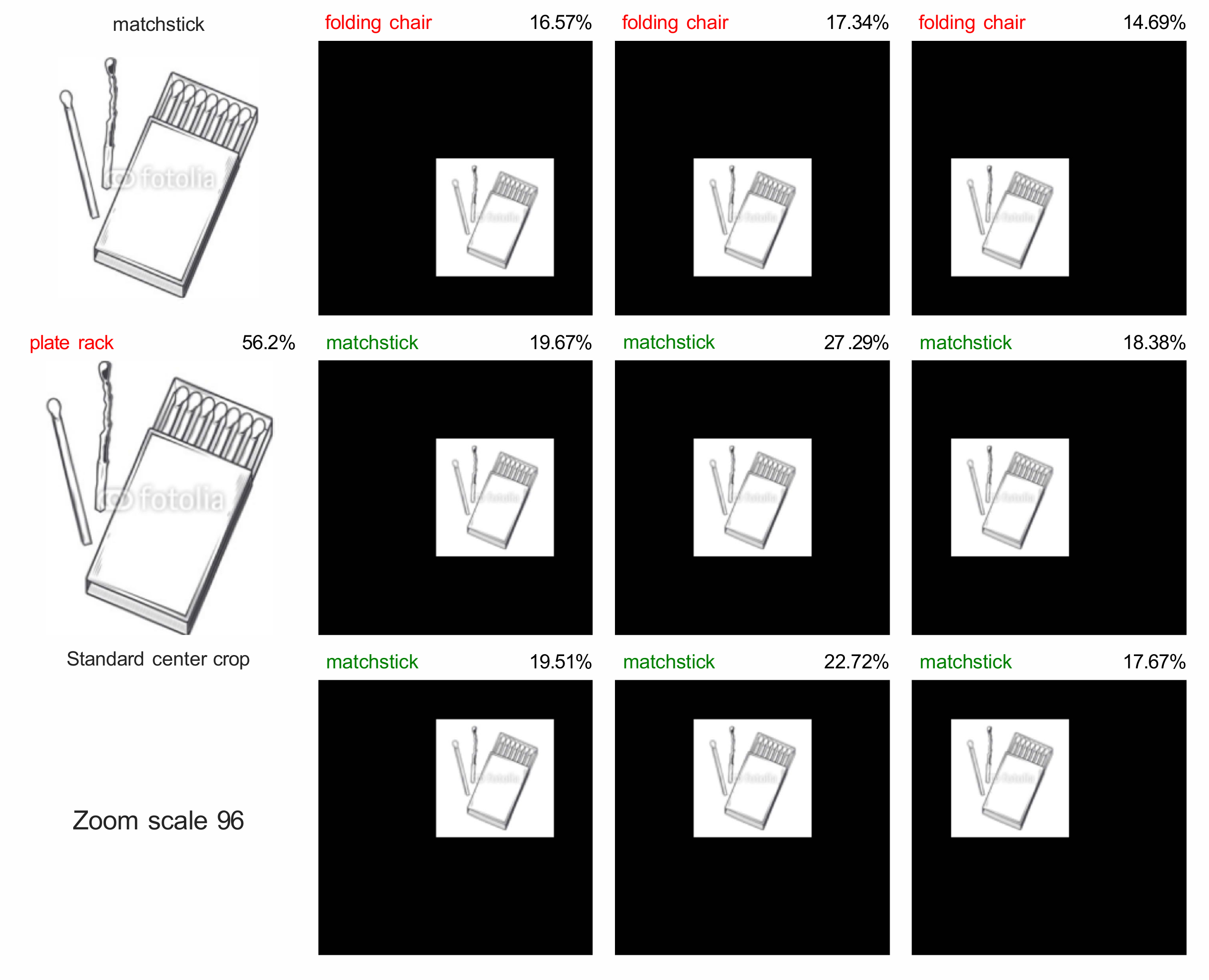}
    \end{subfigure}
    \begin{subfigure}[b]{0.48\linewidth}  
        \centering
        \includegraphics[width=\linewidth]{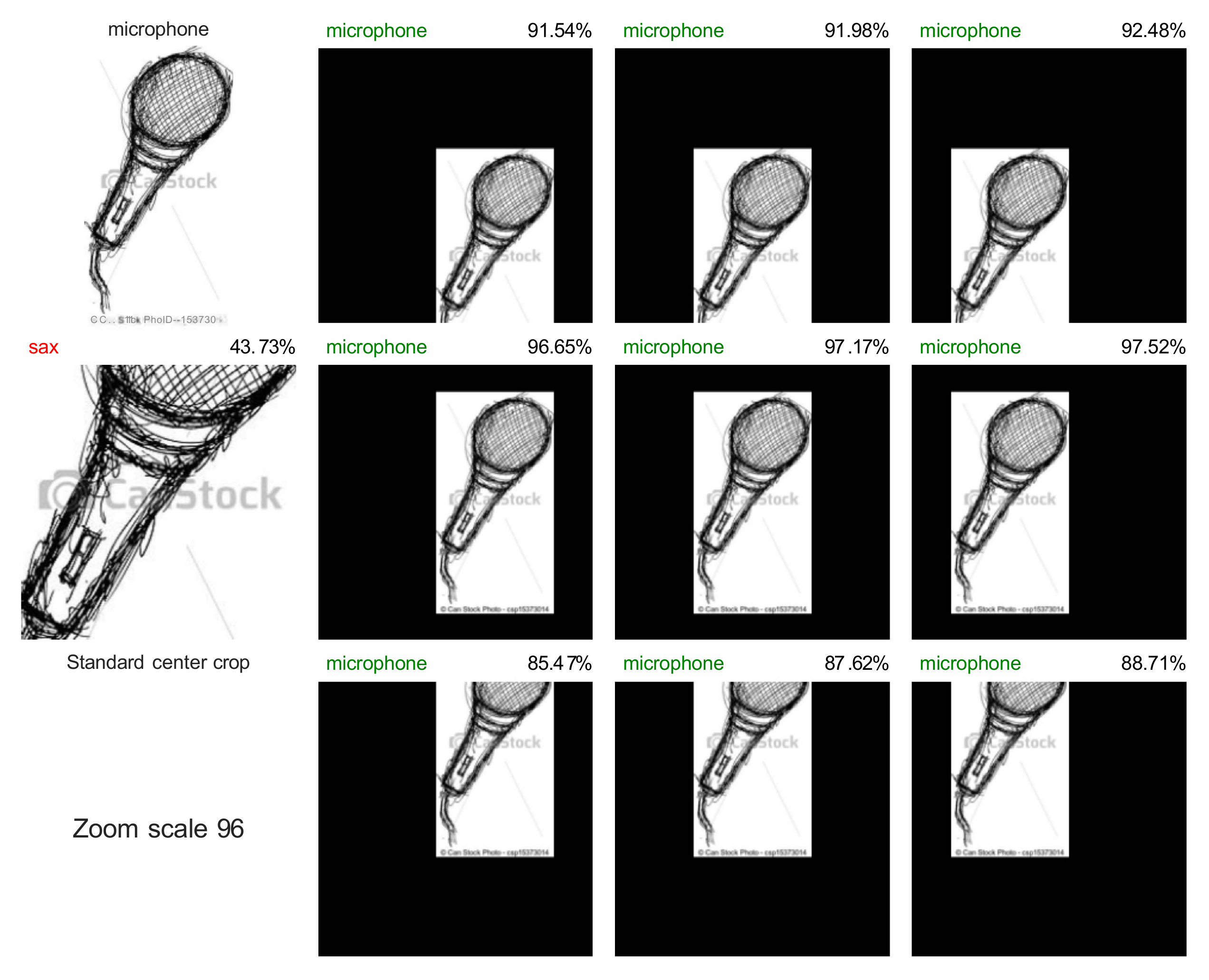}
    \end{subfigure}
    \centering
    \begin{subfigure}[b]{0.48\linewidth}   
        \centering
        \includegraphics[width=\linewidth]{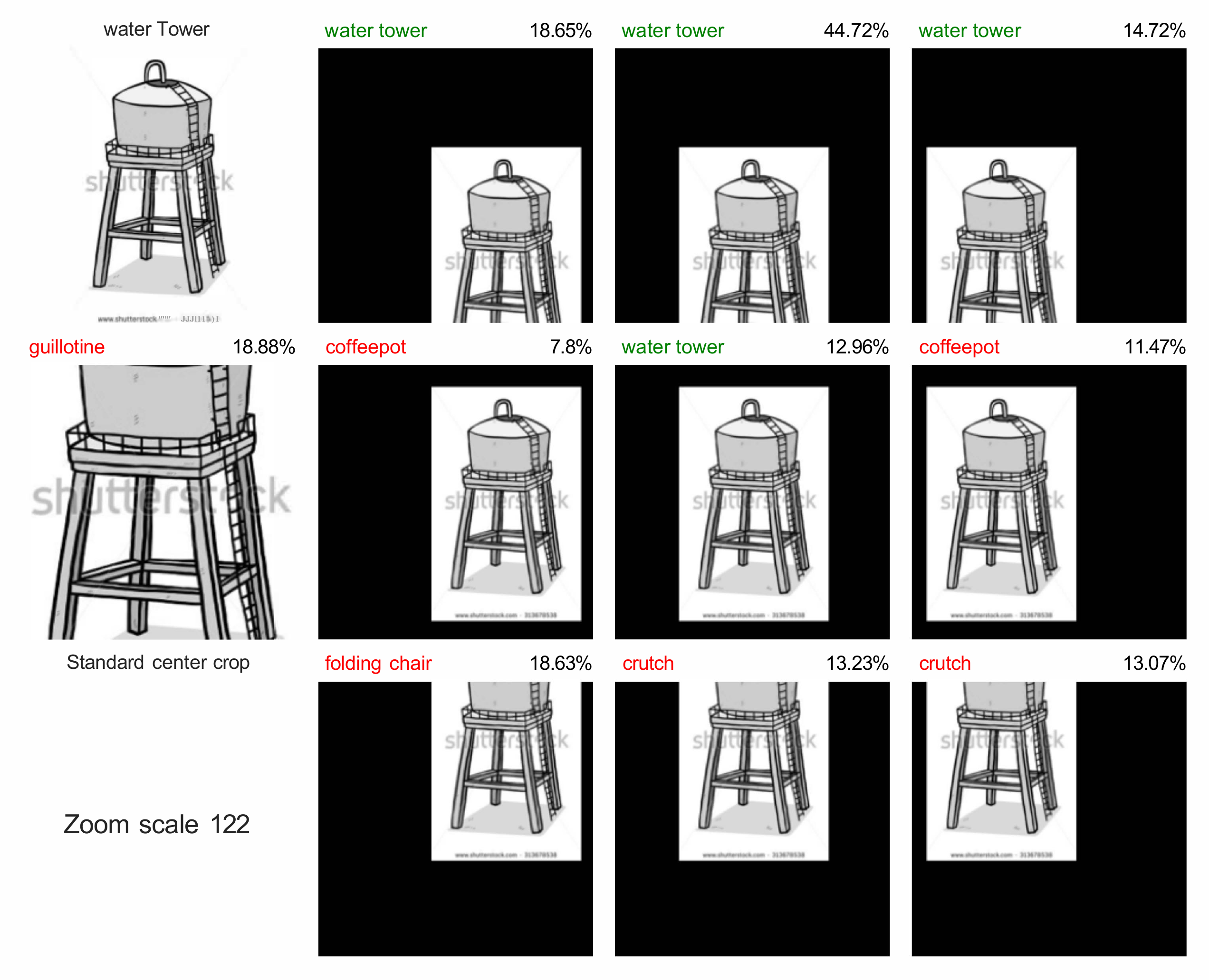}
    \end{subfigure}
    \begin{subfigure}[b]{0.48\linewidth}  
        \centering
        \includegraphics[width=\linewidth]{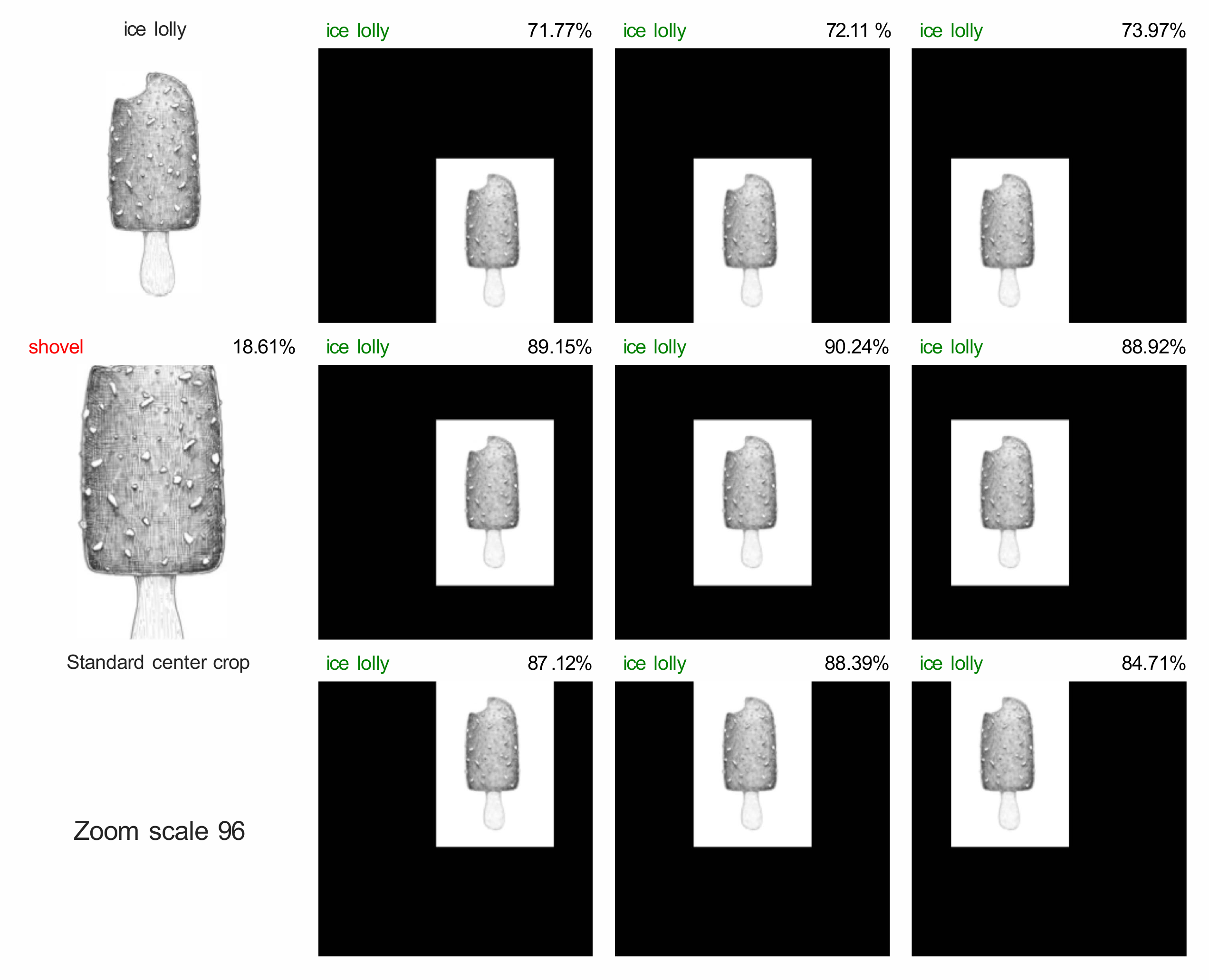}
    \end{subfigure}
    
    \caption{ImageNet-Sketch images that can only be solved using \zoomout. Predictions are from a ResNet-50 classifier.}
    \label{suppfig:hard_samples_sketch}
\end{figure}

\clearpage
\subsubsection{ImageNet-R}
\label{suppsec:imagenet_R_zoomout}

\begin{figure}[!hbt]
    \centering
    \begin{subfigure}[b]{0.48\linewidth}   
        \centering
        \includegraphics[width=\linewidth]{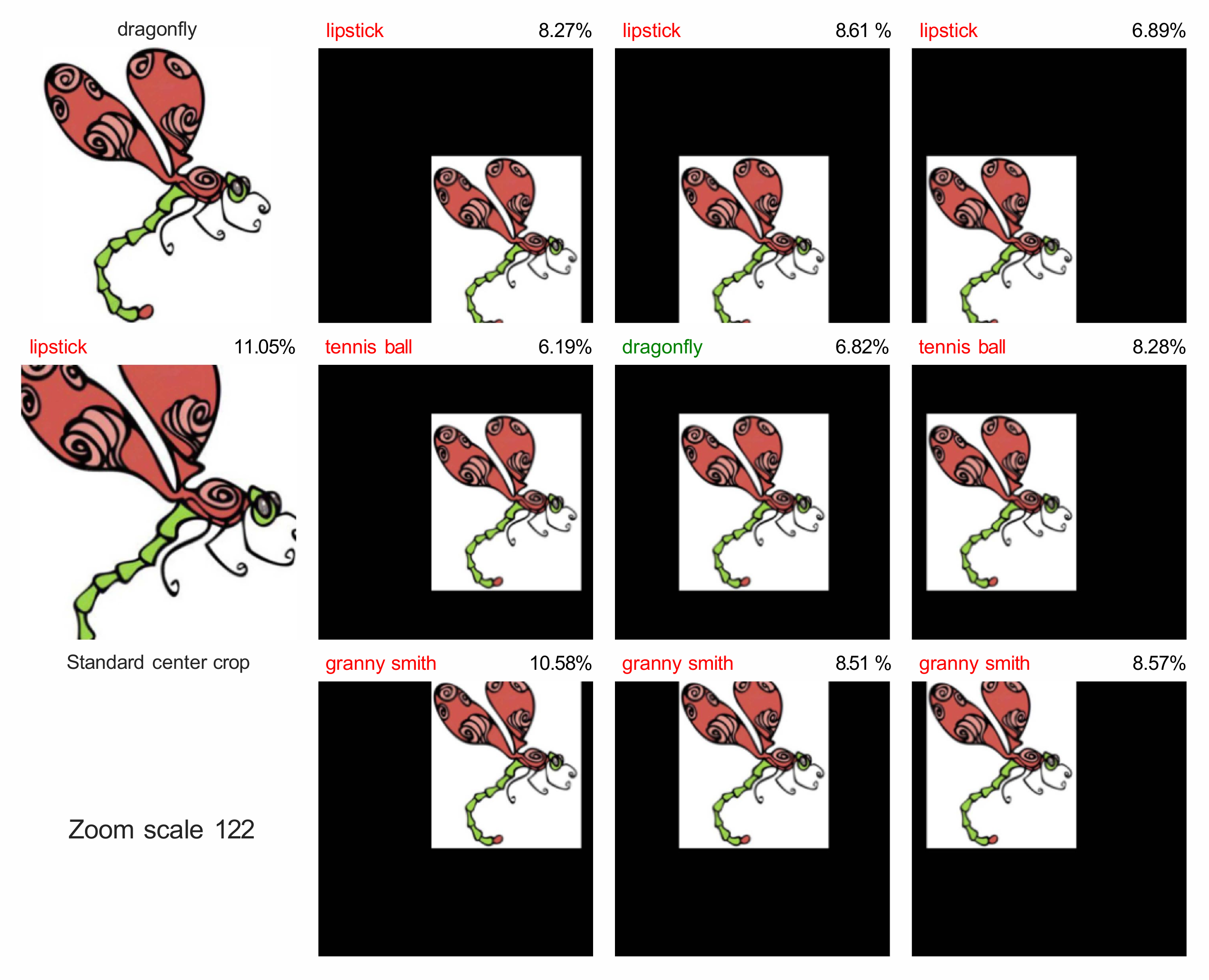}
    \end{subfigure}
    \begin{subfigure}[b]{0.48\linewidth}  
        \centering
        \includegraphics[width=\linewidth]{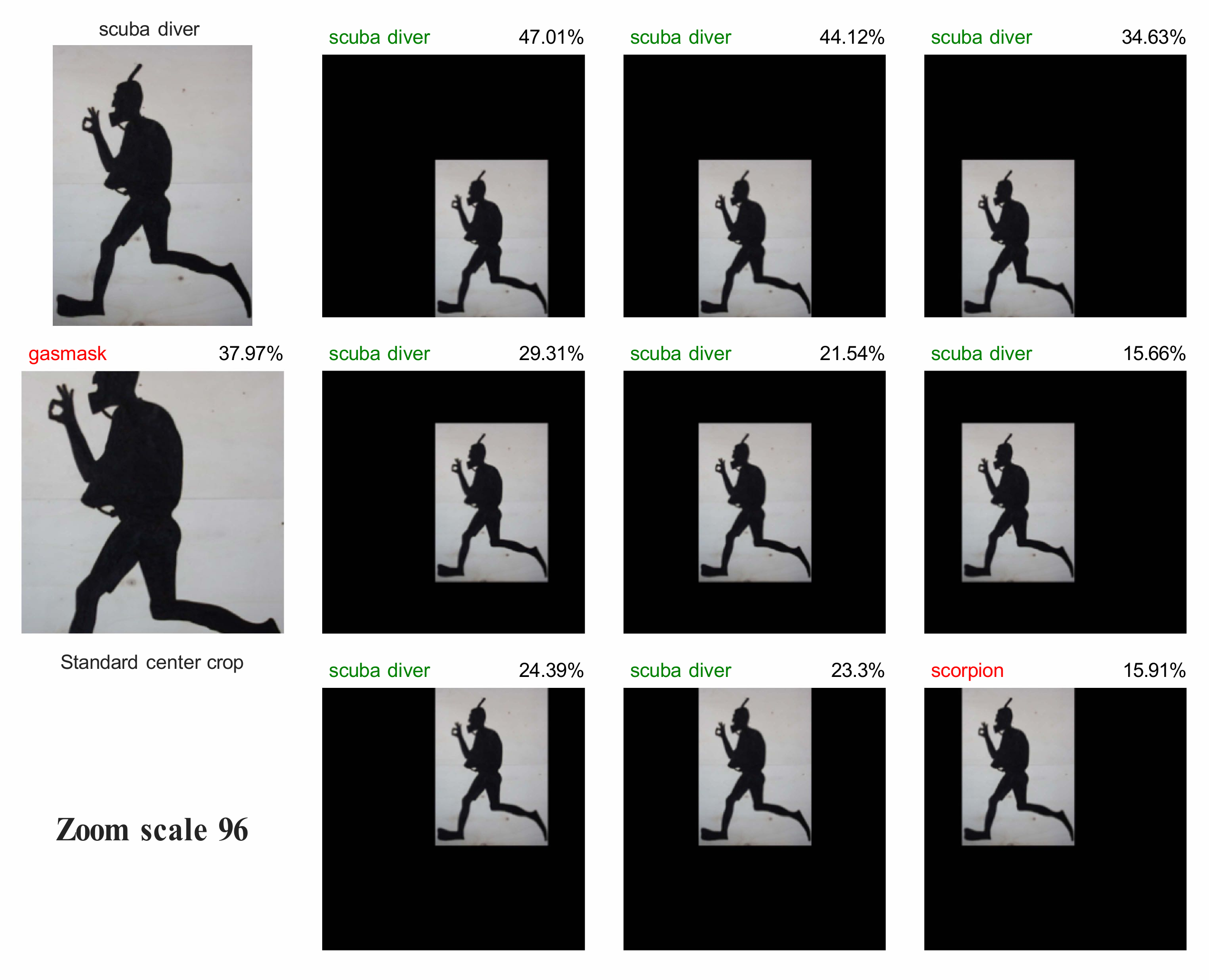}
    \end{subfigure}
    \centering
    \begin{subfigure}[b]{0.48\linewidth}   
        \centering
        \includegraphics[width=\linewidth]{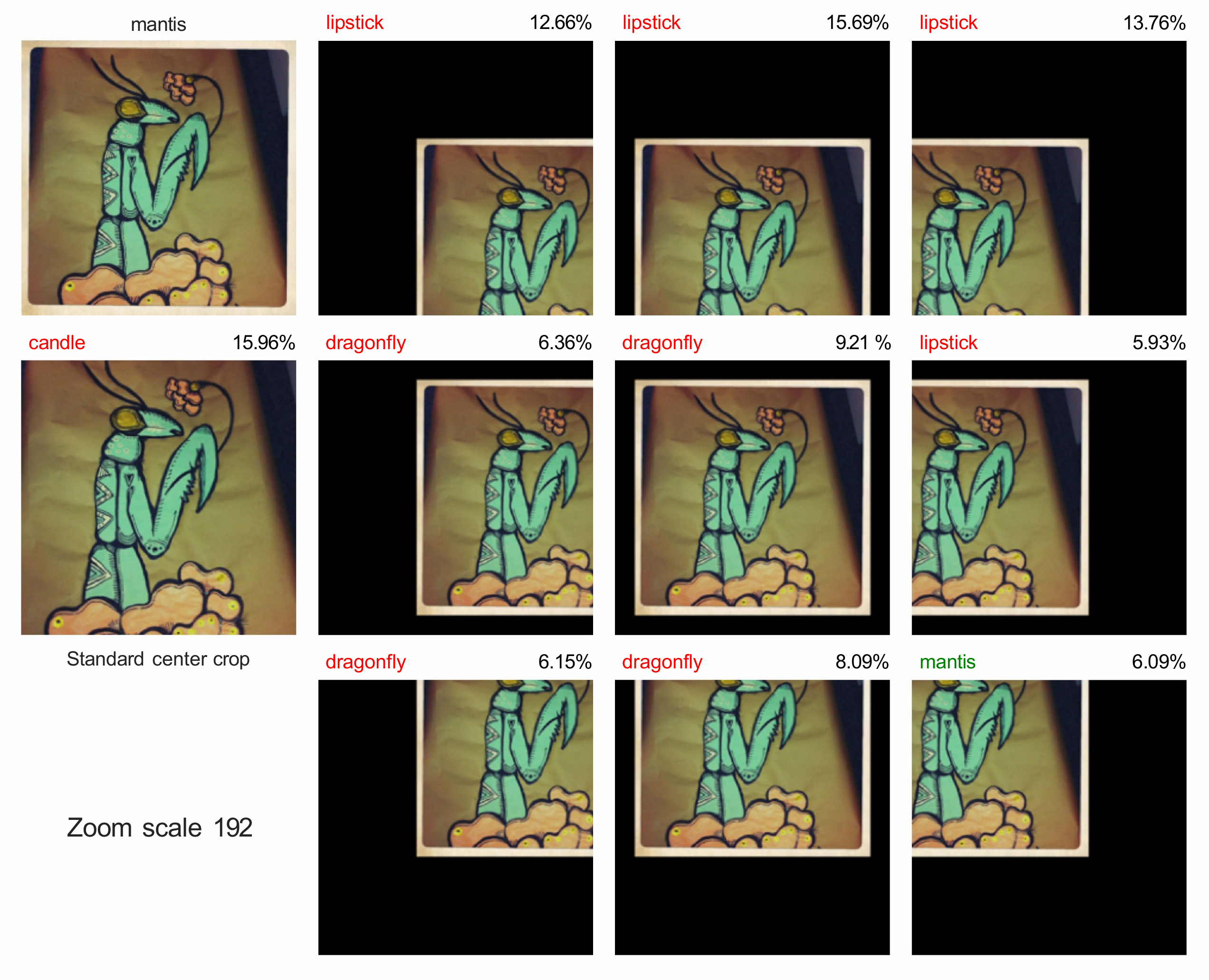}
    \end{subfigure}
    \begin{subfigure}[b]{0.48\linewidth}  
        \centering
        \includegraphics[width=\linewidth]{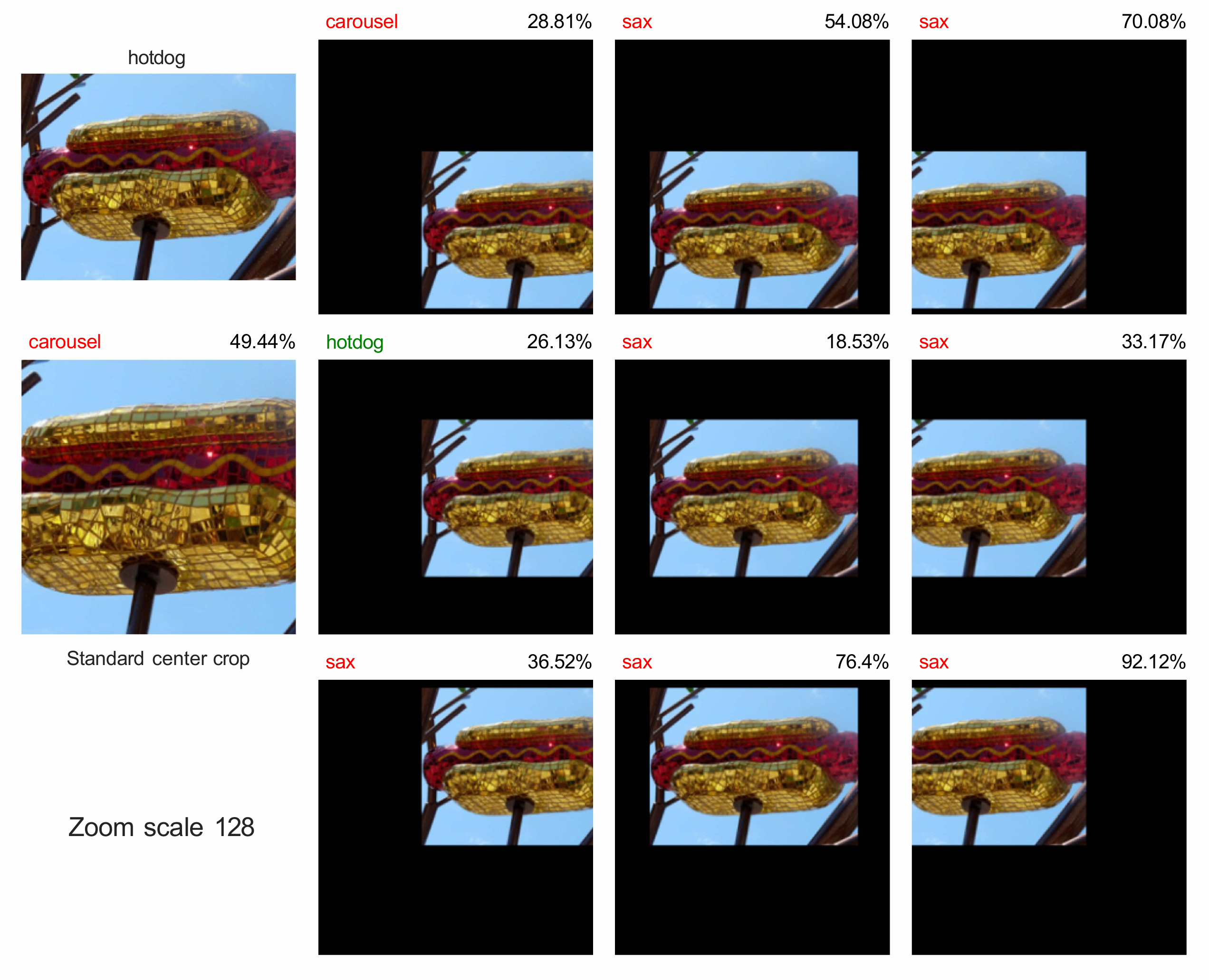}
    \end{subfigure}
    
 \caption{ImageNet-R images that can only be solved using \zoomout. Predictions are from a ResNet-50 classifier.}
    \label{suppfig:hard_samples_r}
\end{figure}

\clearpage
\subsubsection{ObjectNet}
\label{suppsec:imagenet_objectnet}

\begin{figure}[!hbt]
    \centering
    \begin{subfigure}[b]{0.48\linewidth}   
        \centering
        \includegraphics[width=\linewidth]{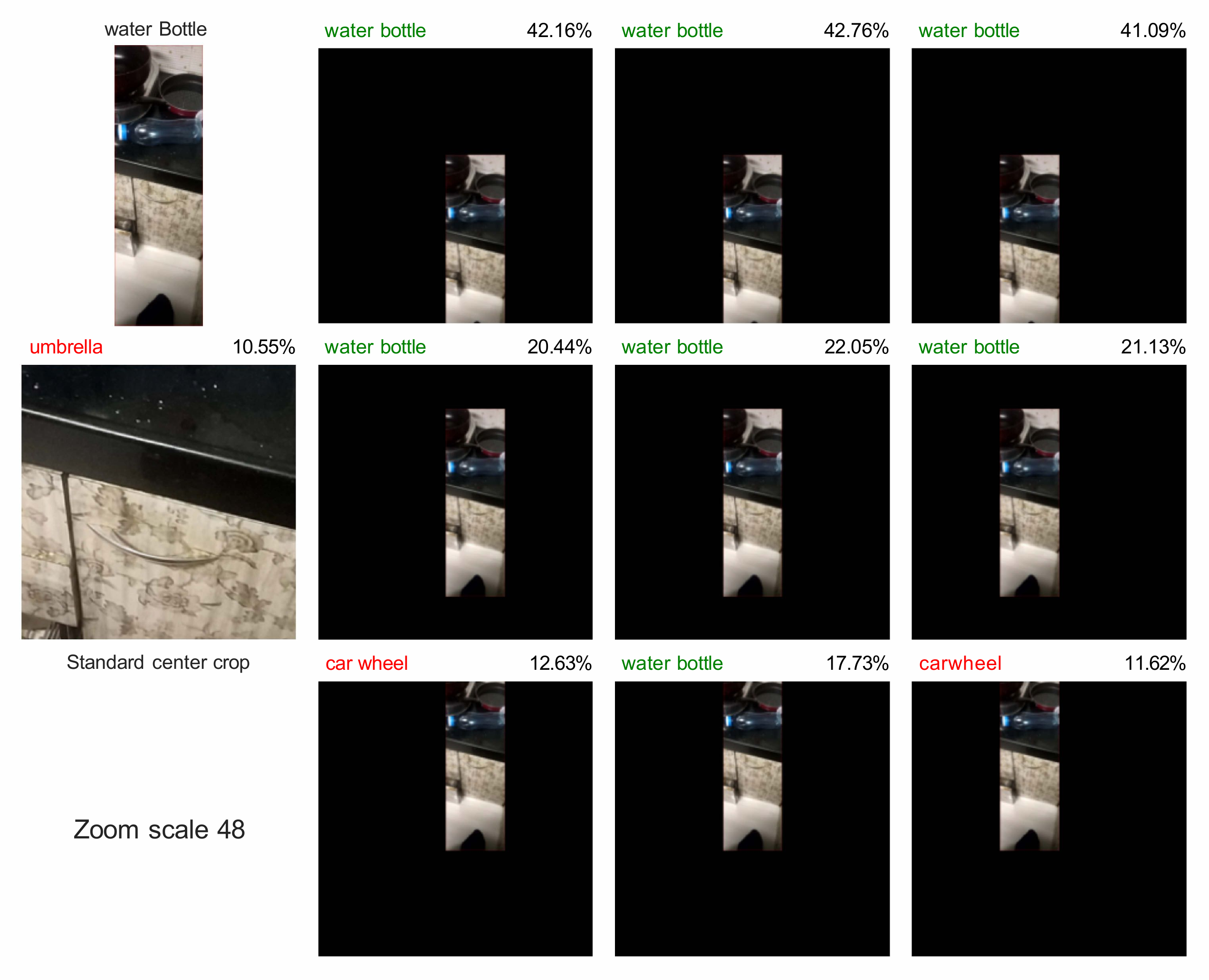}
    \end{subfigure}
    \begin{subfigure}[b]{0.48\linewidth}  
        \centering
         \includegraphics[width=\linewidth]{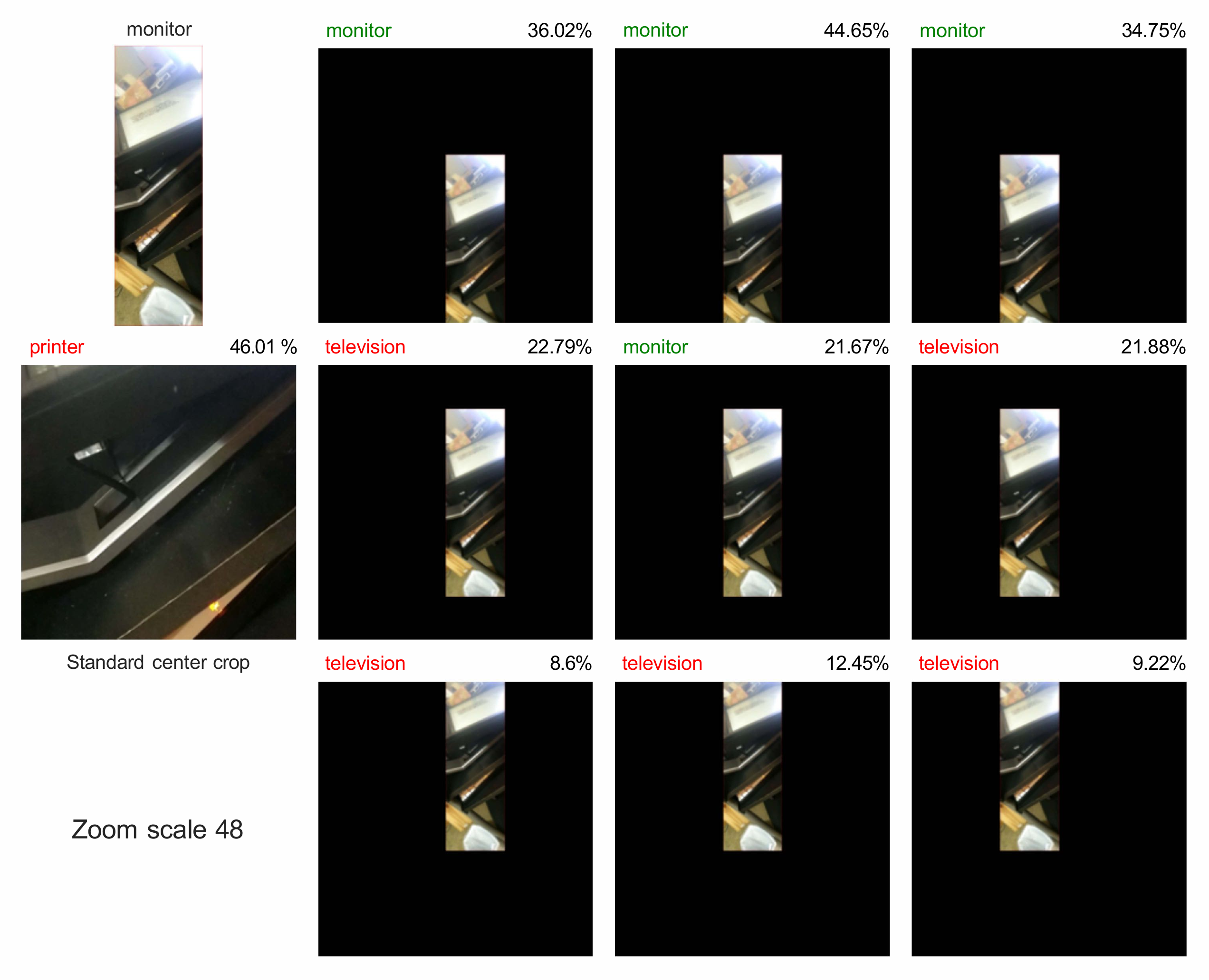}
    \end{subfigure}
    \centering
    \begin{subfigure}[b]{0.48\linewidth}   
        \centering
         \includegraphics[width=\linewidth]{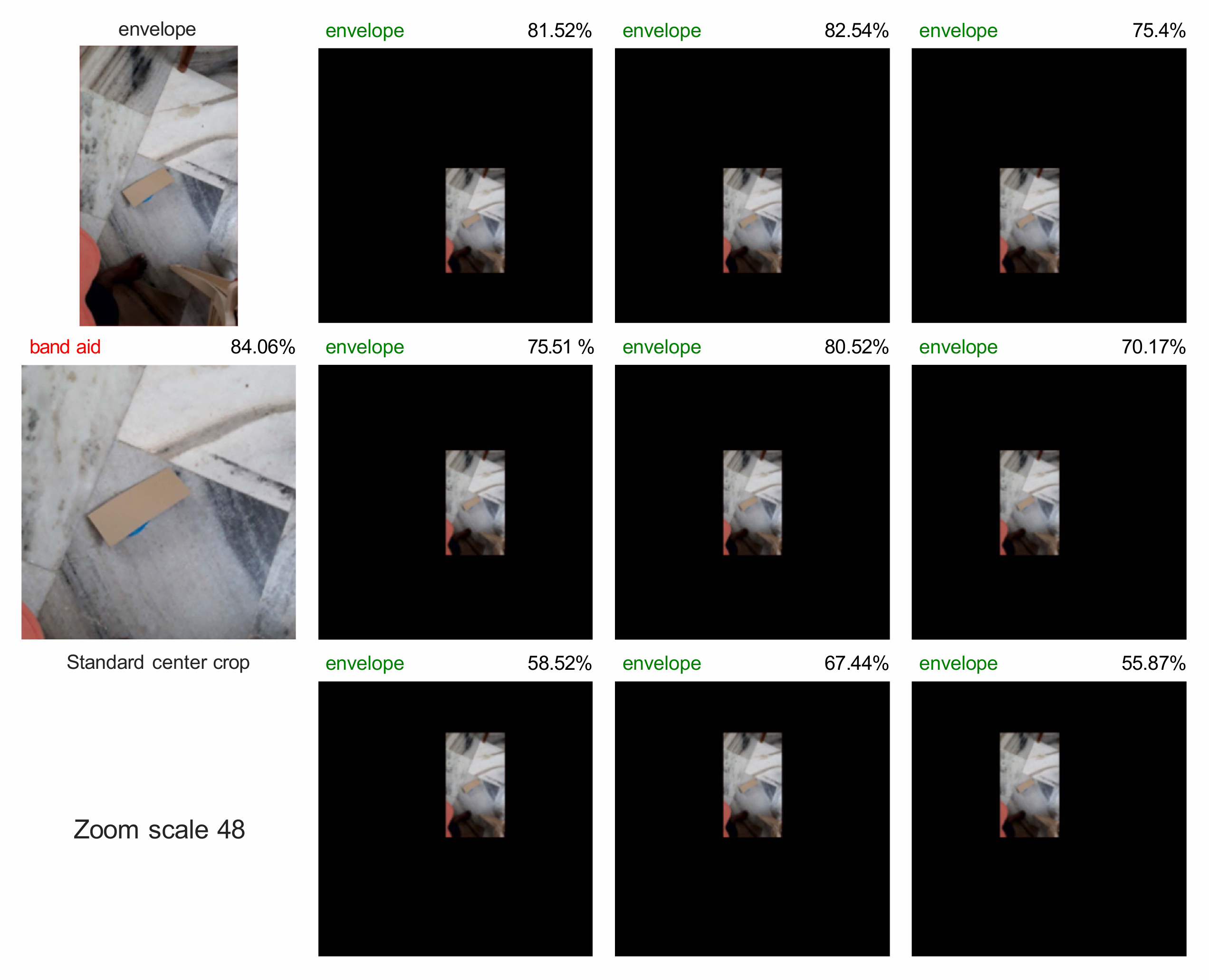}
    \end{subfigure}
    \begin{subfigure}[b]{0.48\linewidth}  
        \centering
         \includegraphics[width=\linewidth]{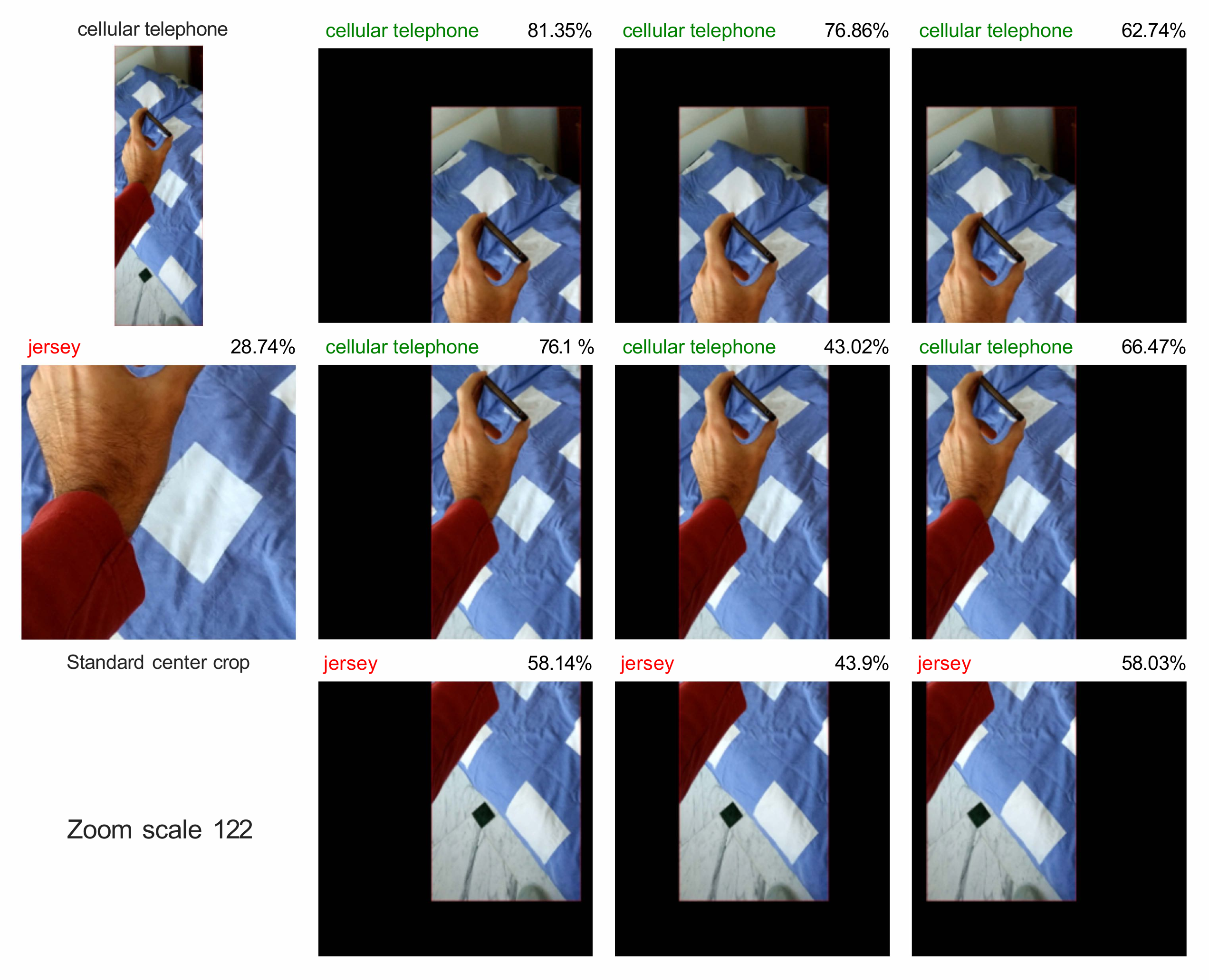}
    \end{subfigure}
    
 \caption{ObjectNet images that can only be solved using \zoomout. Predictions are from a ResNet-50 classifier.}
    \label{suppfig:hard_samples_objectnet}
\end{figure}

\clearpage
\subsection{Only \zoomin solves}
Sample images that required zooming in to be  classified correctly.

\subsubsection{ObjectNet}
\label{suppsec:imagenet_objectnet_zoom_in}

\begin{figure}[!hbt]
    \centering
    \begin{subfigure}[b]{0.48\linewidth}   
        \centering
        \includegraphics[width=\linewidth]{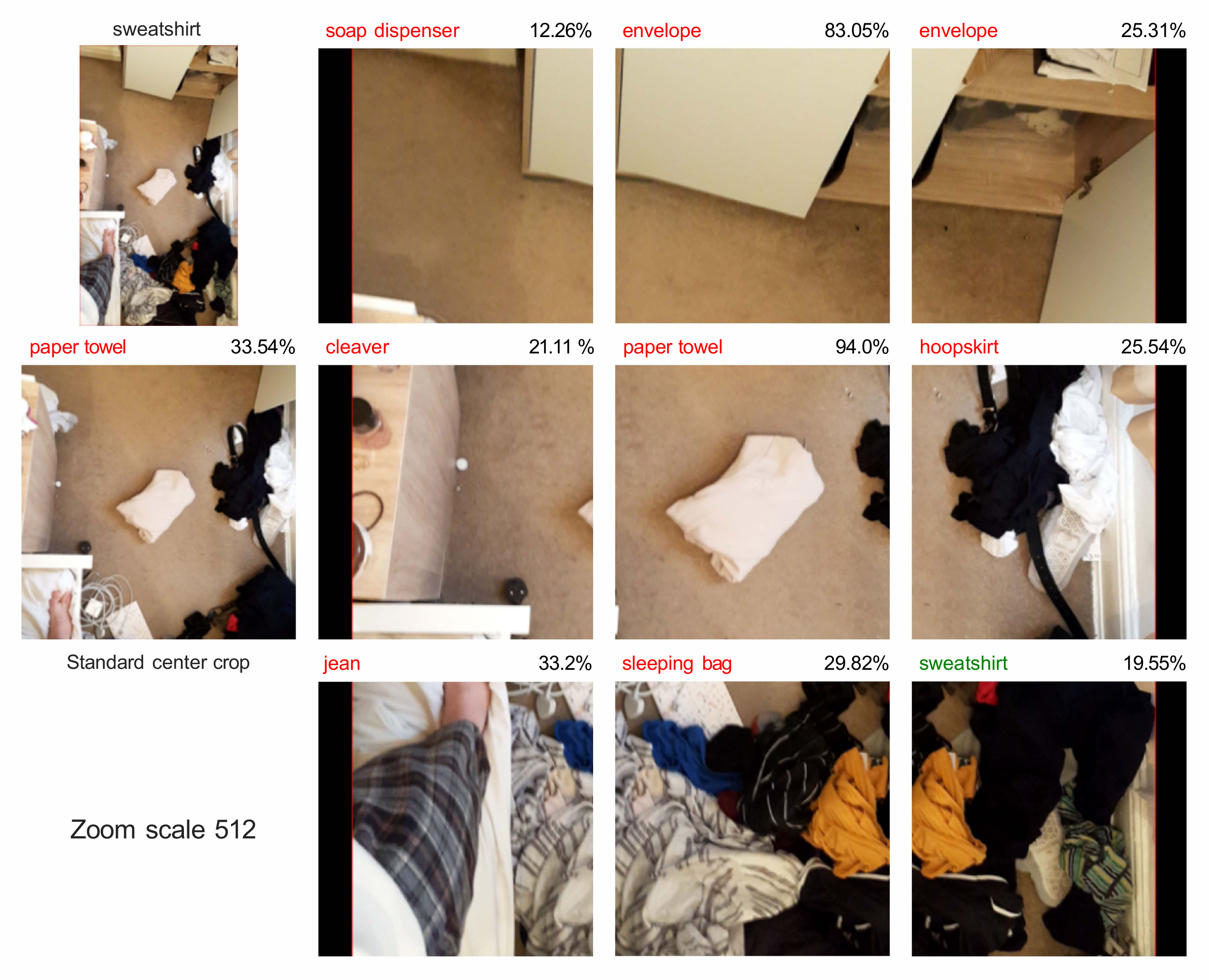}
    \end{subfigure}
    \begin{subfigure}[b]{0.48\linewidth}  
        \centering
         \includegraphics[width=\linewidth]{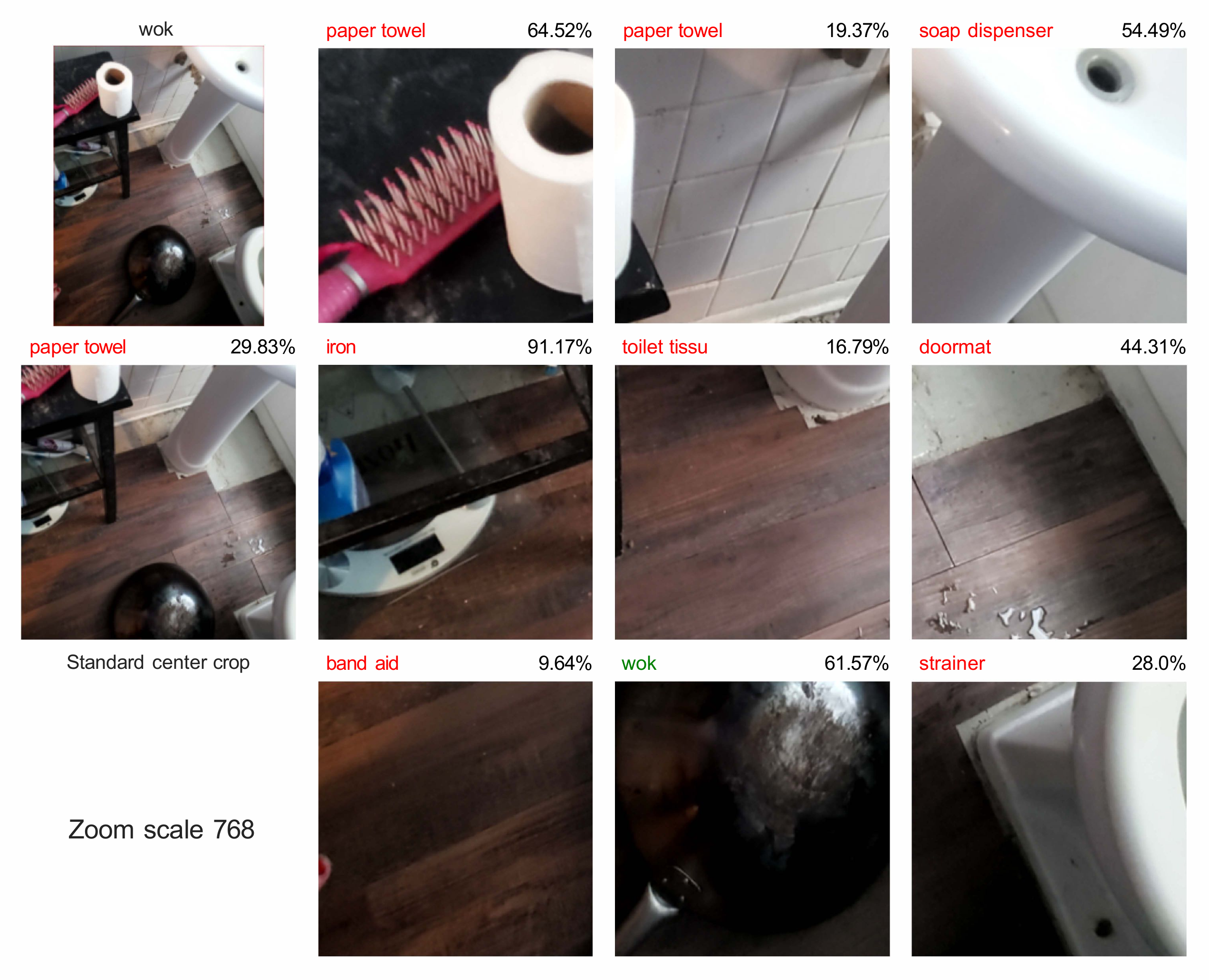}
    \end{subfigure}
    \centering
    \begin{subfigure}[b]{0.48\linewidth}   
        \centering
         \includegraphics[width=\linewidth]{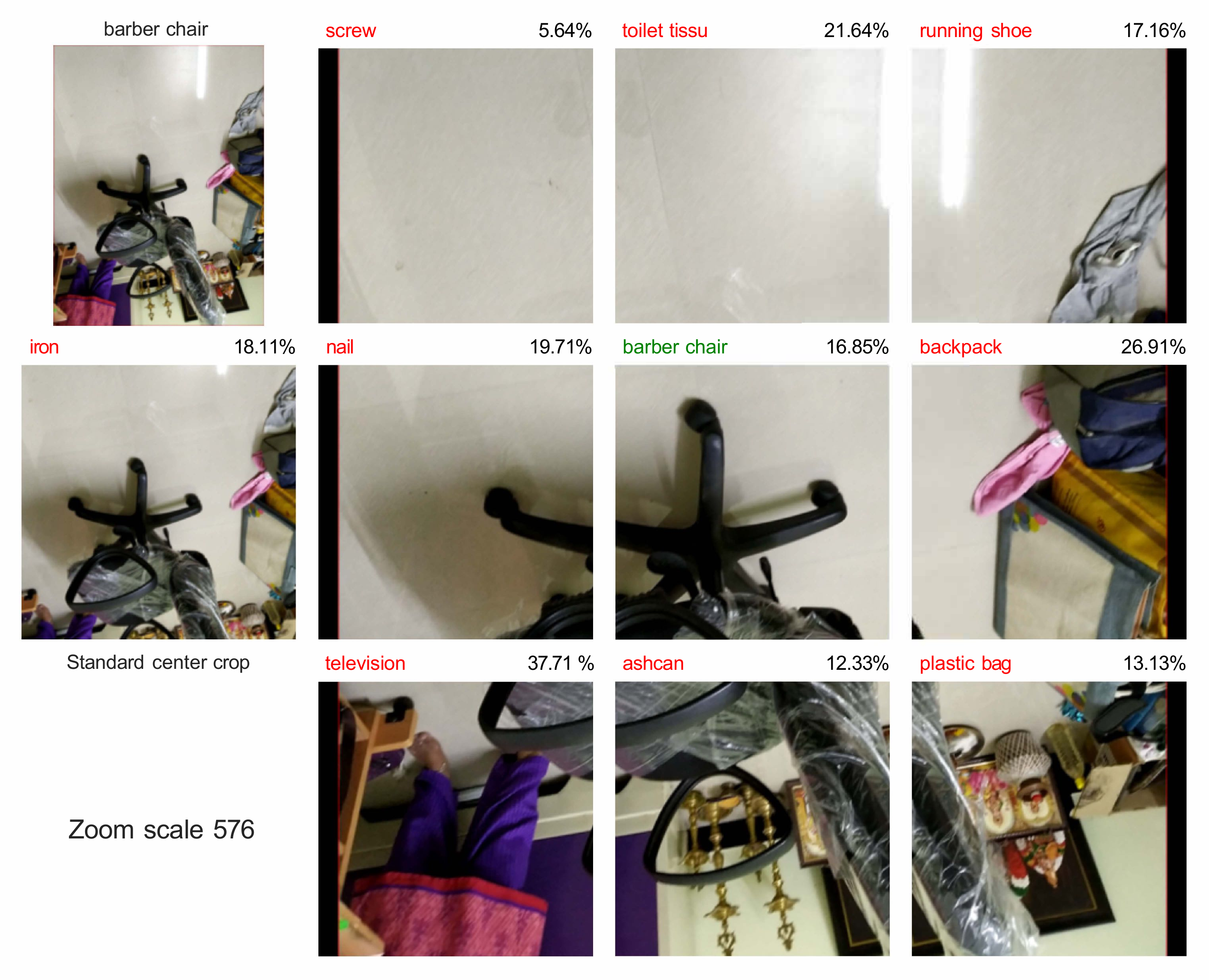}
    \end{subfigure}
    \begin{subfigure}[b]{0.48\linewidth}  
        \centering
         \includegraphics[width=\linewidth]{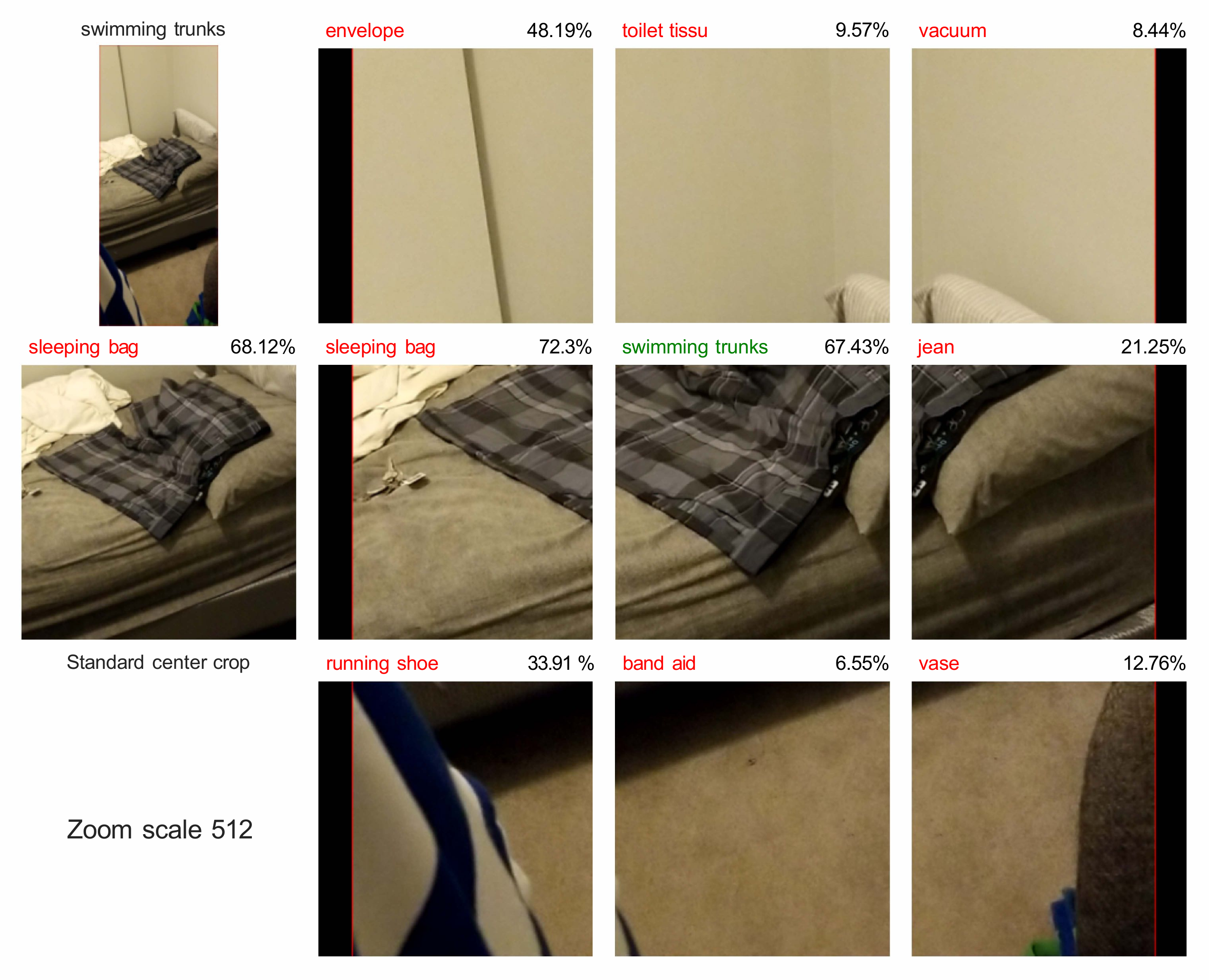}
    \end{subfigure}
    
 \caption{ObjectNet images that can only be solved using \zoomin. Predictions are from a ResNet-50 classifier.}
    \label{suppfig:hard_samples_objectnet2}
\end{figure}


\clearpage
\subsection{AugMix and \texttt{RandomResizedCrop}}
\label{suppviz:augmix-vs_rrc}

\begin{figure}[!hbt]
    \centering
    \includegraphics[width=1\textwidth]{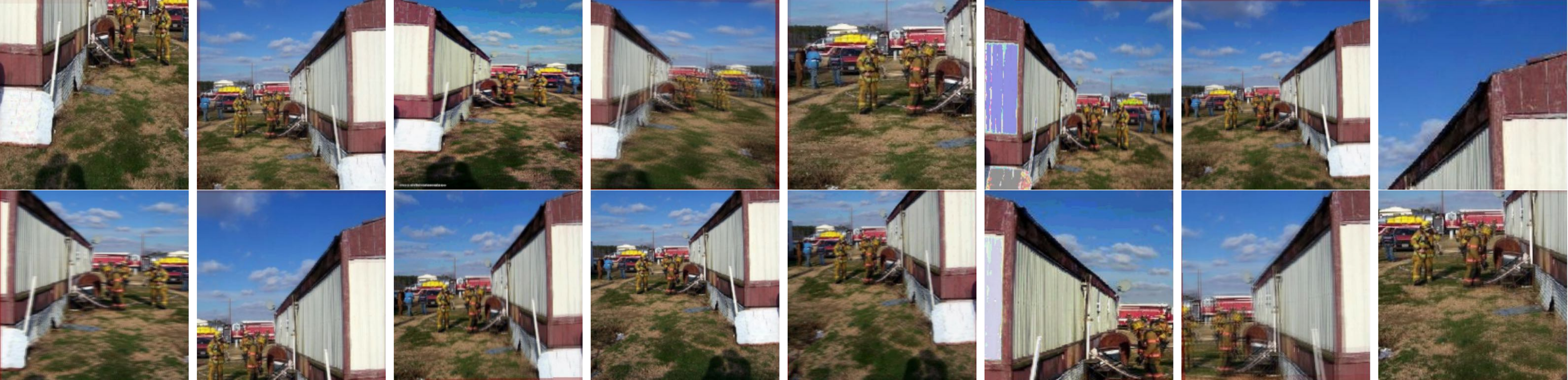}
    \caption{$K = 16$ sample outputs from AugMix~\cite{hendrycks2019augmix} (which yields the results of random sampling from 13 transformations that include both spatial and color distortions).}
    \label{fig:augmix_1}
\end{figure}

\begin{figure}[!hbt]
    \centering
    \includegraphics[width=1\textwidth]{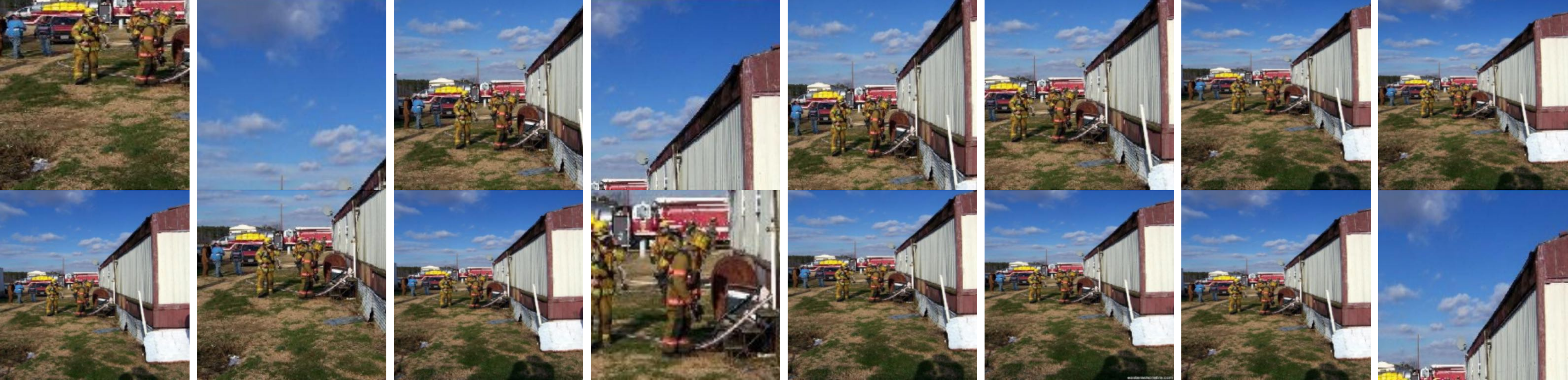}
    \caption{$K = 16$ sample outputs from \texttt{RandomResizedCrop} (\rrc), which basically randomly zooms into an arbitrary region in the input image.}
    \label{fig:rrc_1}
\end{figure}

\begin{figure}[!hbt]
    \centering
    \includegraphics[width=1\textwidth]{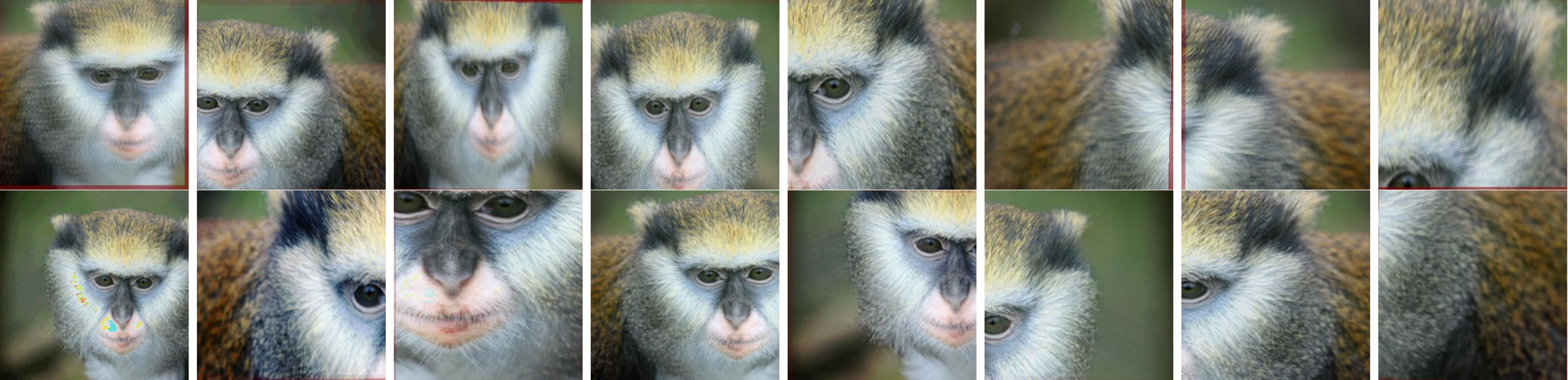}
    \caption{$K = 16$ sample outputs from AugMix~\cite{hendrycks2019augmix} (which yields the results of random sampling from 13 transformations that include both spatial and color distortions).}
    \label{fig:augmix_2}
\end{figure}

\begin{figure}[!hbt]
    \centering
    \includegraphics[width=1\textwidth]{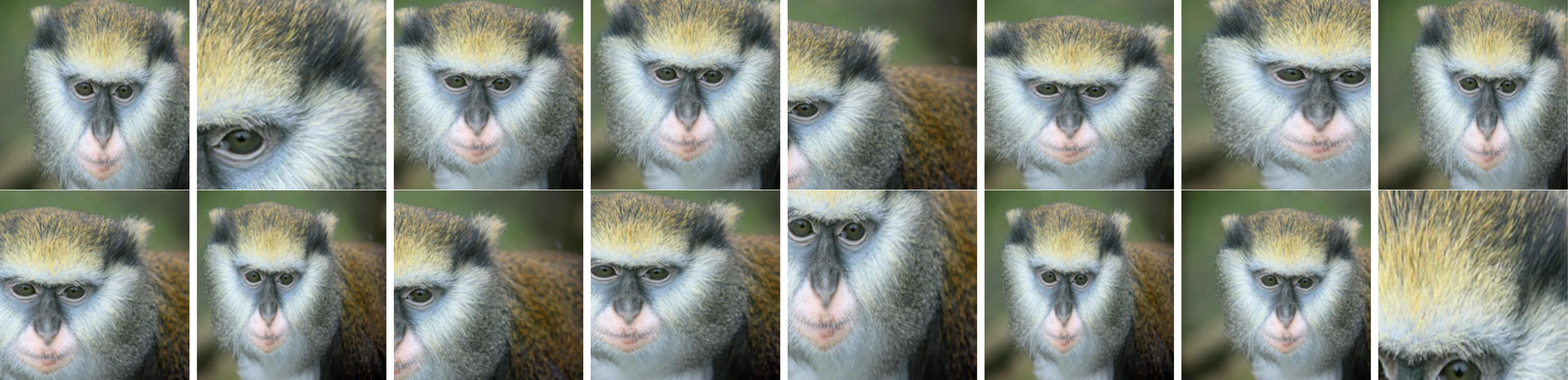}
    \caption{$K = 16$ sample outputs from \texttt{RandomResizedCrop} (\rrc), which basically randomly zooms into an arbitrary region in the input image.}
    \label{fig:rrc_2}
\end{figure}


\clearpage
\section{ImageNet-Hard}
\label{supp:imagenet_hard_section}

In this section, we provide details about the ImageNet-hard dataset.

\subsection{Distribution}
\label{supp:imagenet_hard_section_dist}

\begin{figure}[!hbt]
    \centering
    \includegraphics[width=1\textwidth]{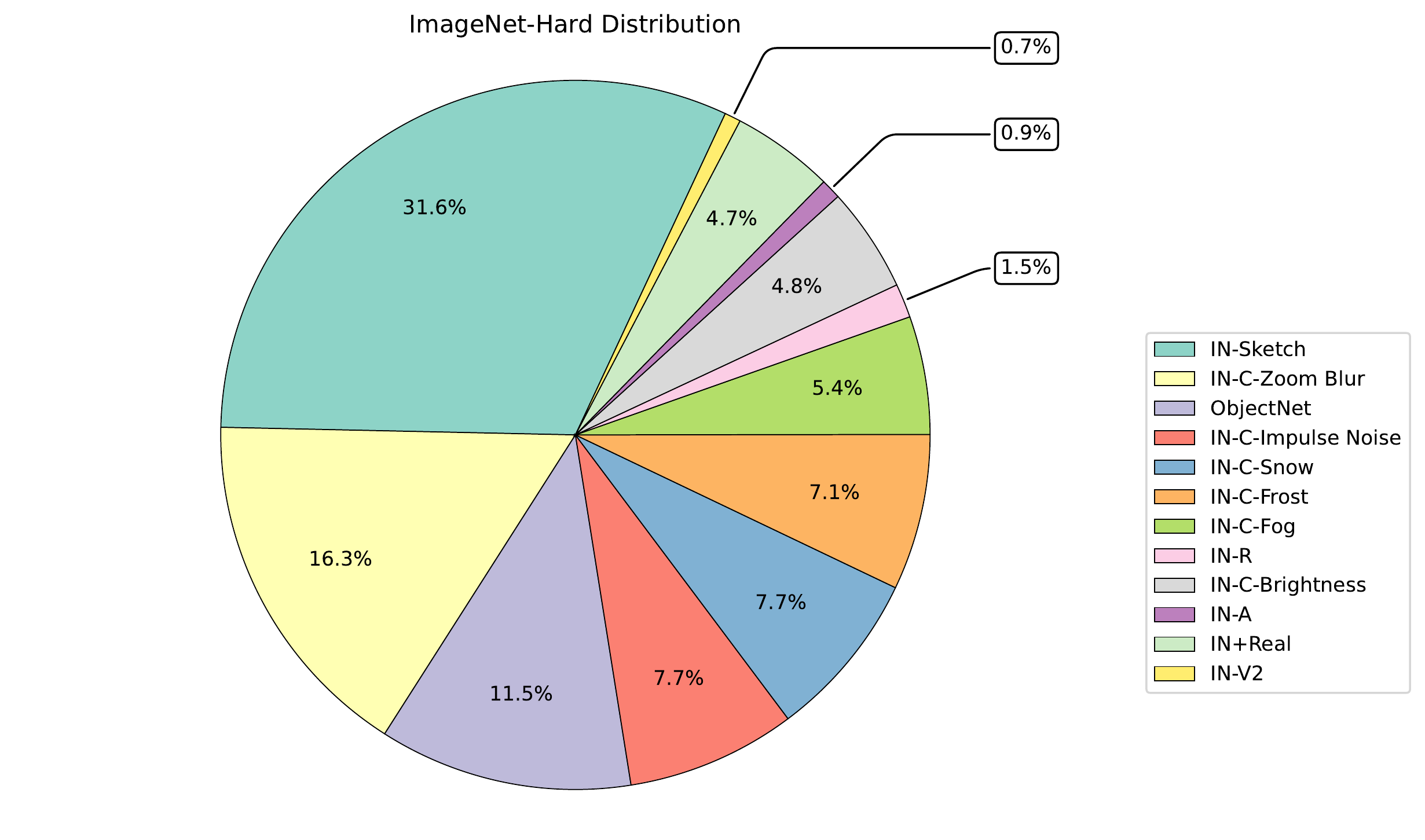}
    \caption{The distribution of the dataset within the ImageNet-Hard Dataset.}
    \label{suppfig:imagenet_hard_dist}
\end{figure}

\FloatBarrier
\subsection{Samples images}

\begin{figure}[!hbt]
    \centering
    \includegraphics[width=\textwidth]{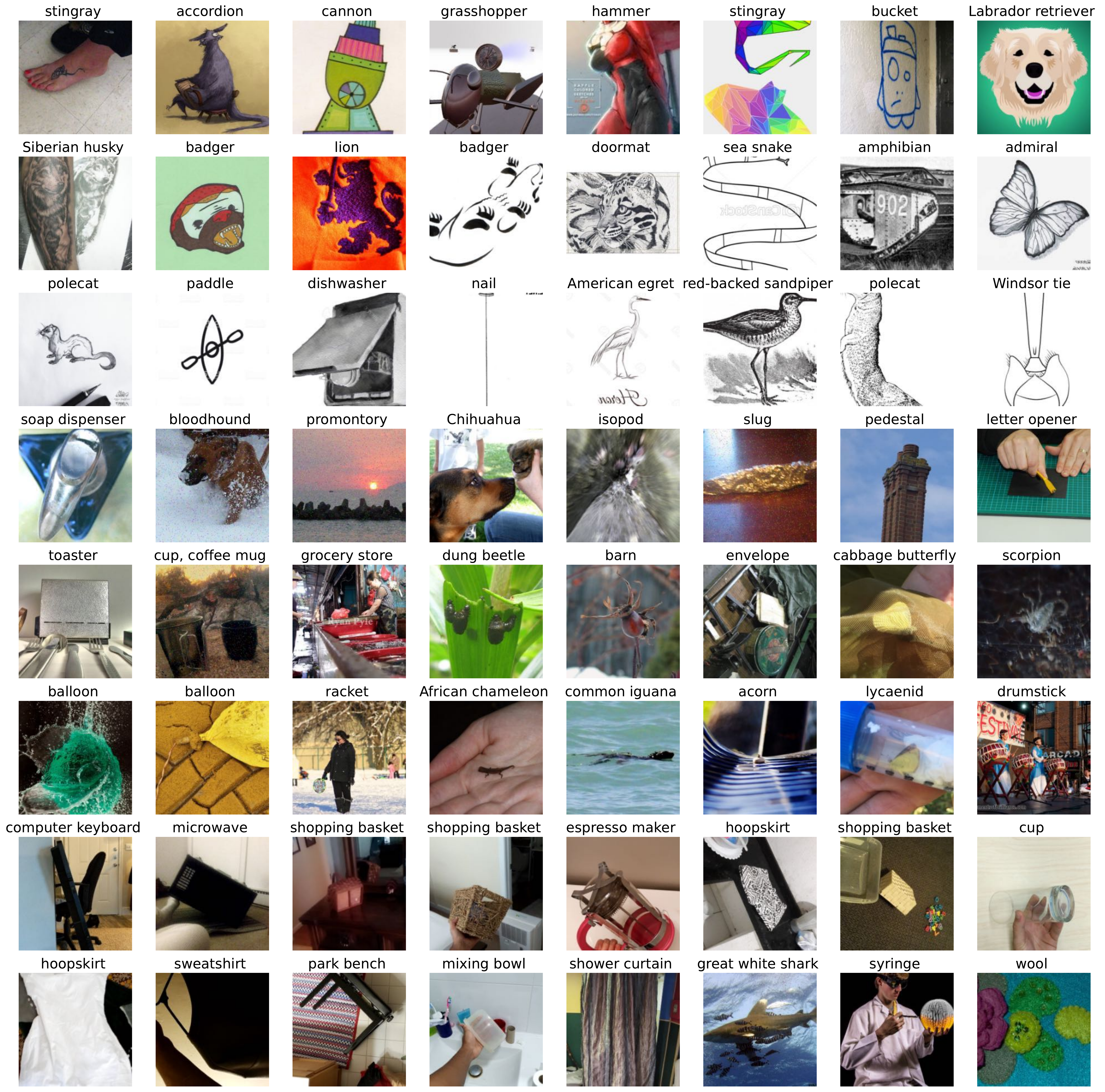}
    \caption{Sample images from ImageNet-Hard dataset with groundtruth labels.}
    \label{fig:imagenet_hard_sample_images}
\end{figure}

\FloatBarrier
\subsection{Analysis of wrong predictions}
\label{supp:analysis_of_gpt}

We used \texttt{gpt-3.5-turbo} to categorize each misprediction made by EfficientNet-L2 into two classes: plausible and implausible, based on the semantic distance between the groundtruth label and the predicted label.

\begin{figure}[htb!]
    \centering
    \includegraphics[width=1\textwidth]{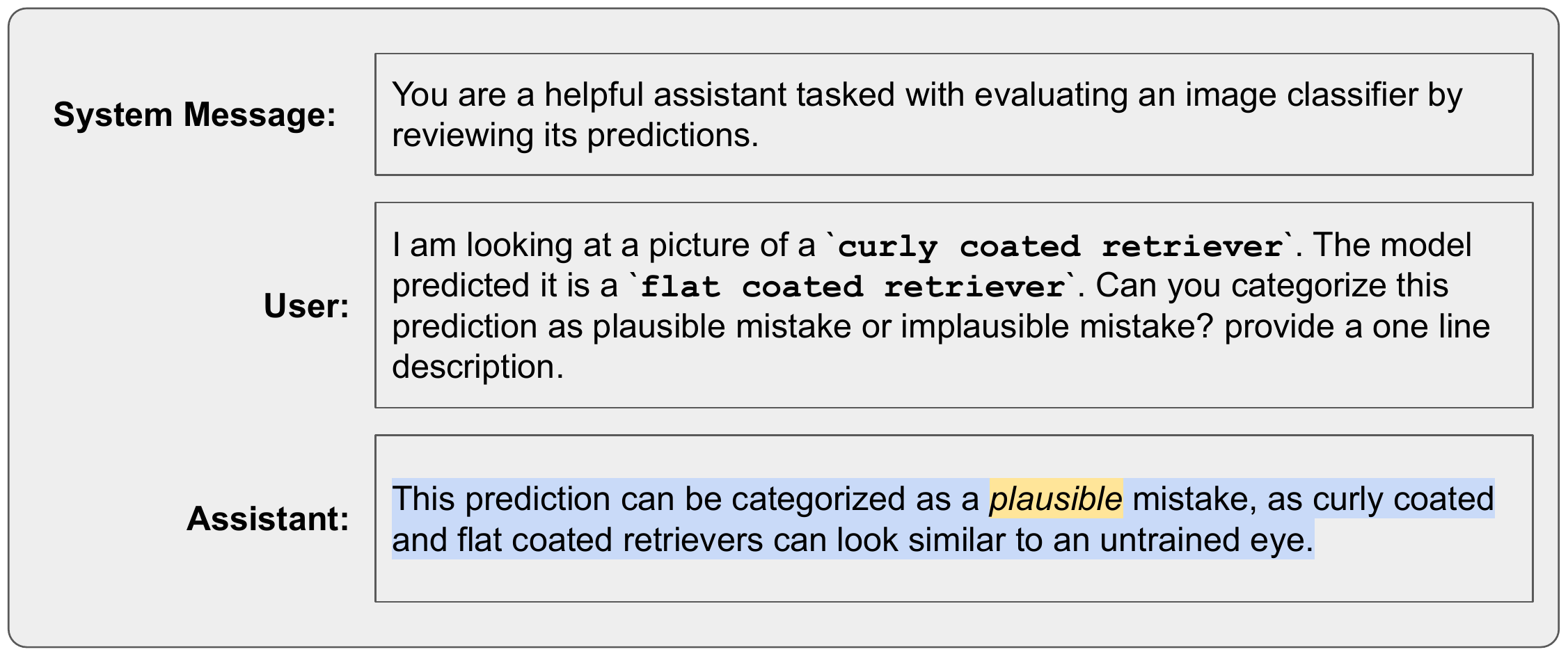}
     \caption{Sample prompt and response of \texttt{gpt-3.5-turbo} for a plausible classification. The text in the \textbf{Assistant} block is the generated response.}
    \label{fig:gpt_sample1}
\end{figure}

\begin{figure}[htb!]
    \centering
    \includegraphics[width=1\textwidth]{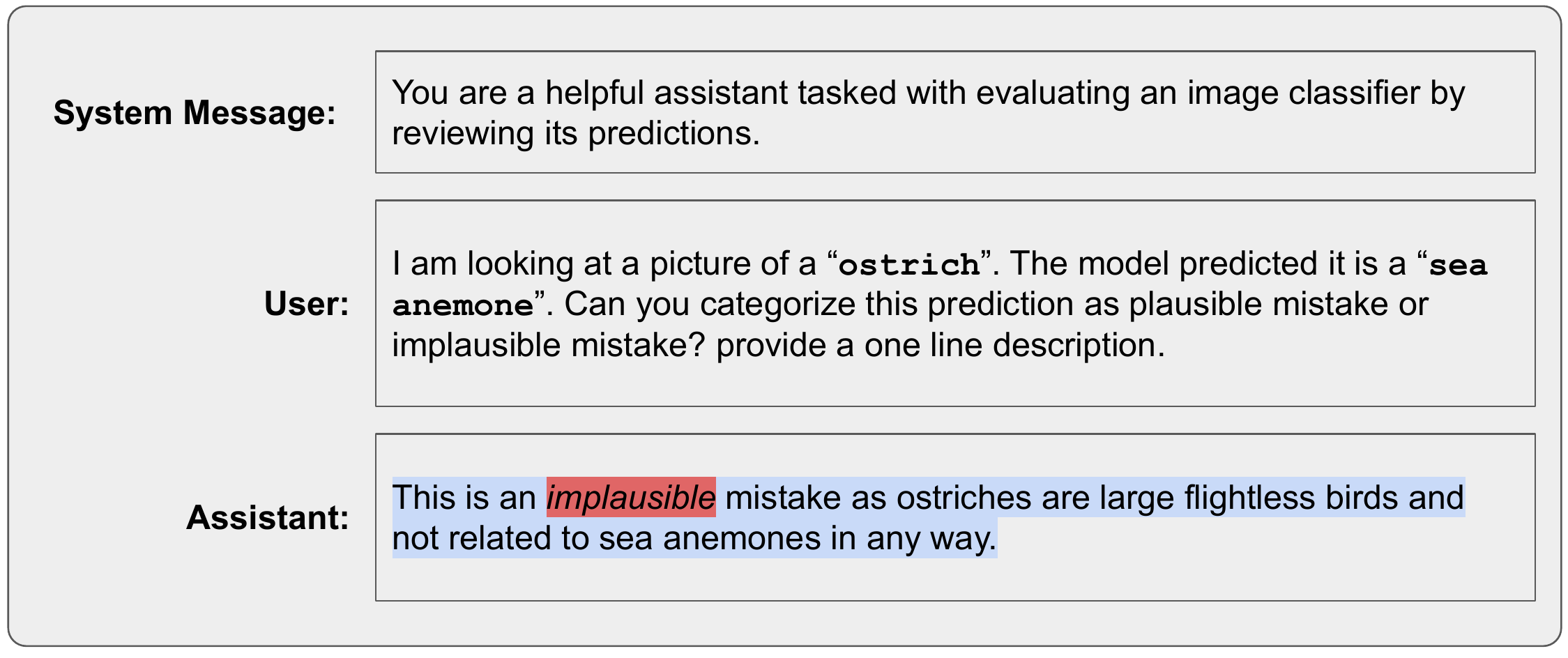}
    \caption{Sample prompt and response of \texttt{gpt-3.5-turbo} for an implausible classification. The text in the \textbf{Assistant} block is the generated response.}
    \label{fig:gpt_sample2}
\end{figure}

\clearpage
\subsection{Confusing classes}
\label{supp:analysis_confusing_classes}

In this section, we present a selection of examples highlighting the errors made by our highest-performing model, EfficientNet-L2.

\begin{figure}[htb!]
    \centering
    \includegraphics[width=\textwidth]{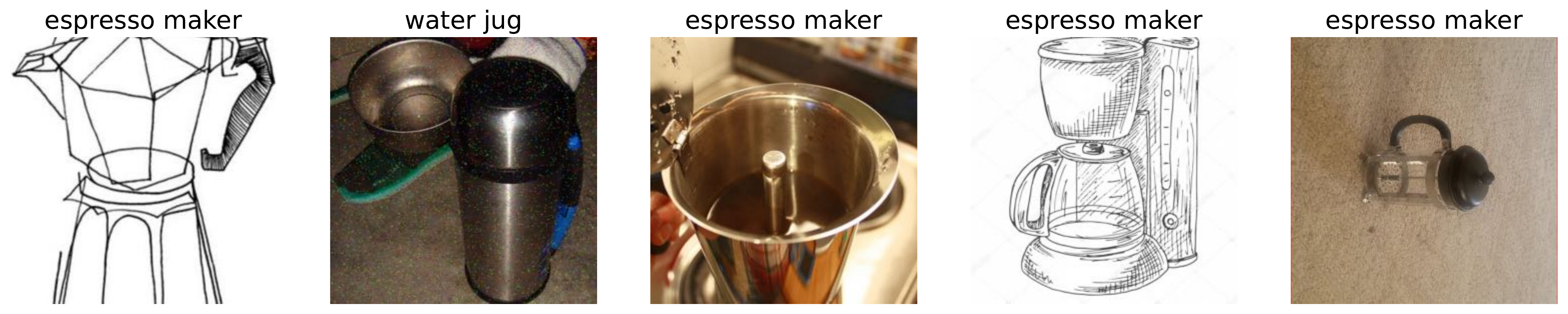}
    \caption{Images \wrong{misclassified} into \class{coffemaker} by EfficientNet-L2}
    \label{fig:misclassified_as_coffeemaker}
\end{figure}

\begin{figure}[htb!]
    \centering
    \includegraphics[width=\textwidth]{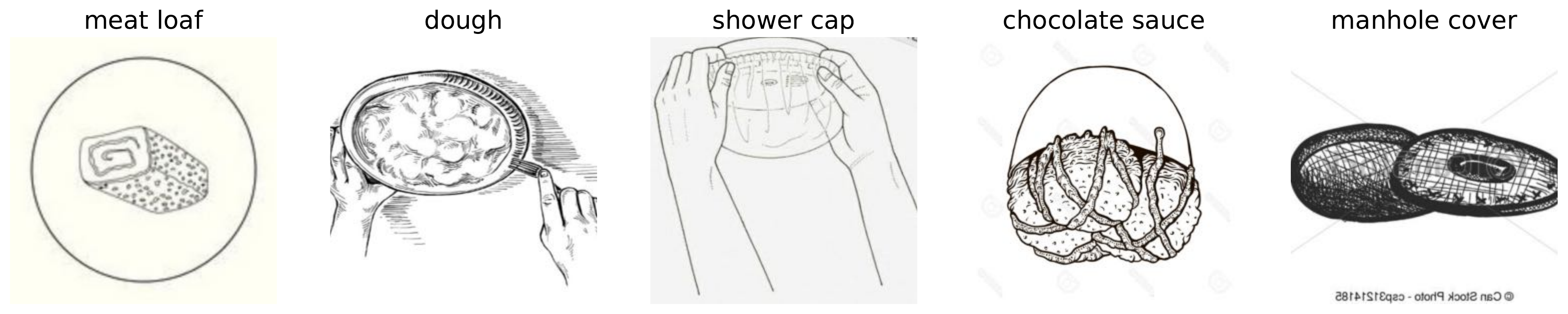}
    \caption{Images \wrong{misclassified} into \class{strainer} by EfficientNet-L2}
    \label{fig:misclassified_as_strainer}
\end{figure}

\begin{figure}[htb!]
    \centering
    \includegraphics[width=\textwidth]{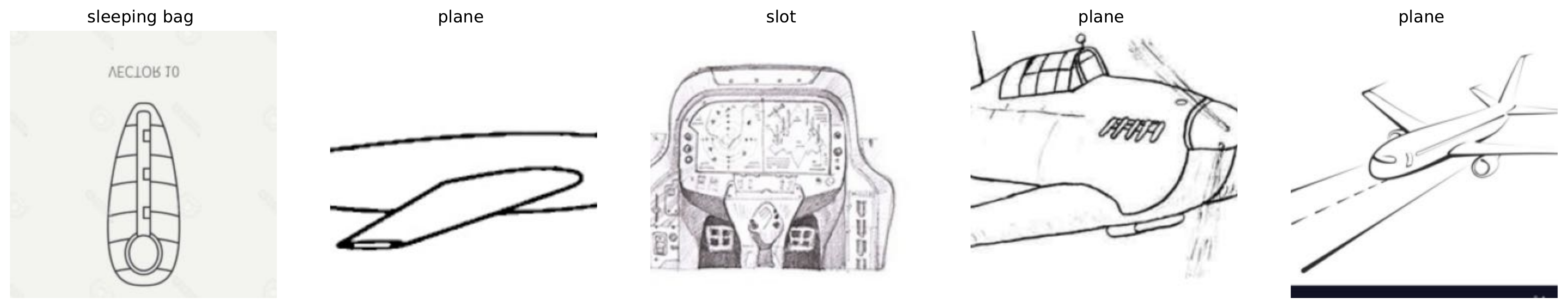}
    \caption{Images \wrong{misclassified} into \class{space shuttle} by EfficientNet-L2}
    \label{fig:misclassified_as_quail}
\end{figure}

\begin{figure}[htb!]
    \centering
    \includegraphics[width=\textwidth]{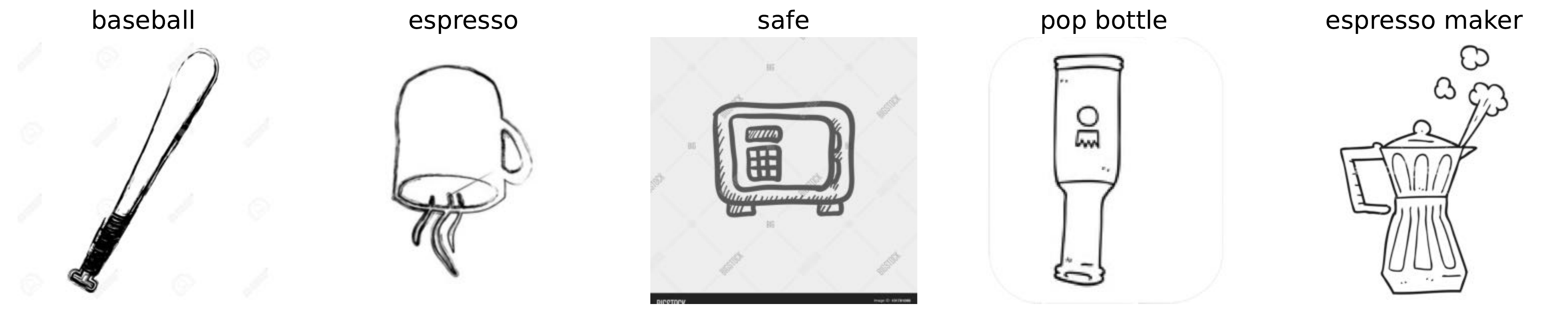}
    \caption{Images \wrong{misclassified} into \class{safety pin} by EfficientNet-L2}
    \label{fig:misclassified_as_safetypin}
\end{figure}

\clearpage
\subsection{Common and rare misclassification}

This section shows some sample misclassification for EfficientNet-L2 and OpenCLIP's ViT-bigG-14 classifiers.

\begin{figure}[htb!]
    \centering
    \includegraphics[width=\textwidth]{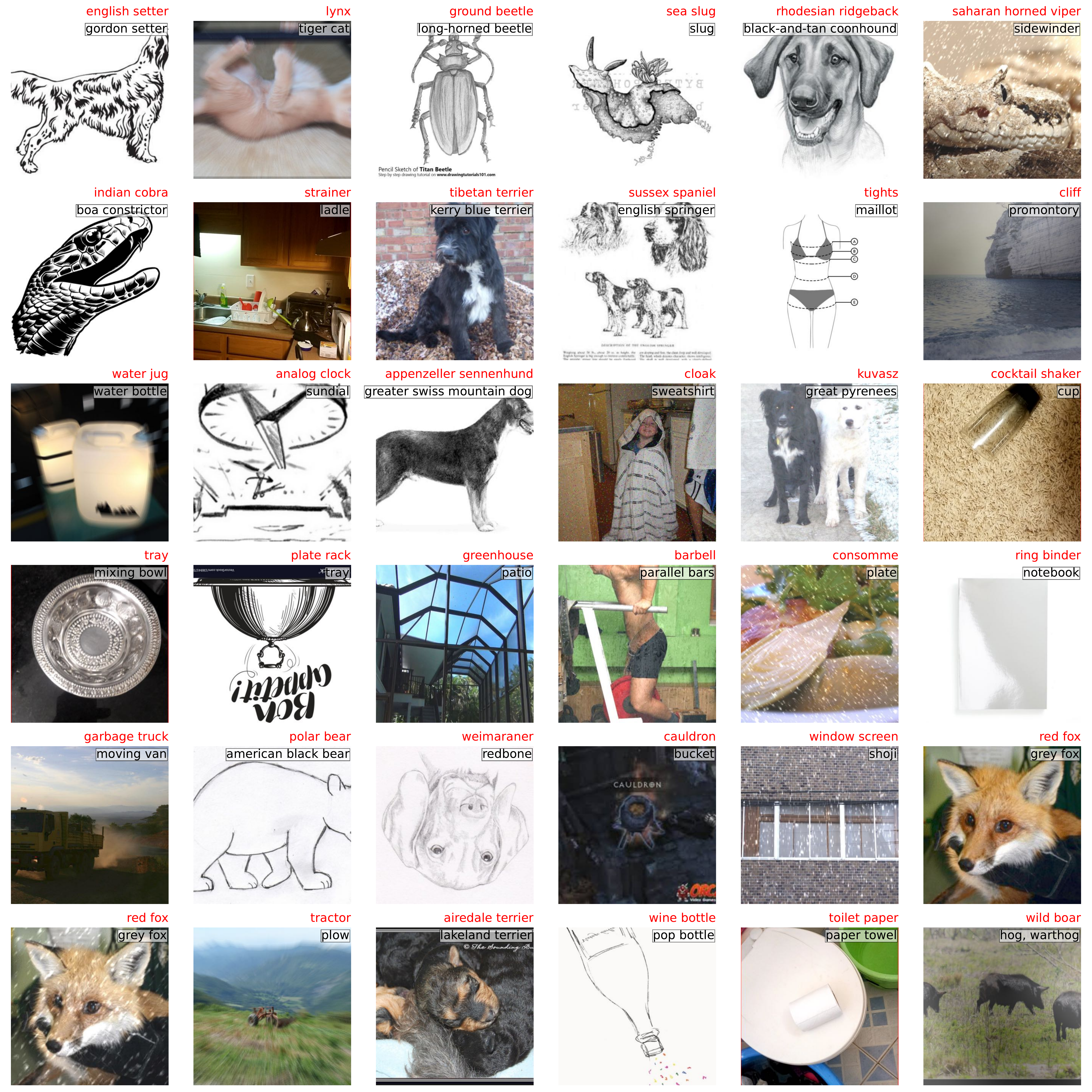}
    \caption{Examples of misclassifications by EfficientNet-L2 under the \textit{Common} category.}
    \label{suppfig:commonmisclassification}
\end{figure}

\begin{figure}[htb!]
    \centering
    \includegraphics[width=\textwidth]{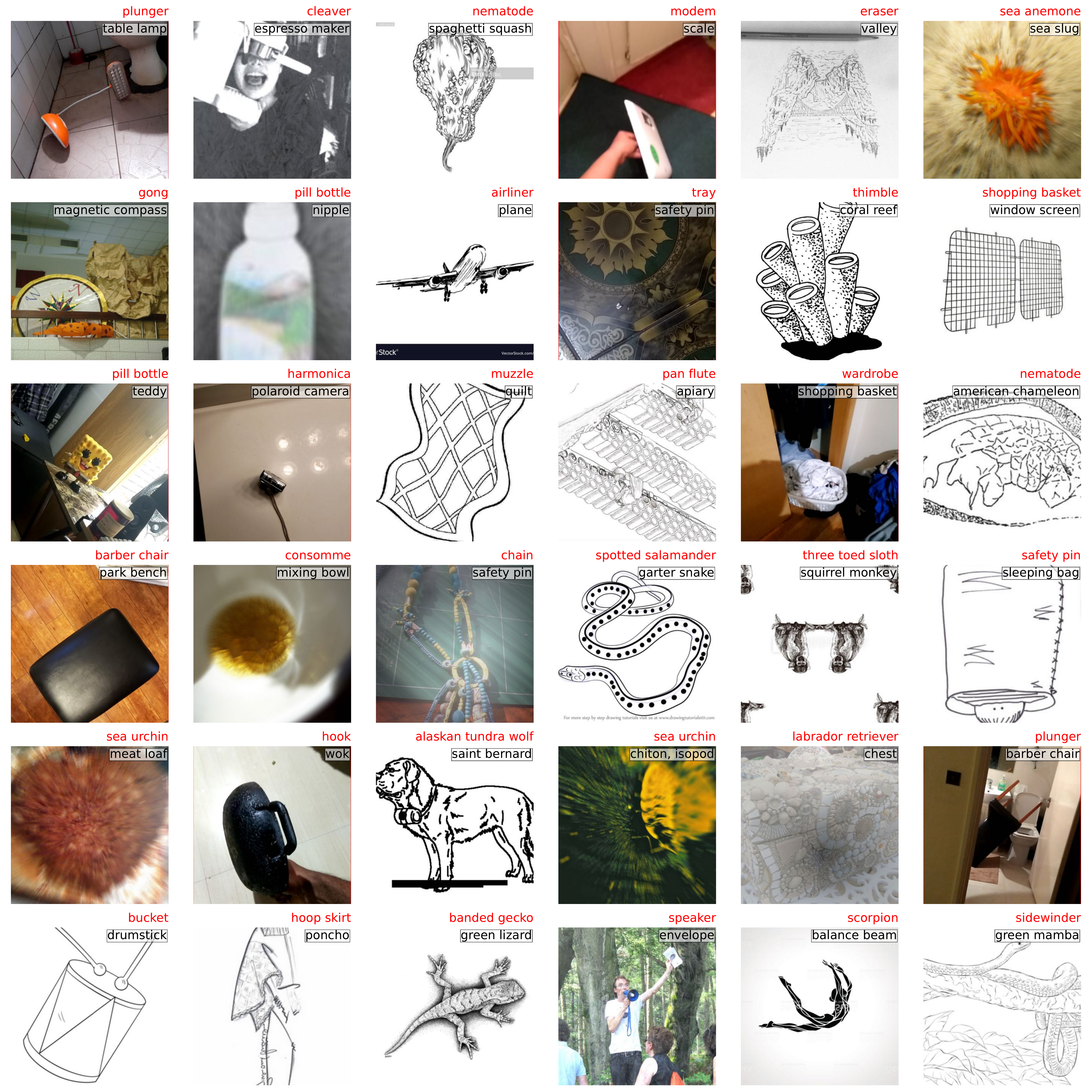}
    \caption{Examples of misclassifications by EfficientNet-L2 under the \textit{Rare} category.}
    \label{suppfig:raremisclassifications}
\end{figure}

\begin{figure}[htb!]
    \centering
    \includegraphics[width=\textwidth]{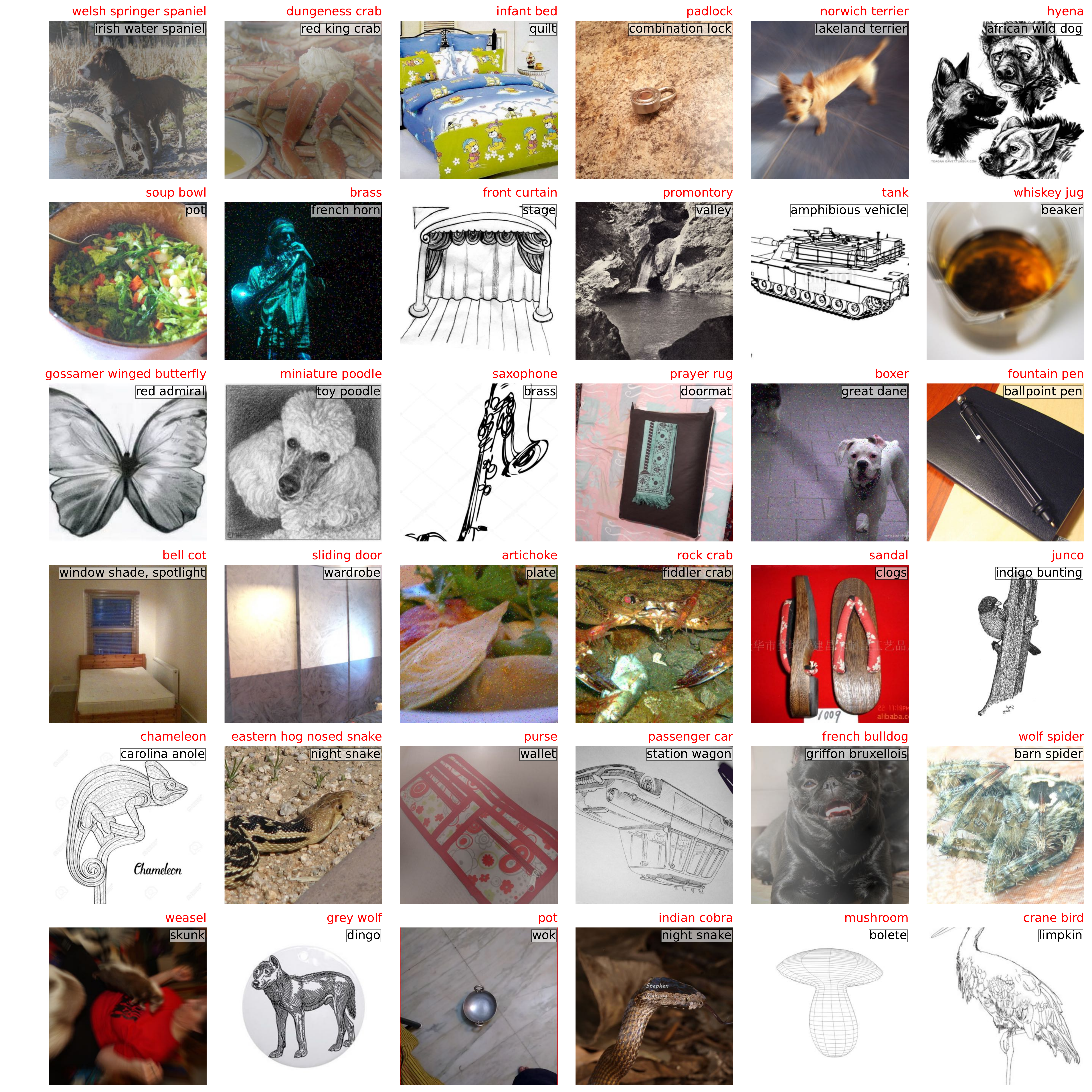}
    \caption{Examples of misclassifications by OpenCLIP under the \textit{Common} category.}
    \label{suppfig:commonmisclassification2}
\end{figure}

\begin{figure}[htb!]
    \centering
    \includegraphics[width=\textwidth]{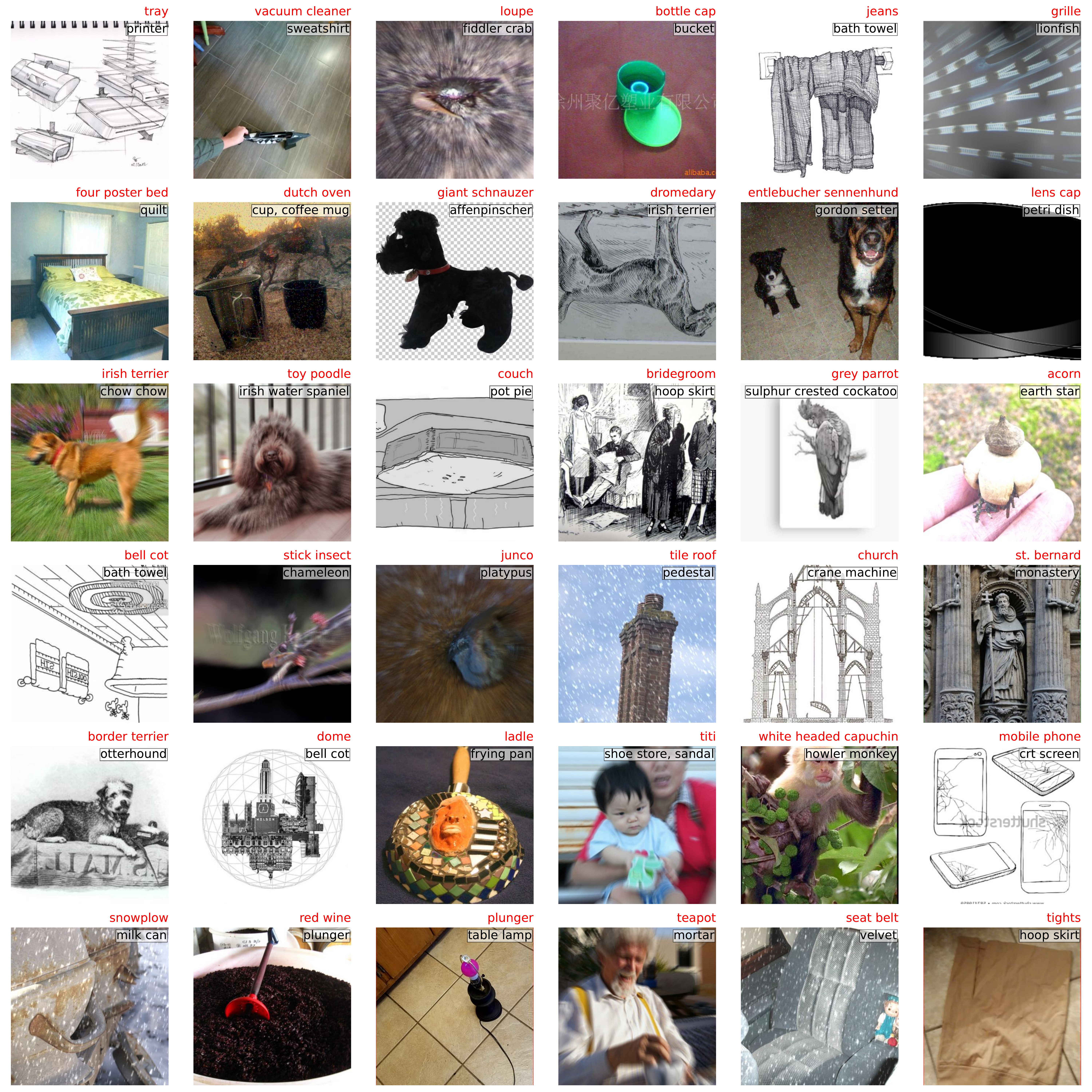}
    \caption{Examples of misclassifications by OpenCLIP under the \textit{Rare} category.}
    \label{suppfig:raremisclassifications2}
\end{figure}

\clearpage
\subsection{Evaluating OpenCLIP models' performance on ImageNet-Hard}
\label{supp:imagenet_hard_openclip}

All the models in this section are downloaded and used from the \href{https://github.com/mlfoundations/open_clip}{\textit{OpenCLIP}} library version \texttt{2.20.0}.

\begin{longtable}[c]{@{}llr@{}}
\caption{Zero-shot performance of OpenCLIP on ImageNet-Hard (\%)}
\label{tab:openclip_all}\\
\toprule
\multicolumn{1}{c}{Model} & \multicolumn{1}{c}{Pre-trained Dataset}       & \multicolumn{1}{c}{Top-1 Accuracy} \\* \midrule
\endfirsthead
\endhead
\bottomrule
\endfoot
\endlastfoot
RN50                       & \texttt{yfcc15m}                                & 0.80  \\
RN50                       & \texttt{cc12m}                                  & 1.18  \\
RN50-quickgelu             & \texttt{yfcc15m}                                & 0.75  \\
RN50-quickgelu             & \texttt{cc12m}                                  & 1.08  \\
RN101                      & \texttt{yfcc15m}                                & 0.65  \\
RN101-quickgelu            & \texttt{yfcc15m}                                & 0.62  \\
ViT-B/32                   & \texttt{laion400m\_e31}                         & 5.34  \\
ViT-B/32                   & \texttt{laion400m\_e32}                         & 5.41  \\
ViT-B/32                   & \texttt{laion2b\_e16}                           & 5.66  \\
ViT-B/32                   & \texttt{laion2b\_s34b\_b79k}                    & 6.13  \\
ViT-B/32                   & \texttt{datacomp\_m\_s128m\_b4k}                & 2.79  \\
ViT-B/32                   & \texttt{commonpool\_m\_clip\_s128m\_b4k}        & 2.50  \\
ViT-B/32                   & \texttt{commonpool\_m\_laion\_s128m\_b4k}       & 2.41  \\
ViT-B/32                   & \texttt{commonpool\_m\_image\_s128m\_b4k}       & 2.72  \\
ViT-B/32                   & \texttt{commonpool\_m\_text\_s128m\_b4k}        & 2.46  \\
ViT-B/32                   & \texttt{commonpool\_m\_basic\_s128m\_b4k}       & 2.23  \\
ViT-B/32                   & \texttt{commonpool\_m\_s128m\_b4k}              & 1.73  \\
ViT-B/32                   & \texttt{datacomp\_s\_s13m\_b4k}                 & 0.61  \\
ViT-B/32                   & \texttt{commonpool\_s\_clip\_s13m\_b4k}         & 0.84  \\
ViT-B/32                   & \texttt{commonpool\_s\_laion\_s13m\_b4k}        & 0.66  \\
ViT-B/32                   & \texttt{commonpool\_s\_image\_s13m\_b4k}        & 0.61  \\
ViT-B/32                   & \texttt{commonpool\_s\_text\_s13m\_b4k}         & 0.77  \\
ViT-B/32                   & \texttt{commonpool\_s\_basic\_s13m\_b4k}        & 0.75  \\
ViT-B/32                   & \texttt{commonpool\_s\_s13m\_b4k}               & 0.43  \\
ViT-B/32-quickgelu         & \texttt{laion400m\_e31}                         & 5.34  \\
ViT-B/32-quickgelu         & \texttt{laion400m\_e32}                         & 5.28  \\
ViT-B/16                   & \texttt{laion400m\_e31}                         & 6.31  \\
ViT-B/16                   & \texttt{laion400m\_e32}                         & 6.46  \\
ViT-B/16                   & \texttt{laion2b\_s34b\_b88k}                    & 7.18  \\
ViT-B/16                   & \texttt{datacomp\_l\_s1b\_b8k}                  & 5.98  \\
ViT-B/16                   & \texttt{commonpool\_l\_clip\_s1b\_b8k}          & 4.92  \\
ViT-B/16                   & \texttt{commonpool\_l\_laion\_s1b\_b8k}         & 4.44  \\
ViT-B/16                   & \texttt{commonpool\_l\_image\_s1b\_b8k}         & 4.75  \\
ViT-B/16                   & \texttt{commonpool\_l\_text\_s1b\_b8k}          & 5.63  \\
ViT-B/16                   & \texttt{commonpool\_l\_basic\_s1b\_b8k}         & 4.44  \\
ViT-B/16                   & \texttt{commonpool\_l\_s1b\_b8k}                & 3.83  \\
ViT-B/16-plus-240          & \texttt{laion400m\_e31}                         & 6.65  \\
ViT-B/16-plus-240          & \texttt{laion400m\_e32}                         & 6.69  \\
ViT-L/14                   & \texttt{laion400m\_e31}                         & 8.83  \\
ViT-L/14                   & \texttt{laion400m\_e32}                         & 8.72  \\
ViT-L/14                   & \texttt{laion2b\_s32b\_b82k}                    & 10.13 \\
ViT-L/14                   & \texttt{datacomp\_xl\_s13b\_b90k}               & 15.60 \\
ViT-L/14                   & \texttt{commonpool\_xl\_clip\_s13b\_b90k}       & 11.58 \\
ViT-L/14                   & \texttt{commonpool\_xl\_laion\_s13b\_b90k}      & 11.42 \\
ViT-L/14                   & \texttt{commonpool\_xl\_s13b\_b90k}             & 12.44 \\
ViT-H/14                   & \texttt{laion2b\_s32b\_b79k}                    & 13.01 \\
ViT-g/14                   & \texttt{laion2b\_s12b\_b42k}                    & 11.47 \\
ViT-g/14                   & \texttt{laion2b\_s34b\_b88k}                    & 14.03 \\
ViT-bigG-14                & \texttt{laion2b\_s39b\_b160k}                   & 15.93 \\
roberta-ViT-B/32           & \texttt{laion2b\_s12b\_b32k}                    & 5.21  \\
xlm-roberta-base-ViT-B/32  & \texttt{laion5b\_s13b\_b90k}                    & 5.72  \\
xlm-roberta-large-ViT-H/14 & \texttt{frozen\_laion5b\_s13b\_b90k}            & 12.95 \\
convnext\_base             & \texttt{laion400m\_s13b\_b51k}                  & 4.74  \\
convnext\_base\_w          & \texttt{laion2b\_s13b\_b82k}                    & 6.09  \\
convnext\_base\_w          & \texttt{laion2b\_s13b\_b82k\_augreg}            & 7.25  \\
convnext\_base\_w          & \texttt{laion\_aesthetic\_s13b\_b82k}           & 5.57  \\
convnext\_base\_w\_320     & \texttt{laion\_aesthetic\_s13b\_b82k}           & 5.50  \\
convnext\_base\_w\_320     & \texttt{laion\_aesthetic\_s13b\_b82k\_augreg}   & 7.14  \\
convnext\_large\_d         & \texttt{laion2b\_s26b\_b102k\_augreg}           & 10.39 \\
convnext\_large\_d\_320    & \texttt{laion2b\_s29b\_b131k\_ft}               & 10.69 \\
convnext\_large\_d\_320    & \texttt{laion2b\_s29b\_b131k\_ft\_soup}         & 11.20 \\
convnext\_xxlarge          & \texttt{laion2b\_s34b\_b82k\_augreg}            & 14.27 \\
convnext\_xxlarge          & \texttt{laion2b\_s34b\_b82k\_augreg\_rewind}    & 14.23 \\
convnext\_xxlarge          & \texttt{laion2b\_s34b\_b82k\_augreg\_soup}      & 14.68 \\
coca\_ViT-B/32             & \texttt{laion2b\_s13b\_b90k}                    & 5.83  \\
coca\_ViT-B/32             & \texttt{mscoco\_finetuned\_laion2b\_s13b\_b90k} & 0.20  \\
coca\_ViT-L/14             & \texttt{laion2b\_s13b\_b90k}                    & 10.79 \\
coca\_ViT-L/14             & \texttt{mscoco\_finetuned\_laion2b\_s13b\_b90k} & 9.28  \\* 
\bottomrule
\end{longtable}

\begin{table}[htb!]
\centering
\caption{Zero-shot performance of CommonPool and DataComp models on ImageNet-Hard (\%)}
\label{tab:datacomp_all}
\begin{tabular}{@{}cllr@{}}
\toprule
Scale                   & \multicolumn{1}{c}{Model} & \multicolumn{1}{c}{Pretrained} & \multicolumn{1}{c}{Top-1 Accuracy} \\ \midrule
\multirow{4}{*}{\rotatebox[origin=c]{90}{xlarge}} & ViT-L/14                  & \texttt{datacomp\_xl\_s13b\_b90k}       & 15.60                              \\
                        & ViT-L/14 & \texttt{commonpool\_xl\_clip\_s13b\_b90k}  & 11.58 \\
                        & ViT-L/14 & \texttt{commonpool\_xl\_laion\_s13b\_b90k} & 11.42 \\
                        & ViT-L/14 & \texttt{commonpool\_xl\_s13b\_b90k}        & 12.44 \\ \midrule
\multirow{7}{*}{\rotatebox[origin=c]{90}{large}}  & ViT-B/16 & \texttt{datacomp\_l\_s1b\_b8k}             & 5.98  \\
                        & ViT-B/16 & \texttt{commonpool\_l\_clip\_s1b\_b8k}     & 4.92  \\
                        & ViT-B/16 & \texttt{commonpool\_l\_laion\_s1b\_b8k}    & 4.44  \\
                        & ViT-B/16 & \texttt{commonpool\_l\_image\_s1b\_b8k}    & 4.75  \\
                        & ViT-B/16 & \texttt{commonpool\_l\_text\_s1b\_b8k}     & 5.63  \\
                        & ViT-B/16 & \texttt{commonpool\_l\_basic\_s1b\_b8k}    & 4.44  \\
                        & ViT-B/16 & \texttt{commonpool\_l\_s1b\_b8k}           & 3.83  \\ \midrule
\multirow{7}{*}{\rotatebox[origin=c]{90}{medium}} & ViT-B/32 & \texttt{datacomp\_m\_s128m\_b4k}           & 2.79  \\
                        & ViT-B/32 & \texttt{commonpool\_m\_clip\_s128m\_b4k}   & 2.50  \\
                        & ViT-B/32 & \texttt{commonpool\_m\_laion\_s128m\_b4k}  & 2.41  \\
                        & ViT-B/32 & \texttt{commonpool\_m\_image\_s128m\_b4k}  & 2.72  \\
                        & ViT-B/32 & \texttt{commonpool\_m\_text\_s128m\_b4k}   & 2.46  \\
                        & ViT-B/32 & \texttt{commonpool\_m\_basic\_s128m\_b4k}  & 2.23  \\
                        & ViT-B/32 & \texttt{commonpool\_m\_s128m\_b4k}         & 1.73  \\ \midrule
\multirow{7}{*}{\rotatebox[origin=c]{90}{small}}  & ViT-B/32 & \texttt{datacomp\_s\_s13m\_b4k}            & 0.61  \\
                        & ViT-B/32 & \texttt{commonpool\_s\_clip\_s13m\_b4k}    & 0.84  \\
                        & ViT-B/32 & \texttt{commonpool\_s\_laion\_s13m\_b4k}   & 0.66  \\
                        & ViT-B/32 & \texttt{commonpool\_s\_image\_s13m\_b4k}   & 0.61  \\
                        & ViT-B/32 & \texttt{commonpool\_s\_text\_s13m\_b4k}    & 0.77  \\
                        & ViT-B/32 & \texttt{commonpool\_s\_basic\_s13m\_b4k}   & 0.75  \\
                        & ViT-B/32 & \texttt{commonpool\_s\_s13m\_b4k}          & 0.43  \\ \bottomrule
\end{tabular}
\end{table}

\clearpage
\subsection{Evaluating classifiers on ImageNet-Hard-4K}

\begin{table}[ht]
\centering
\caption{Top-1 accuracy (\%) on ImageNet-Hard-4K. 
Most models obtain a \textcolor{red}{lower} accuracy compared to their corresponding accuracy on ImageNet-Hard.
}
\label{tab:imagenet_hard4k_perforamnce}
\resizebox{1\textwidth}{!}{
    \begin{tabular}{@{}lrlrlr@{}}
    \toprule
    \multicolumn{1}{c}{\textbf{Classifier}} & \multicolumn{1}{c}{\textbf{Accuracy}} & \multicolumn{1}{c}{\textbf{Classifier}} & \multicolumn{1}{c}{\textbf{Accuracy}} & \multicolumn{1}{c}{\textbf{Classifier}} & \multicolumn{1}{c}{\textbf{Accuracy}} \\ \midrule
    \textcolor{specialcolor}{AlexNet} & 7.08    \decrease{$0.16$}     & \textcolor{specialcolor}{ViT-B/32} & 18.12   \decrease{$0.40$}              & \clip-ViT-L/14@224px & 1.81 \decrease{$0.05$}  \\
    \textcolor{specialcolor}{VGG-16} & 11.32    \decrease{$0.68$}     & \textcolor{specialcolor}{EfficientNet-B0@224px} & 12.94 \decrease{$3.63$}        & \clip-ViT-L/14@336px & 1.88 \decrease{$0.14$} \\
    \textcolor{specialcolor}{ResNet-18} & 10.42 \decrease{$0.44$}     & \textcolor{specialcolor}{EfficientNet-B7@600px} & 18.67  \decrease{$4.53$}       & OpenCLIP-ViT-bigG-14 & 14.33 \decrease{$1.60$} \\
    \textcolor{specialcolor}{ResNet-50} & 13.93  \decrease{$0.81$}    & EfficientNet-L2@800px & 28.42   \decrease{$10.58$}                   & OpenCLIP-ViT-L-14  & 13.04  \decrease{$2.56$}   \\ \bottomrule
    \end{tabular}%
}
\end{table}

\FloatBarrier
\subsection{Obviously ill-posed samples from ImageNet-Sketch}

\begin{figure}[htb!]
    \centering
    \includegraphics[width=0.85\textwidth]{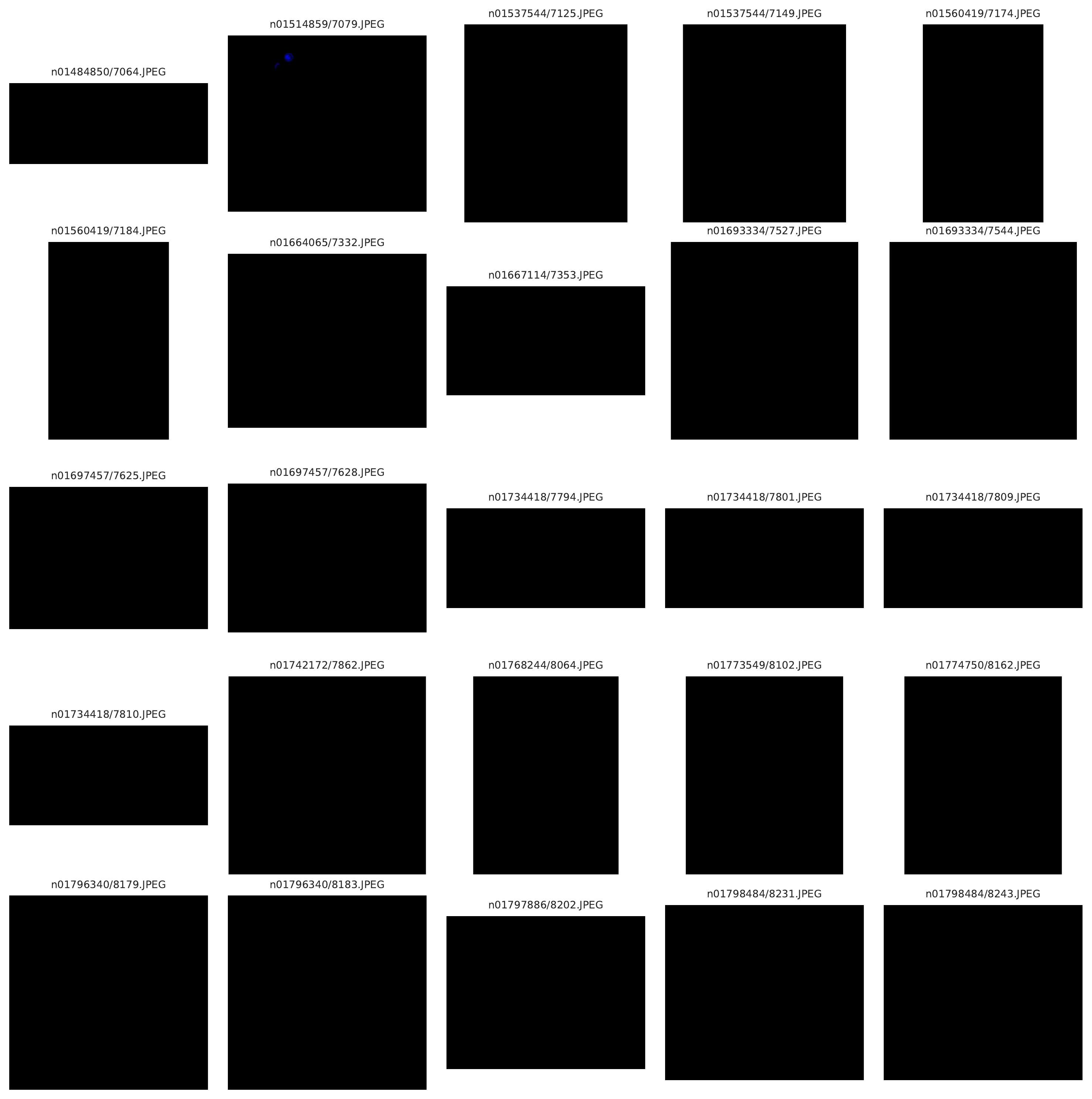}
    \caption{Sample images from ImageNet-Sketch that are completely black.}
    \label{fig:blcakimages_sketch}
\end{figure}

\clearpage
\section{Additional Details}

\subsection{Additional results for performance of classifiers using maximum possible accuracy}

In this section, we present the delta values for \cref{tab:1_main_results}, which represent the difference in relation to the 1-crop accuracy for each cell.

\begin{table}[htb!]
\centering
\captionof{table}{
On in-distribution data (IN \& ReaL) there exists a substantial improvement when models are provided with an optimal zoom, either selected from $36$ (b) or $324$ pre-defined zoom crops (c).
In contrast, \colorbox{MyLightGray}{OOD benchmarks} still pose a significant challenge to \textcolor{specialcolor}{IN-trained models} even with optimal zooming (i.e., all upper-bound accuracy scores < $80\%$). \textcolor{ForestGreen}{Improvements} are respected to the standard 1-crop accuracy.
}
\label{tab:1_main_results_supp}
\resizebox{\textwidth}{!}{%
\begin{tabular}{lrrcrrrr}
 & \multicolumn{1}{c}{IN} & \multicolumn{1}{c}{ReaL} & \multicolumn{1}{l}{IN+ReaL} & \multicolumn{1}{c}{\cellcolor{MyLightGray} IN-A} & \multicolumn{1}{c}{\cellcolor{MyLightGray} IN-R} & \multicolumn{1}{c}{\cellcolor{MyLightGray} IN-S} & \multicolumn{1}{c}{\cellcolor{MyLightGray} ON} \\ \bottomrule
\multicolumn{8}{l}{(a) \textit{Standard top-1 accuracy based on $N$ = \textbf{1} crop}} \\ 
{\textcolor{specialcolor}{AlexNet}} & 56.16 & 62.67 & 61.76 & 1.75 & 21.10 & 10.05 & 14.23 \\ 
{\textcolor{specialcolor}{VGG-16}} & 71.37 & 78.90 & 78.52 & 2.69 & 26.98 & 16.78 & 28.32 \\
{\textcolor{specialcolor}{ResNet-18}} & 69.45 & 76.94 & 76.47 & 1.37 & 32.14 & 19.41 & 27.59 \\
{\textcolor{specialcolor}{ResNet-50}} & 75.75 & 82.63 & 82.97 & 0.21 & 35.39 & 22.91 & 36.18 \\
{\textcolor{specialcolor}{ViT-B/32}} & 75.75 & 81.89 & 82.59 & 9.64 & 41.29 & 26.83 & 30.89 \\
\textbf{\footnotesize{\clip-ViT-L/14}} & 75.03 & 80.68 & 81.95 & 71.28 & 87.74 & 58.23 & 66.32 \\  \bottomrule 
\multicolumn{8}{l}{(b) \textit{Upper-bound accuracy using $N$ = \textbf{36} crops}} \\ 
{\textcolor{specialcolor}{AlexNet}}       & 85.19 (\increasenoparent{29.03}) & 90.30 (\increasenoparent{27.63}) & 89.74 (\increasenoparent{27.98}) & 31.37 (\increasenoparent{29.62}) & 47.04 (\increasenoparent{25.94}) & 24.40 (\increasenoparent{14.35}) & 49.17 (\increasenoparent{34.94}) \\
{\textcolor{specialcolor}{VGG-16}}        & 92.30 (\increasenoparent{20.93}) & 96.08 (\increasenoparent{17.18}) & 95.81 (\increasenoparent{17.29}) & 46.69 (\increasenoparent{44.00}) & 52.86 (\increasenoparent{25.88}) & 34.34 (\increasenoparent{17.56}) & 62.94 (\increasenoparent{34.62}) \\
{\textcolor{specialcolor}{ResNet-18}}     & 92.08 (\increasenoparent{22.63}) & 95.97 (\increasenoparent{19.03}) & 95.73 (\increasenoparent{19.26}) & 47.48 (\increasenoparent{46.11}) & 58.85 (\increasenoparent{26.71}) & 37.91 (\increasenoparent{18.50}) & 63.08 (\increasenoparent{35.49}) \\
{\textcolor{specialcolor}{ResNet-50}}     & 94.46 (\increasenoparent{18.71}) & 97.36 (\increasenoparent{14.73}) & 97.40 (\increasenoparent{14.43}) & 55.68 (\increasenoparent{55.47}) & 61.42 (\increasenoparent{26.03}) & 41.71 (\increasenoparent{18.80}) & 69.60 (\increasenoparent{33.42}) \\
{\textcolor{specialcolor}{ViT-B/32}}      & 95.05 (\increasenoparent{19.30}) & 97.61 (\increasenoparent{15.72}) & 97.88 (\increasenoparent{15.29}) & 68.43 (\increasenoparent{58.79}) & 68.77 (\increasenoparent{27.48}) & 49.10 (\increasenoparent{22.27}) & 70.30 (\increasenoparent{39.41}) \\
\textbf{\footnotesize{\clip-ViT-L/14}}    & 94.19 (\increasenoparent{19.16}) & 97.32 (\increasenoparent{16.64}) & 97.56 (\increasenoparent{15.61}) & 97.16 (\increasenoparent{25.88}) & 98.60 (\increasenoparent{10.86}) & 83.77 (\increasenoparent{25.54}) & 89.59 (\increasenoparent{23.27}) \\
\bottomrule
\multicolumn{8}{l}{(c) \textit{Upper-bound accuracy using $N$ = \textbf{324} crops}} \\
{\textcolor{specialcolor}{AlexNet}}   & 90.03 (\increasenoparent{33.87}) & 93.85 (\increasenoparent{31.18}) & 93.48 (\increasenoparent{31.72}) & 42.23 (\increasenoparent{40.48}) & 55.52 (\increasenoparent{34.42}) & 29.53 (\increasenoparent{19.48}) & 59.65 (\increasenoparent{45.42}) \\
{\textcolor{specialcolor}{VGG-16}}    & 95.30 (\increasenoparent{23.93}) & 97.90 (\increasenoparent{19.00}) & 97.66 (\increasenoparent{19.14}) & 58.27 (\increasenoparent{55.58}) & 60.88 (\increasenoparent{33.90}) & 39.90 (\increasenoparent{23.12}) & 71.85 (\increasenoparent{43.53}) \\
{\textcolor{specialcolor}{ResNet-18}} & 95.15 (\increasenoparent{25.70}) & 97.76 (\increasenoparent{20.82}) & 97.55 (\increasenoparent{21.08}) & 58.87 (\increasenoparent{57.50}) & 66.89 (\increasenoparent{34.75}) & 43.68 (\increasenoparent{24.27}) & 71.44 (\increasenoparent{43.85}) \\
{\textcolor{specialcolor}{ResNet-50}} & 96.78 (\increasenoparent{21.03}) & 98.62 (\increasenoparent{15.99}) & 98.57 (\increasenoparent{15.60}) & 66.68 (\increasenoparent{66.47}) & 68.84 (\increasenoparent{33.45}) & 47.64 (\increasenoparent{24.73}) & 76.83 (\increasenoparent{40.65}) \\
{\textcolor{specialcolor}{ViT-B/32}}  & \textbf{97.19} (\increasenoparent{21.44}) & \textbf{98.75} (\increasenoparent{16.86}) & \textbf{98.91} (\increasenoparent{16.32}) & 78.03 (\increasenoparent{68.39}) & 75.58 (\increasenoparent{34.29}) & 55.99 (\increasenoparent{29.16}) & 79.28 (\increasenoparent{48.39}) \\
\textbf{\footnotesize{\clip-ViT-L/14}} & 96.78 (\increasenoparent{21.75}) & 98.69 (\increasenoparent{18.01}) & 98.80 (\increasenoparent{16.85}) & \textbf{98.45} (\increasenoparent{27.17}) & \textbf{99.20} (\increasenoparent{11.46}) & \textbf{89.00} (\increasenoparent{30.77}) & \textbf{93.13} (\increasenoparent{26.81}) \\
\bottomrule
\end{tabular}
}
\end{table}     

\FloatBarrier
\subsection{Comparing modern architectures with regard to their maximum possible accuracy.}

In this section, we repeated the experiment presented in \cref{tab:1_main_results} in order to include a broader range of architectures and conduct a rank analysis. Specifically, we added MaxViT~\cite{tu2022maxvit}, Swin Transformer~\cite{liu2021swin}, ConvNext~\cite{liu2022convnet}, EfficientNet~\cite{tan2019efficientnet}, and MobileNetV3~\cite{howard2019searching}.

\begin{figure}[htb!]
\centering
\begin{subfigure}{0.525\columnwidth}
\includegraphics[width=\columnwidth,keepaspectratio]{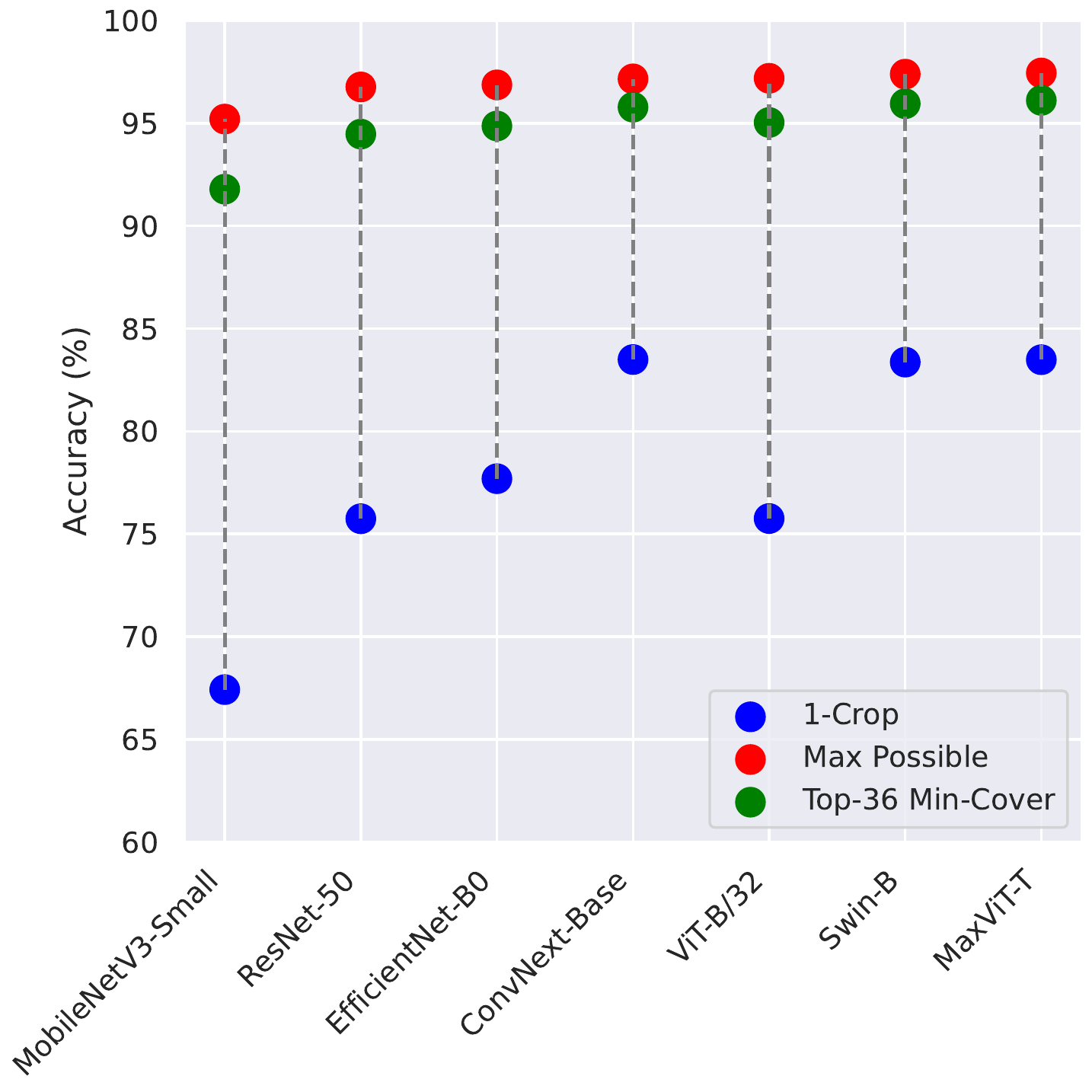}
\caption{\scriptsize{ImageNet}}
\end{subfigure}
\hfill
\begin{subfigure}{0.525\columnwidth}
\includegraphics[width=\columnwidth,keepaspectratio]{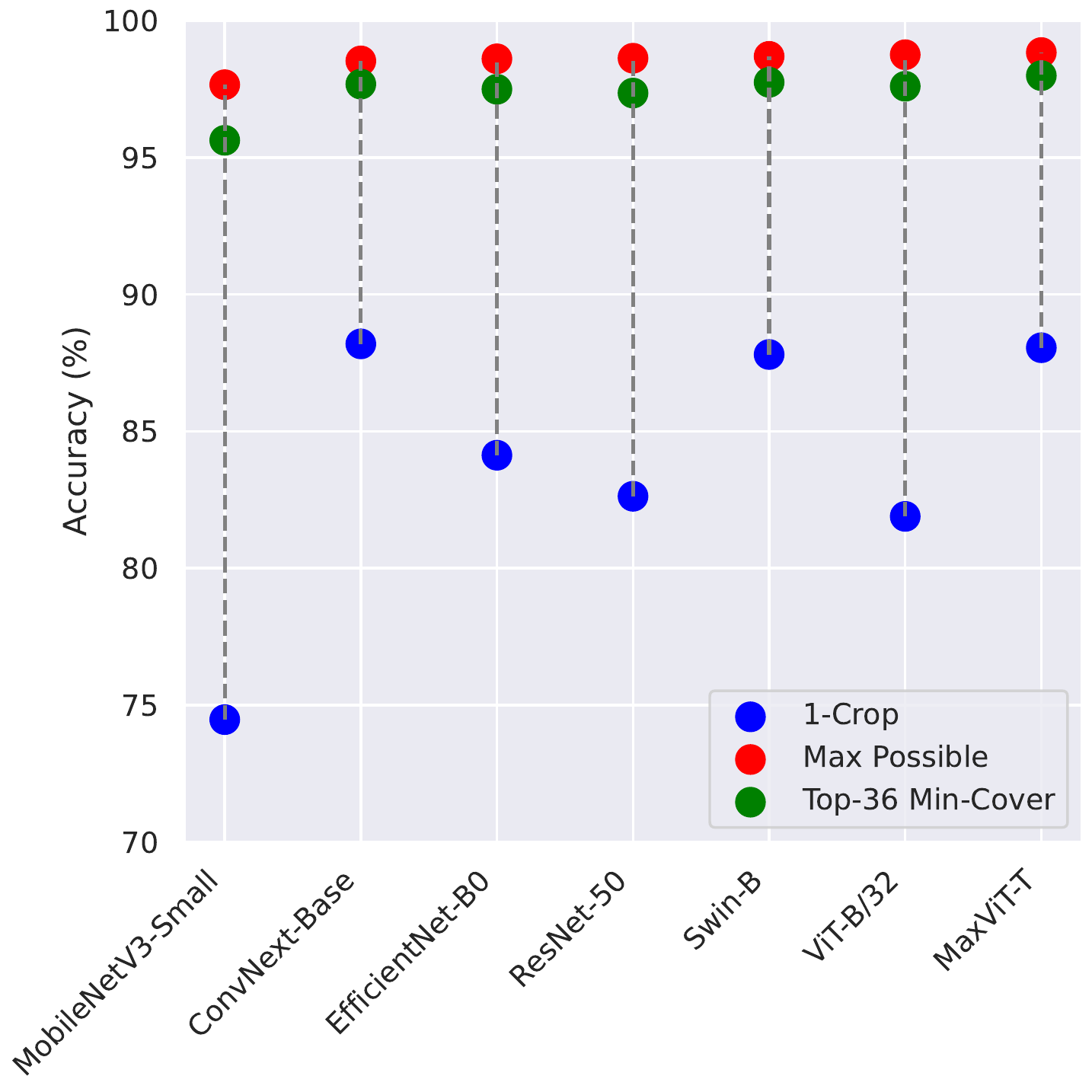}
\caption{\scriptsize{ImageNet-ReaL}}
\end{subfigure}
\hfill
\begin{subfigure}{0.525\columnwidth}
\includegraphics[width=\columnwidth,keepaspectratio]{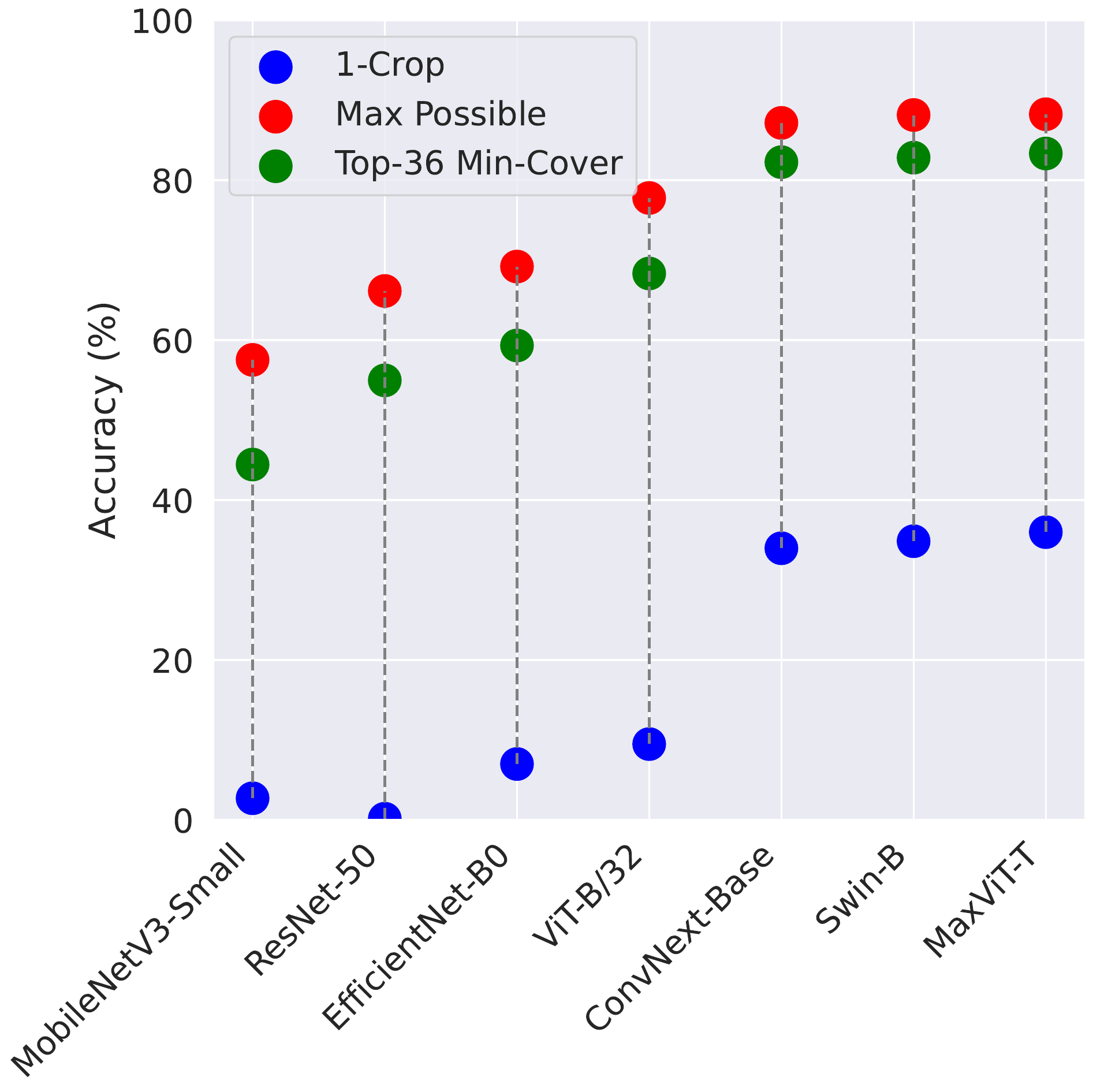}
\caption{\scriptsize{ImageNet-A}}
\end{subfigure}
\caption{Comparing modern architectures using maximum possible accuracy}
\label{fig:comparing_modern_arch}
\end{figure}

\begin{table}[htb!]
\centering
\captionof{table}{
Repeating the experiment in \cref{tab:1_main_results} for various architectures, please note that the small variations of ResNet-50 and ViT-B/32 models arise from the use of different CUDA and PyTorch versions.
}
\label{tab:1_modern_results_supp}
\begin{tabular}{lrrr}
\toprule
\multicolumn{4}{l}{(a) \textit{Standard top-1 accuracy based on \( N = \mathbf{1} \) crop}} \\
\textbf{Classifier}       & \multicolumn{1}{c}{ImageNet} & \multicolumn{1}{c}{ImageNet-ReaL} & \multicolumn{1}{c}{\cellcolor{MyLightGray} ImageNet-A} \\ \midrule
\textcolor{specialcolor}{MobileNetV3-Small} & 67.42 & 74.47 & 2.73 \\
\textcolor{specialcolor}{ResNet-50} & 75.74 & 82.63 & 0.17 \\
\textcolor{specialcolor}{EfficientNet-B0} & 77.69 & 84.13 & 7.01 \\
\textcolor{specialcolor}{ViT-B/32} & 75.75 & 81.89 & 9.51 \\
\textcolor{specialcolor}{ConvNext-Base} & 83.49 & 88.19 & 33.99 \\
\textcolor{specialcolor}{Swin-B} & 83.37 & 87.80 & 34.87 \\
\textcolor{specialcolor}{MaxViT-T} & 83.49 & 88.05 & 36.00 \\
\midrule
\multicolumn{4}{l}{(b) \textit{Upper-bound accuracy using \( N = \mathbf{36} \) crops}} \\
\textcolor{specialcolor}{MobileNetV3-Small} & 91.79 & 95.63 & 44.45 \\
\textcolor{specialcolor}{ResNet-50} & 94.47 & 97.35 & 54.97 \\
\textcolor{specialcolor}{EfficientNet-B0} & 94.87 & 97.49 & 59.33 \\
\textcolor{specialcolor}{ViT-B/32} & 95.04 & 97.60 & 68.33 \\
\textcolor{specialcolor}{ConvNext-Base} & 95.78 & 97.68 & 82.25 \\
\textcolor{specialcolor}{Swin-B} & 95.95 & 97.75 & 82.81 \\
\textcolor{specialcolor}{MaxViT-T} & 96.11 & 97.99 & 83.33 \\
\midrule
\multicolumn{4}{l}{(c) \textit{Upper-bound accuracy using $N$ = \textbf{324} crops}} \\
\textcolor{specialcolor}{MobileNetV3-Small} & 95.20 & 97.66 & 57.53 \\
\textcolor{specialcolor}{ResNet-50} & 96.77 & 98.63 & 66.15 \\
\textcolor{specialcolor}{EfficientNet-B0} & 96.87 & 98.60 & 69.20 \\
\textcolor{specialcolor}{ViT-B/32} & 97.19 & 98.75 & 77.77 \\
\textcolor{specialcolor}{ConvNext-Base} & 97.16 & 98.53 & 87.16 \\
\textcolor{specialcolor}{Swin-B} & 97.39 & 98.69 & 88.15 \\
\textcolor{specialcolor}{MaxViT-T} & 97.45 & 98.83 & 88.24 \\
\bottomrule
\end{tabular}
\end{table}

\begin{figure}[htb!]
\centering
\includegraphics[width=\columnwidth,keepaspectratio]{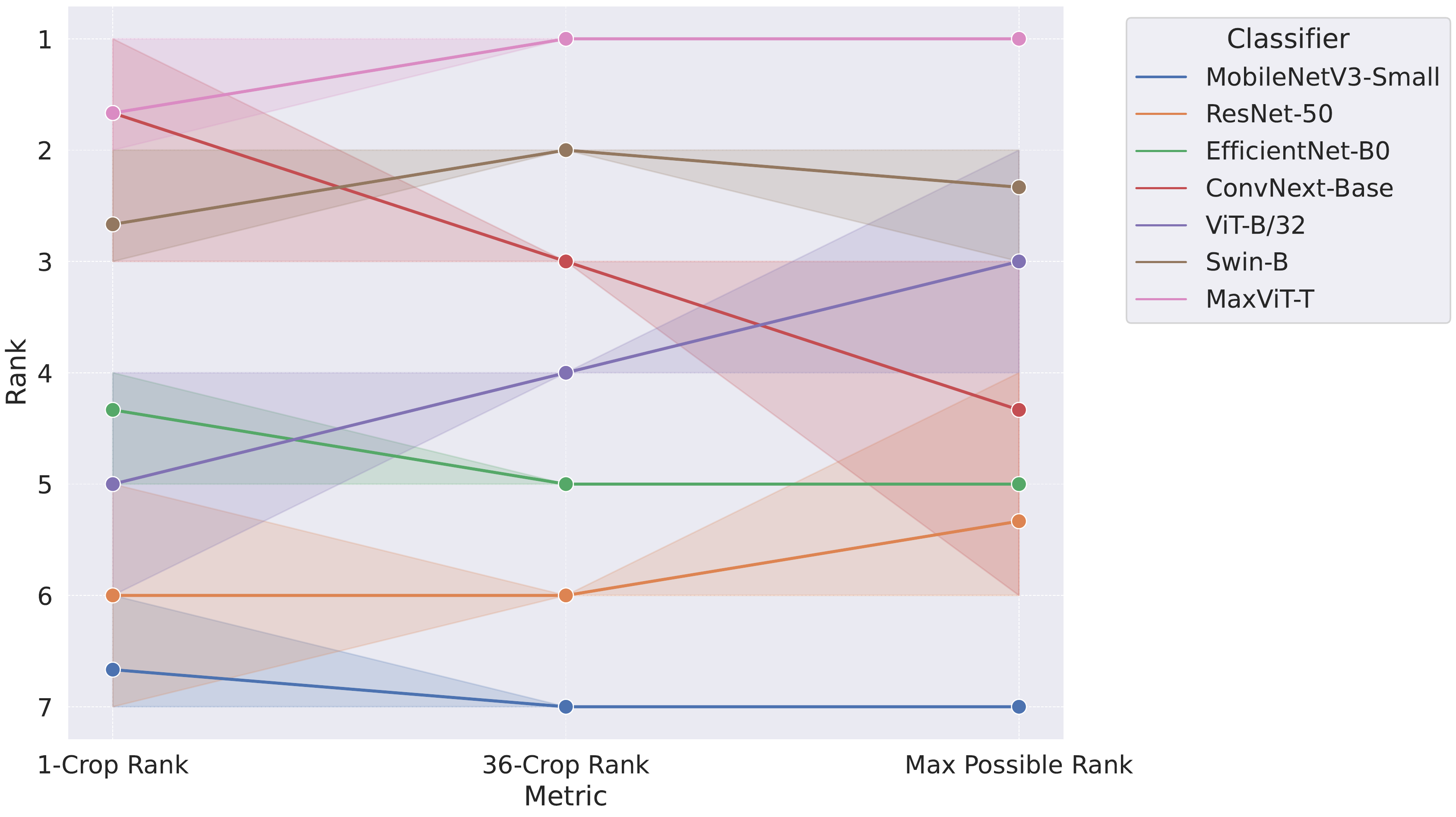}
\caption{Average rank analysis of various architectures on ImageNet, ImageNet-ReaL, and ImageNet-A using different metrics. Max-ViT consistently emerges as the top performer across all experiments. Despite its high 1-crop accuracy, ConvNext's rank declined when more crops were included. Conversely, ViT-B/32 exhibited a reverse trend; while its 1-crop rank was not robust, the inclusion of more crops elevated its accuracy beyond that of ConvNext.}
\label{fig:Rank_Analysis}
\end{figure}

\FloatBarrier
\subsection{Error-analysis for MEMO results}

In this section, we repeated the MEMO experiment using various random seeds to evaluate the consistency of the results. Our findings demonstrate a consistent trend between MEMO and \rrc, indicating that this relationship holds steady regardless of variations in individual runs.

\begin{figure}[htb!]
\centering
\begin{subfigure}{0.325\columnwidth}
\includegraphics[width=\columnwidth,keepaspectratio]{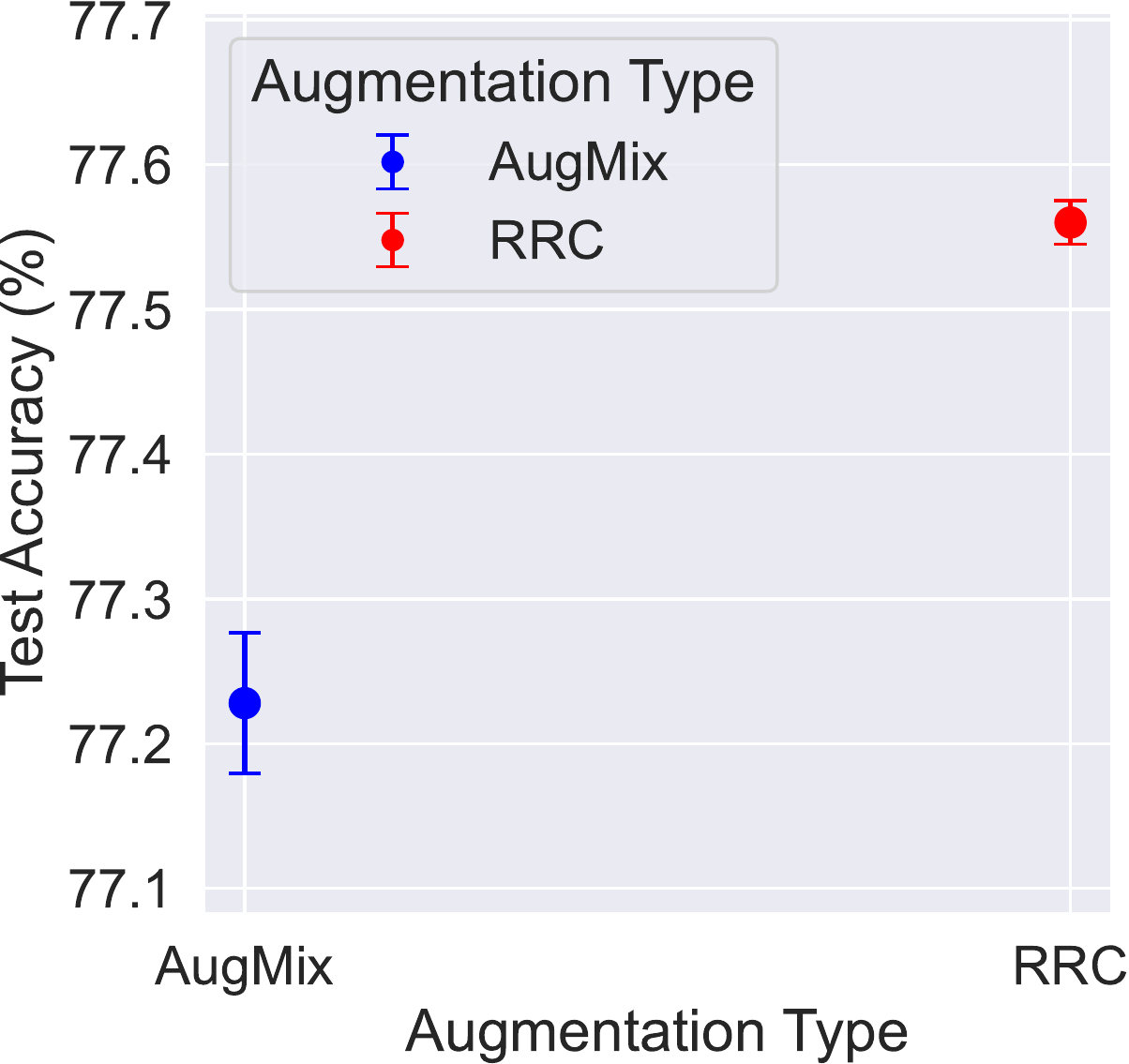}
\caption{\scriptsize{ResNet-50~\cite{he2016deep}}}
\end{subfigure}
\hfill
\begin{subfigure}{0.325\columnwidth}
\includegraphics[width=\columnwidth,keepaspectratio]{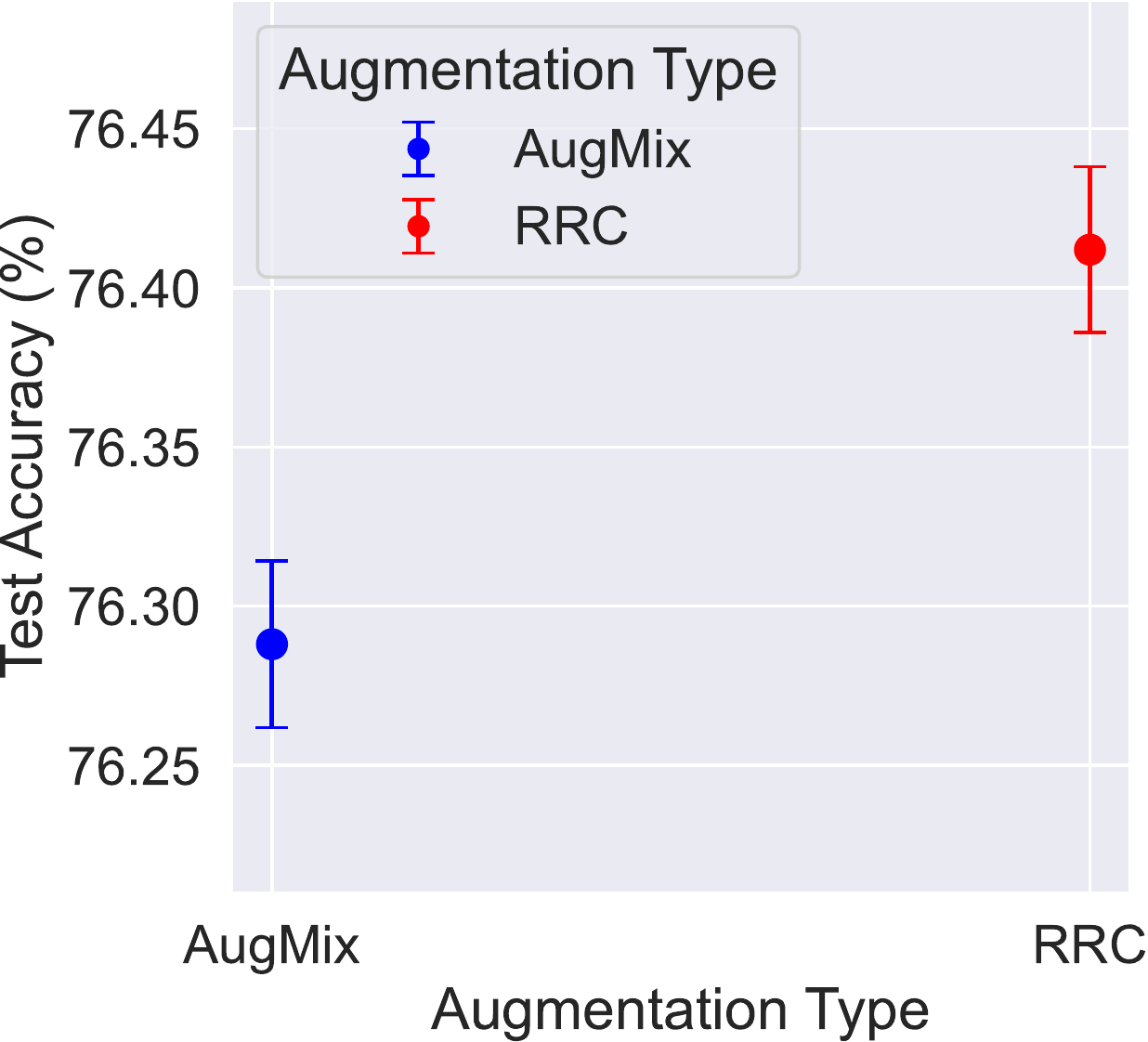}
\caption{\scriptsize{DeepAug+AugMix~\cite{hendrycks2021many}}}
\end{subfigure}
\hfill
\begin{subfigure}{0.325\columnwidth}
\includegraphics[width=\columnwidth,keepaspectratio]{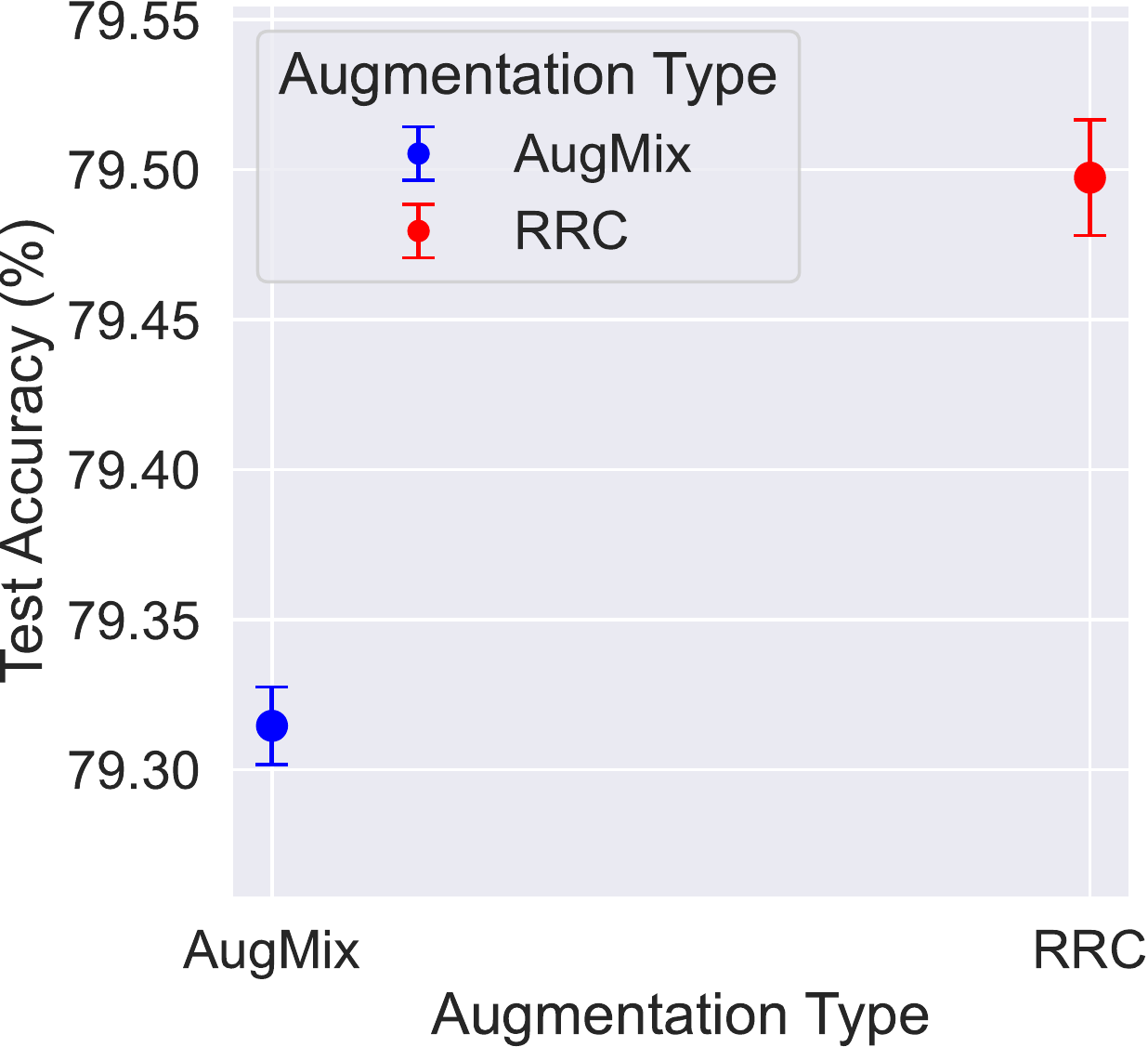}
\caption{\scriptsize{MoEx+CutMix~\cite{li2021feature} }}
\end{subfigure}
\caption{Error bars representing the outcomes of the MEMO experiment on ImageNet, conducted with three distinct random seeds.}
\label{fig:additional_results_MEMO_IN}
\end{figure}

\begin{figure}[htb!]
\centering
\begin{subfigure}{0.325\columnwidth}
\includegraphics[width=\columnwidth,keepaspectratio]{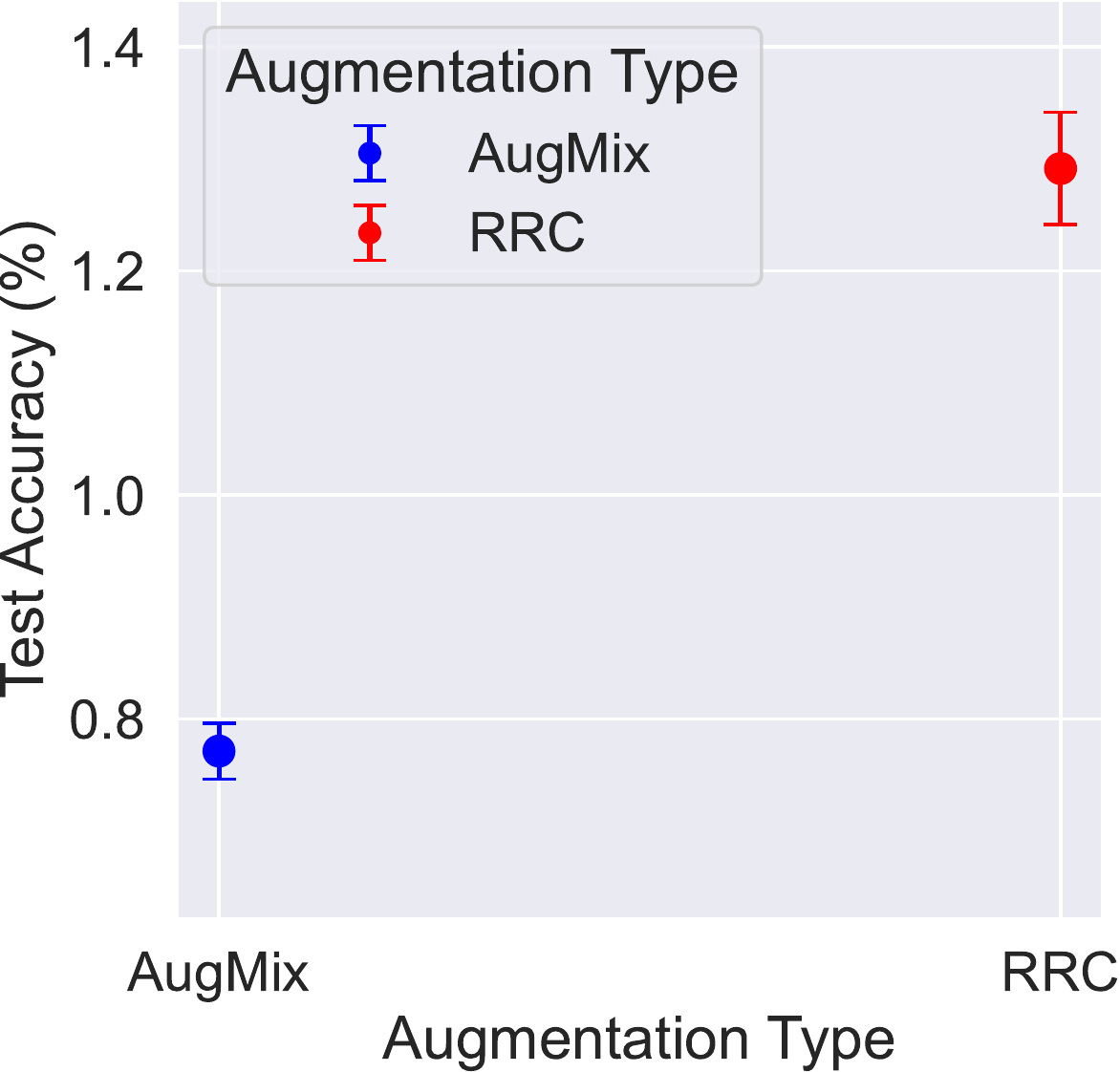}
\caption{\scriptsize{ResNet-50~\cite{he2016deep}}}
\end{subfigure}
\hfill
\begin{subfigure}{0.325\columnwidth}
\includegraphics[width=\columnwidth,keepaspectratio]{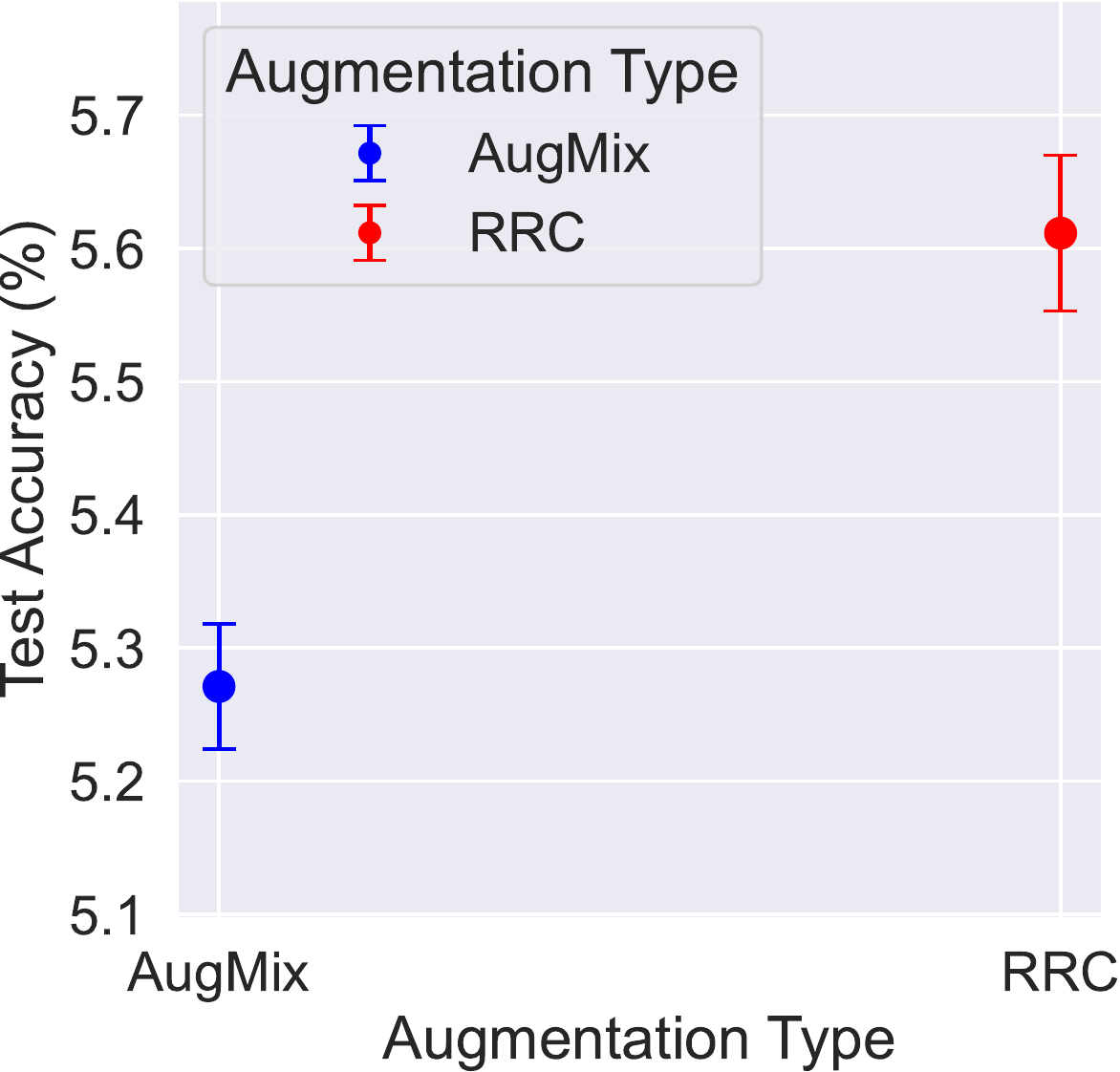}
\caption{\scriptsize{DeepAug+AugMix~\cite{hendrycks2021many}}}
\end{subfigure}
\hfill
\begin{subfigure}{0.325\columnwidth}
\includegraphics[width=\columnwidth,keepaspectratio]{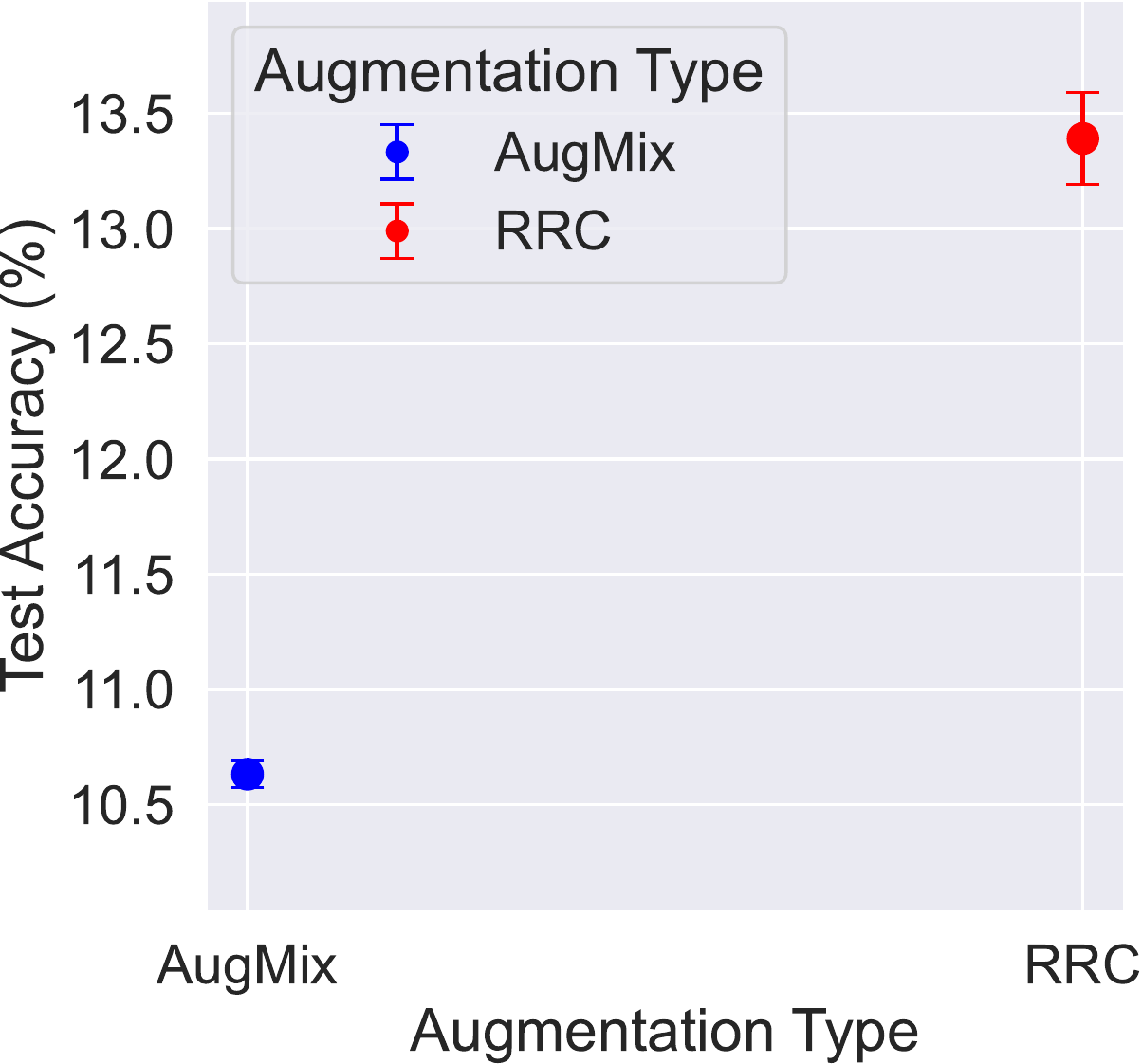}
\caption{\scriptsize{MoEx+CutMix~\cite{li2021feature} }}
\end{subfigure}
\caption{Error bars representing the outcomes of the MEMO experiment on ImageNet-A, conducted with three distinct random seeds.}
\label{fig:additional_results_MEMO_INA}
\end{figure}

\FloatBarrier
\subsection{Grad-CAM visualizations for MEMO + \rrc}

\begin{figure}[htb!]
\centering
\begin{subfigure}[b]{0.495\textwidth}
\includegraphics[height=2.3cm,keepaspectratio]{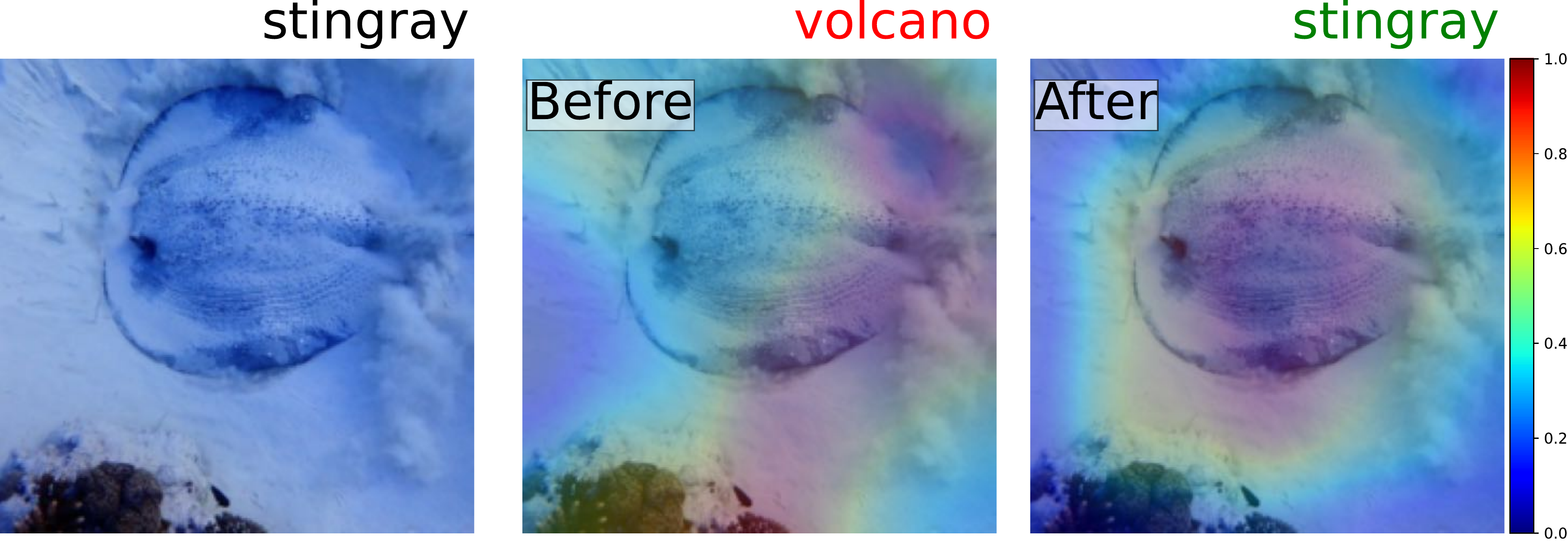}
\end{subfigure}
\hfill
\begin{subfigure}[b]{0.495\textwidth}
\includegraphics[height=2.3cm,keepaspectratio]{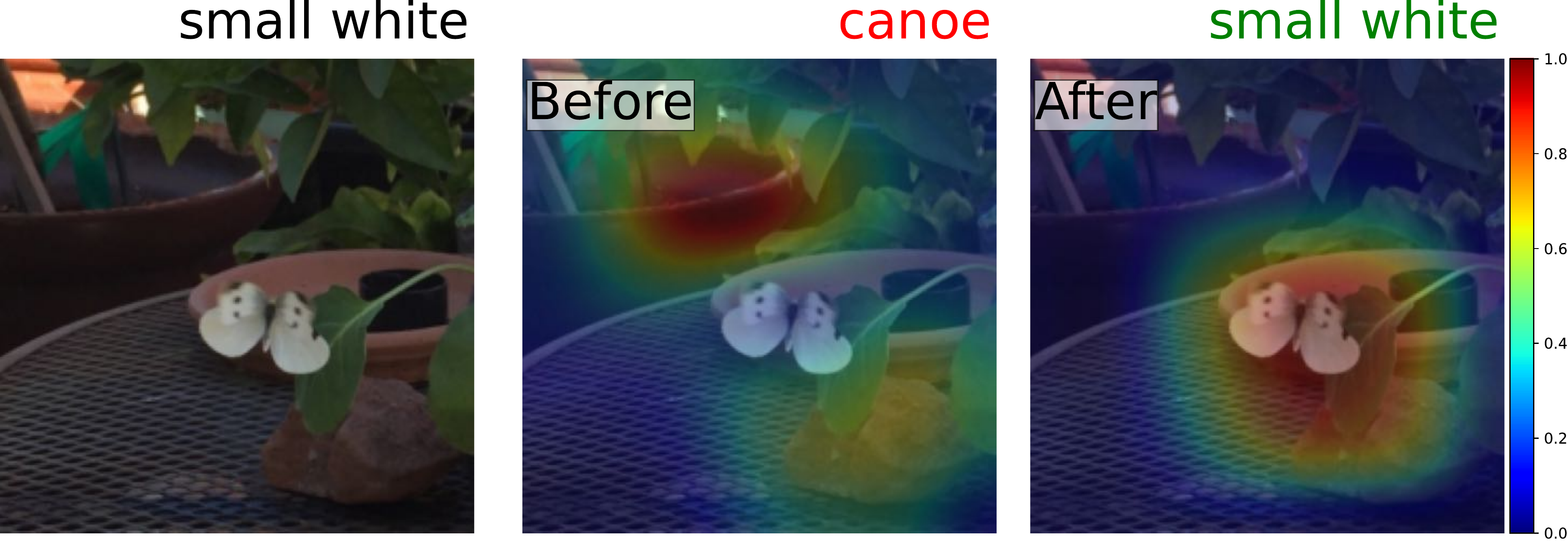}
\end{subfigure}
\hfill
\begin{subfigure}[b]{0.495\textwidth}
\includegraphics[height=2.3cm,keepaspectratio]{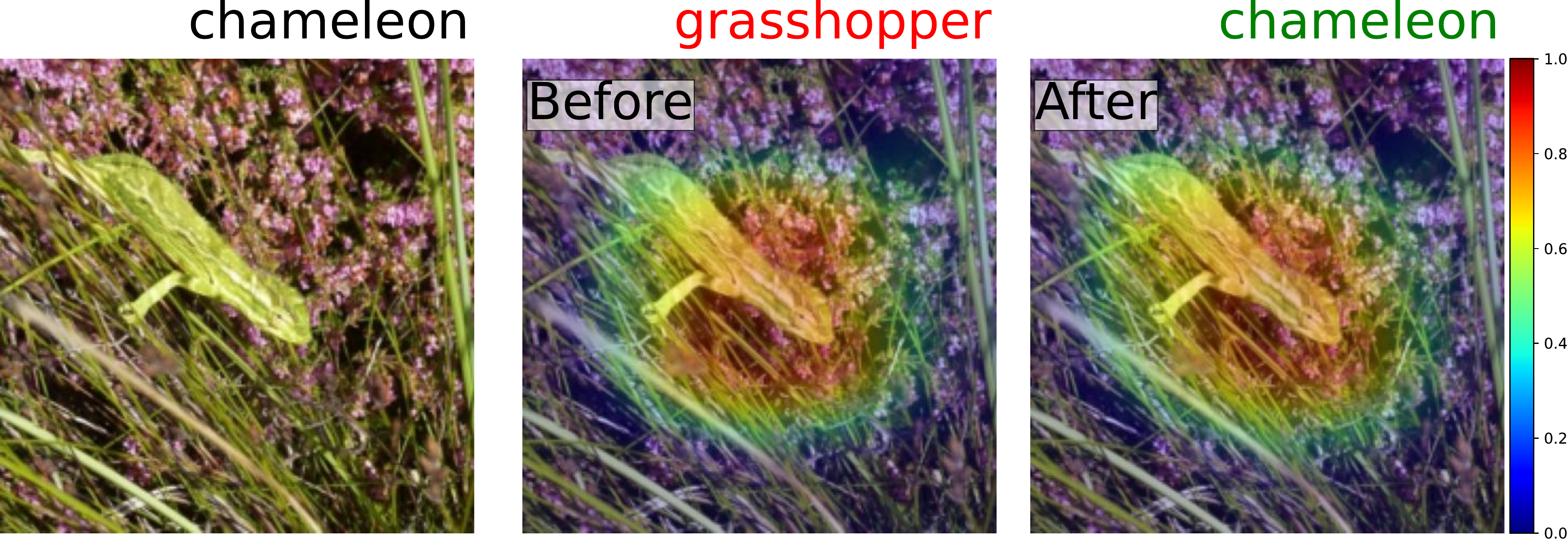}
\end{subfigure}
\hfill
\begin{subfigure}[b]{0.495\textwidth}
\includegraphics[height=2.3cm,keepaspectratio]{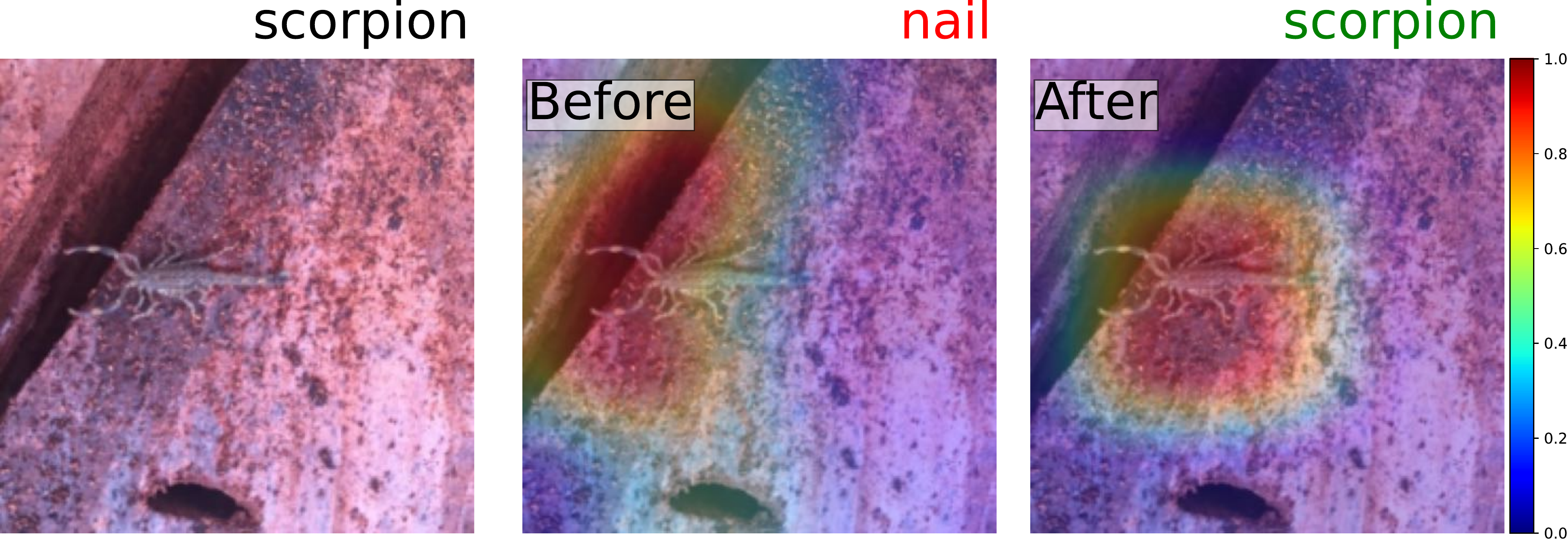}
\end{subfigure}
\caption{Additional Grad-CAM visualization for the final convolutional layer of a ResNet-50 before and after MEMO + \rrc update.}
\label{fig:memo_before_after_supp1}
\end{figure}

\begin{figure}[htb!]
\centering
\begin{subfigure}[b]{0.495\textwidth}
\includegraphics[height=2.3cm,keepaspectratio]{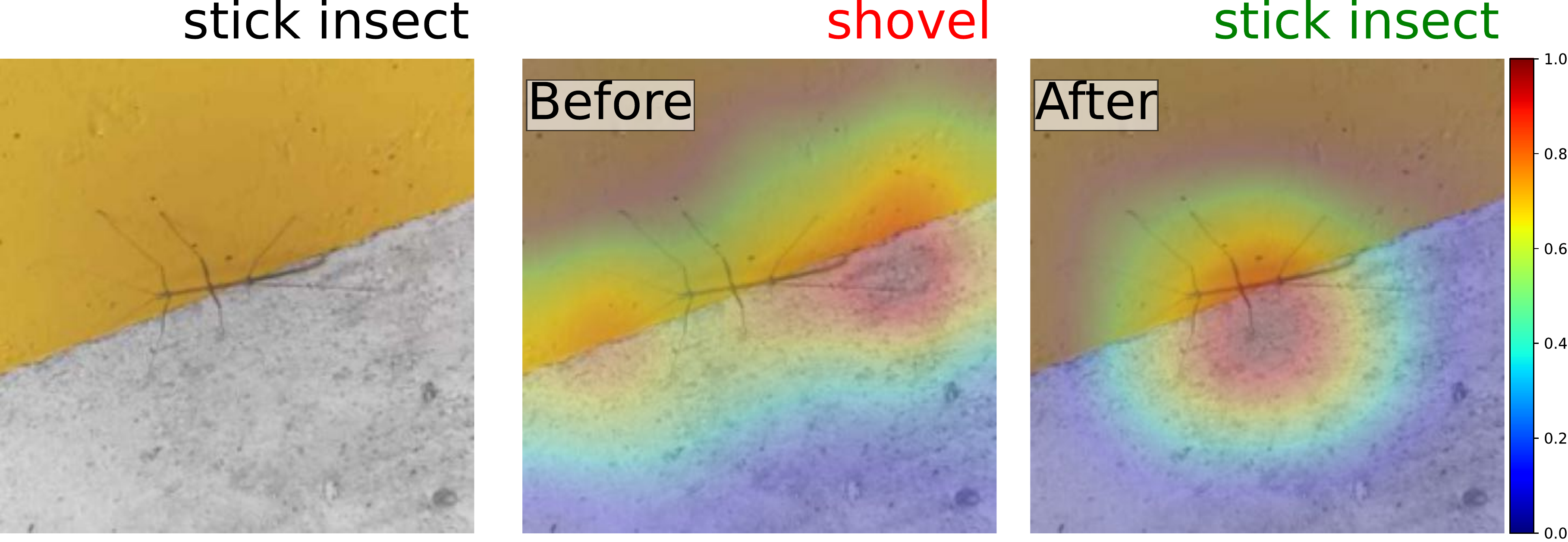}
\end{subfigure}
\hfill
\begin{subfigure}[b]{0.495\textwidth}
\includegraphics[height=2.3cm,keepaspectratio]{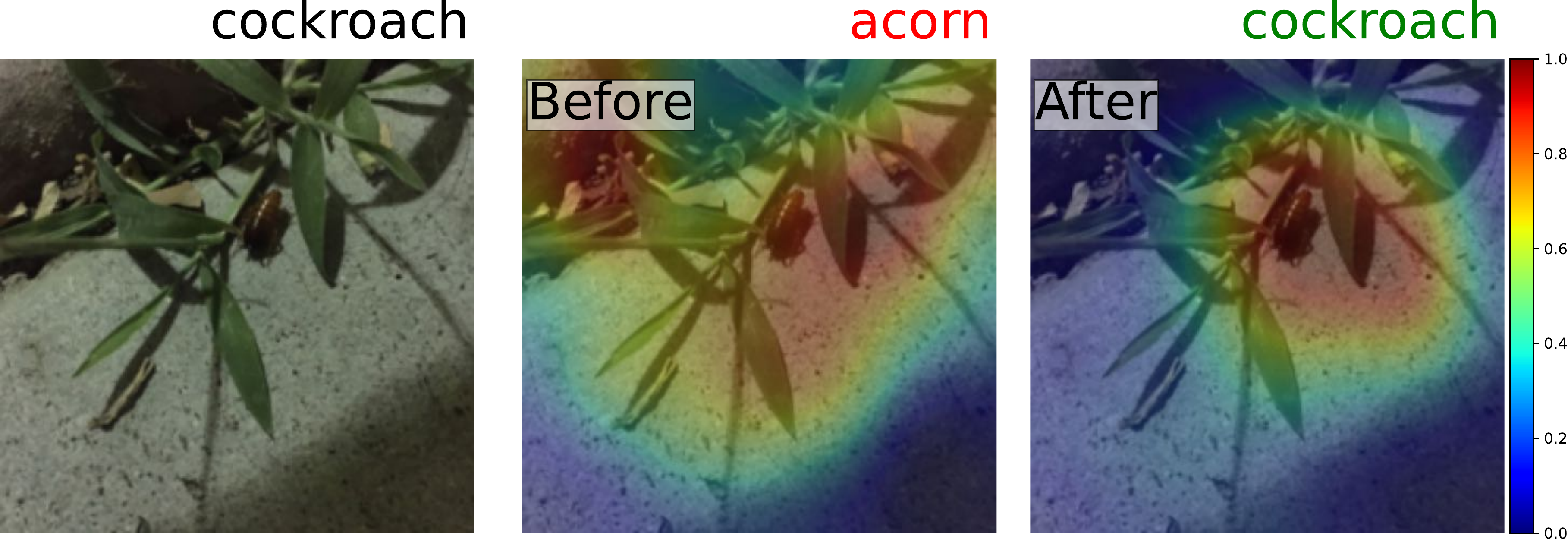}
\end{subfigure}
\hfill
\begin{subfigure}[b]{0.495\textwidth}
\includegraphics[height=2.3cm,keepaspectratio]{rebuttal/memo_before_after/4207.pdf}
\end{subfigure}
\hfill
\begin{subfigure}[b]{0.495\textwidth}
\includegraphics[height=2.3cm,keepaspectratio]{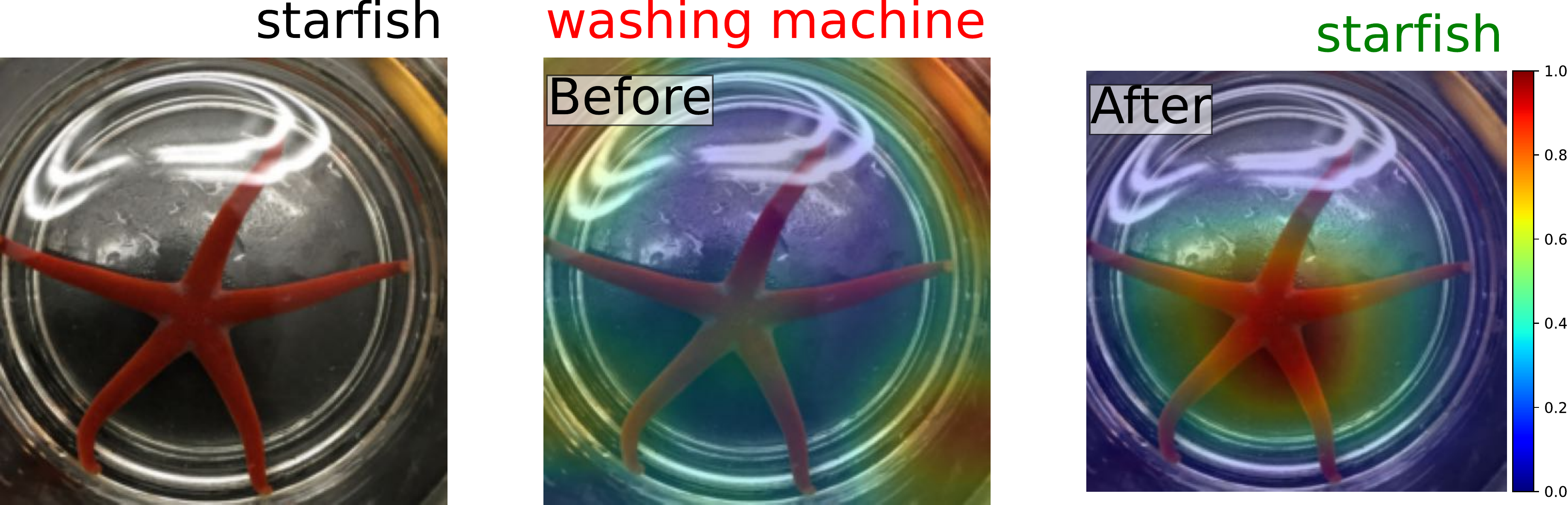}
\end{subfigure}
\hfill
\begin{subfigure}[b]{0.495\textwidth}
\includegraphics[height=2.3cm,keepaspectratio]{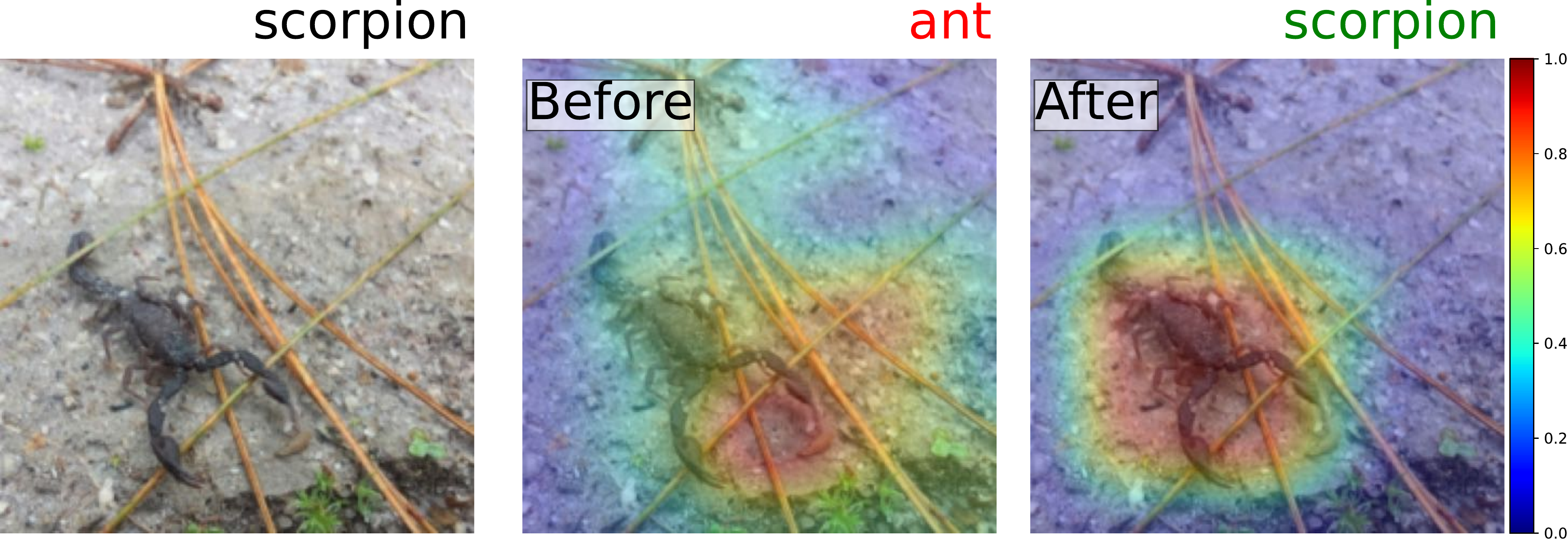}
\end{subfigure}
\hfill
\begin{subfigure}[b]{0.495\textwidth}
\includegraphics[height=2.3cm,keepaspectratio]{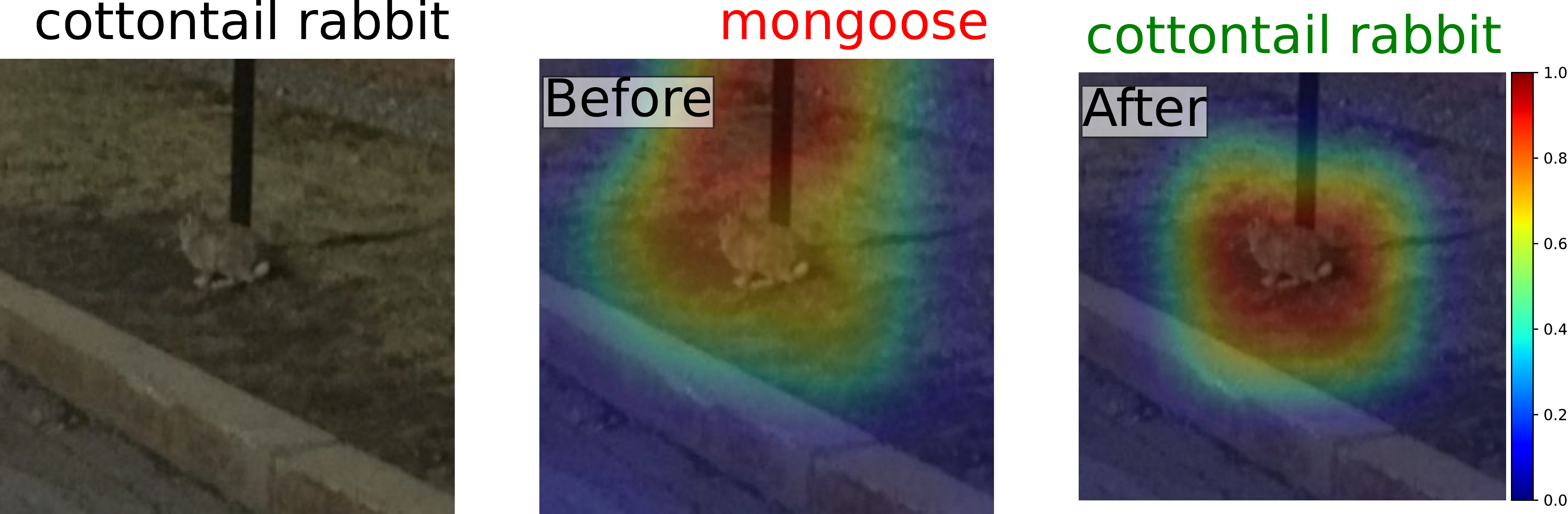}
\end{subfigure}
\hfill
\begin{subfigure}[b]{0.495\textwidth}
\includegraphics[height=2.3cm,keepaspectratio]{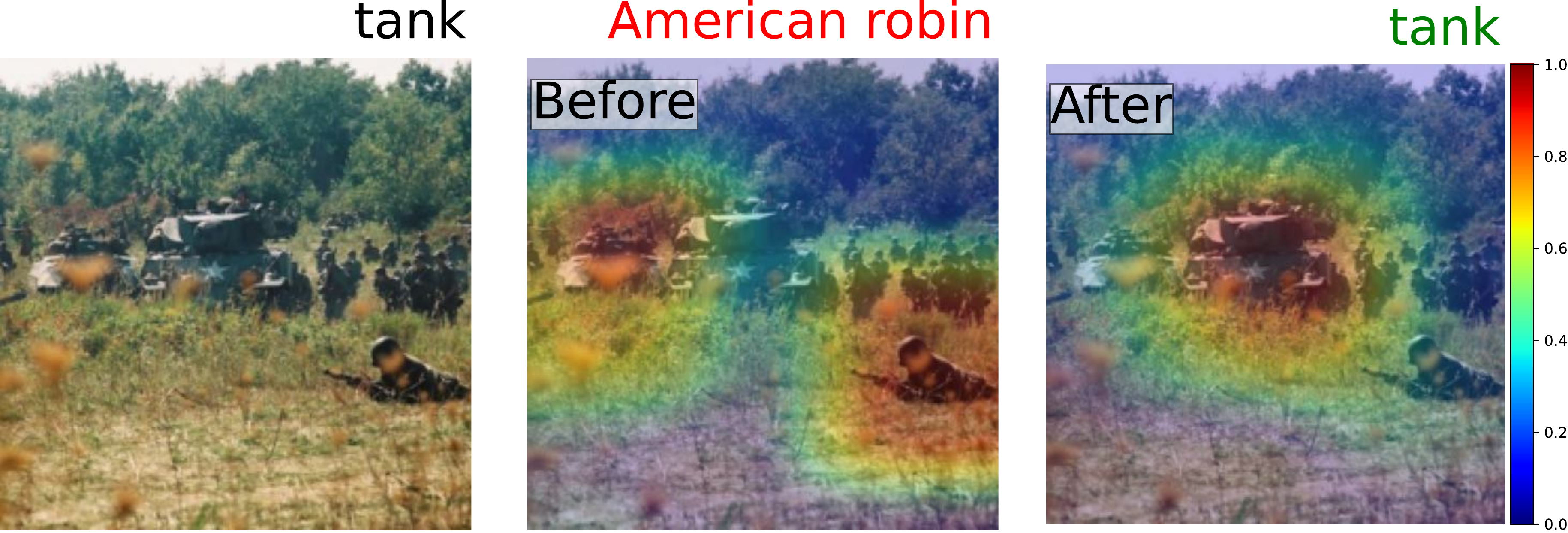}
\end{subfigure}
\hfill
\begin{subfigure}[b]{0.495\textwidth}
\includegraphics[height=2.3cm,keepaspectratio]{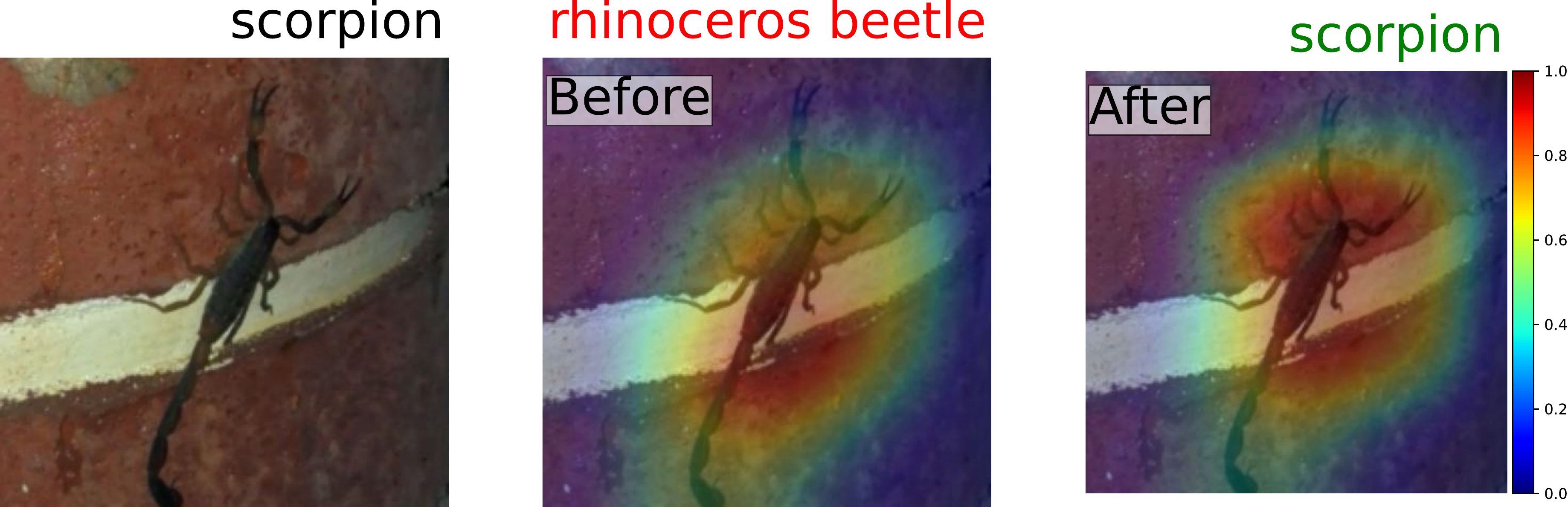}
\end{subfigure}
\caption{Additional Grad-CAM visualization for the final convolutional layer of a ResNet-50 before and after MEMO + \rrc update.}
\label{fig:memo_before_after_supp2}
\end{figure}

\clearpage
\section{Datasheet for ImageNet-Hard}

\subsection{Motivation}

The questions in this section are primarily intended to encourage
dataset creators to clearly articulate their reasons for creating the
dataset and to promote transparency about funding interests.
The latter may be particularly relevant for datasets created for
research purposes.\\

\begin{itemize}

\item \textbf{For what purpose was the dataset created?} Was there a specific task in mind? Was there a specific gap that needed to be filled? Please provide a description.

The ImageNet-Hard is a new benchmark to test the robustness of state-of-the-art image classifiers. 
It comprises an array of challenging images collected from \emph{six} validation datasets of ImageNet. 
This dataset challenges state-of-the-art image classification models because even by perfectly localizing the key objects, the state-of-the-art classifiers still fail to correctly recognize.

\item \textbf{Who created the dataset (e.g., which team, research group) and on behalf of which entity (e.g., company, institution, organization)?}

The dataset was created in collaboration efforts between the University of Alberta, Canada, and Auburn University, USA; mostly by, Mohammad Reza Taesiri and Anh Nguyen.

\item \textbf{Who funded the creation of the dataset?} If there is an associated grant, please provide the name of the grantor and the grant name and number.

Anh Nguyen was supported by NSF Grant No. 2145767, and donations from NaphCare Foundation, and Adobe Research

\item \textbf{Any other comments?}

No.

\end{itemize}

\subsection{Composition}

Dataset creators should read through \textit{these questions} prior to
any data collection and then provide answers once \textit{data} collection is
complete. Most of the questions \textit{in this section} are intended to
provide dataset consumers with the information they need to make
informed decisions about using the dataset for their chosen
tasks. Some of the questions are \textit{designed to elicit} information
about compliance with the EU's General Data Protection Regulation
(GDPR) or comparable regulations in other jurisdictions.

\textit{Questions that apply only to datasets that relate to people are
grouped together at the end of the section. We recommend taking a
broad interpretation of whether a dataset relates to people. For
example, any dataset containing text that was written by people
relates to people.}\\

\begin{itemize}

\item \textbf{What do the instances that comprise the dataset
    represent (e.g., documents, photos, people, countries)?} Are there
  multiple types of instances (e.g., movies, users, and ratings;
  people and interactions between them; nodes and edges)? Please
  provide a description.

  Each instance of the ImageNet-Hard dataset corresponds to an image and at least one groundtruth label that will be used to assess image classifiers.

\item \textbf{How many instances are there in total (of each type, if appropriate)?}

There are 10,980 images in this dataset.

\item \textbf{Does the dataset contain all possible instances or is it
    a sample (not necessarily random) of instances from a larger set?}
  If the dataset is a sample, then what is the larger set? Is the
  sample representative of the larger set (e.g., geographic coverage)?
  If so, please describe how this representativeness was
  validated/verified. If it is not representative of the larger set,
  please describe why not (e.g., to cover a more diverse range of
  instances, because instances were withheld or unavailable).

ImageNet-Hard is a combination of various publicly available datasets. 
We tried multiple refinement steps to make sure to get the best possible samples for the intended purpose.
Then, it is not representative of any larger sets but a selective combination of multiple sets.

\item \textbf{What data does each instance consist of?} ``Raw'' data
  (e.g., unprocessed text or images) or features? In either case,
  please provide a description.

  The dataset contains both raw and processed images.
  The processed images come from ImageNet-C.
  Details can be found in \cref{sec:imagenet_hard}.

\item \textbf{Is there a label or target associated with each
    instance?} If so, please provide a description.

    Yes. Each sample has the label that is the folder name the image belongs to.
    Basically, we follow the structure of the ImageNet paper \cite{russakovsky2015imagenet}.

\item \textbf{Is any information missing from individual instances?}
  If so, please provide a description, explaining why this information
  is missing (e.g., because it was unavailable). This does not include
  intentionally removed information, but might include, e.g., redacted
  text.

  No.

\item \textbf{Are relationships between individual instances made
    explicit (e.g., users' movie ratings, social network links)?} If
  so, please describe how these relationships are made explicit.

  No. The individual instances has no relationships.

\item \textbf{Are there recommended data splits (e.g., training,
    development/validation, testing)?} If so, please provide a
  description of these splits, explaining the rationale behind them.

  No. This dataset is created for the testing purposes.

\item \textbf{Are there any errors, sources of noise, or redundancies
    in the dataset?} If so, please provide a description.

    To the best of our knowledge, No. We tried our best efforts to filter any errors, sources of noise, or redundancies
    to create the ImageNet-Hard dataset.

\item \textbf{Is the dataset self-contained, or does it link to or
    otherwise rely on external resources (e.g., websites, tweets,
    other datasets)?} If it links to or relies on external resources,
    a) are there guarantees that they will exist, and remain constant,
    over time; b) are there official archival versions of the complete
    dataset (i.e., including the external resources as they existed at
    the time the dataset was created); c) are there any restrictions
    (e.g., licenses, fees) associated with any of the external
    resources that might apply to a \textit{dataset consumer}? Please provide
    descriptions of all external resources and any restrictions
    associated with them, as well as links or other access points, as
    appropriate.

    Yes. It does link and inherits from existing image datasets and was detailed in \cref{sec:imagenet_hard}.

\item \textbf{Does the dataset contain data that might be considered
    confidential (e.g., data that is protected by legal privilege or
    by doctor\textit{--}patient confidentiality, data that includes the content
    of individuals' non-public communications)?} If so, please provide
    a description.

    No.

\item \textbf{Does the dataset contain data that, if viewed directly,
    might be offensive, insulting, threatening, or might otherwise
    cause anxiety?} If so, please describe why.

    No.
\end{itemize}

\textit{If the dataset does not }relate to people, you may skip the remaining questions in this section.

\begin{itemize}

\item \textbf{Does the dataset identify any subpopulations (e.g., by
    age, gender)?} If so, please describe how these subpopulations are
  identified and provide a description of their respective
  distributions within the dataset.

  N/A.

\item \textbf{Is it possible to identify individuals (i.e., one or
    more natural persons), either directly or indirectly (i.e., in
    combination with other data) from the dataset?} If so, please
    describe how.

    N/A.

\item \textbf{Does the dataset contain data that might be considered
    sensitive in any way (e.g., data that reveals race or ethnic
    origins, sexual orientations, religious beliefs, political
    opinions or union memberships, or locations; financial or health
    data; biometric or genetic data; forms of government
    identification, such as social security numbers; criminal
    history)?} If so, please provide a description.

    N/A.
\item \textbf{Any other comments?}

No.

\end{itemize}

\subsection{Collection Process}

As with the \textit{questions in the} previous section, dataset creators should
read through these questions prior to any data collection to flag
potential issues and then provide answers once collection is complete.
\textit{In addition to the goals outlined in the previous section, the
questions in this section are designed to elicit information that may
help researchers and practitioners to create alternative datasets with
similar characteristics. Again, questions that apply only to datasets
that relate to people are grouped together at the end of the
section.}\\

\begin{itemize}

\item \textbf{How was the data associated with each instance
    acquired?} Was the data directly observable (e.g., raw text, movie
  ratings), reported by subjects (e.g., survey responses), or
  indirectly inferred/derived from other data (e.g., part-of-speech
  tags, model-based guesses for age or language)? If the data was reported
  by subjects or indirectly inferred/derived from other data, was the
  data validated/verified? If so, please describe how.

  The dataset is linked from other 6 datasets.
  Please find the contribution of original daatasets in \cref{supp:imagenet_hard_section_dist})

\item \textbf{What mechanisms or procedures were used to collect the
    data (e.g., hardware apparatuses or sensors, manual human
    curation, software programs, software APIs)?} How were these
    mechanisms or procedures validated?

    We used both algorithm and human efforts to collect the data.
    Algorithms were used to choose hard samples from various datasets.
    We then used two human groups and their agreement to make sure the high quality of the process. 
    Details for the human validation in ~\cref{sec:imagenet_hard}.
    Finally, we removed samples that have debatable labels (e.g. sunglass vs. sunglasses).

\item \textbf{If the dataset is a sample from a larger set, what was
    the sampling strategy (e.g., deterministic, probabilistic with
    specific sampling probabilities)?}

    No, the dataset was not a subset of a larger set.

\item \textbf{Who was involved in the data collection process (e.g.,
    students, crowdworkers, contractors) and how were they compensated
    (e.g., how much were crowdworkers paid)?}

    In the data collection process, we involved students who voluntarily participated.

\item \textbf{Over what timeframe was the data collected?} Does this
  timeframe match the creation timeframe of the data associated with
  the instances (e.g., recent crawl of old news articles)?  If not,
  please describe the timeframe in which the data associated with the
  instances was created.

  Feedback data was collected from April 20 2023 -- May 4 2023.

\item \textbf{Were any ethical review processes conducted (e.g., by an
    institutional review board)?} If so, please provide a description
  of these review processes, including the outcomes, as well as a link
  or other access point to any supporting documentation.
  
    N/A. In this study, humans are not the subjects. Their voluntary feedback, however, is used to filter out incorrectly labelled samples from the original 6 datasets.
    
\end{itemize}

\textit{If the dataset does not relate to people, you may skip the remaining questions in this section.}

\begin{itemize}

\item \textbf{Did you collect the data from the individuals in
    question directly, or obtain it via third parties or other sources
    (e.g., websites)?}

    We only involved individuals in the label verification step (i.e. 3133 samples).
    The answers from individuals directly affect if one of those 3133 samples will be kept or not.

\item \textbf{Were the individuals in question notified about the data
    collection?} If so, please describe (or show with screenshots or
  other information) how notice was provided, and provide a link or
  other access point to, or otherwise reproduce, the exact language of
  the notification itself.

    N/A. In this study, humans are not the subjects. Their voluntary feedback, however, is used to filter out incorrectly labelled samples from the original 6 datasets.

\item \textbf{Did the individuals in question consent to the
    collection and use of their data?} If so, please describe (or show
  with screenshots or other information) how consent was requested and
  provided, and provide a link or other access point to, or otherwise
  reproduce, the exact language to which the individuals consented.

    N/A. In this study, humans are not the subjects. Their voluntary feedback, however, is used to filter out incorrectly labelled samples from the original 6 datasets.

\item \textbf{If consent was obtained, were the consenting individuals
    provided with a mechanism to revoke their consent in the future or
    for certain uses?} If so, please provide a description, as well as
  a link or other access point to the mechanism (if appropriate).

    N/A. In this study, humans are not the subjects. Their voluntary feedback, however, is used to filter out incorrectly labelled samples from the original 6 datasets.

\item \textbf{Has an analysis of the potential impact of the dataset
    and its use on data subjects (e.g., a data protection impact
    analysis) been conducted?} If so, please provide a description of
  this analysis, including the outcomes, as well as a link or other
  access point to any supporting documentation.

    N/A. In this study, humans are not the subjects. Their voluntary feedback, however, is used to filter out incorrectly labelled samples from the original 6 datasets.

\item \textbf{Any other comments?}

No.

\end{itemize}

\subsection{Preprocessing/cleaning/labeling}

Dataset creators should read through these questions prior to any
preprocessing, cleaning, or labeling and then provide answers once
these tasks are complete. The questions in this section are intended
to provide dataset consumers with the information they need to
determine whether the ``raw'' data has been processed in ways that are
compatible with their chosen tasks. For example, text that has been
converted into a ``bag-of-words'' is not suitable for tasks involving
word order.\\

\begin{itemize}

\item \textbf{Was any preprocessing/cleaning/labeling of the data done
    (e.g., discretization or bucketing, tokenization, part-of-speech
    tagging, SIFT feature extraction, removal of instances, processing
    of missing values)?} If so, please provide a description. If not,
  you may skip the remaining questions in this section.

  Yes, we did cleaning up the data.
  We removed 370 images associated with the labels \class{sunglass}, \class{sunglasses}, \class{tub}, \class{bathtub}, \class{cradle}, \class{bassinet}, \class{projectile}, and \class{missile}, \ie, the classes that often contain similar images that belong to more than one class.
  Also, the agreement human setup in \cref{sec:imagenet_hard} helps us remove bad samples.

\item \textbf{Was the ``raw'' data saved in addition to the preprocessed/cleaned/labeled data (e.g., to support unanticipated future uses)?} If so, please provide a link or other access point to the ``raw'' data.

No, we did not save. However, the ``raw'' data (i.e. six image datasets) can be found on the Internet.

\item \textbf{Is the software \textit{that was} used to preprocess/clean/label the data available?} If so, please provide a link or other access point.

Yes, we provided the source code for the algorithms that can be found at \href{https://github.com/taesiri/ZoomIsAllYouNeed}{here}.
\item \textbf{Any other comments?}

No.

\end{itemize}

\subsection{Uses}

The questions in this section are intended to encourage dataset
creators to reflect on the tasks for which the dataset should and
should not be used. By explicitly highlighting these tasks, dataset
creators can help dataset consumers to make informed decisions,
thereby avoiding potential risks or harms.\\

\begin{itemize}

\item \textbf{Has the dataset been used for any tasks already?} If so, please provide a description.

The dataset was used for image classification in this paper.

\item \textbf{Is there a repository that links to any or all papers or systems that use the dataset?} If so, please provide a link or other access point.

The papers that use the dataset could be found at this \href{https://paperswithcode.com/dataset/imagenet-hard}{paperwithcode} repository.

\item \textbf{What (other) tasks could the dataset be used for?}

Studying the effects of upscaling on image classification.

\item \textbf{Is there anything about the composition of the dataset or the way it was collected and preprocessed/cleaned/labeled that might impact future uses?} For example, is there anything that a \textit{dataset consumer} might need to know to avoid uses that could result in unfair treatment of individuals or groups (e.g., stereotyping, quality of service issues) or other risks or harms (e.g., \textit{legal risks,} financial harms)? If so, please provide a description. Is there anything a \textit{dataset consumer} could do to mitigate these risks or harms?

No. 

\item \textbf{Are there tasks for which the dataset should not be used?} If so, please provide a description.

No.

\item \textbf{Any other comments?}

No.
\end{itemize}

\subsection{Distribution}

Dataset creators should provide answers to these questions prior to
distributing the dataset either internally within the entity on behalf
of which the dataset was created or externally to third parties.\\

\begin{itemize}

\item \textbf{Will the dataset be distributed to third parties outside of the entity (e.g., company, institution, organization) on behalf of which the dataset was created?} If so, please provide a description.

The dataset will be shared with the common public.

\item \textbf{How will the dataset will be distributed (e.g., tarball on website, API, GitHub)?} Does the dataset have a digital object identifier (DOI)?

It has a \href{https://github.com/taesiri/ZoomIsAllYouNeed}{GitHub repository} and \href{https://huggingface.co/datasets/taesiri/imagenet-hard}{Hugging Face page}. 
It has no DOI.

\item \textbf{When will the dataset be distributed?}

Until June 5 2023, the dataset had been already distributed.

\item \textbf{Will the dataset be distributed under a copyright or other intellectual property (IP) license, and/or under applicable terms of use (ToU)?} If so, please describe this license and/or ToU, and provide a link or other access point to, or otherwise reproduce, any relevant licensing terms or ToU, as well as any fees associated with these restrictions.

No.

\item \textbf{Have any third parties imposed IP-based or other restrictions on the data associated with the instances?} If so, please describe these restrictions, and provide a link or other access point to, or otherwise reproduce, any relevant licensing terms, as well as any fees associated with these restrictions.

There are no such restrictions.

\item \textbf{Do any export controls or other regulatory restrictions apply to the dataset or to individual instances?} If so, please describe these restrictions, and provide a link or other access point to, or otherwise reproduce, any supporting documentation.

N/A.

\item \textbf{Any other comments?}

No.

\end{itemize}

\subsection{Maintenance}

As with the questions in the previous section, dataset creators
should provide answers to these questions prior to distributing the
dataset. The questions \textit{in this section} are intended to
encourage dataset creators to plan for dataset maintenance and
communicate this plan to dataset consumers.\\

\begin{itemize}

\item \textbf{Who \textit{will be} supporting/hosting/maintaining the dataset?}

The authors of the paper will be supporting/hosting/maintaining the dataset.

\item \textbf{How can the owner/curator/manager of the dataset be contacted (e.g., email address)?}

Please reach out to \href{mailto:mtaesiri@gmail.com}{mtaesiri@gmail.com}  and \href{mailto:anhng8@gmail.com}{anh.ng8@gmail.com}.

\item \textbf{Is there an erratum?} If so, please provide a link or other access point.

No, there is no erratum.

\item \textbf{Will the dataset be updated (e.g., to correct labeling
    errors, add new instances, delete instances)?} If so, please
  describe how often, by whom, and how updates will be communicated to
  \textit{dataset consumers} (e.g., mailing list, GitHub)?

Yes, we anticipate either correcting the labels or removing some images entirely to reduce the noise in the dataset.
We manage the versioning through the  \href{https://huggingface.co/docs/datasets/index}{Hugging Face Dataset} platform.

\item \textbf{If the dataset relates to people, are there applicable
    limits on the retention of the data associated with the instances
    (e.g., were the individuals in question told that their data would
    be retained for a fixed period of time and then deleted)?} If so,
    please describe these limits and explain how they will be
    enforced.

    N/A

\item \textbf{Will older versions of the dataset continue to be
    supported/hosted/maintained?} If so, please describe how. If not,
  please describe how its obsolescence will be communicated to \textit{dataset
  consumers}.

  No, the older versions will not be updated. Users are encouraged to use the latest version.

\item \textbf{If others want to extend/augment/build on/contribute to
    the dataset, is there a mechanism for them to do so?} If so,
  please provide a description. Will these contributions be
  validated/verified? If so, please describe how. If not, why not? Is
  there a process for communicating/distributing these contributions
  to \textit{dataset consumers}? If so, please provide a description.

  Please reach out to \href{mailto:mtaesiri@gmail.com}{mtaesiri@gmail.com}  or \href{mailto:anhng8@gmail.com}{anhng8@gmail.com}.

\item \textbf{Any other comments?}

No.

\end{itemize}

\end{document}